\documentclass[final,5p,times,twocolumn]{elsarticle}

\usepackage[hang]{subfigure}

\usepackage{tikz-cd}

\usepackage{lineno,hyperref}
\modulolinenumbers[5]

\journal{Neurocomputing}

\newtheorem{proposition}{Proposition}

\usepackage{caption}

\usepackage{paralist}
\usepackage{amscd}

\usepackage{tikz}
\usepackage{tikz-3dplot}
\usepackage{tkz-euclide}
\usetikzlibrary{calc,through}
\usepackage{calc}% http://ctan.org/pkg/calc
\usetikzlibrary{calc,intersections,through,backgrounds}
\usepackage{tkz-euclide}

\usepackage{amsmath}
\usepackage{amssymb}
\usepackage{amsfonts}
\usepackage{mathrsfs}
\usepackage{subfigure}
\usepackage{epsfig}
\usepackage{graphicx}
\usepackage{color}

\usepackage{float}

\DeclareCaptionFormat{cont}{#1 (cont.)#2#3\par}

\usepackage[linesnumbered,ruled,]{algorithm2e}
\newenvironment{proof}{\par\noindent{\bf Proof: }}{\hfill$\myBox$\par}
\graphicspath{{../RPM_model_occlude/}} 

\begin{document}

\begin{frontmatter}

\title{Hybrid Trilinear and Bilinear  Programming for Aligning Partially Overlapping Point Sets}

%\author{Wei Lian }
%\address{Dept. of Computer Science, Changzhi University, Changzhi, Shanxi, China, 046031 }
%\ead{lianwei3@foxmail.com}
%%\fntext[myfootnote]{Since 1880.}

\author[mymainaddress]{Wei Lian} %\corref{mycorrespondingauthor}}
\ead{lianwei3@gmail.com}

\author[mysecondaryaddress]{Wangmeng Zuo}
\ead{cswmzuo@gmail.com}

%\author[mymainaddress]{Zhesen Cui}
%\ead{cuizhesen@gmail.com}

\address[mymainaddress]{Department of Computer Science, Changzhi University, Changzhi, Shanxi, China, 046011}
\address[mysecondaryaddress]{School of Computer Science and Technology, Harbin Institute of Technology, Harbin 150001, China}

\begin{abstract}	

%Registration problems where
%Partially overlapping point sets that are not initially coarsely aligned are difficult  to register.
%	
In many applications, 
we need  algorithms which can align partially overlapping point sets
and are invariant to the corresponding transformations.
%These algorithms can be used to  provide  initial coarse alignment  for local search  based  methods  such as  ICP.
%Aligning partially overlapping point sets 
%where \textcolor{red}{
%there is no prior information about the values of transformation parameters}
%is a challenging problem 
%in computer vision.
In this work,
a method possessing such properties is realized by  minimizing the objective of the robust point matching (RPM) algorithm.
%To this end,
%Aiming at designing algorithms capable of handling these issues, 
%To this end,
We first  show that the RPM objective  
%the robust point matching (RPM) algorithm 
is a cubic polynomial. %   with few  variables.
We then utilize the convex envelopes of trilinear and bilinear monomials to derive its   lower bound function.
%with the nice property that 
The resulting lower bound problem has the merit that it can be efficiently solved via  linear assignment  and low dimensional convex quadratic programming.
%
%We then show that for the lower bound function to converge to the objective funciton,
%we only need to branch over the transformation variable.	
%
We next develop a branch-and-bound (BnB)  algorithm which only branches over the transformation variables and
%thus has a lower dimensional search space than previously proposed methods.
%converges quickly.
runs efficiently.
%	The branching space of our branch-and-bound algorithm has a dimension equal to the number of transformation  parameters.
%Thus, our method can converge  quickly. 
%
Experimental results demonstrated better robustness   of the proposed method 
against non-rigid deformation, positional noise and outliers in case that outliers are not mixed with inliers
when compared with  the state-of-the-art approaches.
They  also showed  that it has competitive efficiency and   scales well with problem size.
%  in terms of  robustness and speed.
\end{abstract}


\end{frontmatter}

%\end{document}

%\linenumbers

%\begin{figure*}
%\begin{tikzcd}
%	E(\mathbf P,\boldsymbol{\theta})
%	\triangleq\sum_{i,j}p_{ij}\|\mathbf y_j-\mathbf T(\mathbf x_i| \boldsymbol{\theta})\|^2
%=		 
%\text{trilinear term of }\boldsymbol\theta \text{ and } \mathbf P
%+
%\text{bilinear term of }\boldsymbol\theta \text{ and } \mathbf P
%\arrow[rd] \arrow[r, "\phi"]
%		  & B \\
%		& C	 
%%	=\text{quadratic term of }\boldsymbol\theta
%%	+	&\boxed{\text{trilinear term of }\boldsymbol\theta \text{ and } \mathbf P
%%	} +
%%	&\boxed{\text{bilinear term of }\boldsymbol\theta \text{ and } \mathbf P
%%} 
%%+&\text{linear term of } \mathbf P
%%	 \arrow[d]  \\
%%	& & & C
%\end{tikzcd}
%\caption{text}
%\end{figure*}

\begin{figure*}[!ht]
	\begin{equation*} % [H]
		%	\begin{tabular}{@{\hspace{-0mm}}c}	
			%		$
			\begin{CD}
 E(\mathbf P,\boldsymbol{\theta})
 \triangleq\sum_{i,j}p_{ij}\|\mathbf y_j-\mathbf T(\mathbf x_i| \boldsymbol{\theta})\|^2
				%E(\mathbf p,\boldsymbol \theta)
				=				@.
\textcolor{red}{				\boxed{\text{trilinear term of }\boldsymbol\theta \text{ and } \mathbf P
				}} 
				@. + 
\textcolor{blue}{				\boxed{\text{bilinear term of }\boldsymbol\theta \text{ and } \mathbf P }}
				@. +
				\text{quadratic term of }\boldsymbol\theta +\text{linear term of } \mathbf P	
				\\@.
				 @V\text{trilinear}V \text{relaxation}V  @V\text{bilinear}V\text{relaxation}V @.
				\\
				E_l(\mathbf P,\boldsymbol{\theta})=		
				@.					
\textcolor{red}{				 \boxed{
				\text{linear term  of }\boldsymbol\theta \text{ and } \mathbf P
				}}
				@.
				+
\textcolor{blue}{				\boxed{
				\text{linear term of }\boldsymbol\theta \text{ and } \mathbf P
				}}
			@.+
			\text{quadratic term of }\boldsymbol\theta +\text{linear term of } \mathbf P
			\end{CD}
			%		$
			%	\end{tabular}		
	\end{equation*} 
			\caption{
				Derivation of  a  lower bound function of the RPM objective 
%				$E(\mathbf P,\boldsymbol{\theta})$ 
				via blinear and trilinear relaxation.
%	The resulting lower bound function can be efficiently minimized by  linear assignment  and low dimensional convex quadratic programming.
Here 
$\mathbf x_i$ and $\mathbf y_j$ denote model point $i$ and scene point $j$, respectively.
$\mathbf P =\{p_{ij} \}$ denotes the correspondence matrix
 with $p_{ij} = 1$ indicating that there is a correspondence between $\mathbf x_i$ and $\mathbf y_j$ and $p_{ij} = 0$ otherwise,
 $\mathbf T(\cdot|\boldsymbol\theta)$ denotes the transformation with parameters
 $\boldsymbol\theta$.
%  denotes the transformation parameters.
%We choose $\mathbf T(\mathbf x_i|\boldsymbol\theta)=\mathbf J(\mathbf x_i)\boldsymbol{\theta}$.
		\label{RPM_lb_idea}		%
	}
\end{figure*}

\section{Introduction}

%\begin{tikzcd}AA\arrow[rd]\arrow[r, "\phi"] & B \\& C\end{tikzcd}
%
%
%\begin{tikzcd}AA\arrow[rd]\arrow[rd, "\phi"] & B \\& C\end{tikzcd}

%\begin{tikzpicture}[commutative diagrams/every diagram]\node(P0)at (90:2.3cm){$X\otimes(Y\otimes (Z\otimes T))$};\node(P1)at (90+72:2cm){$X\otimes((Y\otimes Z)\otimesT))$} ;\node(P2)at (90+2*72:2cm){\makebox[5ex][r]{$(X\otimes (Y\otimes Z))\otimesT$}};\node(P3)at (90+3*72:2cm){\makebox[5ex][l]{$((X\otimes Y)\otimesZ)\otimesT$}};\node(P4)at (90+4*72:2cm){$(X\otimes Y)\otimes(Z\otimes T)$};\path[commutative diagrams/.cd,every arrow,every label](P0)edgenode[swap] {  $1\otimes\phi$} (P1)(P1)edgenode[swap] {  $\phi$} (P2)(P2)edgenode{$\phi\otimes1$} (P3)(P4)edgenode{$\phi$} (P3)(P0)edgenode{$\phi$} (P4);
%\end{tikzpicture}

Registration of 2D/3D point sets is  a crucial step in many applications of  computer vision, robotics and remote sensing.
%aims to find a spatial  transformation that best aligns two point sets
%to a common coordinate system.
%It has extensive applications in 3D reconstruction, 3D localization, pose estimation, panorama stitching and  medical imaging. 
%Point set alignment is a fundamental  problem
%and medical image analysis.
%with applications such as structure-from-motion, tracking, 
%image retrieval and object recognition \cite{book_multiple_view}. 
% aiming at matching the data (target) point set to the model (template) point set [5]. 
Apart from difficulties such as non-rigid deformation and positional noise,
this problem is particularly challenging when  two point sets
 only partially overlap and 
 %the poses of two point sets are unknown.
are not initially coarsely aligned.
%Disturbances such as non-rigid deformation, positional noise, partial overlap and unknown poses %in the point sets 
%render this problem rather  challenging.
%
%

Point set registration can be accomplished 
by minimizing the objective of the robust point matching (RPM) algorithm \cite{RPM_TPS}. 
Under the condition that one point set can be embedded in another  set,
Lian \textit{et al.}  reduced the RPM objective  to a concave quadratic function of  point correspondence \cite{RPM_concave_PAMI}.
The resulting function has the important  property that the Hessian matrix has a low rank.
Thus the branch-and-bound (BnB) algorithm with a low dimensional branching space can be employed for efficient optimization.
%after  eliminating the transformation variable
%and then employed the normal rectangular algorithm
%(a variant of the branch-and-bound (BnB) algorithm) 
%for optimization   
%The method has good  robustness.
%\cite{RPM_concave_PAMI} also proposes an efficient lower bound scheme which involves solving a linear assignment problem. % and can be efficiently done.
%

This  algorithm  is  generalized  to the case when  two  point sets only partially overlap in \cite{RPM_model_occlude}.
%by allowing both point sets to have outliers,
The RPM objective   is reduced to a  concave function of point correspondence,
which, although not quadratic, still has a low rank structure.
% by eliminating the transformation variable.
%The resulting function, 
%which, albeit  not quadratic, still has 
This enables efficient 
%normal simplex algorithm
%(a variant of
 BnB based optimization. 
%\cite{RPM_model_occlude}. 
%There are several  drawbacks with the method.
%
%First, 
But the branching space of the algorithm is  a projected space of point correspondence,
whose dimensionality is, different from  \cite{RPM_concave_PAMI},
higher than the  number of   transformation parameters,
leading to slow convergence.
%causing the method to converge slowly.
%the method 
Furthermore,
to meet the requirement that 
%the  algorithm requires that 
the objective  is  concave over the initial simplices,
%which is larger than the concavity region of the objective function.
%a sufficiently large region,
%whereas it is only concave over a finite region.
%
% whose union includes the feasible region,
%
%is not necessarily concave outside the feasible region.
%For this,
%ensure the concavity of the objective function over the covering simplexes, 
the method enlarges the concavity region of the objective  
by imposing  regularization  on transformation 
where prior information about the transformation is required.
However, this  causes the method  biased in favor of generating solutions only consistent with the prior information. 
%Thus, the method is not invariant to the corresponding transformation.
%So the method is unsuitable for the matching problems 
%where  \textcolor{red}{there is no prior information about the values of transformation parameters.}
%Finally,
%the number of matches need to be a priori known.
%This requirement may not be satisfied in certain applications.

%To address these issue, in this paper, 
%The key reason that the BnB algorithm  of \cite{RPM_model_occlude} has a high dimensional branching space is that it is 
%branching over a projected space of the 
% point correspondence variable,
%whereas the latter space has a dimension much higher than the number of the transformation parameters.

To address the issue that \cite{RPM_model_occlude} requires regularization on transformation, 
%the method of \cite{RPM_model_occlude}  
%is not invariant to the corresponding transformation,
%needs prior information about the values of transformation parameters,
Lian and Zhang utilized
the  constraints of  rigid transformations to derive a new objective  of point correspondence 
which is concave over the entire vector space in \cite{RPM_model_occlude_PR},
thus regularization on transformation is not needed when applying the BnB algorithm to the minimization of the objective.
%avoiding the need of imposing additional 
% and 
%the method 
%is  invariant to the corresponding transformation.
% instead of regularization on transformation.
%
From a different perspective,
Lian \textit{et al.}  
applied the polyhedral annexation (PA) algorithm to the minimization of the objective of RPM \cite{lian2021polyhedral}.
Different from BnB,
PA has the advantage of only operating in the concavity region of the  objective, thus 
avoiding the need of  regularizing transformation.
Nevertheless, like  \cite{RPM_model_occlude},
%the same as \cite{RPM_model_occlude},
both \cite{RPM_model_occlude_PR} and \cite{lian2021polyhedral} still  branch   a large dimensional projected space of point correspondence,
leading to slow convergence. 
Besides, unlike BnB,
PA has the disadvantage of having to cut an increasingly complex polygon as the algorithm progresses. This  further worsens  the  convergence speed of the algorithm. 

%by proposing using an alternative  algorithm, PA has advantage over BnB in that 
%The method 

\begin{tabular}{l|l}
	
\end{tabular}

%optimizing the resulting objective function of the correspondence variable by eliminating transformation variable. 
%However, the dimension of the projected  space of the correspondence variable cannot be further reduced.

%Since the use of the BnB algorithm leads to  
%regularization on transformation in \cite{RPM_model_occlude},
%\cite{RPM_inner_approx}  proposes to instead use the  inner approximation algorithm \cite{book_concave_intro} for optimization,
%which has the advantage of only operating within the feasible region of a projected space of   point correspondence,
%where 
%the objective  function is naturally concave regardless of whether regularization on transformation is used.
%But 
%the projected space %of point correspondence
%has a dimension equal to that of \cite{RPM_model_occlude},
%causing the method to converge slowly.
%Thus, 
%no full 3D affine transformation is demonstrated.

Aiming at aligning partially overlapping point sets which are not initially  coarsely aligned,
in this work,
we propose an alternative BnB based approach to minimize the objective  of RPM.
We argue that it is advantageous to retain the transformation variable instead of eliminating them as are the practices of Ref. \cite{RPM_concave_PAMI, RPM_model_occlude, RPM_model_occlude_PR,lian2021polyhedral}.
The reasons as well as the contributions of this work are listed as follows.
\begin{itemize}
	\item 
	The objective  of RPM is a cubic polynomial.
	Thus, the convex envelopes of trilinear and bilinear monomials can be  utilized to derive the lower bound function,
	as illustrated in Fig. \ref{RPM_lb_idea}.
	
	\item
	The resulting lower bound problem can be decomposed into separable
	linear assignment and 
	low dimensional convex quadratic program,  both of which can be efficiently solved.
	\item
	For the lower bound function to converge to the objective function,
	we only need to  branch over the transformation variable,
%	Thus,	
%	the branching space of the proposed BnB algorithm has a dimensionality equal to  the number of transformation variables,
which is in contrast to 
	 previously attempts \cite{RPM_model_occlude,RPM_model_occlude_PR,lian2021polyhedral}
with large branching spaces.
%can slow   the  convergence of the algorithms.
Consequently, 
the proposed method   converges quickly in comparison with previous attempts.
%	to converge  quickly.
%	than those of 
	\item
	Unlike \cite{RPM_model_occlude},
%	for relatively easy registration problems,
	our method %can %be enabled not 
 needs not  to   
	impose   requirements on   transformation. % or point correspondence,
	Consequently,
	the proposed method is  invariant to the corresponding transformation
	and can be applied to situations where initial coarse alignment is not available.
\end{itemize}

%Thus,
%it can  handle  the  case where \textcolor{red}{prior information about the values of transformation  parameters is unknown.}
%Our  contributions are summarized as follows:
%\begin{enumerate}
%\item 
%The convex envelopes of trilinear and bilinear monomials are utilized to develop a novel  lower bound function of the objective  of RPM.	
%
%	\item 
%The proposed BnB algorithm directly branches over the transformation variable.	
%%The branching space of our BnB algorithm has a dimension equal to the number of transformation parameters.
%This enables our method to converge more quickly than
%\cite{RPM_model_occlude,RPM_model_occlude_PR}.
%
%\item
%The lower bound problem can be efficiently solved via linear assignment and 
%low dimensional convex quadratic programming.	
%	
%
%
%\item 
%Our method can impose no  requirements on   transformation, % or point correspondence,
%thus,
%it can be made  invariant to the corresponding transformation.	
%\end{enumerate}

%Our method is conceptionally simple and elegant.
%To the best of our knowledge,
%our method is the first to achieve practical globally optimal 3D affine alignment.
%Majority of the state-of-the-art 3D alignment methods only focus on the case of 3D rigid transformation,  
%causing them to struggle   with  nonrigid deformation.
%By allowing more degrees of transformation freedom, our method can better handle nonrigid deformation.

The remainder of the paper is organized as follows:
We first review related work (Sec.\ \ref{sec:relate_work}).
We then present  theoretical framework needed for the development of our lower bound functions (Sec.\ \ref{sec:cov_env}).
We  next discuss our objective functions and optimization strategies under  two different types of transformations (Sec.\ \ref{sec:case_one} and \ref{sec:case_two}).
%In Sec.\ \ref{sec:BnB}, we present the BnB algorithm.
We finally  present experimental results (Sec.\ \ref{sec:exp}) 
and conclude the paper (Sec. \ref{sec:conclude}).

\section{Related work \label{sec:relate_work}}
A survey of point set registration methods can be found in 
\cite{10.1007/s11263-020-01359-2}.
Here we only mention work close related with ours.

\subsection{Local    methods} 

\textbf{
	Simultaneous pose and correspondence.
%Methods modeling both transformation and  correspondence.
} %ICP based methods.}
%The  methods closely related to ours are those modeling both transformation and point correspondence.
%The  iterative closest point (ICP) method 
%is  well known because of its simplicity and speed.
 The well known ICP method \cite{ICP}
iterates between recovering  point correspondence 
%based on nearest-neighbor relationship 
and updating spatial transformation.
%as a least square problem. % based on the estimated correspondence.
% iterates between recovering  point correspondence %based on  nearest neighbor relationship 
%and
%updating transformation. %as a least square problem.
Despite being  efficient,
the discrete nature of point correspondence causes the method to be  easily trapped in local minima. 
%due to  the discrete nature of point correspondence.
% causes this method prone to be trapped in local minima.
% and only good initial alignment can guarantee success.
%To address this issue,
Zhang \textit{et al.} 
improved the robustness of ICP   by bounding the rotation angle at each iterative step \cite{ICP_2D_rot}.
%, which reduces the variety of the transformation.
%
Liang  supplemented RGB information to the registration process of ICP 
%and  used the correntropy measurement criterion to overcome noise
 \cite{ICP_rgb_correntropy}.
%precise iterative closest point algo-
%rithm for RGB-D data registration which not only supplement the
%Dong \textit{et al.}  improved the Trimmed Iterative Closest Point (TrICP) algorithm by representing the transformations in the registration problem as elements in a Lie group \cite{LieTrICP}.
%
%To address this issue,
The RPM method  \cite{RPM_TPS} 
relaxes point correspondence to be fuzzily valued and uses deterministic annealing (DA) to gradually recover  point correspondence.
%improves ICP by
%adopting the strategies of soft assign and    deterministic annealing (DA). %for optimization
%to gradually determine point assignment.
%But the fuzzinness of correspondences generated by DA is  homogeneous,
%causing the method  biased towards aligning  the mass  centers of  point sets.
%
Instead of DA,
Sofka \textit{et al.} use
the covariance matrix of the transformation parameters   to  
determine  the fuzziness  of point correspondence,
%which is more anisotropic
% the CDC algorithm 
resulting  in  better robustness to missing or extraneous structures  \cite{CDC}.
Ma \textit{et al.} used
L2E  to robustly estimate transformation with application to non-rigid registration \cite{ieee7001713}.
%But the method is only suitable for simple transformation 
%since  the size of the covariance matrix is square times the number of transformation parameters.
Our method is  related with the above methods by also  modeling  spatial transformation and correspondences.
But our objective function is globally minimized.
%  instead of by local   methods,
This enables  our method to be more robust to disturbances.

%We introduce a new transformation estimation algorithm using the L 2 E estimator and apply it to non-rigid registration for building robust sparse and dense correspondences. 

%The $L_2E$ estimator is introduced 
%in \cite{L2E_mismatch}
%to more robustly estimate the spatial transformation.
%What's  common  with the above methods is that they are all heuristically based.
%Thus,  these methods may fail if the employed heuristics don't fit well the alignment problem.

\textbf{Correspondence-free methods.}
%	GMM based methods.}
%
%Majority of the recently proposed  methods are  based on Guassian mixture model (GMM).
%modeling only  transformation.
%Methods in this category generally utilize the idea that
%a point set can be viewed  as the result of sampling from a distribution.
%Among these methods,
The CPD method \cite{CPD_match} casts  point matching  
as the problem of fitting a Guassian mixture model (GMM)  representing one point set to another point set.
%The expectation-maximization algorithm is used for optimization.
%To eliminate the need for solving for point  correspondence,
%Glaunes \textit{et al.} \cite{diffeo_match} formulated point matching as 
%aligning two weighted sum of Dirac measures representing two point sets.
%But  Dirac measures  is difficult to numerically compute.
% To address this issue,
GMMREG   \cite{kernel_Gaussian_journal} uses two GMMs to represent two point sets  
and minimizes the $L_2$-norm distance between them.
Support vector parameterized Gaussian mixtures (SVGMs) is used to  represent point sets  using sparse  Gaussian components in
\cite{mixture_SVC}.
The efficiency of GMM based methods is improved by  using filtering 
to solve the  correspondence  problem
in \cite{GMM_filtering}.
%
%Yang \textit{et al.} used the global and local mixture distance for non-rigid point set registration in \cite{GLMDTPS},
%achieving  promising results.
%Feature guided  GMM is used for key point matching  in  \cite{FG_GMM}.
%%dual-feature finite mixture
%%model and global-local structural preservation is proposed in .
%The dual-feature mixture model is used  for point set registration in \cite{dual_feat_GMM}, where the global-local
%structure is preserved by using two regularization terms.
%%
%Reweighted discriminative optimization is used to 
%%(RDO), an asymmetrical parameter treatment scheme to
%improve the accuracy of parameter estimation in least-squares problems
%%point cloud registration 
%in \cite{reweig_discrim_opt_LS}.
%
The density variation problem of point sets is addressed in 
\cite{mixture_average}  by modeling the structure of scene as well.
Hierarchical Gaussian mixture representation is used in \cite{tree_Gaussian_mixture}  to improve the speed and accuracy of registration.
Ma et al. proposed a
Non-Rigid Point Set Registration method by Preserving Global and Local Structures \cite{ieee7185406}.

\subsection{Global  methods}

\textbf{BnB based methods.}
The  BnB algorithm is a popular global optimization technique.
%  widely used in computer vision.
Based on the Lipschitz  theory,
BnB is used to align  3D shapes in
 \cite{Lipschitz_3D_align}.
% by exploiting the special structure of the 3D rotation space.
But the method does not allow
%can only apply when there are no
  outliers or occlusion.
BnB is also used to solve for 3D rigid transformation in \cite{branch_bound_align}.
But the 
%point-to-point/line/plane
correspondences need to be known a priori,
which limits the applicability  of the method.
% An alternative optimization method for the RPM objective function was proposed in  \cite{RPM_concave}
% by assuming  the transformation is linear with respect to  its parameters.
% Under this requirement, 
% the optimization problem can be reduced to a low rank concave quadratic program
% in the point correspondence variable.
% This method was later extended to allow partial overlap between two point sets in \cite{RPM_model_occlude},
% where the objective function, although not being quadratic, is still concave and has a low rank structure.
By leveraging  the  structure of the geometry of 3D rigid motions, %$SE(3)$.
Go-ICP   \cite{Go-ICP_pami} globally optimizes the objective  of ICP.
GOGMA \cite{BnB_mixture_Gaussian}
achieves point set registration by aligning the GMMs constructed from the original point sets.
% The method is quite efficient.
% But the method requires that each point in one point set has a counterpart in another point set.
% which excludes cases such as outliers exist in both point sets.
%BnB was recently  applied to  the problem of  
%consensus set maximization (CSM) \cite{BnB_consensus_integer},
%% This is a problem with a similar goal as RANSAC seeking the transformation maximizing the number of inliers.
%which seeks the best transformation maximizing the number of inliers.
%% largest subset of consistent correspondences.
%% which has the same goal as that of RANSAC by seeking the best 
%The CSM framework was used for the correspondence and grouping problems 
%% A general framework for CSM was proposed 
%in \cite{BnB_consensus_group}.
%But 
%the method is quite slow due to lack of efficient optimization techniques 
%for the resulting bounding problems.
%In the case that there is only 3D rotation between two point sets,
Based on stereographic projection,
the fast rotation search (FRS) method 
%Efficient methods have been proposed 
efficiently recovers 3D rotation between two point sets \cite{BnB_consensus_project}.
%Then, rigid registration is achieved via a nested BnB algorithm.
%
%The two GMMs with 50 components were generated by
%support-vector-parameterized Gaussian mixtures (SVGMs)
%[46], which are also time-consuming processes.
%
%Straub et
%al. [36] proposed a two-stage BnB algorithm using a data
%structure based on surface normals. 
%Liu \textit{et al.} proposed a
A rotation-invariant feature is proposed in   \cite{BnB_rigid_regist_decom},
resulting in an efficient BnB based registration algorithm.
%	 decomposed  the search of
%	a 6D rigid transformation into separate searches for a 3D
%Rotation is subsequently recovered. 
%This results in a more efficient BnB algorithm.
%	
%Straub \textit{et al.}
%separate 3D rotation from translation by using surface normals.
%They also
%proposed 
%an   alignment algorithm
%based on 
%A new way of representing point sets based on
Bayesian nonparametric mixture and   
a novel approach of tessellating rotation space
are proposed in \cite{BnB_Bayesian_mixture},
leading to an efficient alignment algorithm.
Our BnB based method differs from the above methods in that we also model point correspondences.
This enables our  method to better handle non-rigid deformation.

\textbf{Mismatch removal methods.}
Another line of research focuses on recovering spatial transformation given putative point correspondences.
%The robustness of BnB based methods is often achieved at the expense  of decreased efficiency.
%%There are also global registration methods 
%Efficient global registration methods can be developed if candidate matches are also available.
% and the resulting methods can be  fast.
%For example,
The fast global registration (FGR) method \cite{fast_global_regist}
%successively
 optimizes a robust objective
based on 
%Black-Rangarajan
duality between line processes and  robust estimation.
GORE  \cite{GORE_outlier_removal}
removes large portion of outliers based on geometric operations before RANSAC is invoked.
TEASER++  \cite{TEASER}
uses a graph-theoretic method to
decouple  rotation, scale   and translation estimation, and adopts a truncated least squares formulation for each subproblem. 
%

%where rotation estimation can be relaxed to a semidefinite program.

%Inner approximation algorithm  is used to optimize the RPM objective in \cite{RPM_inner_approx},
%waiving the need of regularization on transformation.
%Bayesian nonparametric point cloud representation 
%and 
%a
%novel way of tessellating  rotation space in 
%leading to  more efficient optimization 
%for solving for 3D rotations.
%based on surface normals.
%
%The above methods  are all  targeted at rigid registration. 
%Therefore, they  may not cope well with scaling  and non-rigid deformation.

%BnB is also used to optimize the RPM objective function in \cite{B_B_RPM},
%where branching over the correspondence variable 
%and over the transformation variable are both considered.
%However, due to the lack of good  structures for optimization,
%the  methods are only suitable for small scale problems. 

%Our algorithm distinctly 
%differs  from the above methods 
%in that 
%instead of using BnB,
%we use an alternative global optimization  algorithm,
%the inner approximation algorithm, for optimization,
%which is particularly suited for our problem formulation
%where optimization over a finite  concave region is required.

\subsection{Deep learning methods}
%Deep learning is particularly good at extracting features. 
%To extract features from point sets,
%methods like PointNet, DGCNN have been proposed.

\textbf{Correspondence-free methods.}
%PointNetLK   employs
The Lucas–Kanade  algorithm is used to align  global features
of  point sets generated by a PointNet \cite{point_net} network in \cite{PointNetLK}.
%thus achieving point set registration.
%  to extract   global features of two point sets and uses the .
% Registration is achieved by LK algorithm.
%
%In a similar vein,
This method is  improved in \cite{feature_metric_regist}  by  using  
%using  an encoder to generate global features from  point sets and 
a  decoder to convert  the generated global features back into point sets.
%LK  is used to align the features
The fidelity of the global features is guaranteed by minimizing the Chamfer distances between the decoded point sets  and the original point sets.

\textbf{Correspondence-based methods.}
The FCGF  method \cite{full_conv_feat} uses sparse high dimensional convolutions to extract dense feature descriptors
from point clouds.
This method outperforms many hand-crafted features and PointNet-based methods. % by a large margin.
 It also outperforms a  3D convolution-based method \cite{perfect_match}.
The deep global registration method \cite{deep_global_reg} uses FCGF
feature descriptors for point cloud registration.
\cite{metric_learn_reality} uncovered several flaws of  metric learning based methods.
DCP  \cite{DeepClosestPoint} uses DGCNN \cite{DGCNN} and Transformer \cite{Transformer} to extract features for each point set
%Similarities of the features are used to determine point correspondences and
and SVD is used to recover rigid transformation.
PRNet \cite{PRNet}  extends DCP to handle  partially overlapping point sets  in an iterative way.
%It also employs Gumbel–Softmax 
%obtain  sharp 
%point correspondences.
%
IDAM \cite{li2020iterative} considers both geometric and distance features during its point matching process.
RIENet \cite{shen2022reliable} calculates
feature-to-feature correspondences with neighborhood consensus. 
\cite{shu2022partial} performs parital point set registration via matching credibility generation on the global semantic level and geometric
feature learning on the local structural level.

\section{Convex envelope linearization \label{sec:cov_env}}

In this section,
%we present  theoretical framework 
we discuss the convex envelopes of bilinear (Sec. \ref{sec:bil_lb}) and trilinear monomials (Sec. \ref{sec:lb_tri}) and their  linearizations.
The linearization results  play a crutial  role in 
the derivation  of the lower bound functions of our objective functions in section \ref{sec:case_one} and \ref{sec:case_two}.

\subsection{ Bilinear monomial %$-2Ap$
	\label{sec:bil_lb}
}

The upper part of Eq. \ref{conv_bilinear}
shows
the convex envelope
% (i.e.,  tightest convex underestimator)
of a bilinear monomial $xy$.
% within  ranges
%$\underline{x}\le x\le \overline{x}$,
%$\underline{y}\le y\le \overline{y}$
%is presented in.
%\begin{align}
%(xy)_{l1}
%\triangleq \underline{x}y+x\underline{y}-\underline{x}\underline{y}  \le xy\\
%(xy)_{l2}\triangleq \overline{x}y+x\overline{y}-\overline{x}\overline{y}  \le xy 
%\end{align}
%
If we directly adopt this form of convex envelope
when  designing the lower bound function of our objective function,
%(to be presented in Sec.\ \ref{subsec:obj_aff} and \ref{subsec:obj_3d_rigid}),
either term in the form of $\max\{(xy)_{l1},(xy)_{l2}\}$ 
needs to be incorporated into the lower bound function
or
subproblem in the form  of
$\min \{\alpha| \alpha\ge (xy)_{l1}, \alpha\ge (xy)_{l2} \}$
needs to be incorporated into the lower bound problem.
%But the lower bound function will not  be linear.
%Employing gradient descent like algorithms to optimize the lower bound function  will entail multiple steps of optimization which can be slow.
% is no longer a linear function of $\mathbf P$ 
%
For the former case,
the lower bound function is convex but not linear,
thus requiring  multiple steps of optimization,
% to solve the lower bound problem 
which is slow
and  susceptible to numerical inaccuracy issue.
For the  latter case,
the lower bound problem will not take the form of 
a combinatorial optimization problem,
%the linear assignment problem.
causing
% no  combinatorial optimization algorithms can be employed  and our
the resulting algorithm to be inefficient.
%
%This form of lower convex envelope is still too complex to be directly used in the design of the lower bound function of $E$.
%
\begin{equation} %[H]
		\begin{CD}
			xy@>\text{ convex envelope}>
		\begin{matrix}
\scriptstyle	\underline{x}\le x\le \overline{x},\\
\scriptstyle	\underline{y}\le y\le \overline{y}
			\end{matrix} 
			>\begin{cases}
	(xy)_{l1}\triangleq \underline{x}y+x\underline{y}-\underline{x}\underline{y}  \le xy ,\\
 (xy)_{l2}\triangleq \overline{x}y+x\overline{y}-\overline{x}\overline{y}  \le xy 
			\end{cases}\\
		@. @V\text{ average}VV \\		
%				@>\text{average}>>  
@.		(xy)_{l} \triangleq
			\frac{1}{2} \sum_i (xy)_{li}\le xy
		\end{CD}		
	\label{conv_bilinear}
\end{equation} 

To address the above issues, 
similar to the practice of  \cite{upper_bound_linear},
%in this work,
we  construct a new linear lower bound function $(xy)_l$ 
of $xy$ by  averaging  the convex envelope  inequalities,
as illustrated in the lower part of Eq. \ref{conv_bilinear}.
%\begin{gather}
%(xy)_{l} \triangleq \frac{1}{2} \sum_i (xy)_{li}\le xy
%%(xy)_{l} \triangleq
%%%\frac{1}{2}((xy)_{l1} + (xy)_{l2}) \\ 
%%\frac{1}{2}(\underline{x}+\overline{x})y+ \frac{1}{2}(\underline{y}+\overline{y})x-\frac{1}{2}(\underline{x}\underline{y}+\overline{x}\overline{y})
%%\le xy
%\label{bil_single_lb}
%\end{gather}
%$(xy)_l$ has the advantage of being linear.
%It can be easily verified that $(xy)_{lb}$ is a valid lower bound function of $xy$.
%We note that similar idea has been employed in 
%for upper bound linearization.

Chandraker and Kriegman
show that 
for a bilinear monomial,
for its convex envelope to converge to it,
%it is sufficient that 
we only need to branch  over one  variable that makes up the bilinear monomial \cite{bilinear_one_var}.
Similar conclusion can also be drawn about our linear lower bound function. % \eqref{bil_single_lb}.
The proof is presented in \ref{bil_tri_converge}.
% \cite{bilinear_one_var} shows that 
%for bilinear forms,
%it is sufficient that we only  branch  over one set of variables that make up the bilinear forms.

\subsection{Trilinear monomial
	\label{sec:lb_tri}
}
%\parbox{9pt}{tex\\t}

The convex envelope of a trilinear monomial $x_ix_jx_k$ depends on the signs of the bounds on variables  \cite{trilinear_lb0,trilinear_lb}.
Denoting a permutation of $x_i,x_j,x_k$ by the symbols $x,y,z$, 
%(here $\underline{z}\le z\le  \overline{z}$),
where $\underline{x}\le x\le \overline{x}$,
$\underline{y}\le y\le \overline{y}$ and
$\underline{z}\le z\le \overline{z}$.
For the case $\underline{x}\le 0, \underline{y}\ge 0, \underline{z}\le 0, \overline{z}\ge 0$,
we can get the  convex envelope of $xyz$ as shown in the upper part of Eq. \ref{conv_trilinear}.
%\begin{align}
%(xyz)_{l1}\triangleq&\overline{y}\overline{z}x+ \overline{x}\overline{z}y+ \overline{x}\overline{y}z -2\overline{x}\overline{y}\overline{z}\le xyz \\%checked
%(xyz)_{l2}\triangleq&\overline{y}\underline{z}x+ \underline{x}\overline{z}y+ \underline{x}\overline{y}z -\underline{x}\overline{y}\underline{z}-\underline{x}\overline{y}\overline{z}\le xyz \\%checked
%(xyz)_{l3}\triangleq&\overline{y}\underline{z}x+ \underline{x}\underline{z}y+ \underline{x}\underline{y}z -\underline{x}\overline{y}\underline{z}-\underline{x}\underline{y}\underline{z}\le xyz  \\ %checked
%(xyz)_{l4}\triangleq&\underline{y}\overline{z}x+ \overline{x}\underline{z}y+ \overline{x}\underline{y}z -\overline{x}\underline{y}\overline{z}-\overline{x}\underline{y}\underline{z}\le xyz  \\%checked
%(xyz)_{l5}\triangleq&\underline{y}\underline{z}x+ \overline{x}\underline{z}y+ \underline{x}\underline{y}z -\overline{x}\underline{y}\underline{z}-\underline{x}\underline{y}\underline{z}\le xyz  \\%checked
%(xyz)_{l6}\triangleq&\underline{y}\overline{z}x+ \underline{x}\overline{z}y+ \phi/(\overline{z}-\underline{z})z -\phi\underline{z}/(\overline{z}-\underline{z}) \notag\\
%&-\underline{x}\overline{y}\overline{z}-\overline{x}\underline{y}\overline{z}+\overline{x}\overline{y}\underline{z} \le xyz
%\label{tril_lb}
%\end{align}
%where
%$
%\phi =\underline{x}\overline{y}\overline{z}-\overline{x}\overline{y}\underline{z}-\underline{x}\underline{y}\overline{z}+\overline{x}\underline{y}\overline{z} %checked
%$.
%
\begin{equation}
	\begin{CD}
	xyz@>\begin{matrix}\scriptsize\text{convex}\\\scriptsize\text{envelope}\end{matrix}	>
	\begin{matrix}
\scriptstyle	\underline{x}\le x\le \overline{x},\\
\scriptstyle	\underline{y}\le y\le \overline{y},\\
\scriptstyle	\underline{z}\le z\le \overline{z},	\\
\scriptstyle	\underline{x}\le 0, \underline{y}\ge 0,\\ 
\scriptstyle	\underline{z}\le 0, \overline{z}\ge 0
	\end{matrix}
	>
	\begin{cases}
%\normstyle
 (xyz)_{l1}{\triangleq}\overline{y}\overline{z}x+ \overline{x}\overline{z}y+ \overline{x}\overline{y}z -2\overline{x}\overline{y}\overline{z}\le xyz, \\%checked
%\scriptstyle
	(xyz)_{l2}\triangleq\overline{y}\underline{z}x+ \underline{x}\overline{z}y+ \underline{x}\overline{y}z -\underline{x}\overline{y}\underline{z}-\underline{x}\overline{y}\overline{z}\le xyz, \\%checked
%\scriptstyle
	(xyz)_{l3}\triangleq\overline{y}\underline{z}x+ \underline{x}\underline{z}y+ \underline{x}\underline{y}z -\underline{x}\overline{y}\underline{z}-\underline{x}\underline{y}\underline{z}\le xyz,  \\ %checked
%\scriptstyle
	(xyz)_{l4}\triangleq\underline{y}\overline{z}x+ \overline{x}\underline{z}y+ \overline{x}\underline{y}z -\overline{x}\underline{y}\overline{z}-\overline{x}\underline{y}\underline{z}\le xyz,  \\%checked
%\scriptstyle
	(xyz)_{l5}\triangleq\underline{y}\underline{z}x+ \overline{x}\underline{z}y+ \underline{x}\underline{y}z -\overline{x}\underline{y}\underline{z}-\underline{x}\underline{y}\underline{z}\le xyz,  \\%checked
%\scriptstyle
	(xyz)_{l6}\triangleq\underline{y}\overline{z}x+ \underline{x}\overline{z}y + \phi/(\overline{z}-\underline{z})z  -\phi\underline{z}/(\overline{z}-\underline{z}) \\
%\scriptstyle
\qquad \qquad 	-\underline{x}\overline{y}\overline{z}-\overline{x}\underline{y}\overline{z}+\overline{x}\overline{y}\underline{z} \le xyz
	\end{cases} 
\\
@. @V\text{ average}VV \\
@. 	(xyz)_l \triangleq\frac{1}{6}\sum_i (xyz)_{li} \le xyz
	\end{CD}
	\label{conv_trilinear}
\end{equation}

Based on the same consideration as stated in Sec.\ \ref{sec:bil_lb},
we do not directly utilize this form of  convex envelope.
Instead,
we construct a new linear  lower bound function $(xyz)_l$ 
of $xyz$ by averaging  the convex envelope inequalities, 
as shown in the lower part of Eq. \ref{conv_trilinear}.
%\begin{gather}
%(xyz)_l\triangleq\frac{1}{6}\sum_i (xyz)_{li} \le xyz
%\label{tri_lb_avg}
%%(xyz)_l=\text{average of the LHS of \ref{tril_lb}}
%\end{gather}

%resulting in the following form of linear lower bound function:
%\begin{gather}
%l_1 x + l_2 y + l_3 z \le xyz
%\end{gather}
%
For other cases of the bounds on $x,y,z$,
we conduct similar operations to  linearize the corresponding convex envelopes. 

%
% viewing it as the matrix product of $\theta^\top$, $\text{mat} ( \mathbf F\mathbf B_2\mathbf p   
%)$ and $\theta$
%and computing the lower bound functions of corresponding element products.

In a similar vein as in Sec.\ \ref{sec:bil_lb},
for a trilinear monomial,
for its linear lower bound function 
%(as defined in Sec.\ \ref{sec:lb_tri}) 
to converge to it,
we only need to branch over two  variables that make up the trilinear monomial.
The proof is presented in \ref{bil_tri_converge}.

\section{Case one: spatial transformation is linear with respect to  its parameters
	\label{sec:case_one}
}
%\subsection{Objective function
%\label{subsec:obj_aff}
%}
%In this section,
%we mainly follow the work of \cite{RPM_model_occlude} to derive our objective function.
%In  \cite{RPM_model_occlude},
% the goal  is to derive a  lower bound function
%and only  the nonlinear part of the RPM objective function is converted into a function of a low dimensional variable.
%In contrast,
%our goal is to apply the inner approximation algorithm in a low dimensional space,
%and thus, the entire RPM objective function is
%converted.
% into a function of a low dimensional variable.
%

In this section, first, 
we    derive  our objective function for the case that the spatial transformation is linear with respect to  its parameters.
%In the following,
Then, we  reformulate the objective function due to a requirement when using  the convex envelope of a trilinear monomial \cite{trilinear_lb} (Sec.\ \ref{subsec:obj_reform_case_one}).
Next, we  discuss the  branching strategy (Sec.\ \ref{subsec:branch_case_one})
and use the lower bound functions of bilinear and trilinear  terms 
to derive the lower bound function of our objective function (Sec.\ \ref{subsec:lb_case_one}).
Finally, we  discuss  computation of the upper bound (Sec.\ \ref{subsec:ub_case_one})
and  present the  pseudo code of our BnB algorithm (Sec.\ \ref{sec:BnB_aff}).
%together with  its convergence property (Sec.\ \ref{sec:converge_2d}).

Suppose there are two point sets 
$\mathscr{X}=\{ \mathbf x_i,1\le i\le n_x\}$
%is the model point set
and
$\mathscr{Y}=\{ \mathbf y_j, 1\le j \le n_y\}$ 
in space $\mathbb R^{n_d}$,
where the column vectors $\mathbf x_i$ and $\mathbf y_j$ denote coordinates of points $i$ and $j$, repectively.
%
%=\begin{bmatrix}
%x_i^1,\ldots,x_i^{n_d}
%\end{bmatrix}^\top$
%and =\begin{bmatrix}
%y_j^1,\ldots,y_j^{n_d}
%\end{bmatrix}^\top$.
%
Following \cite{RPM_model_occlude},
we model registration  of these two point sets
as a mixed linear assignment$-$least square problem:
\begin{gather}
\min E(\mathbf P,\boldsymbol{\theta})=\sum_{i,j}p_{ij}\|\mathbf y_j-\mathbf J(\mathbf x_i) \boldsymbol{\theta}\|^2 \label{energy_case_one_generic}
%= \mathbf 1_{n_x}^\top \mathbf P \widetilde{\mathbf y} \notag \\
%+{\boldsymbol\theta}^\top \mathbf J^\top  (\text{diag}(\mathbf P  \mathbf 1_{n_y}) \otimes \mathbf I_d) \mathbf J  {\boldsymbol\theta}   
%-2{\boldsymbol\theta}^\top \mathbf J^\top (\mathbf P\otimes \mathbf I_{d}) \mathbf y 
\\
s.t.\ \mathbf P  \mathbf 1_{n_y}\le  \mathbf 1_{n_x},\
\mathbf 1_{n_x}^\top \mathbf P\le  \mathbf 1_{n_y}^\top,\
\mathbf 1_{n_x}^\top \mathbf P  \mathbf 1_{n_y}=n_p,\  %\notag\\
\mathbf P\ge 0, \
\underline{\boldsymbol{\theta}}_0\le \boldsymbol{\theta}\le \overline{\boldsymbol{\theta}}_0
% \}
\label{k_card_P_const}
\end{gather}
where\\
\begin{tabular}{rl}
$\boldsymbol\theta$ :& parameters of the spatial transformation 
$J(\textbf x_i)\boldsymbol\theta$.\\
$J(\textbf x_i)$  :&the Jacobian matrix of $\mathbf x_i$.\\
$P$ :& correspondence matrix,
with  element $p_{ij}{=}1$ \\
& or $0$ indicating that  $\mathbf x_i$ matches $\mathbf y_j$  or not.\\
$\mathbf 1_{n_x}^\top \mathbf P  \mathbf 1_{n_y}{=}n_p$:&  the cardinality of matches 
is required to be \\ & equal to  a constant integer $n_p$.\\
$\underline{\boldsymbol{\theta}}_0$ (resp. 
$\overline{\boldsymbol{\theta}}_0$) :& 
 the lower (resp.  upper) bound of $\boldsymbol{\theta}$.
\end{tabular}

Based on Eq. \cite{RPM_model_occlude},
$E$ can be written in a concise form as:
\begin{gather}
E(\mathbf p,\boldsymbol \theta)=\boldsymbol\theta^\top\text{mat} (\mathbf B\mathbf p) \boldsymbol{\theta}-
2\boldsymbol\theta^\top \mathbf A\mathbf p  + 
\boldsymbol \rho^\top \mathbf p  
\end{gather}
where\\
\begin{tabular}{r l}
 $\mathbf p\triangleq \text{vec}(\mathbf P)$:& a vector. \\
 $\boldsymbol\rho\triangleq\mathbf 1_{n_x}  \otimes\widetilde{\mathbf y}$:& a vector. \\
$\mathbf A\triangleq(\mathbf{J}^\top \otimes\mathbf y^\top )\mathbf W^{n_x,n_y}_{n_d}$:&  a matrix. \\
$\mathbf B\triangleq(\mathbf{J}_2^\top \otimes \mathbf I_{n_\theta}) \mathbf W^{n_x,1}_{n_\theta} (\mathbf I_{n_x} \otimes\mathbf 1_{n_y}^\top )$:&  a matrix.\\
$\mathbf J\triangleq\begin{bmatrix}
\mathbf J^\top(\mathbf x_1), \dots, \mathbf J^\top(\mathbf x_{n_x})
\end{bmatrix}^\top$:&  a matrix.\\
$
\mathbf{J}_2\triangleq\begin{bmatrix}
\mathbf{J}(\mathbf x_1)^\top \mathbf{J}(\mathbf x_1),       \ldots,      \mathbf{J}(\mathbf x_{n_x})^\top \mathbf{J}(\mathbf x_{n_x})
\end{bmatrix}^\top
$: &a matrix. \\
%The $mnd^2\times mn$ constant matrix
$
\mathbf W^{m,n}_d\triangleq
\mathbf I_m \otimes  \begin{bmatrix}
\mathbf I_n \otimes (\mathbf e_d^1)^\top,
\dots,
\mathbf I_n \otimes (\mathbf e_d^d)^\top
\end{bmatrix}^\top
$:& a matrix. \\
$\mathbf y\triangleq\begin{bmatrix}
\mathbf y_1^\top, \dots, \mathbf y_{n_y}^\top
\end{bmatrix}^\top$: & a vector. \\
$\widetilde{\mathbf {y}}\triangleq \begin{bmatrix}
\|\mathbf y_1\|_2^2, \dots, \|\mathbf y_{n_y}\|_2^2
\end{bmatrix}^\top$: & a vector.
\end{tabular}
\begin{tabular}{r l}
$\text{vec}(\cdot)$: & 	 vectorization  of a matrix by concatenating  rows.\\
$n_\theta$: &  the dimensionality of ${\boldsymbol{\theta}}$.\\
$\mathbf e_d^i$: & the  $d$-dimensional column vector 
with only \\  & one nonzero unit  element at the $i$-th position.\\
% being $1$ and all other entries being $0$.
%$\mathbf W^{m,n}_d$ & a large but sparse matrix and can be implemented  using the function $speye$ in Matlab.
%
%
$\mathbf 1_{n_x}$: & the $n_x$-dimensional  vector of  all ones.\\
$\mathbf I_{d}$: & the $d\times d$ identity matrix. \\
%$I_d$ & the $d$-dimensional identity matrix, 
$\otimes$: & the Kronecker product. \\ 
%$\text{diag}(\cdot)$: & converts a vector into a diagonal matrix.\\
%
$\text{mat}(\cdot)$: &
reconstructs a  matrix 
from a vector,
which is \\
&  the inverse of the operator $\text{vec}(\cdot)$.
\end{tabular}

% which is the  result of applying $\text{vec}(\cdot)$ to a  matrix.
%Therefore, $\text{mat}(\cdot)$ can be viewed as 

Since $\mathbf  1^\top_{n_xn_y} \mathbf  p=n_p$, a constant,
for rows of $\mathbf B$
containing identical elements,
%equal to scaled versions  of $\mathbf  1^\top_{n_xn_y}$ 
the result of them multiplied by $\mathbf p$  can be replaced by  constants.
Also, redundant rows can  be removed.
Since $\text{mat}(\mathbf B\mathbf p)$ is a symmetric matrix,
$\mathbf B$   will  contain redundant rows.
Based on these  analyses,
we hereby  denote $\mathbf B_2$ 
as the matrix formed as a result of $\mathbf B$ 
removing such  rows.
%It can be verified that for
%2D similarity/affine transformations and 3D scaling + translation transformation
(Please refer to Sec.  \ref{subsec:exp_case_one} for examples of $\mathbf B_2$).
% for different types of transformations. 
%$\mathbf \Xi$ and $\mathbf D$ contains such rows.
%
Consequently,
$E$ can be  rewritten as
\begin{gather}
E(\mathbf p,\boldsymbol \theta)=\boldsymbol\theta^\top \left[\text{mat} ( \mathbf K\mathbf B_2\mathbf p   
){+} \mathbf C\right] \boldsymbol{\theta}  
{-}
2\boldsymbol\theta^\top \mathbf A\mathbf p  {+} 
\boldsymbol \rho^\top \mathbf p  
\label{energy_case_one_final}
\end{gather}             
where the nonzero elements of the constant matrix $\mathbf C$ correspond to the  rows of $\mathbf B$ containing identical elements.
The elements of the constant matrix $\mathbf K$ 
%satisfies $B=F\mathbf B_2$,
take  values  in $\{0,1\}$ and   record the correspondences  between the rows of $\mathbf B_2 $ and those of $\mathbf B$.

\subsection{Objective function
	reformulation 
	\label{subsec:obj_reform_case_one}
}

%Ref. \cite{trilinear_lb}
%%Clifford and Christodoulos
%only presents  
The convex envelope of a trilinear monomial presented in  \cite{trilinear_lb} only holds
under the condition that at least one variable
that makes up the monomial
has a range containing the origin,
whereas for the trilinear term 
$\boldsymbol\theta^\top\text{mat} ( \mathbf K\mathbf B_2\mathbf p   
) \boldsymbol{\theta}$
in our objective function, 
such a condition can be violated.
To address this issue,
since our branching strategy 
(to be presented in 
%(as will be shown in
Sec.\ \ref{subsec:branch_case_one})
is that the range of $\theta_i$ changes during the BnB iterations,
whereas   the range of 
$[\text{mat} ( \mathbf K\mathbf B_2\mathbf p   
)]_{ij}$ remains fixed,
therefore, 
we seek to replace $\text{mat} ( \mathbf K\mathbf B_2\mathbf p   
)$ by a proper matrix such that  this condition can be satisfied.
%we reformulate the objective function so that this condition can be satisfied.
The procedure is as follows:

We build a constant symmetric matrix $\mathbf D$ 
such that 
%$0\in
%\{ [\text{mat} ( \mathbf K\mathbf B_2\mathbf p   
%)-\mathbf D]_{ij}| \mathbf p\in \Omega\}$,
all the elements of
%the range of 
$\text{mat} (\mathbf K\mathbf  B_2\mathbf p )-\mathbf D$
have ranges 
containing the origin.
This can be accomplished  as follows:
We first choose 
$\mathbf p_0\in \Omega$,
where $\Omega$ represents the feasible region of $\mathbf p$, as is defined by \eqref{k_card_P_const}.
%and 
%choose $\mathbf D$ as
%$\mathbf D=\text{mat}(F\mathbf B_2 \mathbf p_0)$.
(In this paper, $\mathbf p_0$ is chosen as $\frac{n_p}{n_x n_y}\mathbf 1_{n_x n_y} $,
corresponding to the case that point correspondence is the fuzziest.)
We then  let $\mathbf D=\text{mat}(\mathbf K \mathbf B_2 \mathbf p_0)$.
We now have 
%the following result:
\begin{proposition}
	%$0\in [\text{mat} (\mathbf K\mathbf  B_2\mathbf p )-\mathbf D]_{ij}$.
	All the  elements of 
	$\text{mat} (\mathbf K\mathbf  B_2\mathbf p )-\mathbf D$
	have ranges containing the origin.	
\end{proposition} 
%\vspace{-2mm}
{\proof
	Since $\mathbf p_0\in \Omega$,
	we have  $\mathbf D\in
	\{\text{mat} ( \mathbf K\mathbf B_2\mathbf p   
	)| \mathbf p\in \Omega\}$.
	% of $\text{mat} ( \mathbf K\mathbf B_2\mathbf p   
	%)$.
	Therefore, 
	$0\in
	\{ [\text{mat} ( \mathbf K\mathbf B_2\mathbf p   
	)-\mathbf D]_{ij}| \mathbf p\in \Omega\}$.
	%$0\in [\text{mat} (\mathbf K\mathbf  B_2\mathbf p )-\mathbf D]_{ij}$.
	%all the  elements of 
	%$\text{mat} (\mathbf K\mathbf  B_2\mathbf p )-\mathbf D$
	%have ranges containing the origin.
}
%\vspace{-2mm}

Our experiments indicate that different choices of $\mathbf p_0$ (and thus $\mathbf D$) lead to almost the same alignment results.
Based on the above analysis,
%the energy function
$E$ is reformulated as:
\begin{gather}
E(\mathbf p,\boldsymbol \theta)=\boldsymbol\theta^\top [\text{mat} (\mathbf K\mathbf  B_2\mathbf p )-\mathbf D] \boldsymbol{\theta}+
\boldsymbol\theta^\top (\mathbf C+\mathbf D)
\boldsymbol{\theta} %\notag\\
-
2\boldsymbol\theta^\top \mathbf A\mathbf p  + 
\boldsymbol \rho^\top \mathbf p  
\label{eng_vec}
\end{gather}

\subsection{Branching strategy
	\label{subsec:branch_case_one}
}

Based on the result in Sec.\ \ref{sec:bil_lb},
for the bilinear term $-2\boldsymbol{\theta}^\top \mathbf A \mathbf p$,
%appeared  in our objective function,
for  its linear lower bound function to converge to it,
we only need to branch the range of $\boldsymbol{\theta}$
and leave the range of $-2\mathbf A \mathbf p$ fixed.
Here  the range of $-2\mathbf A \mathbf p$ can be computed as 
%$\min_{\mathbf p\in \Omega} (-2\mathbf A \mathbf p)_i \le (-2\mathbf A \mathbf p)_i \le \max_{\mathbf p\in \Omega}  (-2\mathbf A \mathbf p)_i$
%via solving  linear assignment problems.
\begin{equation}
	\min_{\mathbf p\in \Omega} (-2\mathbf A \mathbf p)_i \le (-2\mathbf A \mathbf p)_i \le \max_{\mathbf p\in \Omega}  (-2\mathbf A \mathbf p)_i
\end{equation}

%\begin{figure} [H]
%	\begin{gather}
%		%\mbox{\tiny $
%			%	\begin{matrix}
%				\underline{\theta}_{i}\le \theta_i\le \overline{\theta}_{i} ,\quad
%					\notag\\	
%				 \notag
%				%	\end{matrix}
%			%	$}
%	\end{gather}
%	\caption{Computation of the ranges of various variables,
%		where the subproblems are linear assignment problems which can be efficiently solved.}
%	\label{range_determ_aff}
%\end{figure} 

%The advantage of this strategy is that: usually,
%one set of variables in
%a bilinear model has far fewer elements than the other variables.
%If we only branch over this set of variables,
%the method will converge quickly.

In a similar vein,
based on the result in Sec.\ \ref{sec:lb_tri},
for the  trilinear term
$\boldsymbol\theta^\top [\text{mat} (\mathbf K\mathbf  B_2\mathbf p )-\mathbf D] \boldsymbol{\theta}$,
for its linear lower bound function to converge to it,
we only need to branch the range of 
$\boldsymbol{\theta}$
and leave the range of $\text{mat} (\mathbf K\mathbf  B_2\mathbf p )-\mathbf D  $ fixed
since %for our problem,
$\boldsymbol{\theta}$ constitutes two sets of variables that make up 
$\boldsymbol\theta^\top [\text{mat} (\mathbf K\mathbf  B_2\mathbf p )-\mathbf D] \boldsymbol{\theta}$.
%thus,
%we only need to branch the ranges of  
%
Here  the  range of $\text{mat} (\mathbf K\mathbf  B_2\mathbf p )-\mathbf D $ can be computed as 
%$\min_{\mathbf p\in \Omega} [\text{mat} ( \mathbf K\mathbf B_2\mathbf p   
%)-\mathbf D]_{ij} \le 
%[\text{mat} ( \mathbf K\mathbf B_2\mathbf p   
%)-\mathbf D]_{ij}
%\le
%\max_{\mathbf p\in \Omega}  [\text{mat} ( \mathbf K\mathbf B_2\mathbf p   
%)-\mathbf D]_{ij}$
%via solving linear assignment problems.
%We recognize these as linear assignment problems which can be efficiently solved.
%
%
\begin{equation}
	\min_{\mathbf p\in \Omega} [\text{mat} ( \mathbf K\mathbf B_2\mathbf p   
	)-\mathbf D]_{ij} {\le} 
	[\text{mat} ( \mathbf K\mathbf B_2\mathbf p   
	)-\mathbf D]_{ij}
	{\le}
	\max_{\mathbf p\in \Omega}  [\text{mat} ( \mathbf K\mathbf B_2\mathbf p   
	)-\mathbf D]_{ij} 
\end{equation}

In conclusion, in our BnB algorithm,
we only need to branch over $\boldsymbol{\theta}$.
%Therefore, the dimension of branching space of our BnB algorithm is equal to that of $\boldsymbol{\theta}$, which is low.
%Consequently, our algorithm can  converge quickly.
Therefore, the dimension of the branching space of our BnB algorithm is low and the proposed algorithm can converge quickly.

\subsection{\label{subsec:lb_case_one}Lower bound %of the objective function
}

Based on the results in Sec.\ \ref{sec:cov_env},
we can obtain the lower bound function $E_l$ of $E$ by respectively deriving the lower bound functions of the bilinear and trilinear terms,
as illustrated in Eq.  \ref{lb_derive}.
%where the constant vectors $\mathbf g^k$ 
%are computed based on the ranges %$[\underline{\theta}_{i}, \overline{\theta}_{i}]$
%%(\textcolor{red}{please refer to Sec. \ref{subsec:range_case_one}})
%of $\theta_{i}$ and  $(-2 \mathbf A \mathbf p )_{i}$
%and the constant matrices $\mathbf H^k$ are computed  based on the ranges %$[\underline{\theta}_{i}, \overline{\theta}_{i}]$
%% (\textcolor{red}{please refer to Sec. \ref{subsec:range_case_one}})
%of $\theta_{i}$ and  $[\text{mat} ( \mathbf K\mathbf B_2\mathbf p   
%)-\mathbf D]_{ij}$.

\begin{figure*}[!ht]
	\begin{equation} % [H]
		%	\begin{tabular}{@{\hspace{-0mm}}c}	
			%		$
			\begin{CD}
				E(\mathbf p,\boldsymbol \theta)=	\boldsymbol\theta^\top (\mathbf C+\mathbf D)
				\boldsymbol{\theta}
				+ 
				\boldsymbol \rho^\top \mathbf p	
				@. +
\textcolor{red}{				\boxed{\boldsymbol\theta^\top [\text{mat} (\mathbf K\mathbf  B_2\mathbf p )-\mathbf D] \boldsymbol{\theta}
				} }
				@. +
\textcolor{blue}{				\boxed{-2\boldsymbol\theta^\top \mathbf A\mathbf p }}
				\\
				@. @V\text{trilinear}V \text{relaxation }V  @V\text{bilinear}V\text{relaxation}V 
				\\
				E_l(\mathbf p,\boldsymbol{\theta})=
				\boldsymbol\theta^\top (\mathbf C+\mathbf D)
				\boldsymbol{\theta} 	
				+ \boldsymbol \rho^\top \mathbf p
				+@. 
\textcolor{red}{
				\boxed{
					\begin{matrix}		
						\boldsymbol{\theta}^\top \mathbf H^0 \mathbf 1_{n_\theta}   +
						\text{tr}([\text{mat} ( \mathbf K\mathbf B_2\mathbf p   
						)-\mathbf D]^\top \mathbf H^1 ) 
						+\mathbf  1_{n_\theta}^\top \mathbf H^2 \mathbf 1_{n_\theta}
					\end{matrix}
				}}
				@.
				+
\textcolor{blue}{				\boxed{
					\begin{matrix}	
						\boldsymbol{\theta}^\top  \mathbf g^0 
						+ (\mathbf g^1)^\top (-2 \mathbf A \mathbf p  ) 
						+ \mathbf 1_{n_\theta}^\top \mathbf g^2 
					\end{matrix}		
				}}	
			\end{CD}
			%		$
			%	\end{tabular}	
		%	\caption{Derivation of the lower bound function,
			%		where the constant vectors $\mathbf g^k$ 
			%		are computed based on the ranges %$[\underline{\theta}_{i}, \overline{\theta}_{i}]$
			%		%(\textcolor{red}{please refer to Sec. \ref{subsec:range_case_one}})
			%		of $\theta_{i}$ and  $(-2 \mathbf A \mathbf p )_{i}$ and
			%		the constant matrices $\mathbf H^k$ are computed  based on the ranges %$[\underline{\theta}_{i}, \overline{\theta}_{i}]$
			%		% (\textcolor{red}{please refer to Sec. \ref{subsec:range_case_one}})
			%		of $\theta_{i}$ and  $[\text{mat} ( \mathbf K\mathbf B_2\mathbf p   
			%		)-\mathbf D]_{ij}$.
			%	}
		\label{lb_derive}
	\end{equation} 
\end{figure*}

Consequently $E_l$ can be written into the following concise form:
%Combining the results in Sec.\  \ref{subsec:lb_bilinear_case_one} and \ref{subsec:lb_trilinear_case_one},
%we  get  lower bound problem of the original problem \eqref{energy_Case_one}, \eqref{k_card_P_const} as
\begin{gather}
\min E_l(\mathbf p,\boldsymbol{\theta})\triangleq
\boldsymbol\theta^\top (\mathbf C+\mathbf D)
\boldsymbol{\theta} {+}\mathbf m_0^\top \boldsymbol{\theta} {+} \mathbf m_1^\top\mathbf p {+}  m_2
%\\
%s.t. \quad \underline{\boldsymbol{\theta}}\le \boldsymbol{\theta} \le \overline{\boldsymbol{\theta}},
%\
%\mathbf p  \in \Omega
\end{gather}  
where $\mathbf m_0$ and $\mathbf m_1$  are  constant vectors and  $m_2$  is a constant scalar.
%$
%\mathbf m_0\triangleq \mathbf H^0 \mathbf 1_{n_\theta} + \mathbf g^0
%$ and
%$
%\mathbf m_1\triangleq
%\mathbf B_2^\top \mathbf K^\top \text{vec}(\mathbf H^1) - 2\mathbf A^\top \mathbf g^1 +\boldsymbol{\rho}
%$.
%The constant scalar 
%$
%m_2\triangleq \mathbf 1_{n_\theta}^\top \mathbf g^2 -\text{tr}(\mathbf D^\top \mathbf H^1)
%+\mathbf  1_{n_\theta}^\top \mathbf H^2 \mathbf 1_{n_\theta}$.

It is obvious that the
minimization of $E_l$ under the constraints
$\underline{\boldsymbol{\theta}}\le \boldsymbol{\theta} \le \overline{\boldsymbol{\theta}},
\
\mathbf p  \in \Omega$
can be decomposed into  separate optimizations over $\mathbf p$ and $\boldsymbol{\theta}$.
%the optimizations of $\mathbf p$ and $\boldsymbol{\theta}$ are separated.
$\mathbf p$ is recovered by solving the following linear assignment problem:
\begin{equation}
\min_{\mathbf p\in \Omega }  \mathbf m_1^\top \mathbf p 
\end{equation}
$\boldsymbol{\theta}$ is recovered by solving the following low dimensional quadratic optimization problem:
\begin{gather}
\min \ \boldsymbol\theta^\top (\mathbf C+\mathbf D)
\boldsymbol{\theta} + \mathbf m_0^\top \boldsymbol{\theta}\\
s.t. \quad \underline{\boldsymbol{\theta}}\le \boldsymbol{\theta} \le \overline{\boldsymbol{\theta}}
\end{gather}
Solvers such as 
the matlab function $quadprog$ can be employed for this problem.
%We use  an off-the-self solver, 
%the matlab function $quadprog$, to solve this problem.
The following result indicates that this is a convex  problem.

\begin{proposition}
	% if  the  point set $\mathscr{X}$ is centered  at the origin,
	We have
	$\mathbf C+\mathbf D\succeq 0$. 
	%is  positive definite.	
\end{proposition}
%\vspace{-2mm}
{\proof
	For any $\mathbf p\in \Omega$,
	we have	
	$
	\text{mat}(\mathbf B\mathbf p)=
	\mathbf{J}_2^\top  ((\mathbf P  \mathbf 1_{n_y}) \otimes \mathbf I_{n_\theta}) =\sum_i \mathbf J_i^\top \mathbf J_i (\mathbf P\mathbf 1_{n_y})_i
	\succeq 0$.
	Since 
	$\mathbf p_0 \in \Omega$,
	thus we have
	$\text{mat} (\mathbf B\mathbf p_0) =\text{mat}(\mathbf K\mathbf B_2 \mathbf p_0)+\mathbf C=\mathbf D+\mathbf C\succeq 0$.
}

\subsection{\label{subsec:ub_case_one}Upper bound} 
%	of the objective function}
%High quality upper bound computation benefits the convergence of our algorithm.
By plugging the $\mathbf p,\boldsymbol{\theta}$ solutions yielded during the computation of the lower bound
into    $E(\mathbf p,\boldsymbol{\theta})$,
we can get an upper bound.
Nevertheless,
the $\boldsymbol{\theta}$ solution may deviate significantly from the optimal $\boldsymbol{\theta}$ especially during the early iterations of the BnB algorithm,
causing the computed upper bound to be poor.
To address this issue,
based on a result of \cite{RPM_model_occlude},
the objective function $E(\mathbf p,\boldsymbol \theta)$ can be rewritten as a function of $\mathbf p$ by eliminating $\boldsymbol \theta$
(please also refer to   \ref{append_E_wrt_p} for detailed derivation):
\begin{gather}
E(\mathbf p)= -\mathbf p^\top \mathbf A^\top [\text{mat} (\mathbf K \mathbf B_2 \mathbf p)+\mathbf C]^{-1} \mathbf A\mathbf p + \boldsymbol{\rho}^\top \mathbf p
\label{E_p_for_upper_bound}
\end{gather}
Therefore, 
we can plug the computed $\mathbf p$ into this function to get a value of $E$, 
which is used as an upper bound.
%instead,
%we  use the equivalent $E(\mathbf p)$ function \eqref{E_p} to compute the upper bound
%where no $\boldsymbol{\theta}$ solution is needed.

\subsection{Branch-and-Bound \label{sec:BnB_aff}}
Based on the aforementioned  preparations,
we are now ready to employ the BnB algorithm
%a globally optimal algorithm for  non-convex problems, 
to optimize $E$. %(\mathbf p,\boldsymbol{\theta})$.
%During initialization,
%we use the inital range of $\boldsymbol{\theta}$
%to  construct the initial hypercube.
%%
%Then, in each iteration of the algorithm, the hypercube yielding the lowest
%lower bound among all the hypercubes is further subdivided so as to improve
%the global lower bound of the problem. Meanwhile, the upper bound is updated
%by evaluating $E(\mathbf p)$ with $\mathbf p$ solutions generated during the computation of the lower bounds. 
The algorithm  is summarized in 
Algo. \ref{tri_BnB_algo},
where
the initial hypercube $M$ is chosen as
the intial range of $\boldsymbol{\theta}$.
%$[\underline{\boldsymbol{\theta}}_0, \overline{\boldsymbol{\theta}}_0]$.
%

%for the  case that transformation is linear w.r.t. its parameters
%or
%$\{ (\mathbf r,\mathbf t)|\mathbf r \in[
%\underline{\mathbf r}_0, \overline{\mathbf r}_0],\
%\mathbf t \in [\underline{\mathbf t}_0,   \overline{\mathbf t}_0]\}$
%for the case that  transformation is 3D rigid.
%
%Here note that for the  case 
%%of Sec. \ref{sec:case_one},
%that transformation is linear w.r.t. its parameters,
%the vector form  of the point correspondence variable is more convenient. % than the matrix form.
%Thus,
%the notations $\mathbf P$, $\mathbf P^k$ and $\mathbf P^{k-1}$ should to be  replaced by   $\mathbf p$, $\mathbf p^k$ and $\mathbf p^{k-1}$, respectively.
%%Accordingly, 
%%the function 
%%$E(\mathbf P^k)$ is replaced by the function $E(\mathbf p^k)$.
%
Since the optimal $\boldsymbol\theta$ solution %transformation parameters
%$\boldsymbol\theta$ 
depends on the relative size and  position of two point sets,
in this work, 
we rescale  point sets $\mathscr{X}$ and $\mathscr Y$ 
to be unit sized 
%is rescaled accordingly)
and translate 
them 
%$\mathscr{X}$ and $\mathscr Y$
to be centered at the origin
before performing registration.
%(this  is consistent with the requirement in Sec.\ \ref{sec:PT_normalize}).
%In this paper, 
Unless otherwise stated,
we set the initial range of $\boldsymbol{\theta}$ as  $[-3 \mathbf 1_{n_\theta},3 \mathbf 1_{n_\theta}]$.
%$\underline{\mathbf r}_0= -\pi \mathbf 1_3$,
%$\overline{\mathbf r}_0=\pi \mathbf 1_3$,
%$\underline{\mathbf t}_0=-3 \mathbf 1_3$,
%$\overline{\mathbf t}_0=3 \mathbf 1_3$.
%$\{ (\mathbf r,\mathbf t)|
%-\pi \mathbf 1\le \mathbf r \le \pi \mathbf 1,
%-3 \mathbf 1\le \mathbf{t} \le 3 \mathbf 1
%\}$
%
%The tolerance error $\epsilon$ should be set taking into account the cardinalities of two  point sets.
%We  empirically  set it as
%$\epsilon=\min(n_x,n_y)\epsilon_0$,
%where $\epsilon_0$ is a predefined positive scalar. 

\begin{algorithm}
	\caption{A BnB algorithm  for minimizing $E$ \label{tri_BnB_algo}}
	\textbf{Initialization}

	Set tolerance error $\epsilon>0$	and
	initial  hypercube $M$.	
	Let $\mathscr M_1=\mathscr N_1=\{M\}$	
	where $\mathscr M_k$ and $\mathscr N_k$ denote the collection of all hypercubes and the collection of active hypercubes at iteration $k$, respectively.
	
	\For{$k=1,2,\ldots$}{
		
		For each hypercube $M\in \mathscr N_k$,
		minimize the lower bound function $E_l$ 
		%		according to Sec.\ \ref{sec:lb_compute}
		to obtain the optimal point correspondence solution $\mathbf p(M)$ and the optimal value $\beta(M)$.
		$\beta(M)$ is a lower bound for  $M$.
		
		%		Use $\mathbf P(M)$ to compute the value of $E$ according to Sec.\ \ref{sec:comput_ub}
		%		which is  an upper bound for region $M$.

		Let $\mathbf p^k$ be the best among all feasible solutions so far encountered: $\mathbf p^{k-1}$ and all $\mathbf p(M)$ for $M\in\mathscr N_k$.
		Delete all $M\in\mathscr M_k$ such that $\beta(M)\ge E(\mathbf p^k)-\epsilon$.
		Let $\mathscr R_k$ be the remaining collection of hypercubes.
		
		If $\mathscr R_k=\emptyset$,
		terminate: $\mathbf p^k $ is the global $\epsilon-$minimum solution.
		
		Select the hypercube $M$ yielding the lowest lower bound and divide it into two sub-hypercubes $M_1$, $M_2$
		by bisecting  the longest edge.
		
		Let $\mathscr N_{k+1}=\{M_1,M_2\}$
		and $\mathscr M_{k+1}=(\mathscr R_k\backslash M) \cup \mathscr N_{k+1}$.

	}

\end{algorithm}

%\subsection{Computation of the ranges
%\label{subsec:range_case_one}
%}
%The above analysis
%indicates that the ranges of $(-2\mathbf A \mathbf p)_i$ 
%and $[\text{mat} ( \mathbf K\mathbf B_2\mathbf p   
%)-\mathbf D]_{ij}$ 
% need to be computed only once.
%Since  $(-2\mathbf A \mathbf p)_i$
%%Here the range of $\boldsymbol{\theta}_i$ is predefined and
% is a linear function of $\mathbf p$,
%its range can be efficiently computed by solving the following    linear assignment problems:
%$\min_{\mathbf p\in \Omega} (-2\mathbf A \mathbf p)_i $ and
%$\max_{\mathbf p\in \Omega}  (-2\mathbf A \mathbf p)_i$.
%%
%Likewise, the ranges of 
%$[\text{mat} ( \mathbf K\mathbf B_2\mathbf p   
%)-\mathbf D]_{ij}$ 
%can be  obtained by solving the corresponding  linear assignment problems.

%products of $\theta_i$, 
%$\text{mat} ( \mathbf F\mathbf B_2\mathbf p   
%)_{ij}$ and $\theta_j$
%and then summing them up.
%Here the range of $\text{mat} ( \mathbf F\mathbf B_2\mathbf p   
%)_{ij}$ can be computed by solving the linear assignment problems:
%$\min \{\text{mat} ( \mathbf F\mathbf B_2\mathbf p   
%)_{ij} | \mathbf p\in \Omega\}$ and
%$\max \{\text{mat} ( \mathbf F\mathbf B_2\mathbf p   
%)_{ij} | \mathbf p\in \Omega\}$.

\subsection{Convergence of the BnB algorithm
	\label{sec:converge_2d}
	%	Tightness of  lower bounds
}

\begin{figure} [t]
	\centering
	\newcommand{\scale}{0.5}

	\begin{tabular}{@{}c@{} c@{} c@{} c}	
		\includegraphics[width=\scale\linewidth]{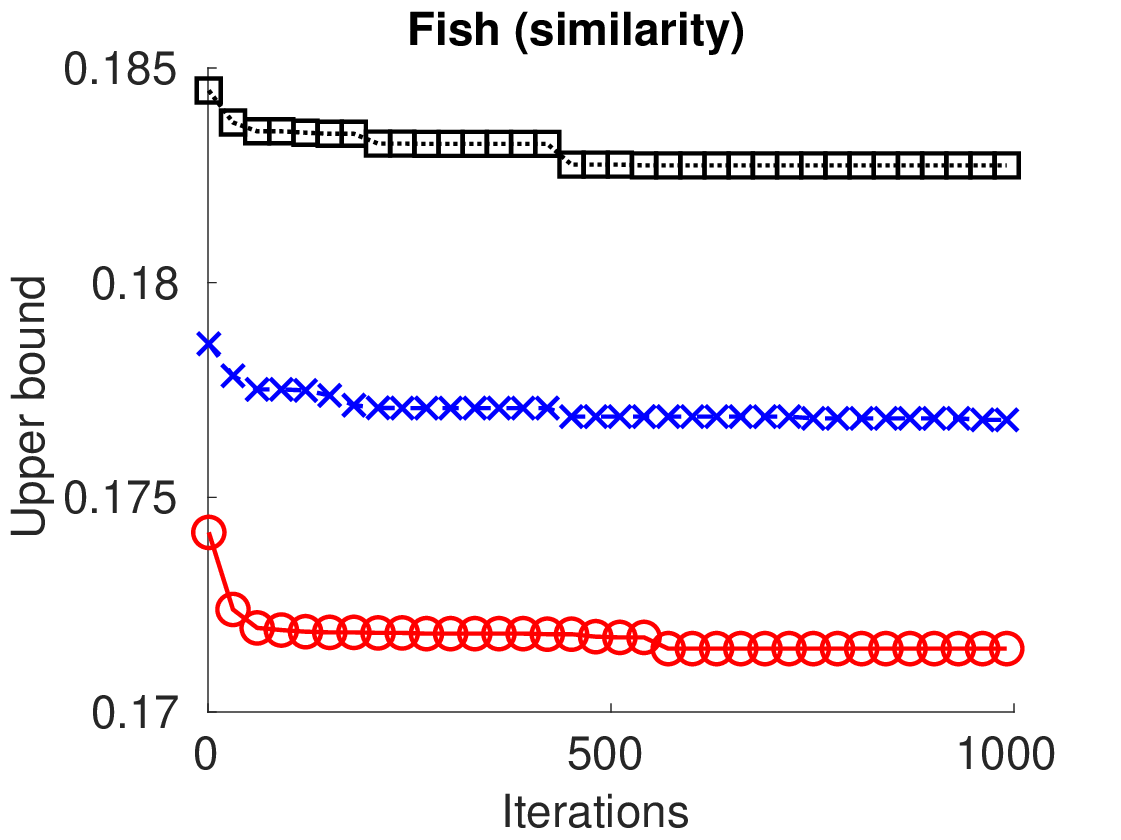}&	
		\includegraphics[width=\scale\linewidth]{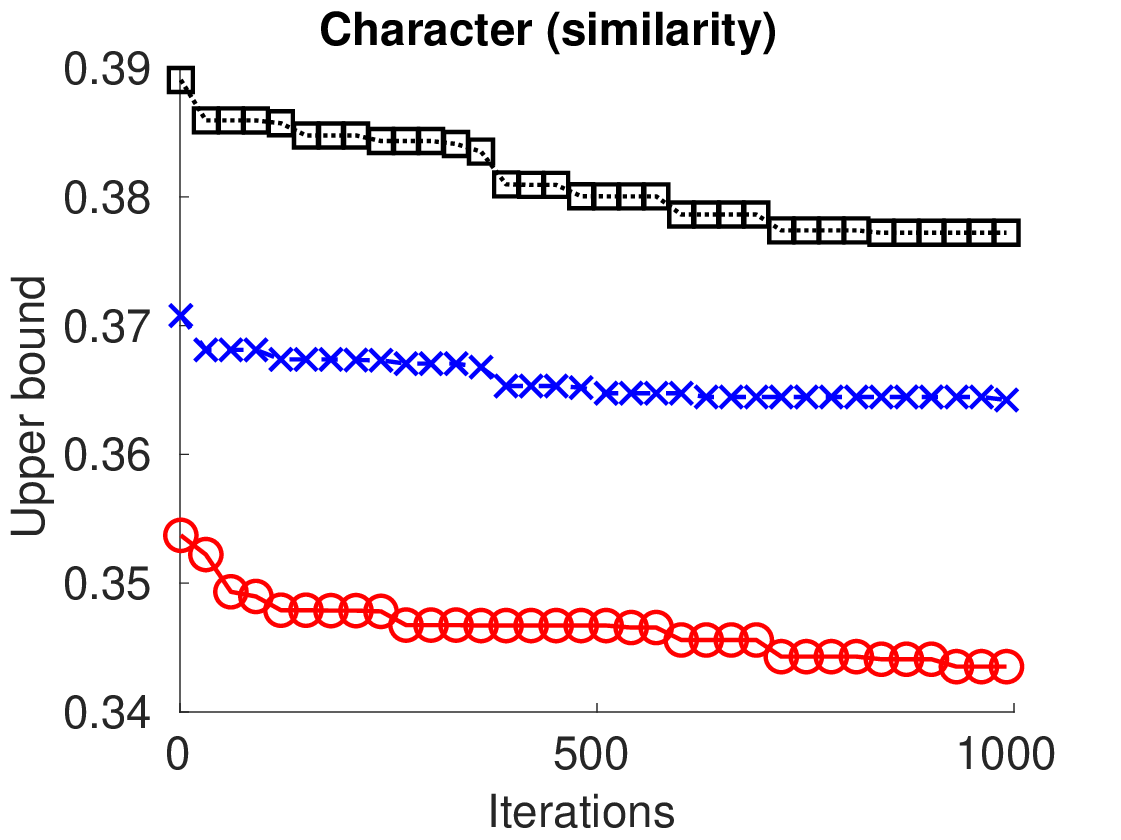}\\		
		\includegraphics[width=\scale\linewidth]{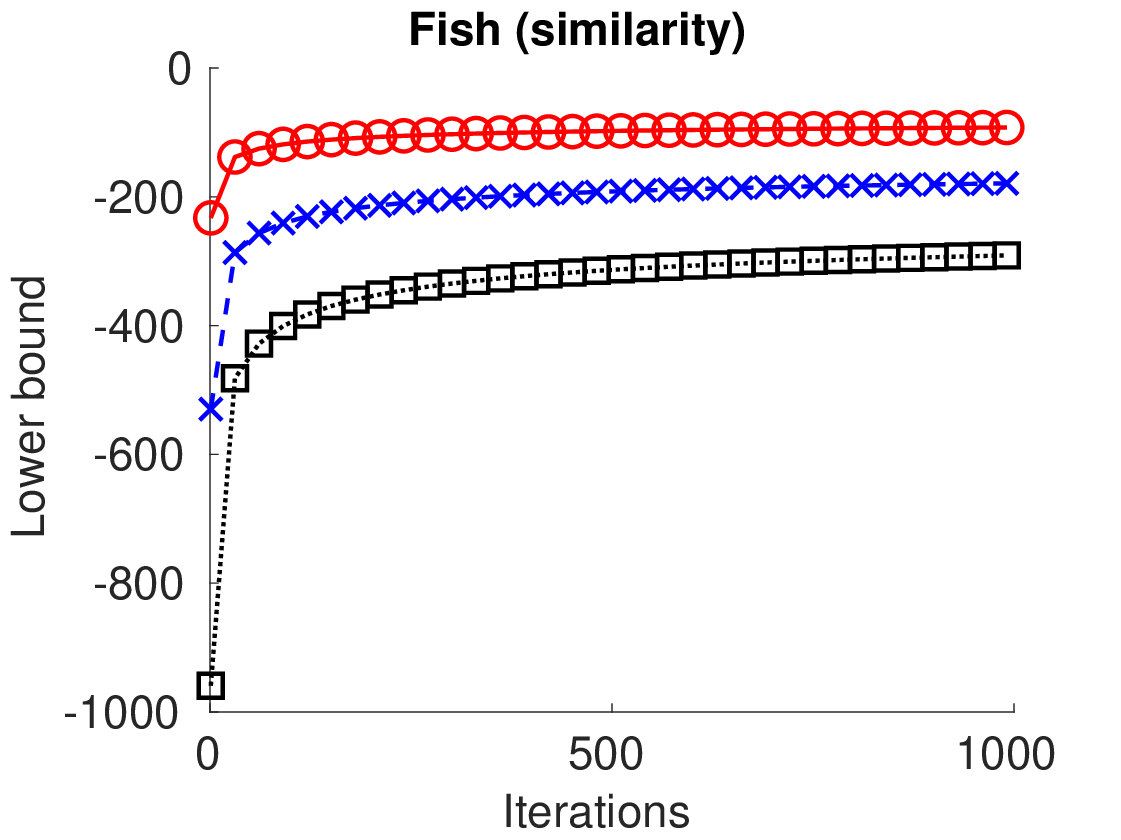}&		
		\includegraphics[width=\scale\linewidth]{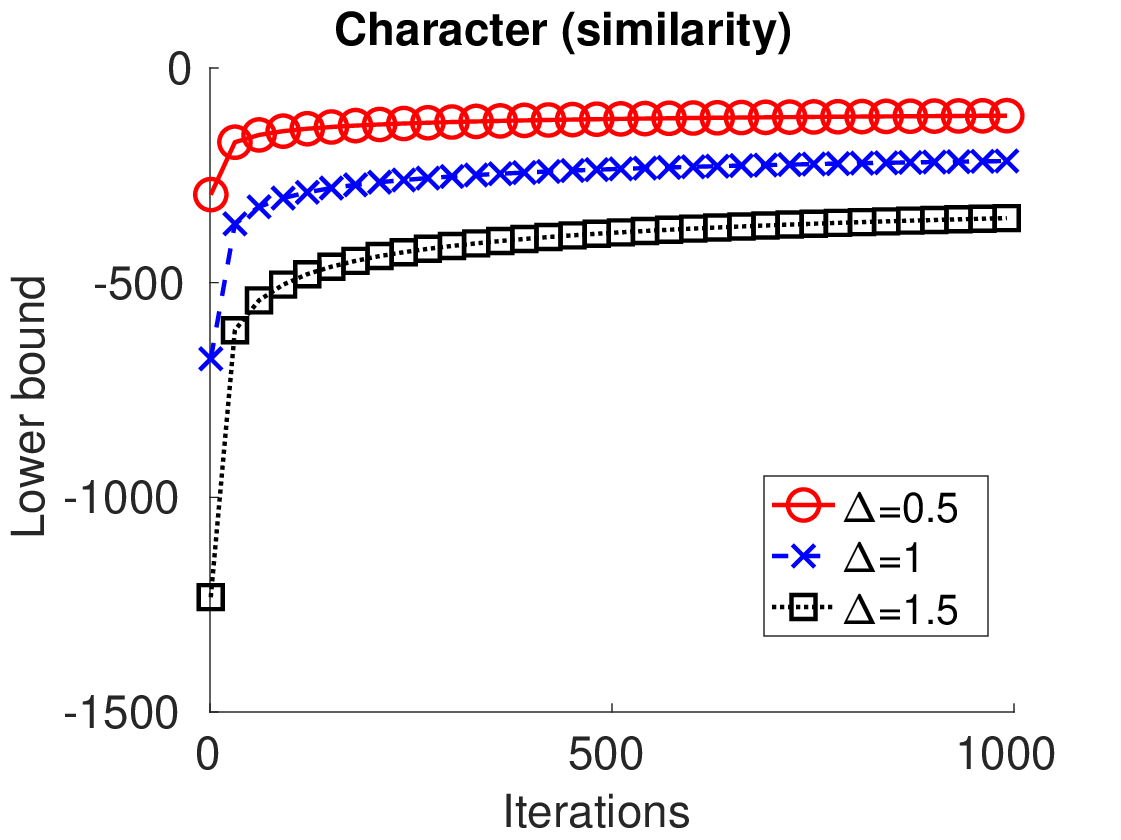}
	\end{tabular}
	\caption{Upper (first row) and
		lower bounds (second row)    generated in each  iteration of the proposed algorithm.
		We test our method with $n_p$ value  chosen as the ground truth and with different initial ranges  
		$[\boldsymbol\theta_{gt}-\Delta, \boldsymbol\theta_{gt}+ \Delta]$
		%$[\theta_{gt}-1, \theta_{gt}+1]$ and
		%$[\theta_{gt}-1.5, \theta_{gt}+1.5]$,
		of $\boldsymbol\theta$,
		where $\boldsymbol\theta_{gt}$ denotes the ground truth $\boldsymbol\theta$ solution and the margin $\Delta$ is respectively chosen as $0.5$, $1$ and $1.5$.
		\label{LB_UB_2D}}
\end{figure}

To evaluate  convergence   of the proposed BnB algorithm,
% tightness of the lower bounds generated by the proposed method,
we use the separate outliers and inliers test described in Sec. \ref{sec:2D_synth_test} where the outlier to data ratio is chosen as $0.3$.
%We test our method with different initial range  
%$[\theta_{gt}-\Delta, \theta_{gt}+ \Delta]$
%%$[\theta_{gt}-1, \theta_{gt}+1]$ and
%%$[\theta_{gt}-1.5, \theta_{gt}+1.5]$,
%of $\theta$,
%where $\theta_{gt}$ denotes the ground truth $\theta$ solution and the margin $\Delta$ is respectively chosen as $0.5$, $1$ and $1.5$.
%
The experimental result is reported in Fig. \ref{LB_UB_2D}.
%From this figure,
%one can see that:
It  shows that: 
\begin{inparaenum}[1)] %\bfseries(1)]
	\item
	The duality gap decreases quickly initially
	but   slowly later on,
	and there  is noticeable  duality gap
	even after many iterations.
	%
	%This suggests that in practice,
	%the tolerance error $\epsilon$ shouldn't be set too small,
	%otherwise  it will incur long running time.
	%
	\item 
	Smaller  initial range of $\boldsymbol\theta$  leads to
	tighter duality gap  (with both better upper and  lower bounds).
	So in practice, 
	if there is more information about the range of $\boldsymbol\theta$, 
	this information can be utilized to reduce  duality gap  of the proposed  method.
	\item 
	Easier problems (e.g., the fish test  versus the character test)
	also leads to tighter duality gap.
	%the duality gap is smaller.
	%thus, accordingly,  the tolerance error can be set smaller.
\end{inparaenum}

Properties  1) and 3)  %together with property 2
suggest   that employing  a  fixed  duality gap threshold
as the stopping criterion 
for all types of problems
%setting the stopping criterion purely based on duality gap
may not be a good idea.
%since  problems of different difficulties cause our method to have different  duality gaps
%and a fixed duality gap threshold can not fit them all.
It may happen that a duality gap threshold 
is set too small for a hard problem,
leading  to long running time,
%hen solving a hard problem,
yet  it is set too large for an easy problem,
causing  premature early termination.
% when solving an easy problem.
%causes hard problems to require too long time to solve,
%yet causes  easier problems to stop being solved prematurely.
So in this paper,
instead of using  duality gap threshold as the stopping criterion,
we use the maximum branching depth as the stopping criterion.
With this choice,
our method is no longer an $\epsilon-$globally optimal algorithm.
Nevertheless,
it differs from traditional  heuristic methods in that it can also return  a degree 
(i.e.,  duality gap)
% at the moment of algorithm termination) 
to which the solution is different from the optimal solution. 

%The above properties  
%In Sec.,
%we will explore  alternative stopping criterion of our algorithm.

\section{Case two: 3D rigid transformation
	\label{sec:case_two}
}

In this section, 
we first discuss the derivation of our objective function for the case that the transformation is 3D rigid.
%In the following, first,
Then, we discuss the branching strategy (Sec. \ref{subsec:branch_case_two}).
Next, we utilize
the lower bound functions of bilinear and trilinear  terms to derive the lower bound function of the objective (Sec.\ \ref{subsec:lb_compute_rigid}).
Then, we discuss the computation of upper bound and computation of the range of $\mathbf R$ from the range of $\mathbf r$ (Sec.\ \ref{subsec:ub_case_two} and \ref{subsec:range_R_case_two}).
Next, we use  point set normalization
to satisfy  a requirement when using  the convex envelope of a trilinear monomial \cite{trilinear_lb} (Sec.\ \ref{subsec:PT_normalize}).
Finally, we present the  BnB algorithm (Sec.\ \ref{sec:BnB_aff}) and discuss its convergence property (Sec.\ \ref{sec:converge_3d}).

A 3D affine transformation has  many parameters,
causing our algorithm to be inefficient.
Therefore,
for the 3D case,
we instead focus on rigid transformation
and develop corresponding objective function
and optimization strategy.

%\subsection{
%	3D rotation	parameterization  and objective function
%}

Using the angle-axis representation,
each 3D rotation can be represented as a 3D vector $\mathbf r$,
with axis $\mathbf r/\|\mathbf r\|$ and angle $\|\mathbf r\|$.
The corresponding $3\times 3$ rotation matrix
$\mathbf R\in \mathbb{SO}_3$
% denotes the group of 3D rotations.
for $\mathbf r$ can be obtained using matrix exponential map as  \cite{Go-ICP_pami}
\begin{gather}
\mathbf R=\mathbf I_3+ \frac{[\mathbf r]_\times \sin\|\mathbf r\|}{\|\mathbf r\|}
+\frac{[\mathbf r]_\times^2(1-\cos \|\mathbf r\|)}{\|\mathbf r\|^2} \label{rot_matrix}
\end{gather}
where $[\cdot]_\times$ denotes the skew-symmetric matrix representation:
\begin{gather}
[\mathbf r]_\times= \begin{bmatrix}
0 & -r_3& r_2\\
r_3&0& -r_1 \\
-r_2 & r_1 &0
\end{bmatrix}
\end{gather}
where $r_i$ is the $i$-th element of $\mathbf r$.

%\subsection{Objective function
%\label{subsec:obj_3d_rigid}
%}
Following 
\cite{RPM_model_occlude_PR},
point set registration is modeled as a linear assignment$-$least square problem:
\begin{gather}
\min 
E(\mathbf P,\mathbf R,\mathbf t)=\sum p_{ij} \| \mathbf y_j -\mathbf R \mathbf x_i -\mathbf t\|^2 \notag\\
=\mathbf 1_{n_x}^\top  \mathbf P \widetilde{\mathbf y} {+} \widetilde{\mathbf x}^\top  \mathbf P \mathbf 1_{n_y} {+} n_p \|\mathbf t\|^2 
{-} 2 \text{tr}(\mathbf R \mathbf X^\top  \mathbf P \mathbf Y )  %\notag\\
{-}2 \mathbf t^\top  (\mathbf Y^\top  \mathbf P^\top \mathbf 1_{n_x}   {-}  \mathbf R \mathbf X^\top  \mathbf P  \mathbf 1_{n_y}) 
\label{energy_Case_two}
\\
s.t.\quad
\mathbf P\in \Omega,\ 
\mathbf R\in \mathbb{SO}_3,\
\underline{\mathbf r}_0\le \mathbf r\le \overline{\mathbf r}_0,\
\underline{\mathbf t}_0 \le \mathbf t\le \overline{\mathbf t}_0
\label{constraint_Case_two}
\end{gather}
%
%\begin{gather}
%\min 
%E(\mathbf P,\mathbf R,\mathbf t)=\sum p_{ij} \| \mathbf n^y_j -\mathbf R \mathbf n^x_i \|^2 \notag\\
%=\mathbf 1^\top  \mathbf P \widetilde{\mathbf n}_y + \widetilde{\mathbf n}_x^\top  \mathbf P \mathbf 1 
%- 2 \text{tr}(\mathbf R \mathbf N_x^\top  \mathbf P \mathbf N_Y )  
%\end{gather}
%
where the transformation  is chosen  as a rigid transformation with rotation matrix $\mathbf R$ and translation $\mathbf t$.
The matrix $\mathbf X\triangleq \begin{bmatrix}
\mathbf x_1,\ldots,\mathbf x_{n_x}
\end{bmatrix}^\top$
and
$\mathbf Y\triangleq\begin{bmatrix}
\mathbf y_1,\ldots,\mathbf y_{n_y}
\end{bmatrix}^\top
$.
The vector $\widetilde{\mathbf x}\triangleq \begin{bmatrix}
\|\mathbf x_1\|_2^2, \dots, \|\mathbf x_{n_x}\|_2^2
\end{bmatrix}^\top$.
%($\widetilde{\mathbf y}$ has been defined.)
%and 
%$\widetilde{\mathbf y}\triangleq \begin{bmatrix}
%\|\mathbf y_1\|_2^2, \dots, \|\mathbf y_n\|_2^2
%\end{bmatrix}^\top$.
%$\mathbf 1_{n_x}$ denotes the $n_x$ dimensional  vector of  all ones.
%Operator
The constant vectors $\underline{\mathbf r}_0$, $\overline{\mathbf r}_0$ (resp. $\underline{\mathbf t}_0$, $\overline{\mathbf t}_0$) define  the  lower and upper bounds of $\mathbf r$ 
(resp.   $\mathbf t$).
%
%Constraint $\mathbf 1_{n_x}^\top \mathbf P  \mathbf 1_{n_y}=n_p$ indicates that the number of matches is required to be equal to   $n_p$,  a predefined  positive integer.
%
%In the following sections,
%we seek to optimize $E$ using the BnB algorithm. 

\subsection{\label{subsec:branch_case_two}
	Branching strategy
}

%\begin{figure} 
%	\begin{tikzcd}
%	E(\mathbf P,\mathbf R,\mathbf t)=\mathbf 1_{n_x}^\top  \mathbf P \widetilde{\mathbf y} + \widetilde{\mathbf x}^\top  \mathbf P \mathbf 1_{n_y} + n_p \|\mathbf t\|^2
%	&	\boxed{	- 2 \text{tr}(\mathbf R \mathbf X^\top  \mathbf P \mathbf Y )  } \arrow[ld,"efe"] 
%	& \boxed{-2 \mathbf t^\top  \mathbf Y^\top  \mathbf P^\top \mathbf 1_{n_x} } \arrow[ld,"lower bound" description]\\
%	E_l=\ldots+ \boxed{ \text{tr}( \mathbf C^0 \mathbf R^\top)
%		-2\text{tr} (\mathbf C^1 \mathbf X^\top \mathbf P\mathbf Y )
%		+ \mathbf 1_3^\top \mathbf C^2 \mathbf 1_3}
%	& +\boxed{\mathbf t^\top \mathbf d^0  -2 \mathbf 1^\top \mathbf P \mathbf Y   \mathbf d^1
%		+ \mathbf 1_3^\top \mathbf d^2}
%	\end{tikzcd}
%\end{figure} 

Based on the result in Sec.\ \ref{sec:bil_lb},
for the bilinear term $-2 \mathbf t^\top  \mathbf Y^\top  \mathbf P^\top  \mathbf 1 $,
for its linear lower bound function to converge to it,
% appeared in our objective function,
we only need to branch  the range of $\mathbf t$
and leave the range of $-2  \mathbf Y^\top  \mathbf P^\top  \mathbf 1 $ fixed.
Here, 
the range of $-2  \mathbf Y^\top  \mathbf P^\top  \mathbf 1 $
can be computed as
%   presented in Fig. \ref{range_determ_rigid}.
%
\begin{equation}
		\min_{\mathbf p\in \Omega} (-2 \mathbf Y^\top \mathbf P^\top \mathbf 1)_i \le
(-2  \mathbf Y^\top  \mathbf P^\top  \mathbf 1)_i
\le \max_{\mathbf p\in \Omega}  (-2 \mathbf Y^\top \mathbf P^\top \mathbf 1)_i 
\end{equation}

%$\min_{\mathbf p\in \Omega} (-2 \mathbf Y^\top \mathbf P^\top \mathbf 1)_i \le
%(-2  \mathbf Y^\top  \mathbf P^\top  \mathbf 1)_i
%\le \max_{\mathbf p\in \Omega}  (-2 \mathbf Y^\top \mathbf P^\top \mathbf 1)_i$
%via solving  linear assignment problems.

%

Likewise,
for the bilinear term $- 2 \text{tr}(\mathbf R \mathbf X^\top  \mathbf P \mathbf Y )$,
we only need to branch  the range of $\mathbf r$ (since the range of $\mathbf r$ determines the range of $\mathbf R$) and leave the range of $- 2  \mathbf X^\top  \mathbf P \mathbf Y $ fixed.
Here, the range of $-2 \mathbf X^\top \mathbf P \mathbf Y $
can be computed as 
% in Fig. \ref{range_determ_rigid}.
%$\min_{\mathbf p\in \Omega} (-2 \mathbf X^\top \mathbf P \mathbf Y)_{ij} \le 
%(-2 \mathbf X^\top \mathbf P \mathbf Y)_{ij} 
%\le
%\max_{\mathbf p\in \Omega}  (-2 \mathbf X^\top \mathbf P \mathbf Y)_{ij}$
%via solving linear assignment problems.
%
\begin{equation}
		\min_{\mathbf p\in \Omega} (-2 \mathbf X^\top \mathbf P \mathbf Y)_{ij} \le 
(-2 \mathbf X^\top \mathbf P \mathbf Y)_{ij} 
\le
\max_{\mathbf p\in \Omega}  (-2 \mathbf X^\top \mathbf P \mathbf Y)_{ij}
\end{equation}

In a similar vein,
based on the result in Sec.\ \ref{sec:lb_tri},
for the trilinear term $2 \mathbf t^\top   \mathbf R \mathbf X^\top  \mathbf P  \mathbf 1 $,
for its linear lower bound function to converge to it,
we only need to branch  the ranges of $\mathbf r$ and $\mathbf t$  and leave the range of $2\mathbf X^\top \mathbf P \mathbf 1$ fixed.
Here, 
the range of $2\mathbf X^\top \mathbf P \mathbf 1 $ can be computed as
% presented in Fig. \ref{range_determ_rigid}.
%$\min_{\mathbf p\in \Omega}  (2\mathbf X^\top \mathbf P \mathbf 1)_j \le
%(2\mathbf X^\top \mathbf P \mathbf 1)_j
%\le
%\max_{\mathbf p\in \Omega}   (2\mathbf X^\top \mathbf P \mathbf 1)_j$
%via solving linear assignment problems.
%
%\begin{figure} [H]
	\begin{equation}
%		\underline{ R}_{ij}\le R_{ij} \le \overline{ R}_{ij}, \quad
%		\underline{ t}_i \le t_i \le \overline{ t}_i
%		, \notag\\
%		\notag\\
%		, \notag\\
		\min_{\mathbf p\in \Omega}  (2\mathbf X^\top \mathbf P \mathbf 1)_j \le
		(2\mathbf X^\top \mathbf P \mathbf 1)_j
		\le
		\max_{\mathbf p\in \Omega}   (2\mathbf X^\top \mathbf P \mathbf 1)_j %\notag
	\end{equation}
%	\caption{Computation of the ranges of various variables, where the subproblems are linear assignment problems which can be efficiently solved.}
%	\label{range_determ_rigid}
%\end{figure} 

In conclusion, in our BnB algorithm,
we only need to branch over  $\mathbf r$ and $\mathbf t$.
Thus, the dimension of the branching space of our BnB algorithm is low and  our algorithm can converge quickly.

\subsection{Lower bound
	%	 of the objective function
	\label{subsec:lb_compute_rigid}}

Based on the results in Sec.\ \ref{sec:cov_env},
we can obtain the lower bound function $E_l$ of $E$ by respectively deriving the lower bound functions of the bilinear and trilinear terms,
as illustrated in Eq. \ref{lb_derive_rigid}.

\begin{figure*}
\begin{equation}  %	[H]
%	\begin{tabular}{@{\hspace{-4mm}}c}	
%	$
	\begin{CD}
	E(\mathbf P,\mathbf R,\mathbf t){=}	
	\begin{matrix}
	\mathbf 1_{n_x}^\top  \mathbf P \widetilde{\mathbf y} 
	 {+} \widetilde{\mathbf x}^\top  \mathbf P \mathbf 1_{n_y} 
	{+} n_p \|\mathbf t\|^2 
	\end{matrix}
	@.\textcolor{blue}{
	\boxed{	- 2 \text{tr}(\mathbf R \mathbf X^\top  \mathbf P \mathbf Y )  }}
	@.\textcolor{blue}{
	\boxed{-2 \mathbf t^\top  \mathbf Y^\top  \mathbf P^\top \mathbf 1_{n_x} } }
	@.
\textcolor{red}{
	\boxed{+2 \mathbf t^\top  \mathbf R \mathbf X^\top  \mathbf P  \mathbf 1_{n_y} }}
	\\
	@.	@V\text{bilinear}V \text{relaxation}V @V\text{bilinear}V\text{relaxation}V @V\text{trilinear}V\text{relaxation}V
	\\
	E_l(\mathbf P,\mathbf R,\mathbf t){=}
	\begin{matrix}
	\mathbf 1_{n_x}^\top  \mathbf P \widetilde{\mathbf y} 
	 {+} \widetilde{\mathbf x}^\top  \mathbf P \mathbf 1_{n_y} 
	{+} n_p \|\mathbf t\|^2 
	\end{matrix}
	@.
	{+}
	\textcolor{blue}{
	\boxed{ 
		\begin{matrix}
		\text{tr}( \mathbf C^0 \mathbf R^\top) 
		{-}2\text{tr} (\mathbf C^1 \mathbf X^\top \mathbf P\mathbf Y ) \\
		+ \mathbf 1_3^\top \mathbf C^2 \mathbf 1_3
		\end{matrix}		
	}}
	@.+
	\textcolor{blue}{
	\boxed{
		\begin{matrix}
		\mathbf t^\top \mathbf d^0  
		{-}2 \mathbf 1^\top \mathbf P \mathbf Y   \mathbf d^1 
		{+} \mathbf 1_3^\top \mathbf d^2
		\end{matrix}	}}
	+@.
	\textcolor{red}{
	\boxed{
		\begin{matrix}
		\mathbf t^\top \mathbf F^0 \mathbf 1_{3}  
		{+} \text{tr}((\mathbf F^1)^\top \mathbf R) 
		{+} 2 \mathbf 1_3^\top \mathbf F^2 \mathbf X^\top \mathbf P \mathbf 1_{n_y} \\
		+\mathbf 1_3^\top \mathbf F^3 \mathbf 1_3
		\end{matrix}
	}}
	\end{CD} 
%	$
%	\end{tabular}
%	\caption{Derivation of the lower bound function,
%		where
%		the constant matrices $\mathbf C^k$ are computed based on the ranges 
%		%$[\underline{ R}_{ij}, \overline{ R}_{ij}]$
%		of $R_{ij}$ and  $(-2 \mathbf Y^\top \mathbf P^\top \mathbf X)_{ij}$,
%		%
%		the constant vectors
%		$\mathbf d^j$ are computed based on the ranges %$[\underline{ t}_i, \overline{ t}_i]$
%		%(\textcolor{red}{please refer to Sec.\ \ref{subsec:range_case_two}})
%		of $t_i$ and  $(-2 \mathbf Y^\top \mathbf P^\top \mathbf 1)_i$
%		%
%		and  the constant matrices $\mathbf F^k$ are computed based on the ranges %$[\underline{ R}_{ij}, \overline{ R}_{ij}]$ 
%		of $R_{ij}$,
%		%$[\underline{ t}_i, \overline{ t}_i]$ 
%		$t_i$  and  $ (2\mathbf X^\top \mathbf P \mathbf 1)_j$.	
%	}
	\label{lb_derive_rigid}
\end{equation} 
\end{figure*}

Consequently, $E_l$ can be written into the following concise form:
%Combining the  results in Sec.\ \ref{subsec:lb_bilinear_case_two} and \ref{subsec:lb_trilinear_case_two},
%we get  lower bound problem of the original problem \eqref{energy_Case_two}, \eqref{constraint_Case_two} as
\begin{gather}
\min E_l(\mathbf P,\mathbf R,\mathbf t)\triangleq
n_p\|\mathbf t\|^2 + \mathbf g_0^\top \mathbf t +
\text{tr}(\mathbf G_1  \mathbf P) %\notag\\
+   \text{tr}(\mathbf G_2 \mathbf  R)	+ g_3 
%s.t.\ \mathbf P\in \Omega,\ 
%\underline{\mathbf R}\le \mathbf R \le \overline{\mathbf R},\
%\mathbf R\in \mathbb{SO}_3,\
%\underline{\mathbf t}\le \mathbf t \le \overline{\mathbf t}
\end{gather}
where $\mathbf g_0$ is a constant  vector,
$\mathbf G_1$ and $\mathbf G_2$ are constant matrices,
and $g_3$ is a constant scalar.
%$\mathbf g_0\triangleq \mathbf d^0+ \mathbf F^0 \mathbf 1_3$,
%the matrices
%$\mathbf G_1\triangleq -2 \mathbf Y \mathbf C^1 \mathbf X^\top - 2 \mathbf Y 
%\mathbf d^1 \mathbf 1_{n_x}^\top + 2 \mathbf 1_{n_y} \mathbf 1_3^\top \mathbf F^2 \mathbf X^\top + \widetilde{\mathbf y} \mathbf 1_{n_x}^\top + \mathbf 1_{n_y}  \widetilde{\mathbf x}^\top$,
%$\mathbf G_2\triangleq (\mathbf C^0)^\top + (\mathbf F^1)^\top$
%and the scalar 
%$g_3\triangleq \mathbf 1_3^\top \mathbf C^2 \mathbf 1_3 + \mathbf 1_3^\top \mathbf d^2 + \mathbf 1_3^\top \mathbf F^3 \mathbf 1_3$.

It is obvious that the minimization of $E_l$ under the constraints 
$\mathbf P\in \Omega,\ 
\underline{\mathbf R}\le \mathbf R \le \overline{\mathbf R},\
\mathbf R\in \mathbb{SO}_3,\
\underline{\mathbf t}\le \mathbf t \le \overline{\mathbf t}$
can be decomposed into separate optimizations over $\mathbf P$, $\mathbf R$  and $\mathbf t$.
$\mathbf P$ is recovered by solving the following linear assignment problem:
\[
\min_{\mathbf P\in\Omega} \text{tr}(\mathbf G_1  \mathbf P) 
\]
%where $\Omega$ denotes the feasible region of $\mathbf P$, as determined by \eqref{k_card_P_const}.
%

$\mathbf R$ is recovered by solving the following optimization problem:
\begin{gather}
\min \text{tr} (\mathbf G_2 \mathbf R) \notag\\
s.t. \ \underline{\mathbf R}\le \mathbf R \le \overline{\mathbf R},\ \mathbf R\in \mathbb{SO}_3
\end{gather}
This problem is still difficult to solve.
We further relax this problem by dropping the constraint $\mathbf R\in \mathbb{SO}_3$.
Then the remaining problem is a low dimensional linear program which can be solved by  solvers such as the Matlab function $linprog$.

$\mathbf t$ is  recovered  by solving the following low dimensional convex quadratic problem:
\begin{gather}
\min \ n_p \|\mathbf t\|^2 + \mathbf g_0^\top \mathbf t \notag\\
s.t. \quad \underline{\mathbf t}\le \mathbf t \le \overline{\mathbf t}
\end{gather}
%which can be solved by 
%We use off-the-self 
Solvers such as 
the matlab function $quadprog$ can be used for this problem.
% to solve this problem.

\subsection{Upper bound
	%	 of the objective function
	\label{subsec:ub_case_two}}

Since the matrix $\mathbf R$ yielded in Sec.\ \ref{subsec:lb_compute_rigid}  is not a valid rotation matrix, 
we can not directly plug the computed  $\mathbf P,\mathbf R,\mathbf t$ 
%computed in Sec.\ \ref{sec:lb_fun} 
into  $E(\mathbf P,\mathbf R,\mathbf t)$ to get an upper bound.
To address this issue, based on a result of \cite{RPM_model_occlude_PR},
the objective  $E(\mathbf P,\mathbf R,\mathbf t)$ can be rewritten as a function of $\mathbf P$ by eliminating $\mathbf t$ and $\mathbf R$:
\begin{gather}
E(\mathbf P) =\mathbf 1^\top  \mathbf P \widetilde{\mathbf y} + \widetilde{\mathbf x}^\top  \mathbf P \mathbf 1  
- \frac{1}{n_p} ( \|\mathbf Y^\top  \mathbf P^\top  \mathbf 1\|^2  +   \| \mathbf X^\top  \mathbf P \mathbf 1\|^2 )
\notag\\
- 2\min_{\mathbf{\widetilde{R}}\in \mathbb{SO}_3}\{ \text{tr}(\mathbf {\widetilde R }( \mathbf X^\top  \mathbf P \mathbf Y - \frac{1}{n_p} \mathbf X^\top \mathbf P\mathbf 1\mathbf 1^\top \mathbf P\mathbf Y))\} 
\end{gather}
Therefore, 
we can plug the computed $\mathbf P$ into this function to get the value of $E$, 
which is used as an upper bound.
Here the optimal $\mathbf R$ can be obtained when we first compute  the singular value decompostion  
$\mathbf U\mathbf S\mathbf V^\top$ of
$( \mathbf X^\top  \mathbf P \mathbf Y - \frac{1}{n_p} \mathbf X^\top \mathbf P\mathbf 1\mathbf 1^\top \mathbf P\mathbf Y)^\top$,
%where
% be the singular value decomposition of $\mathbf G^\top$,
where 
the columns of  $\mathbf U$ and $\mathbf V$ are orthogonal unit vectors, and $\mathbf S$ is a diagonal matrix.
Then,
the optimal  $\mathbf R$ can be computed as
$ \mathbf R^*=\mathbf U\text{diag}(\begin{bmatrix}
1,\ldots,1,\det(\mathbf U\mathbf V^\top)
\end{bmatrix})\mathbf V^\top$ \cite{CPD_match}.

\subsection{The range of $\mathbf R$
	\label{subsec:range_R_case_two}
}

%The  analysis in the previous section
%indicates that the ranges of  $(-2 \mathbf Y^\top \mathbf P^\top\mathbf X)_{ij}$, $(-2 \mathbf Y^\top \mathbf P^\top \mathbf 1)_i$ and
%$ (2\mathbf X^\top \mathbf P \mathbf 1)_j$  need  to be computed only once.
%%
%Since $(-2 \mathbf Y^\top \mathbf P^\top\mathbf X)_{ij}$ is a linear function of $\mathbf  P$,
%its ranges
%% of $(-2 \mathbf X^\top \mathbf P\mathbf Y)_{ij}$
%can be efficiently computed by solving the following linear assignment problems:
%$\min_{\mathbf P\in \Omega} (-2 \mathbf Y^\top \mathbf P^\top\mathbf X)_{ij} $ and
%$\max_{\mathbf P\in \Omega} (-2 \mathbf Y^\top \mathbf P^\top\mathbf X)_{ij} $.
%%
%Likewise,
%the ranges of 
%$(-2 \mathbf Y^\top \mathbf P^\top \mathbf 1)_i$ and
%$ (2\mathbf X^\top \mathbf P \mathbf 1)_j$
%can be  obtained by solving the corresponding linear assignment problems.

%$\min_P \{(-2 \mathbf Y^\top  P^\top  \mathbf 1)_i | P\in\Omega\}$, $\max_P \{(-2 \mathbf Y^\top  P^\top  \mathbf 1)_i | P\in\Omega\}$,
%$\min_P \{(2 \mathbf X^\top   P \mathbf 1)_i | P\in\Omega\}$,
%$\max_P \{(2 \mathbf X^\top P \mathbf 1)_i | P\in\Omega\}$.

Given a range $[\underline{\mathbf r},\overline{\mathbf r}]$  of $\mathbf r$, 
based on  \eqref{rot_matrix},
%we can use interval analysis to compute the range of $R$.
%But  the computed range is often not tight, slowing down the convergence  of our BnB algorithm.
%To address this issue,
%in this paper,
we can compute the range
%$[\underline{\mathbf R},  \overline{\mathbf R}]$
% $[\underline{\mathbf R},\overline{\mathbf R}]$ 
of   $\mathbf R$ as
%solving the following optimization problems
$\min\{  R_{ij}| \underline{\mathbf r}\le \mathbf r\le  \overline{\mathbf r}\}
\le R_{ij} \le
\max\{  R_{ij}| \underline{\mathbf r}\le \mathbf r\le  \overline{\mathbf r}\}$
via solvers such as 
%Off-the-self 
%constrained nonlinear optimization 
%solvers such as 
the Matlab function $fmincon$.
%can be used for this problem.
Nevertheless, this  process is cumbersome,
especially when it is repeatedly executed as a subroutine of the BnB algorithm.
To address this issue,
we  utilize the idea of precomputation to speed up the  process:
Before performing   our BnB algorithm,
we  build a  regular grid (whose width is chosen as $150$ in this work) over the initial cube $[\underline{\mathbf r}_0,\overline{\mathbf r}_0]$.
%where $[\underline{\mathbf r}_0,\overline{\mathbf r}_0]$ denotes the initial range of $\mathbf r$.
We then compute the corresponding $ R_{ij}$ values for all the  grid points and store them.
During the execution of our BnB algorithm,
given a range $[\underline{\mathbf r},\overline{\mathbf r}]$,
we first find the grid points falling in this range.
We then  compute the minimum and maximum of the precomputed $ R_{ij}$ values of these points,
which are used to approximate  
$\min\{ R_{ij}| \underline{\mathbf r}\le \mathbf r\le  \overline{\mathbf r}\}$ and
$\max\{ R_{ij}| \underline{\mathbf r}\le \mathbf r\le  \overline{\mathbf r}\}$.

% respectively computing the range of each element  $(\mathbf R)_{ij}$.

\subsection{Point set normalization
	\label{subsec:PT_normalize}
}
The convex envelope of 
a trilinear monomial presented in  \cite{trilinear_lb} only holds under the condition that at least one variable 
that makes up the monomial
has a range containing the origin,
whereas for the  trilinear term 
$2 \mathbf t^\top    \mathbf R \mathbf X^\top  \mathbf P  \mathbf 1$,
%$ t_i   \mathbf R_{ij}  (2\mathbf X^\top  \mathbf P  \mathbf 1)_j$, 
%in our problem,
such a condition may be violated.
To address this issue,
note that the ranges of $t_i$ and $R_{ij}$ change during the BnB procedure, whereas 
the range of $(\mathbf X^\top \mathbf P \mathbf 1)_j$ remains fixed.
Therefore,
we seek to ensure that the range of $(\mathbf X^\top \mathbf P \mathbf 1)_j$   contains the origin.
This can be accomplished as follows:
%Before  point set alignment,
We translate $\mathscr X$
such that its center locates at the origin:
$\frac{1}{n_x}\mathbf X^\top \mathbf 1 =0$.
Then the mean of the sum of 
the coordinates of 
any $n_p$ number of points in $\mathscr X$ also equals zero:
$\frac{1}{|\Theta|}\sum_{\mathbf P\in \Theta} \mathbf X^\top \mathbf P \mathbf 1=0$,
where the set $\Theta \triangleq\Omega \cap \{\mathbf P|p_{ij}\in\{0,1\}\}$.
Therefore, 
the range of $(\mathbf X^\top \mathbf P \mathbf 1)_j$ will contain the origin.

%Since the optimal $\mathbf t $ depends on the   relative  position of two point sets,
%we also  translate $\mathscr  Y$ 
%so that it is centered at the origin
%before conducting point set registration.

\begin{figure} [t]
	\centering
	\newcommand{\scale}{0.5}

	\begin{tabular}{@{}c@{}  c @{} }
			\includegraphics[width=\scale\linewidth]{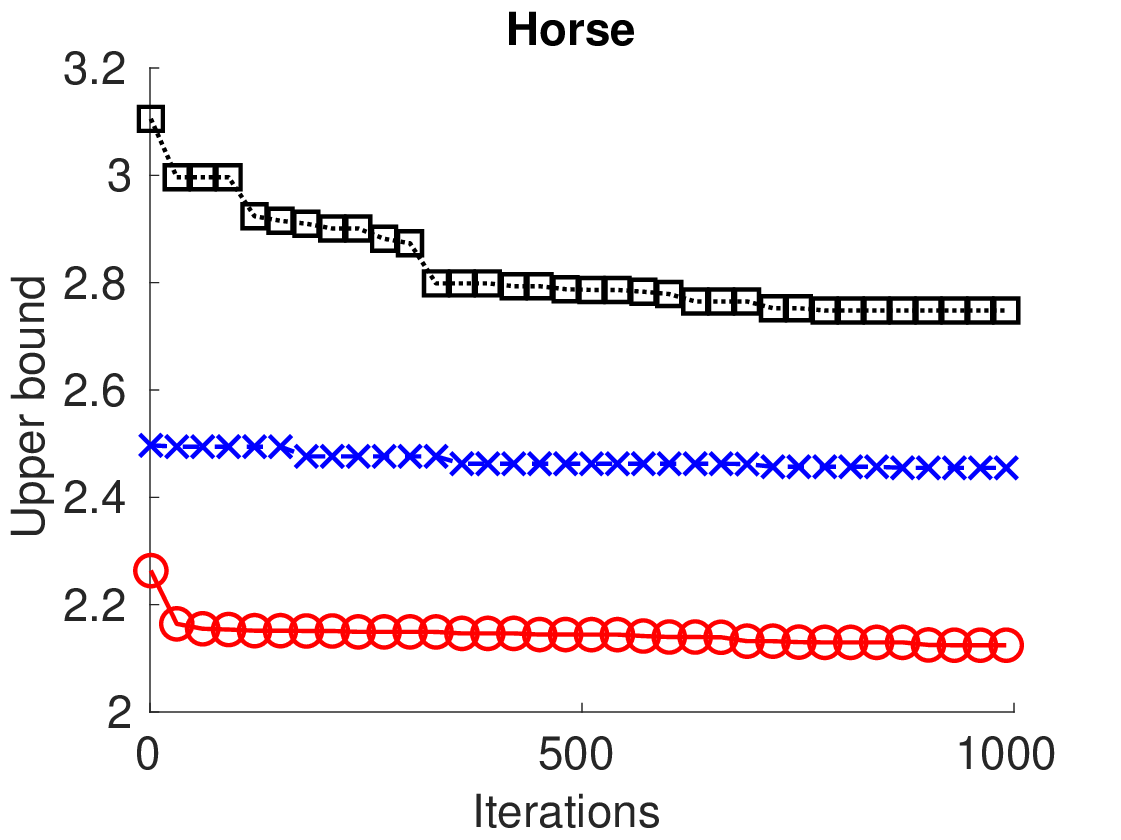}&
			\includegraphics[width=\scale\linewidth]{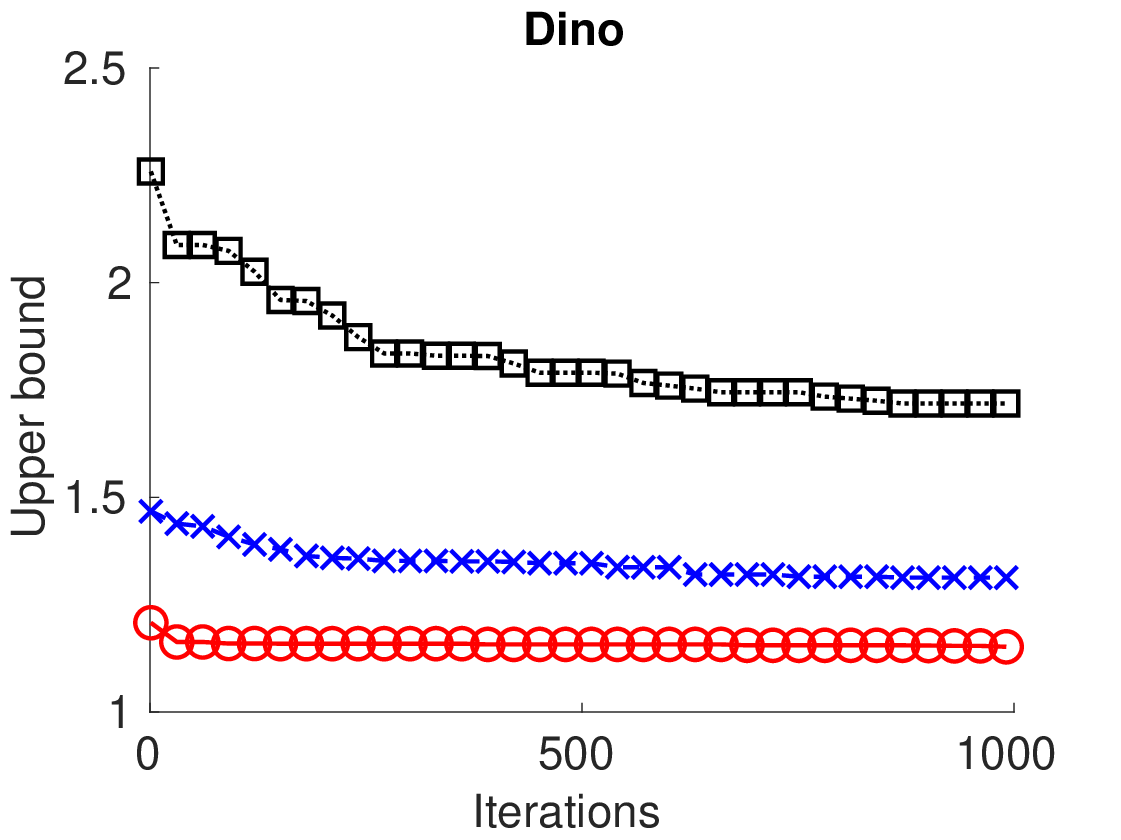}\\
		\includegraphics[width=\scale\linewidth]{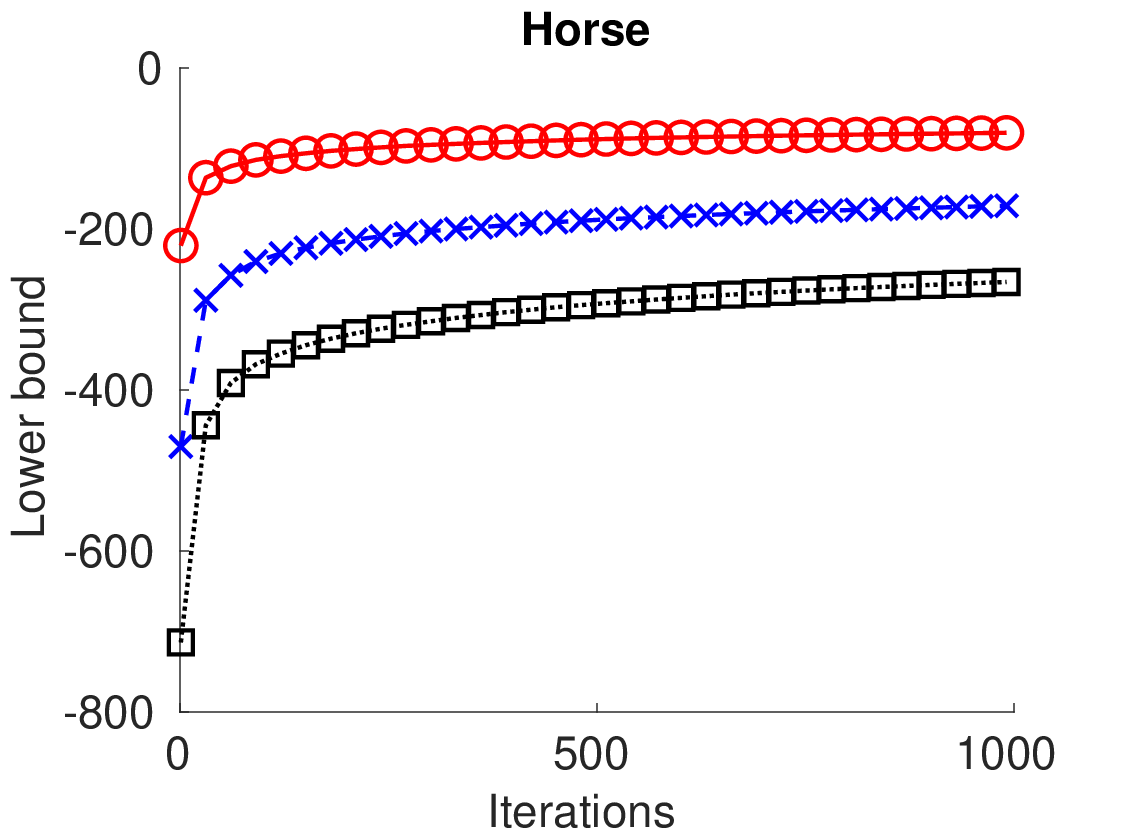}&	
		\includegraphics[width=\scale\linewidth]{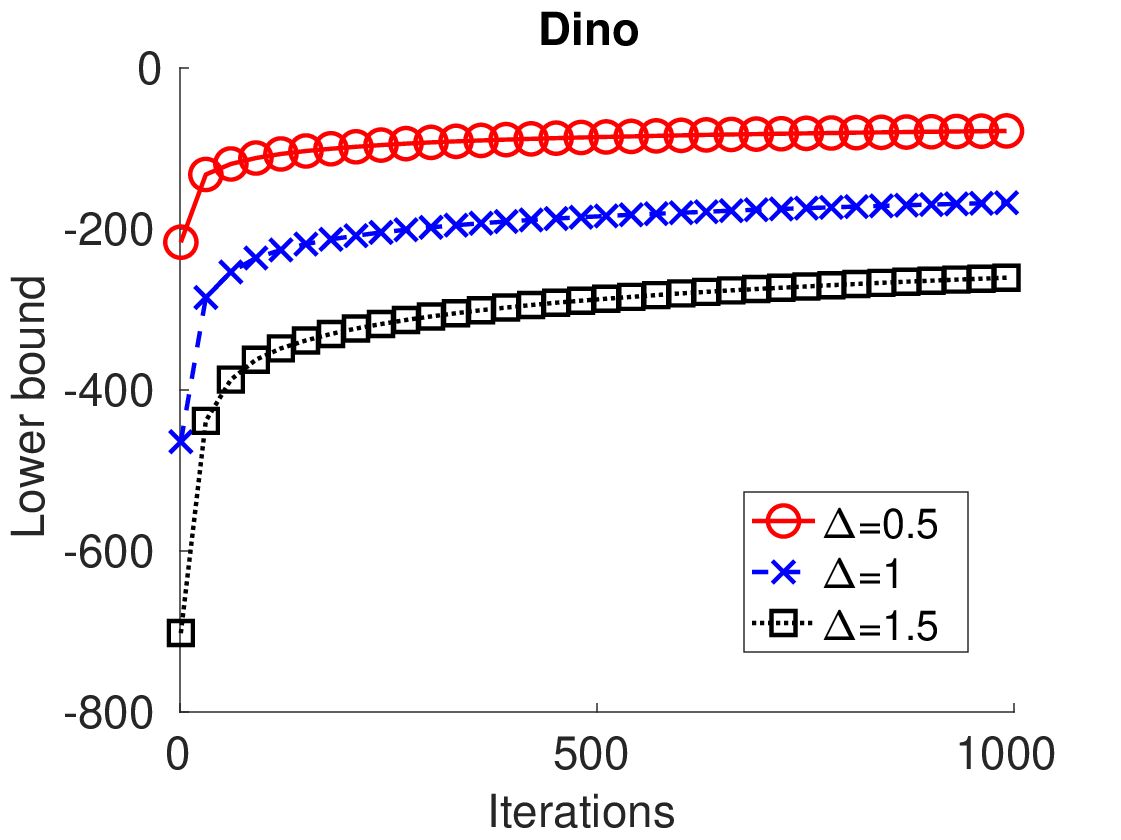} 	 
	\end{tabular}
	\caption{
Upper  (first row) and
lower  bounds (second row)   generated in each  iteration of the proposed algorithm.
		We test our method 
		with $n_p$ value chosen as the ground truth
		and
		with different initial ranges
		$[\mathbf r_{gt}-\Delta, \mathbf r_{gt}+ \Delta]$
		of $\mathbf r$
		and
		$[\mathbf t_{gt}-\Delta, \mathbf t_{gt}+ \Delta]$
		of  $\mathbf t$,
		where $\mathbf r_{gt}$ (resp. $\mathbf t_{gt}$) denotes the ground truth $\mathbf r$ (resp. $\mathbf t$) solution. The margin $\Delta$ is respectively chosen as $0.5$, $1$ and $1.5$.
		\label{LB_UB_3D}}
\end{figure}

\subsection{Branch-and-Bound \label{sec:BnB_rigid}}
Based on the aforementioned preparations,
we are now ready to employ the BnB algorithm
%a globally optimal algorithm for  non-convex problems, 
to optimize $E$. %(\mathbf p,\boldsymbol{\theta})$.
%During initialization,
%we use the inital range of $\boldsymbol{\theta}$
%to  construct the initial hypercube.
%%
%Then, in each iteration of the algorithm, the hypercube yielding the lowest
%lower bound among all the hypercubes is further subdivided so as to improve
%the global lower bound of the problem. Meanwhile, the upper bound is updated
%by evaluating $E(\mathbf p)$ with $\mathbf p$ solutions generated during the computation of the lower bounds. 
The pseudocode  is almost identical to 
Algo. \ref{tri_BnB_algo}
except that 
%where
the initial hypercube $M$ is set as
%$[\underline{\boldsymbol{\theta}}_0, \overline{\boldsymbol{\theta}}_0]$
%for the  case that transformation is linear w.r.t. its parameters
%or
$\{ (\mathbf r,\mathbf t)|\mathbf r \in \text{initial range of }\mathbf r
,\
\mathbf t \in \text{initial range of }\mathbf t\}$
%for the case that  transformation is 3D rigid.
%
and the notations
$\mathbf p$, $\mathbf p^k$ and $\mathbf p^{k-1}$ are replaced by 
$\mathbf P$, $\mathbf P^k$ and $\mathbf P^{k-1}$,
respectively.
% 
%Here note that for the  case 
%%of Sec. \ref{sec:case_one},
%that transformation is linear w.r.t. its parameters,
%the vector form  of the point correspondence variable is more convenient. % than the matrix form.
%Thus,
%the notations $\mathbf P$, $\mathbf P^k$ and $\mathbf P^{k-1}$ should to be  replaced by   $\mathbf p$, $\mathbf p^k$ and $\mathbf p^{k-1}$, respectively.
%Accordingly, 
%the function 
%$E(\mathbf P^k)$ is replaced by the function $E(\mathbf p^k)$.

In a similar vein as in Sec. \ref{sec:BnB_aff},
%Since the optimal transformation parameters
%%$\boldsymbol\theta$ 
%depends on the relative size and  position of two point sets,
%in this work, 
we rescale  point sets $\mathscr{X}$ and $\mathscr Y$ 
to be unit sized 
%is rescaled accordingly)
and translate 
them 
%$\mathscr{X}$ and $\mathscr Y$
to be centered at the origin
before performing alignment.
%(this  is consistent with the requirement in Sec.\ \ref{sec:PT_normalize}).
%In this paper, 
Unless otherwise stated,
we set initial ranges for $\mathbf r$ and $\mathbf t$ as   %$\underline{\boldsymbol{\theta}}_0=-3 \mathbf 1_{n_\theta}$,
%$\overline{\boldsymbol{\theta}}_0=3 \mathbf 1_{n_\theta}$,
$\left[ -\pi \mathbf 1_3,
\pi \mathbf 1_3 \right]$
and
$\left [-3 \mathbf 1_3,
3 \mathbf 1_3\right ]$, respectively.

\subsection{Convergence of the BnB algorithm}
\label{sec:converge_3d}
To evaluate the convergence of the proposed BnB algorithm,
we use
the separate outliers and inliers test described in Sec. \ref{sec:3D_synth_test} where the outlier to data ratio is chosen as $0.3$.
%
%To test  tightness of the lower bounds generated by the proposed method,
%we use the outlier test described in Sec. where the outlier to data ratio is set as $0.3$.
%
The experimental result is presented in Fig. \ref{LB_UB_3D}.
Similar conclusion as in Sec. \ref{sec:converge_2d} can be drawn about the convergence of the proposed BnB algorithm  in 3D case.

\begin{figure*} [t]

	\centering
	\newcommand{\scale}{0.105	}

	\begin{tabular}{@{\hspace{-0mm}}c@{}|@{}c@{}|@{}c@{}|@{}c@{}|@{}c@{}|@{}c@{}|@{}c@{}|@{}c|@{}c }	
		\includegraphics[width=\scale\linewidth]{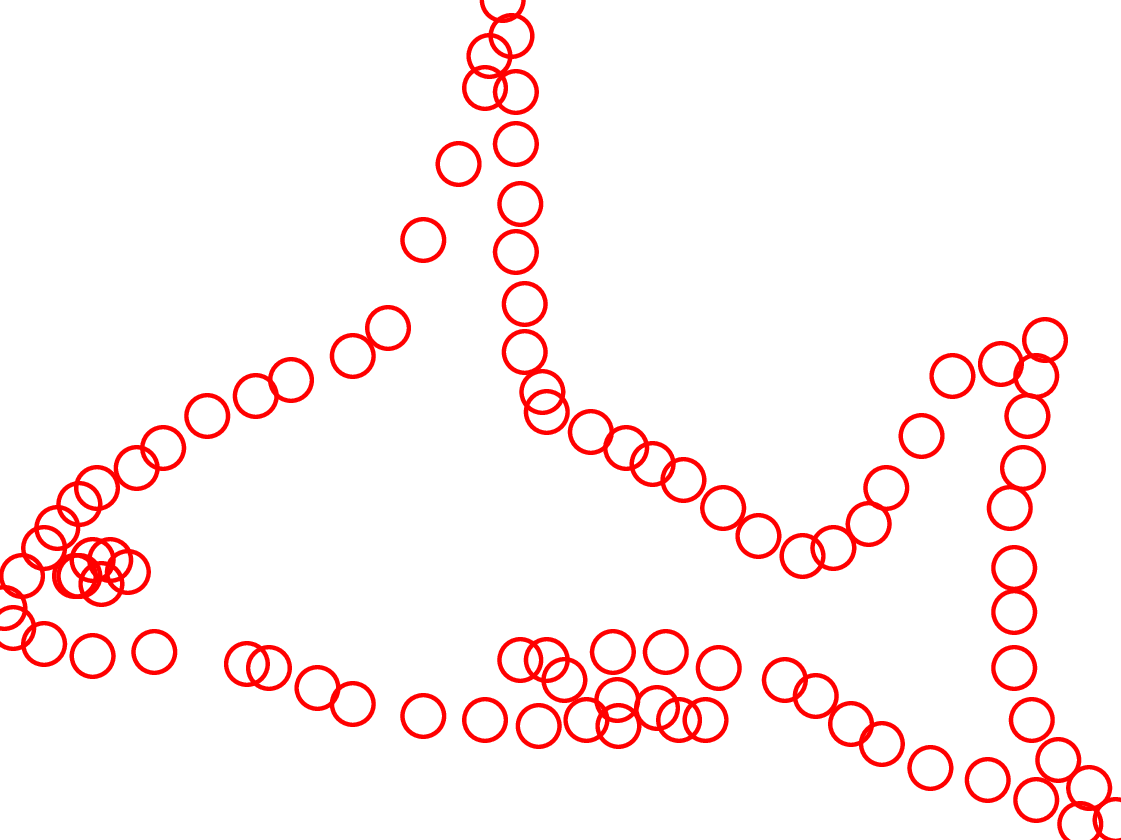}&		
		\includegraphics[width=\scale\linewidth]{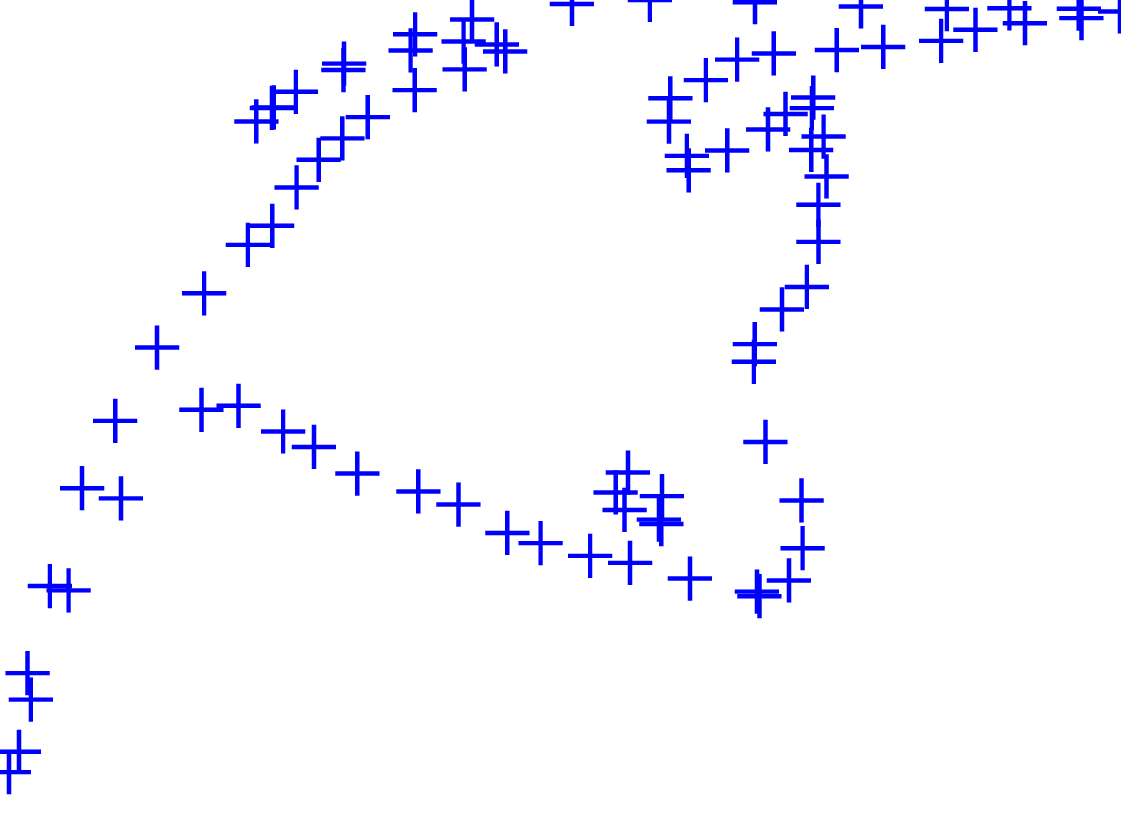}&
		\includegraphics[width=\scale\linewidth]{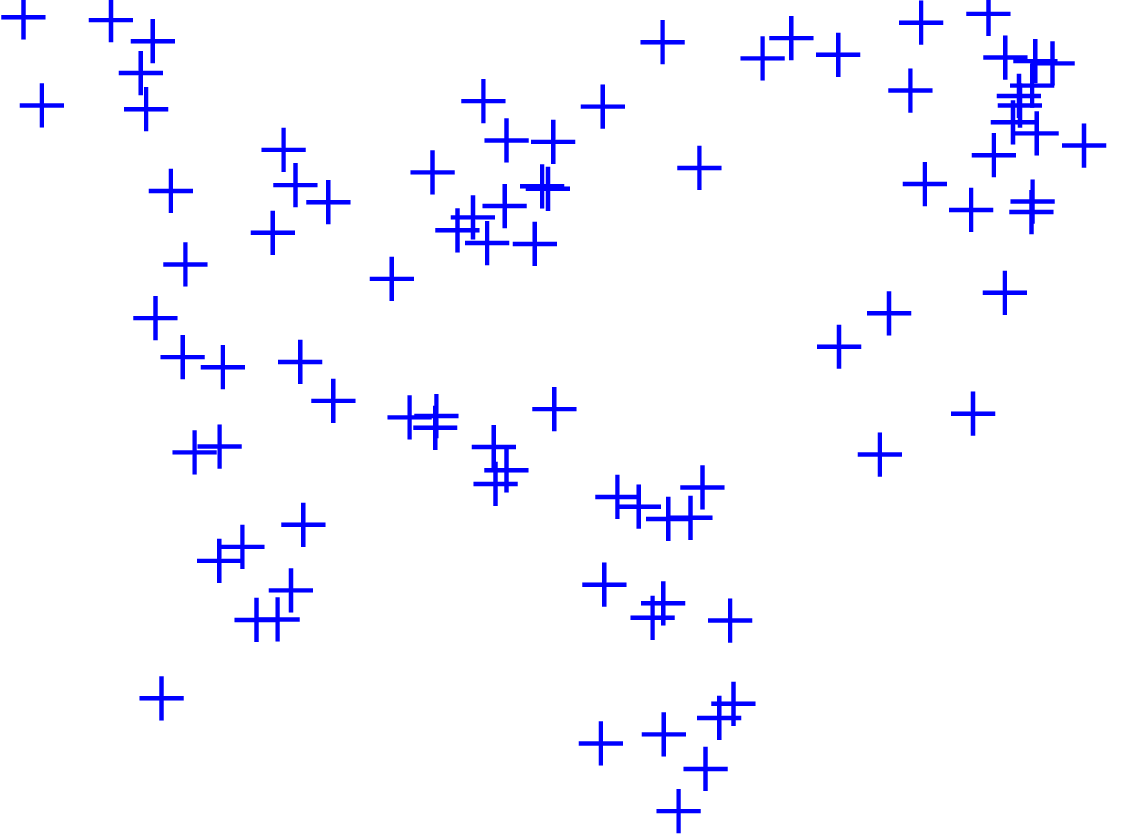}&		
	\includegraphics[width=\scale\linewidth]{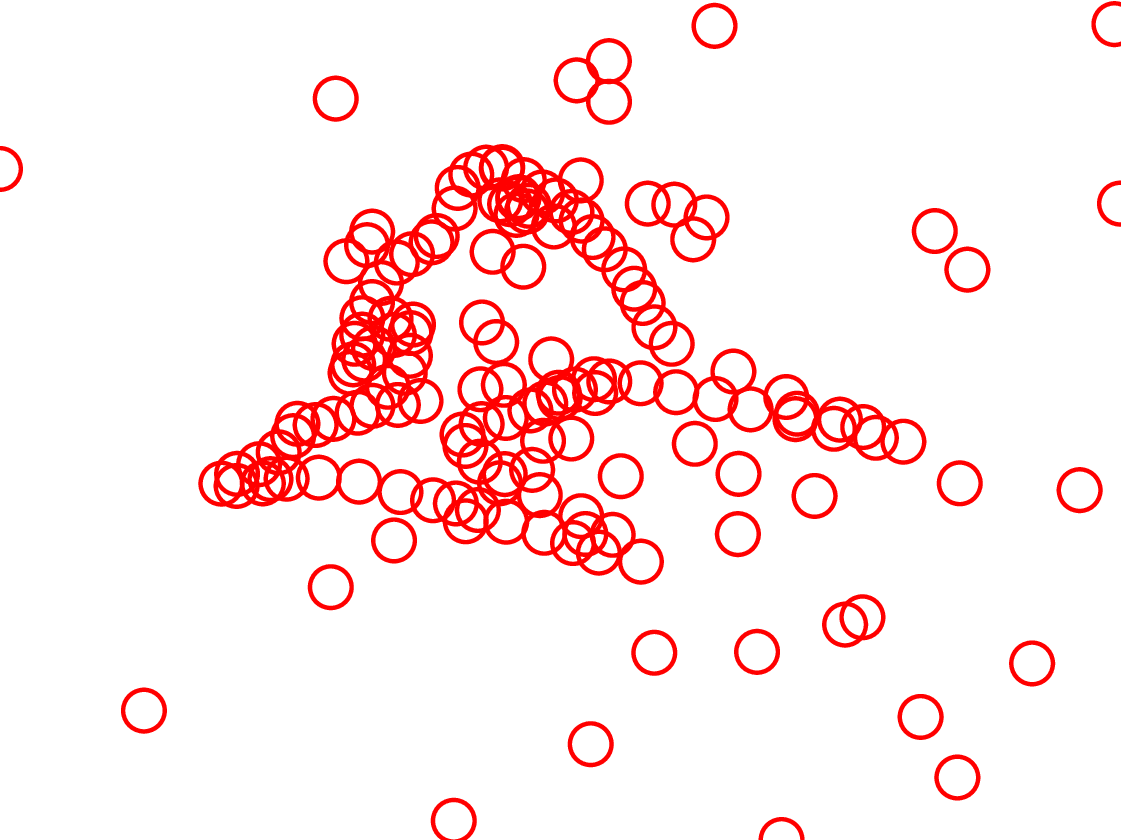}&
		\includegraphics[width=\scale\linewidth]{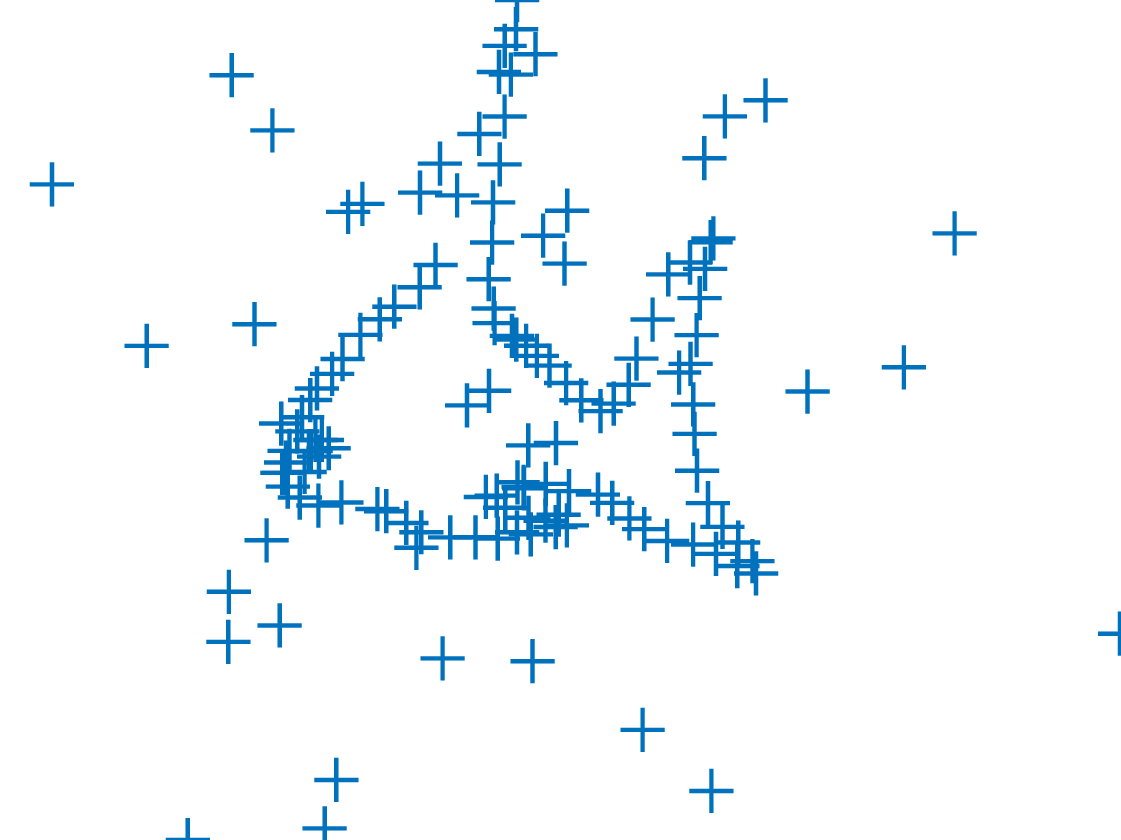}&
		\includegraphics[width=\scale\linewidth]{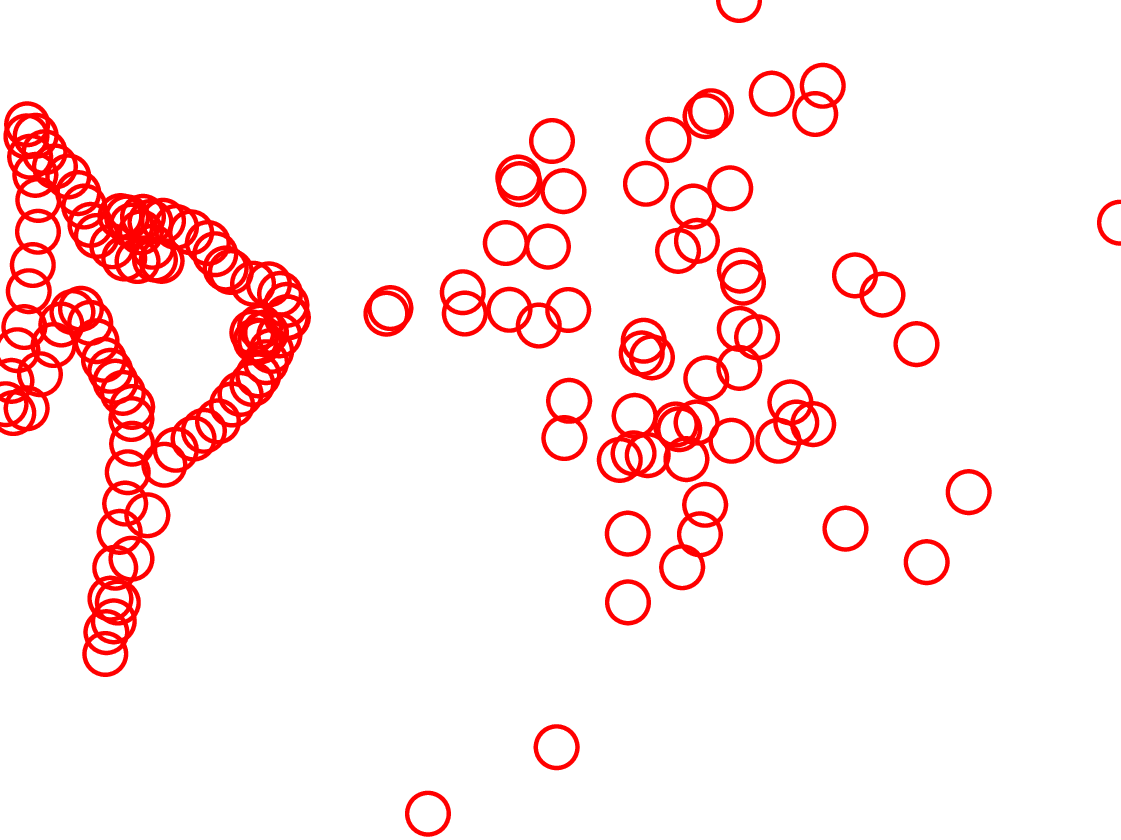}&
		\includegraphics[width=\scale\linewidth]{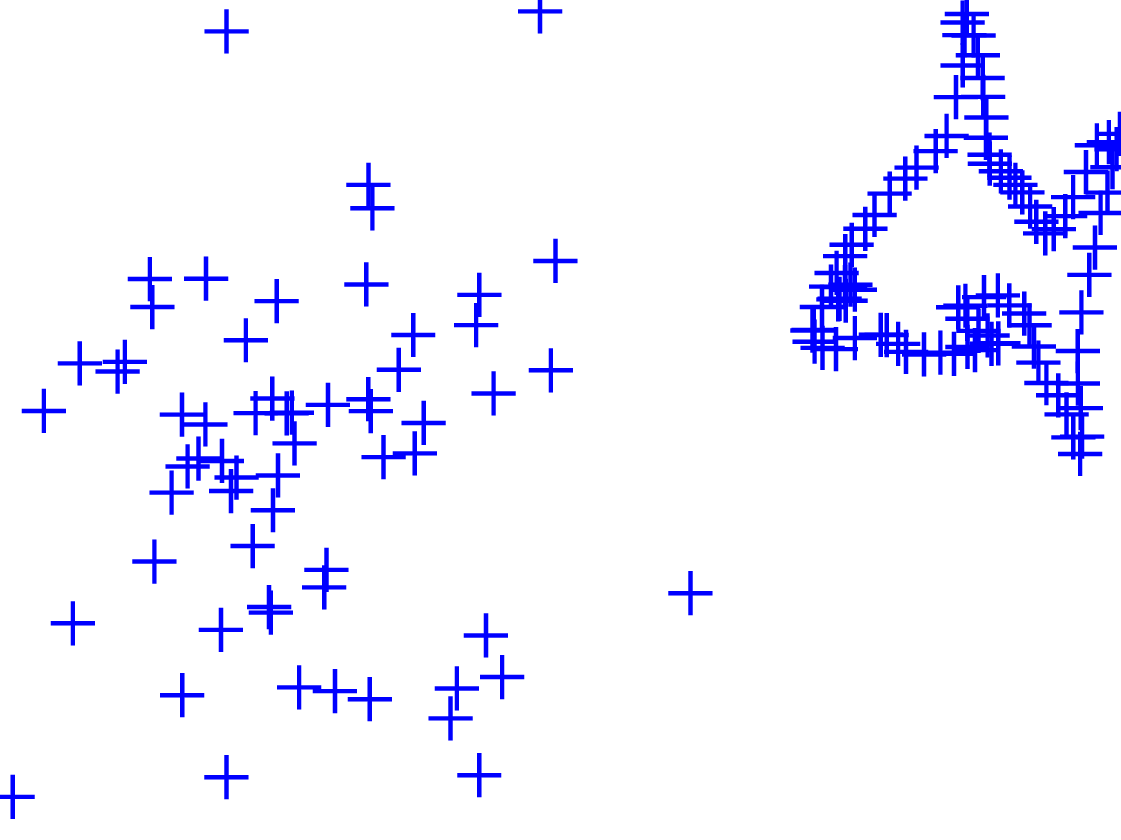}&
		\includegraphics[width=\scale\linewidth]{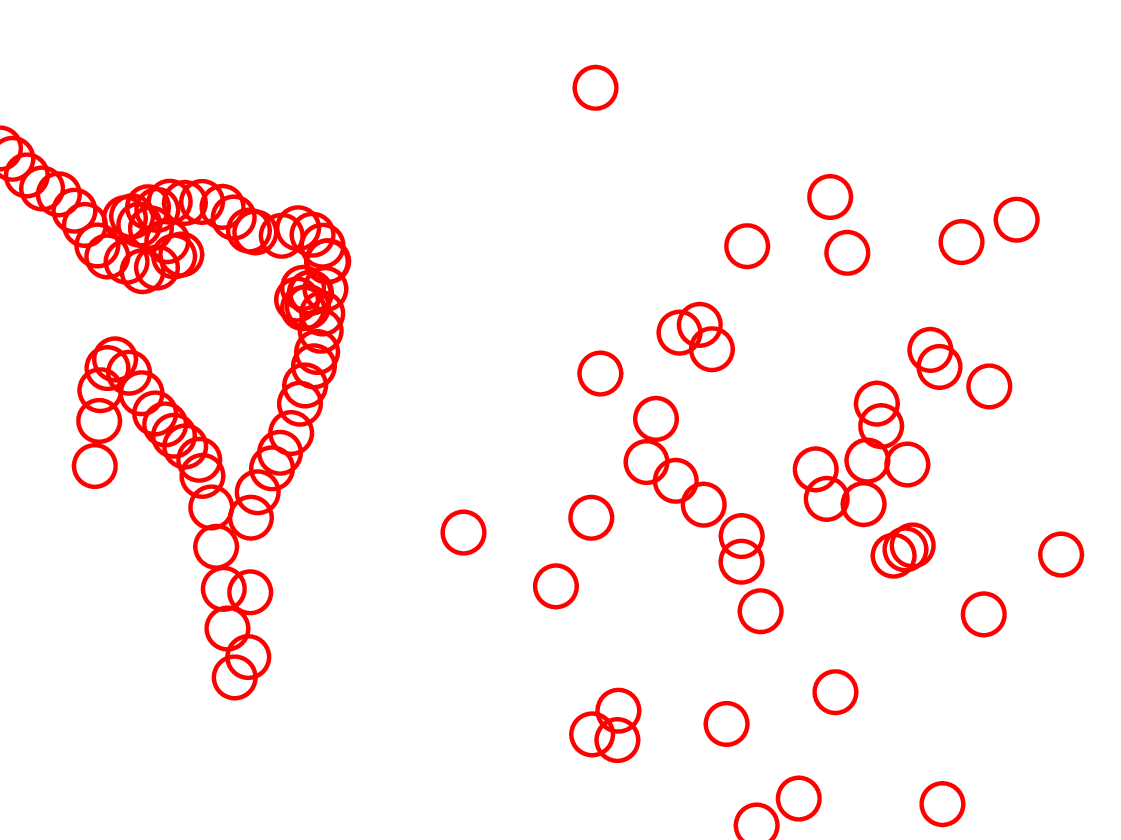}&
		\includegraphics[width=\scale\linewidth]{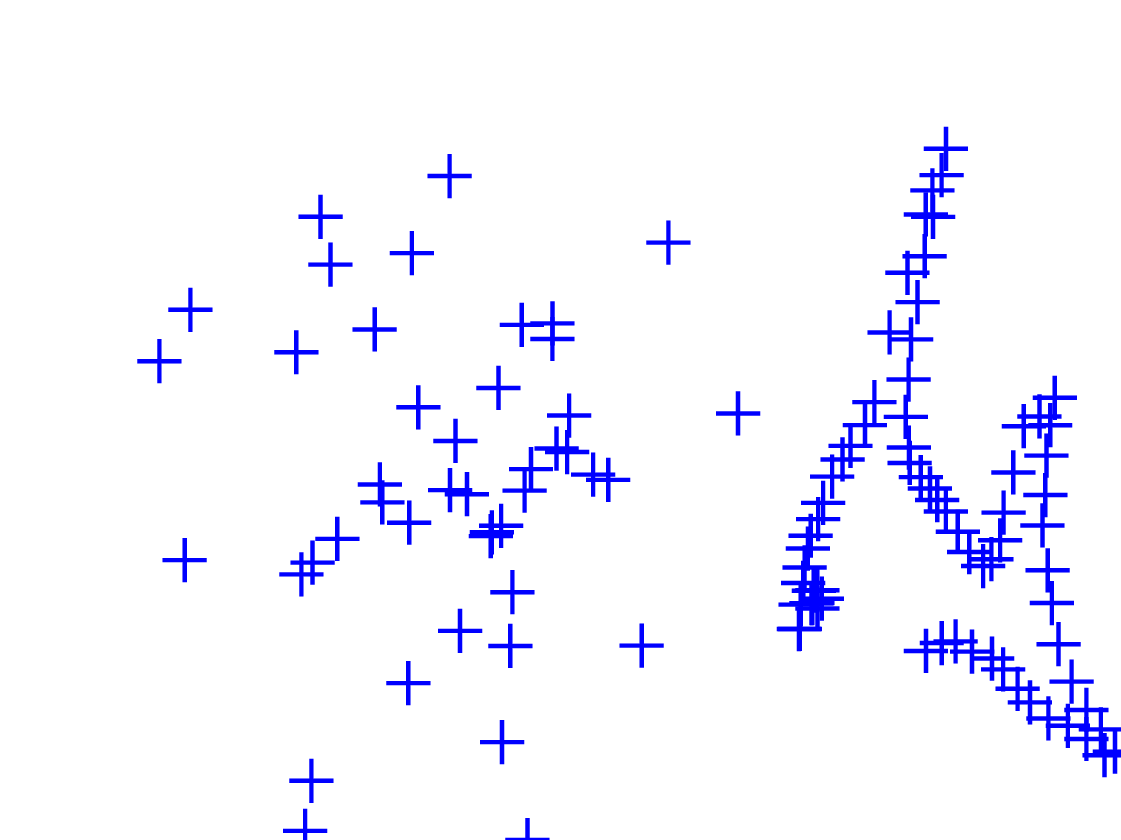} 
		\\	\hline
%				\vspace{1mm}	 
\subfigure[ ]{		\includegraphics[width=\scale\linewidth]{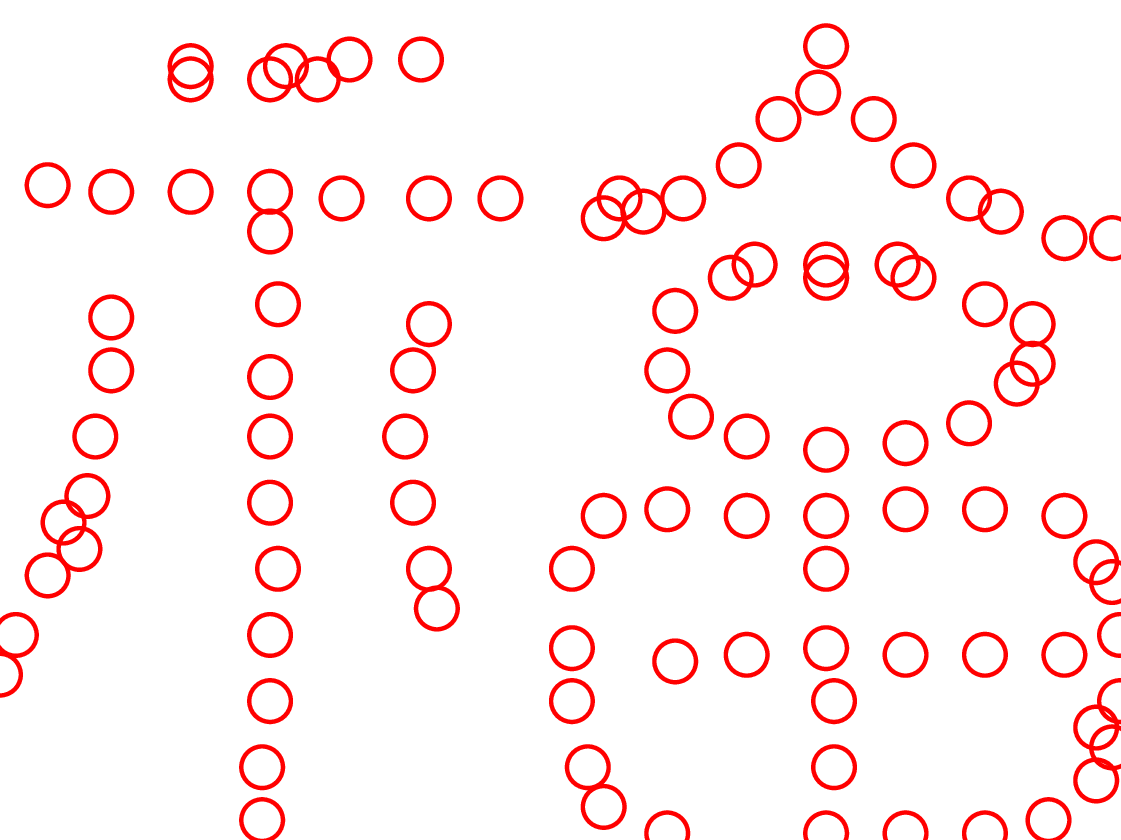}} &
\subfigure[]{		
		\includegraphics[width=\scale\linewidth]{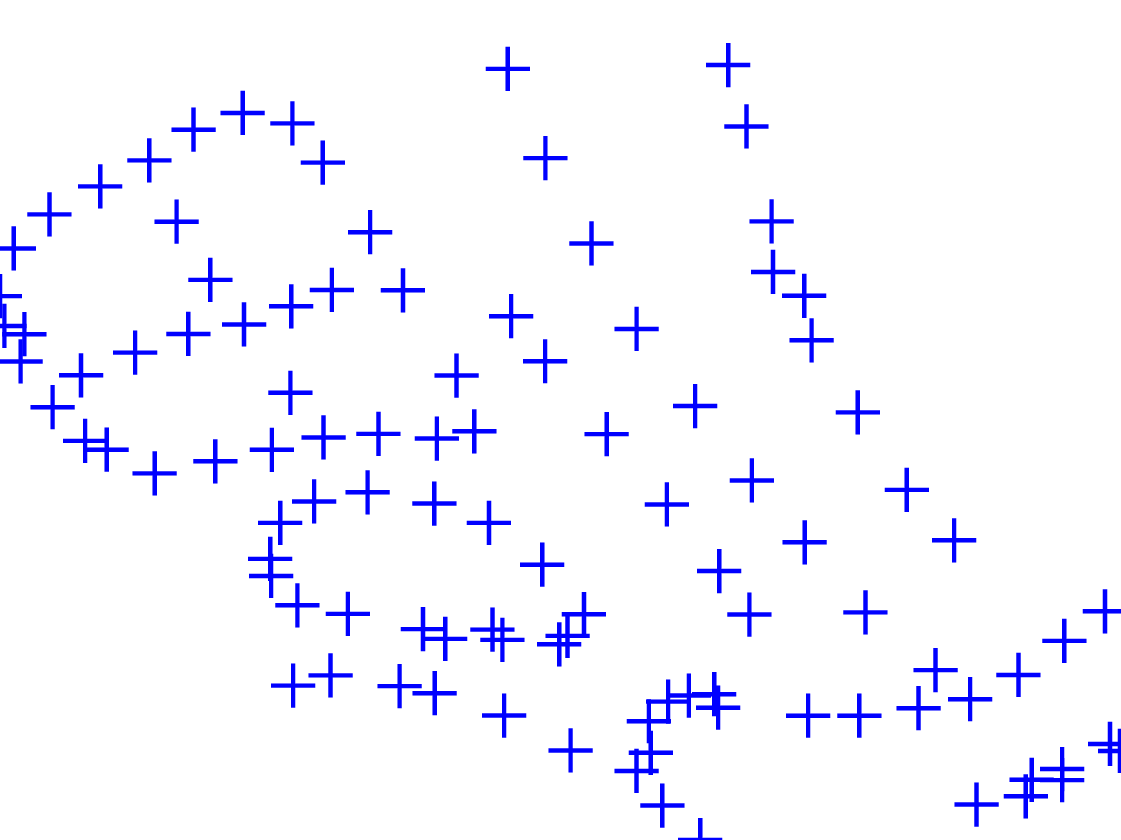}}&
		\subfigure[]{
		\includegraphics[width=\scale\linewidth]{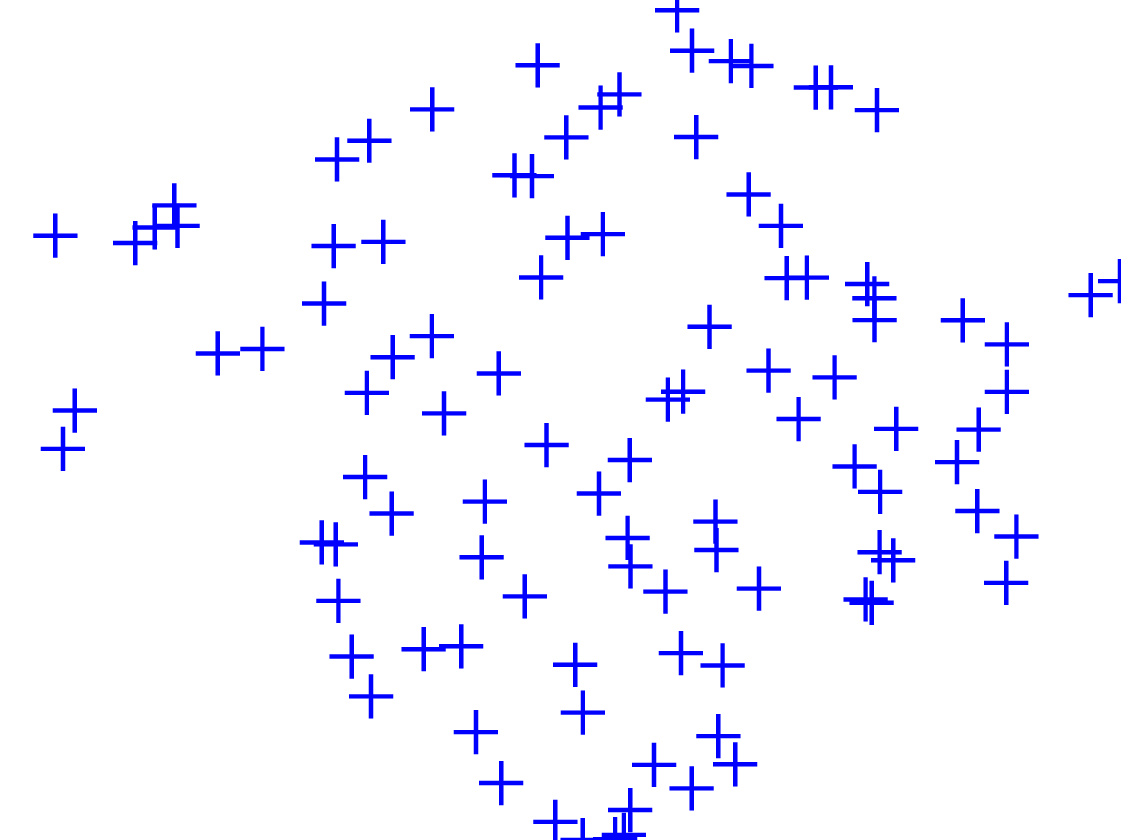}}&	\subfigure[]{		
		  \includegraphics[width=\scale\linewidth]{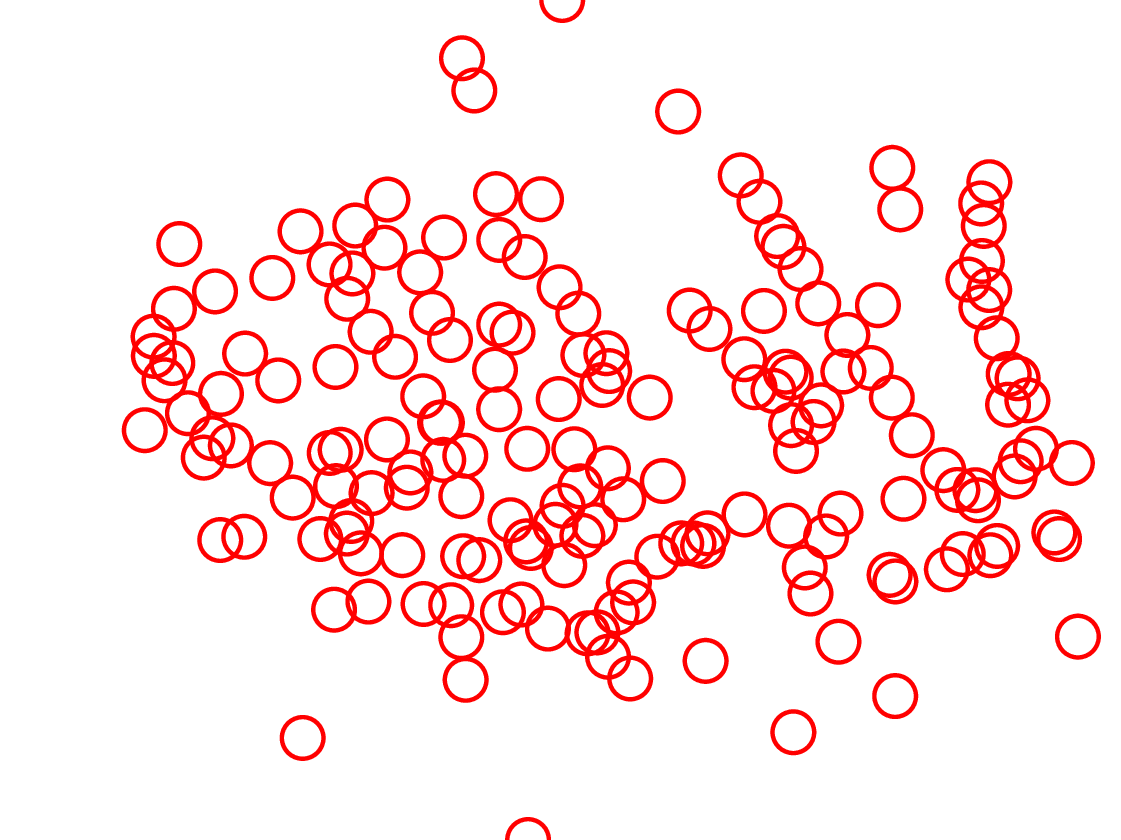}} &
\subfigure[]{	\includegraphics[width=\scale\linewidth]{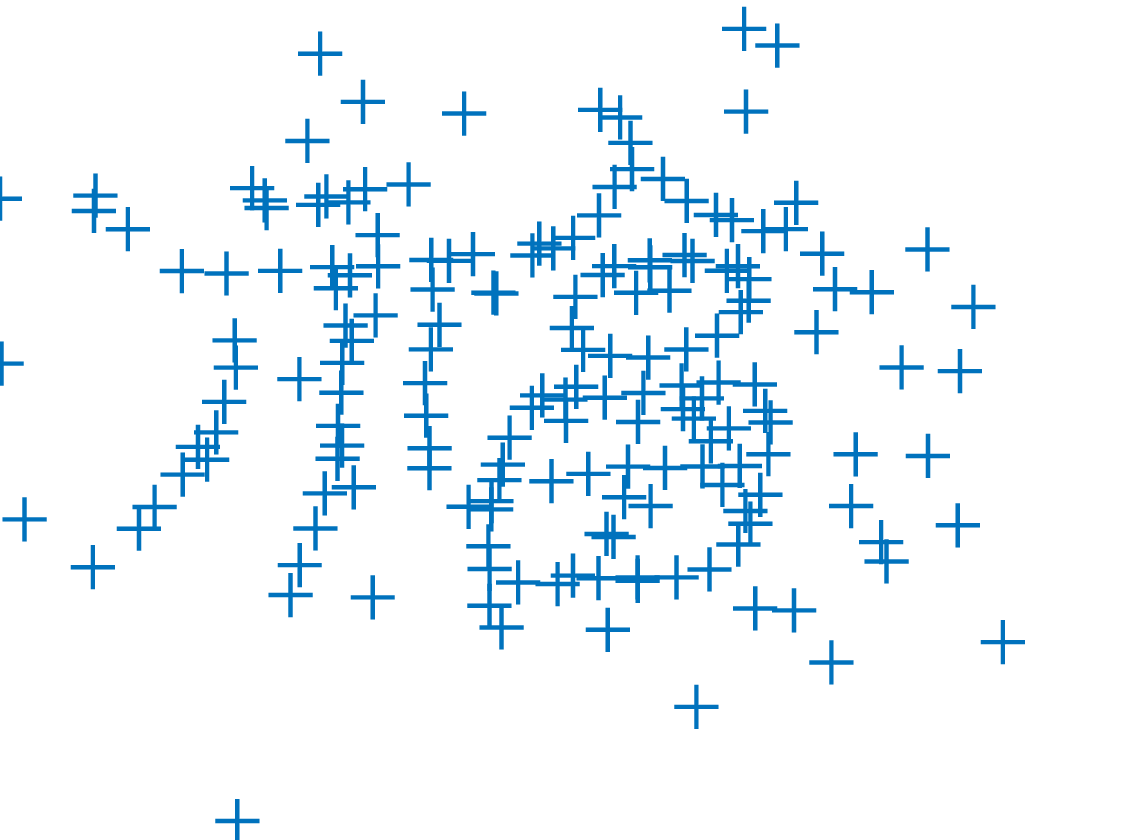}}&	  
		\subfigure[]{ \includegraphics[width=\scale\linewidth]{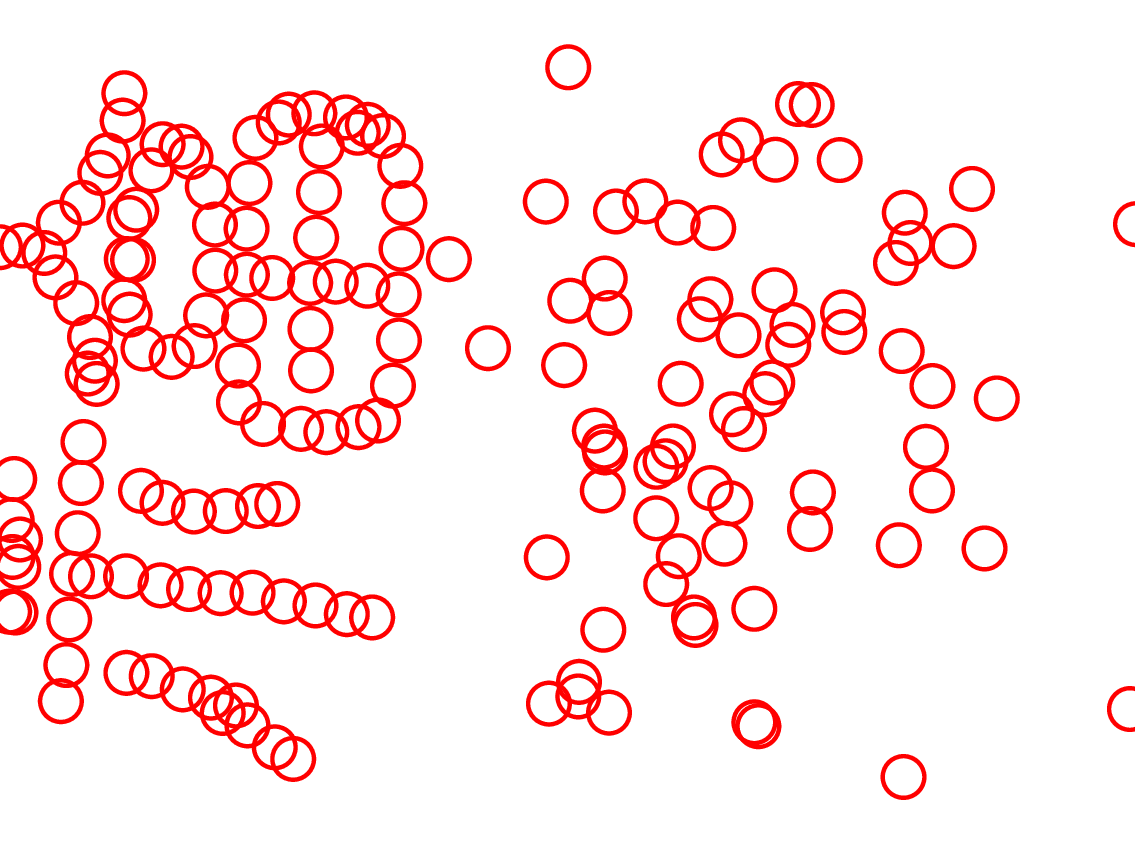}} &
\subfigure[]{ \includegraphics[width=\scale\linewidth]{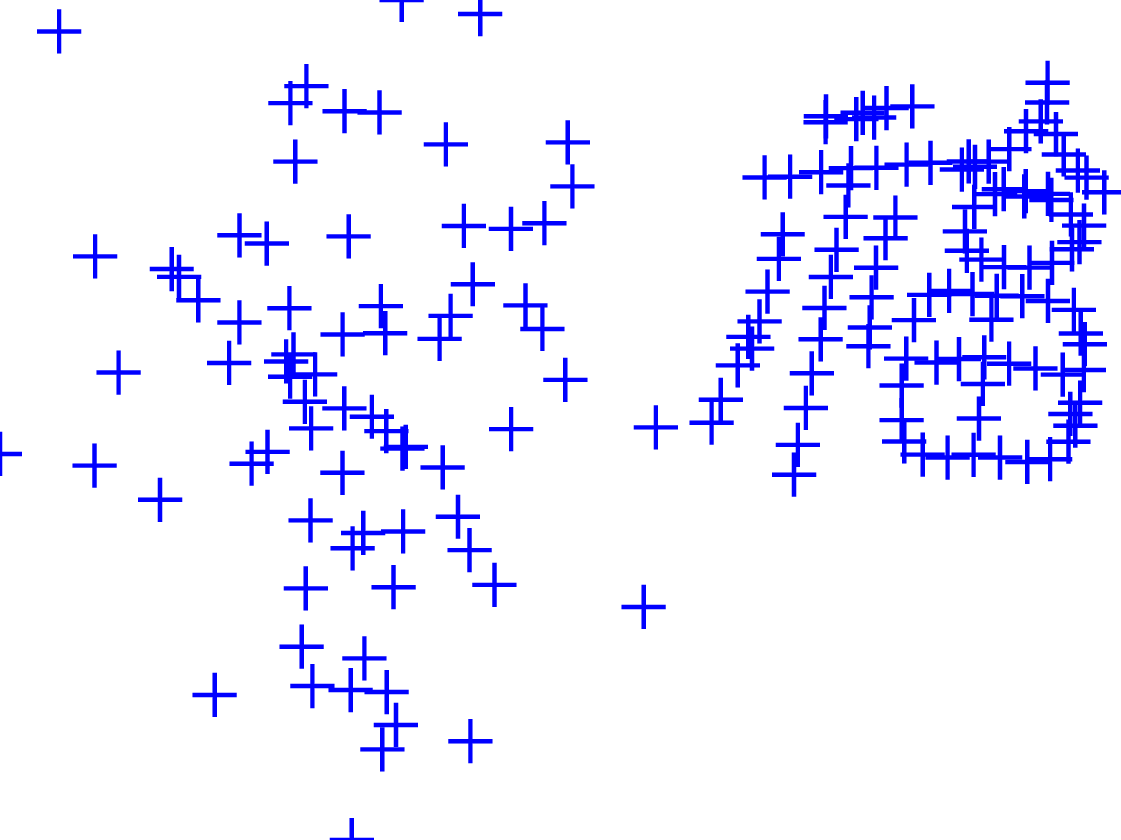}}&
\subfigure[]{ \includegraphics[width=\scale\linewidth]{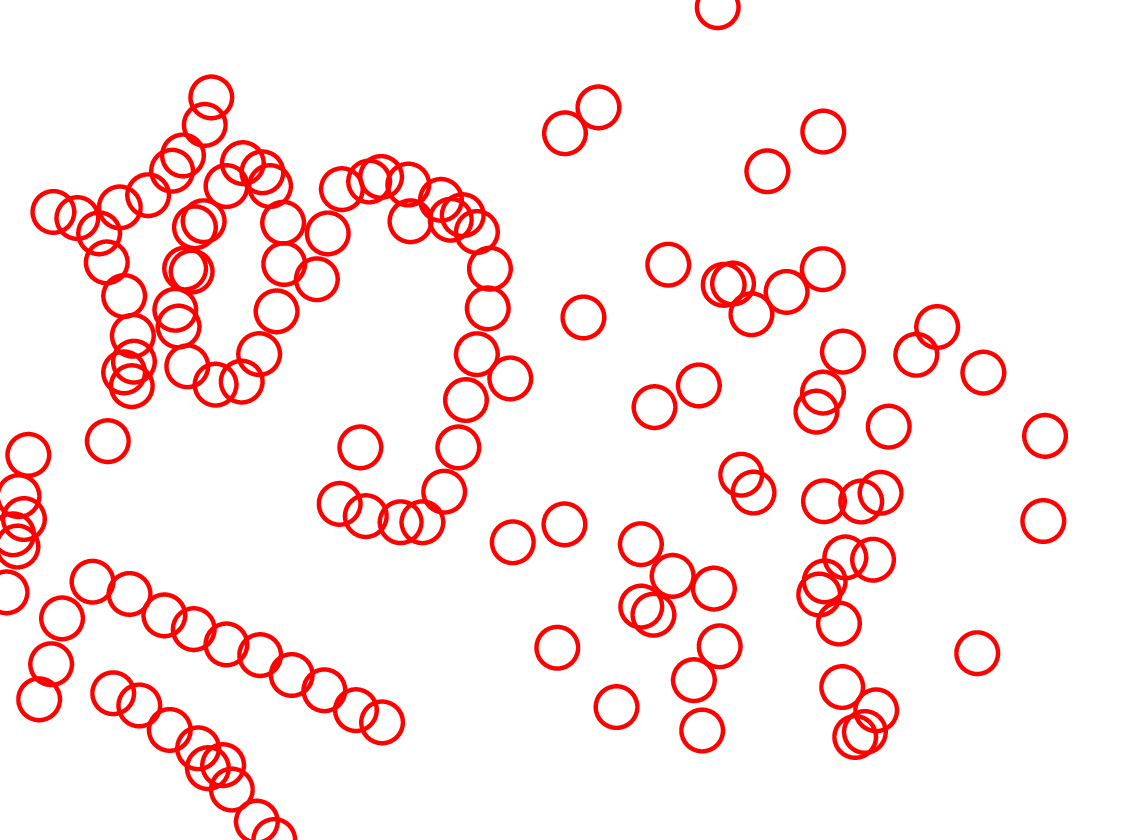}}&
\subfigure[]		{ \includegraphics[width=\scale\linewidth]{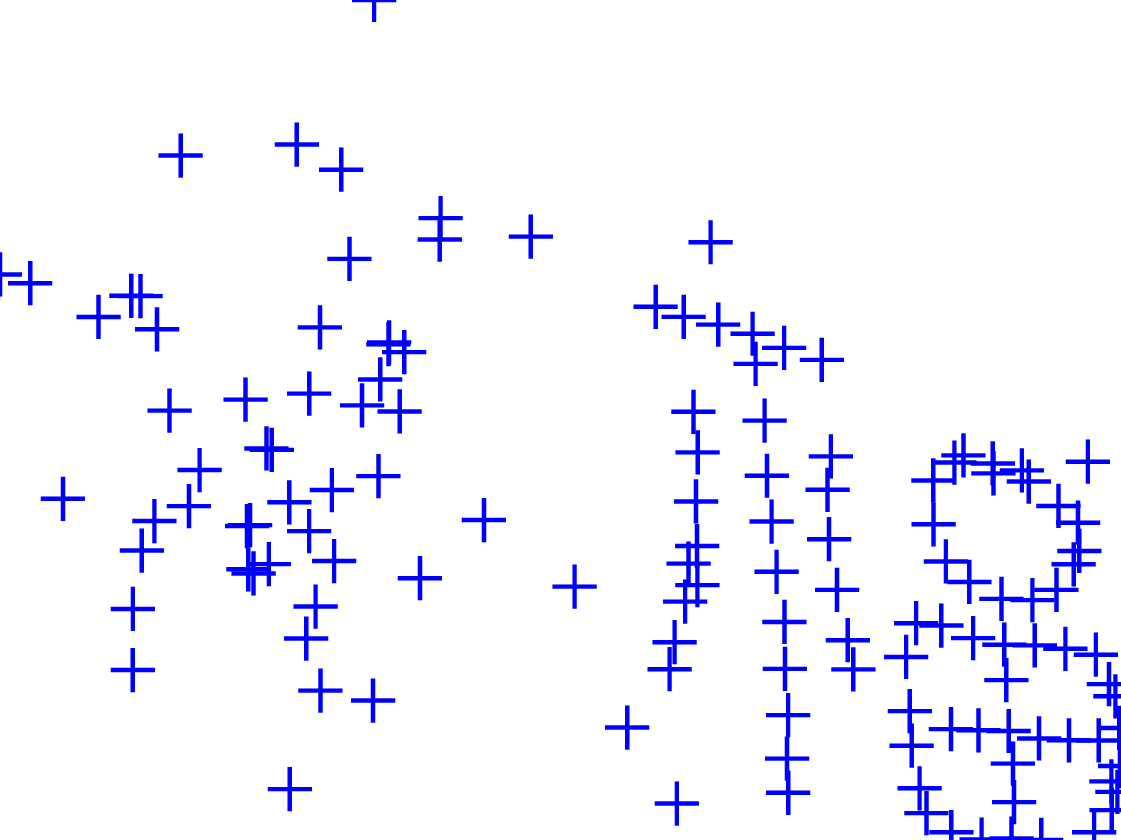}}  
	\end{tabular}
	\caption{
 (a) to (c): model point sets 
and 
examples of scene piont sets  in the deformation and  noise  tests, % (columns 2 to 3), 
respectively.		
		 (d) to (i):
examples of model and scene point sets  in the 
mixed outliers and inliers test ((d), (e)),
separate outliers and inliers test ((f), (g)) and occlusion+outlier  test ((h), (i)), respectively.
Here, the model points are marked by red circles and
the scene points  by blue crosses.
\label{rot_2D_test_data_exa}}
%\end{figure*} 
%\begin{figure*}[t]
	\centering
	\newcommand{\height}{2cm}
	\newcommand\Scale{0.185}	
	\begin{tabular}{@{\hspace{0mm}}c@{\hspace{0mm}}|@{\hspace{0mm}} c@{\hspace{0mm}}|@{\hspace{0mm}} c@{\hspace{0mm}}|@{\hspace{0mm}}  c@{\hspace{0mm} } |c}	
		\includegraphics[height=\height,width=\Scale\linewidth]{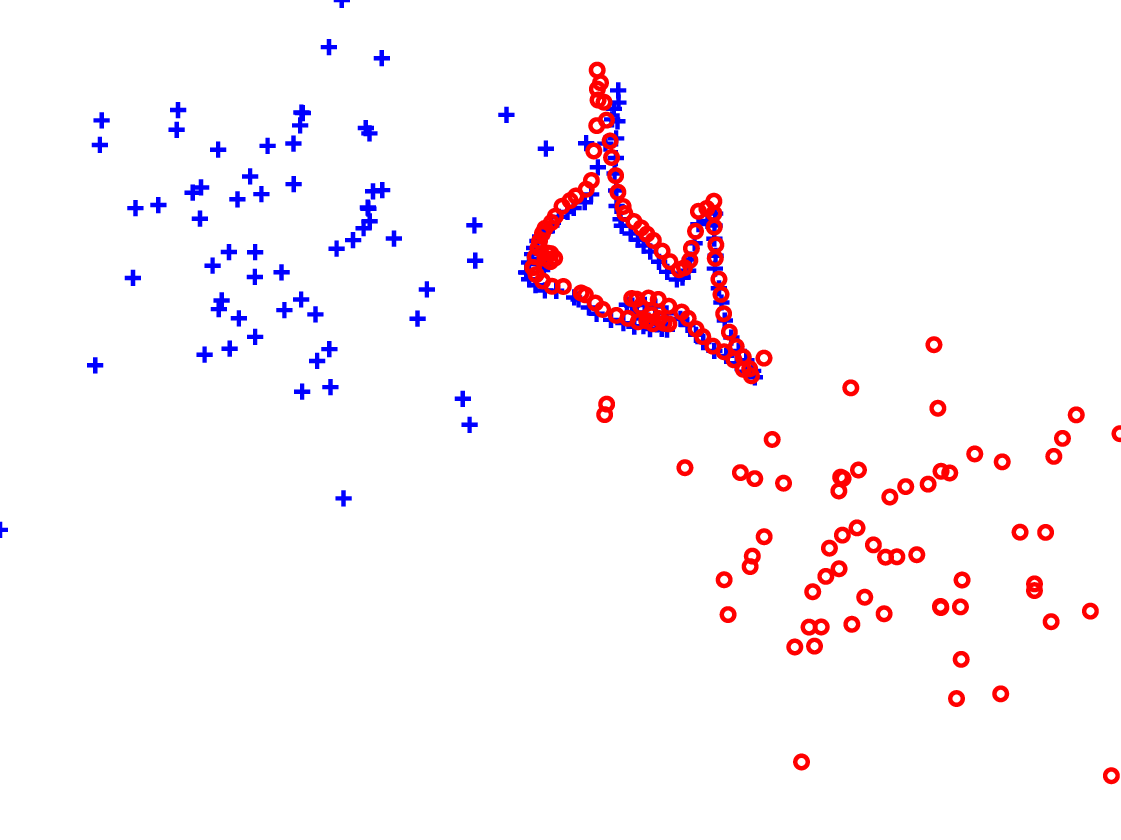} &
		\includegraphics[height=\height,width=\Scale\linewidth]{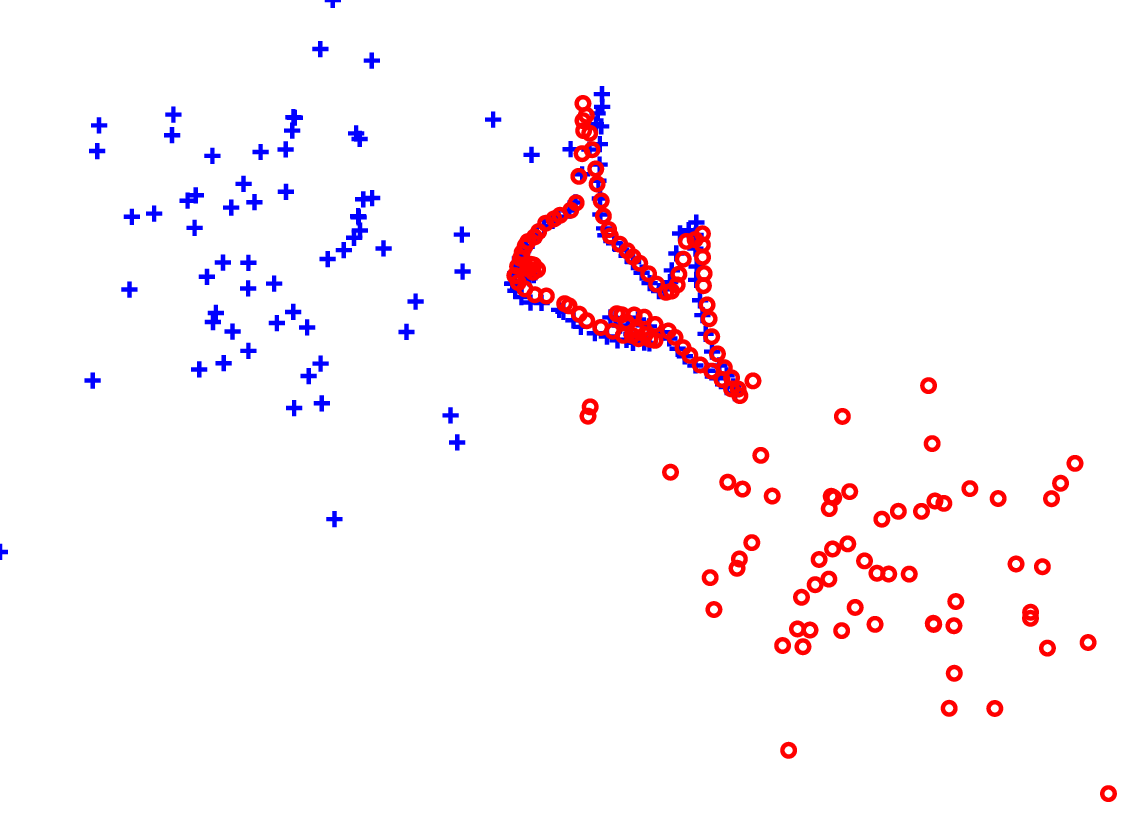} &
		\includegraphics[height=\height,width=\Scale\linewidth]{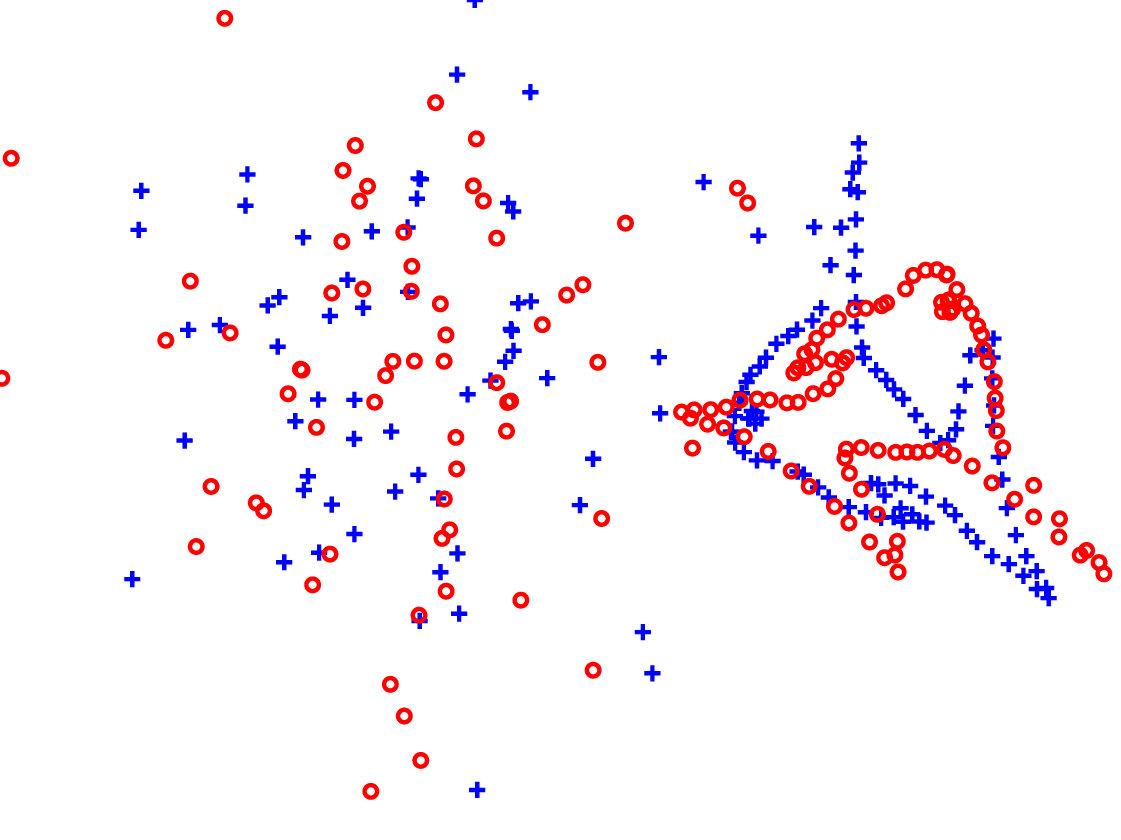}&
		\includegraphics[height=\height,width=\Scale\linewidth]{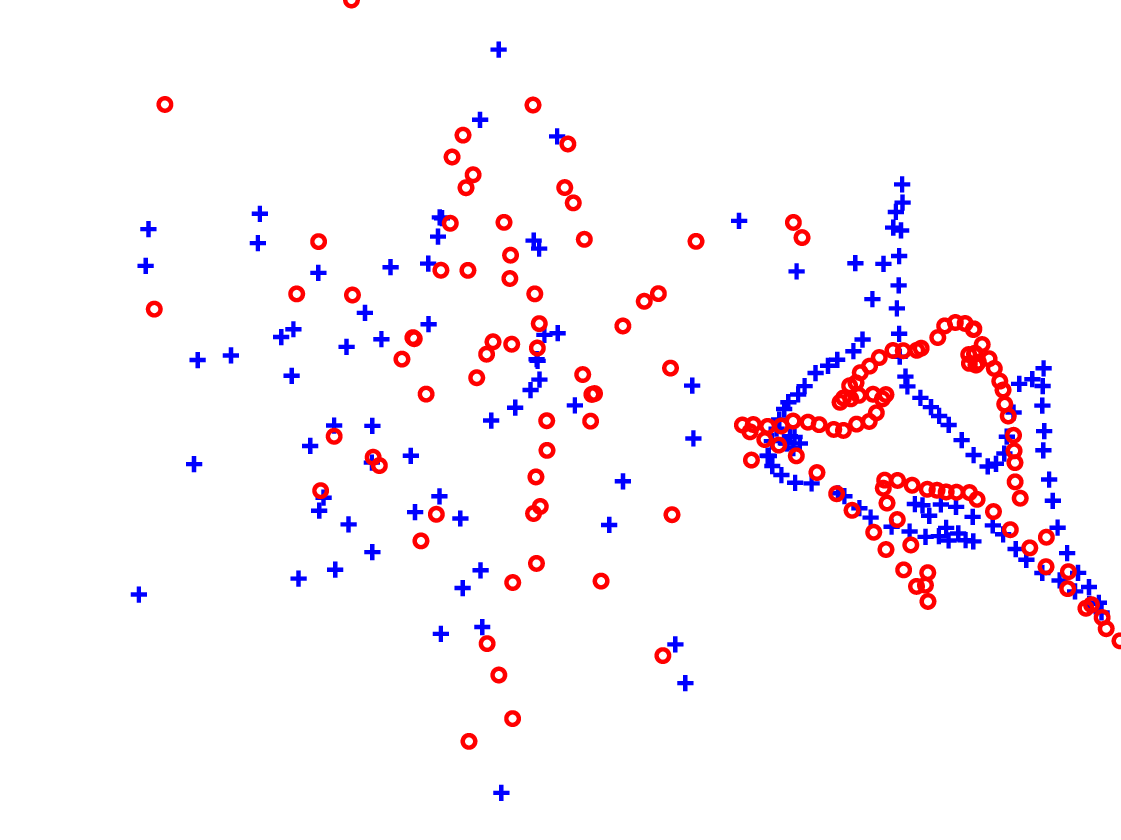} &
		\includegraphics[height=\height,width=\Scale\linewidth]{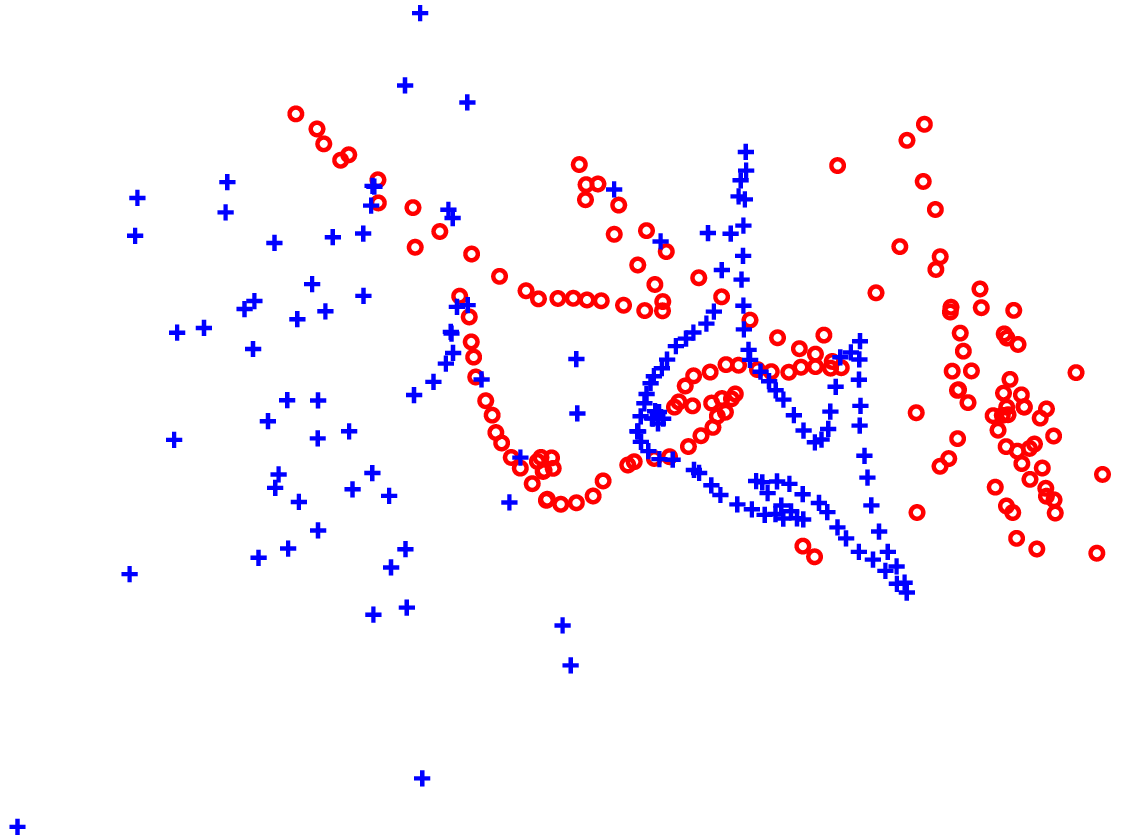}
		\vspace{-2mm}
		\\
		\hline
		%\vspace{-2mm}
		\subfigure[RPM-HTB (simi)]{
			\includegraphics[height=\height,width=\Scale\linewidth]{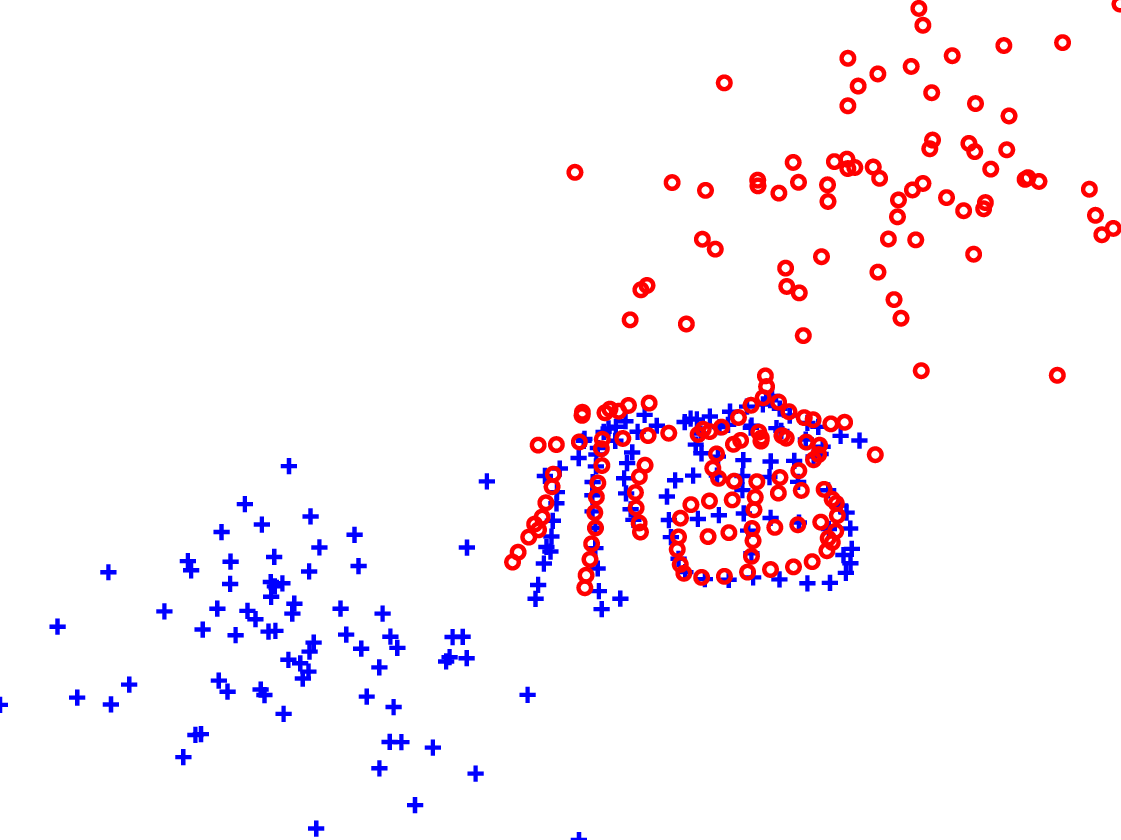} 
		}&\subfigure[RPM-HTB (aff)]{
			\includegraphics[height=\height,width=\Scale\linewidth]{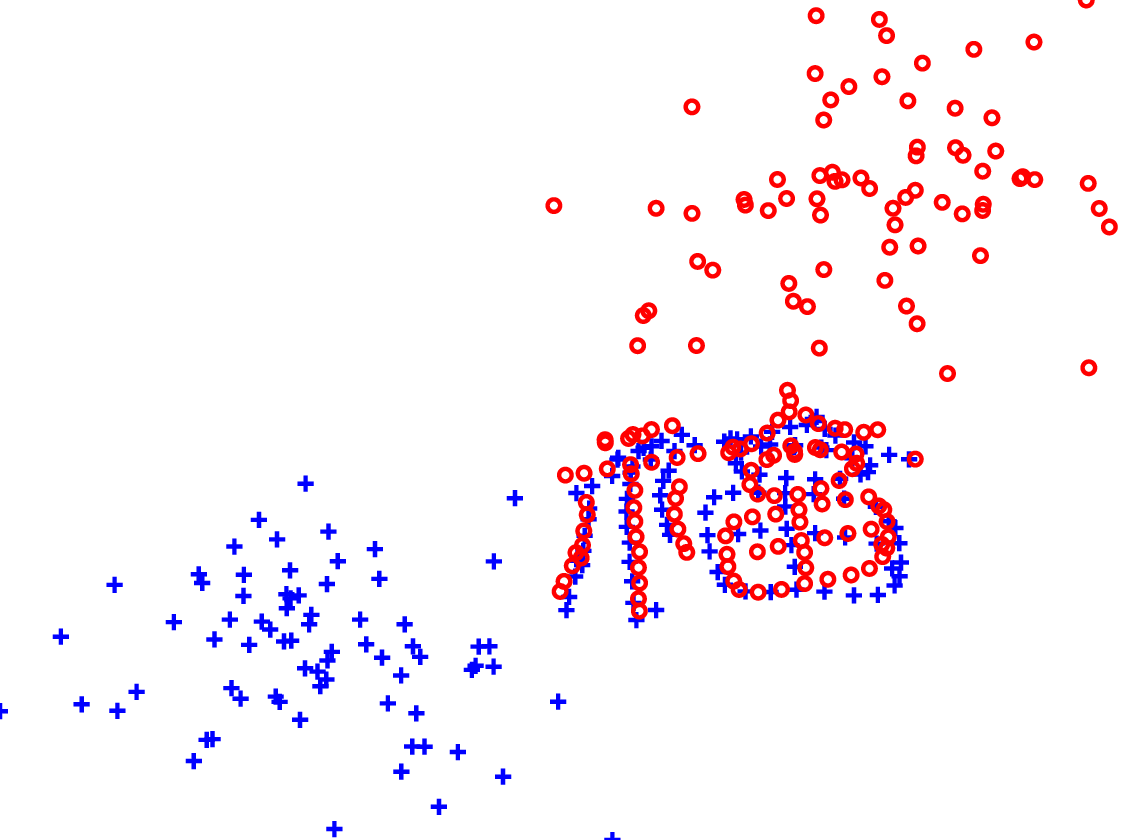}
		} &\subfigure[RPM-PA (simi)]{
			\includegraphics[height=\height,width=\Scale\linewidth]{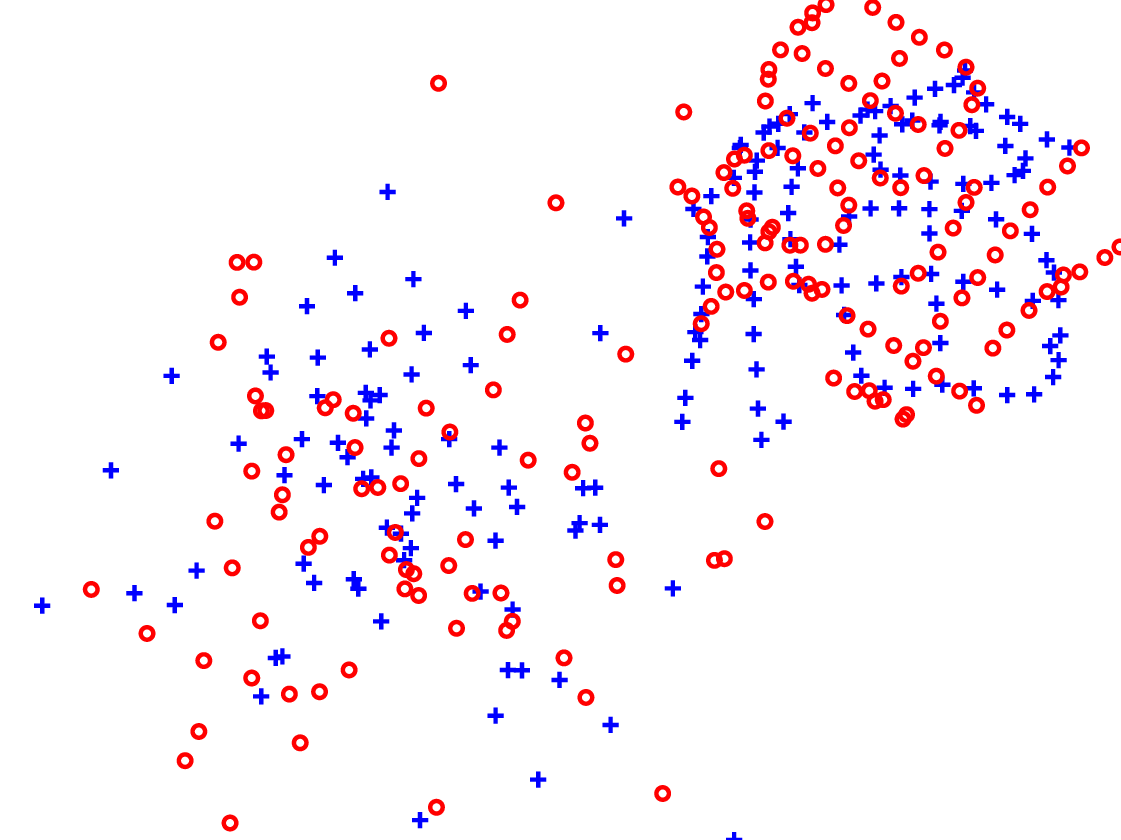}
		}&\subfigure[RPM-CAV (simi)]{
			\includegraphics[height=\height,width=\Scale\linewidth]{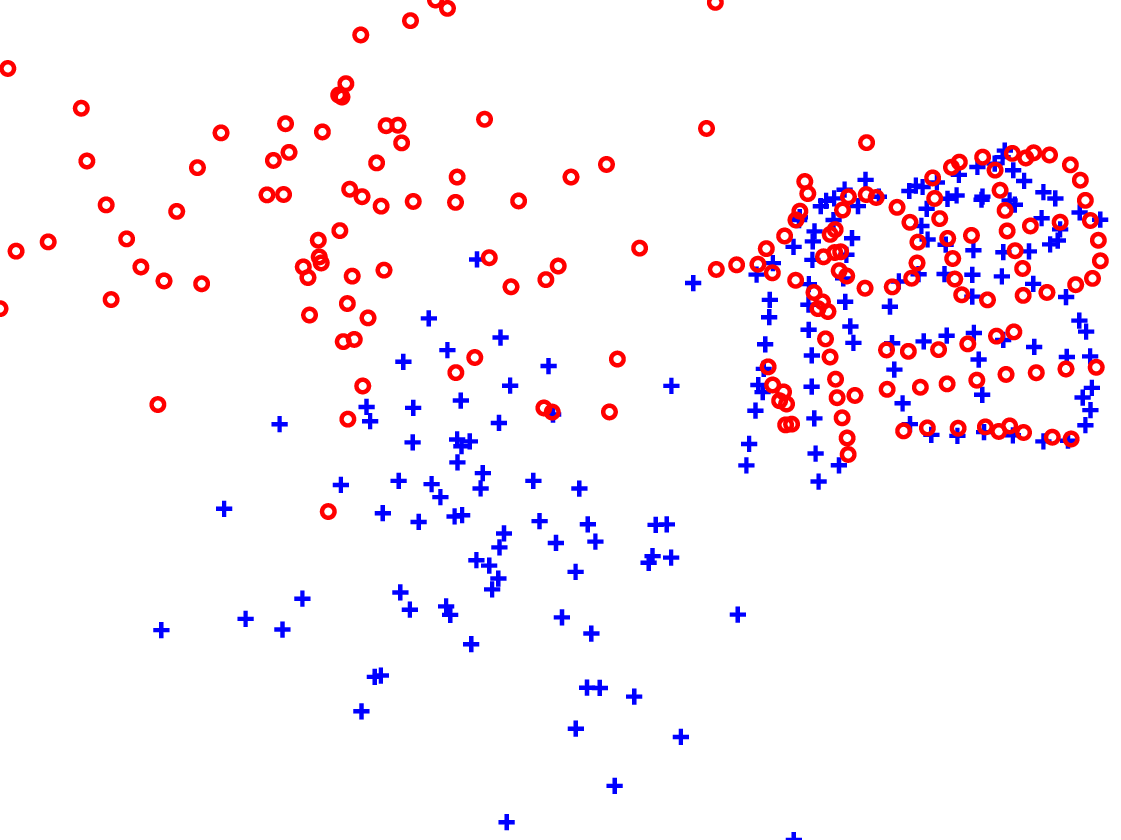} 
		}&
		\subfigure[PR-GLS]{
			\includegraphics[height=\height,width=\Scale\linewidth]{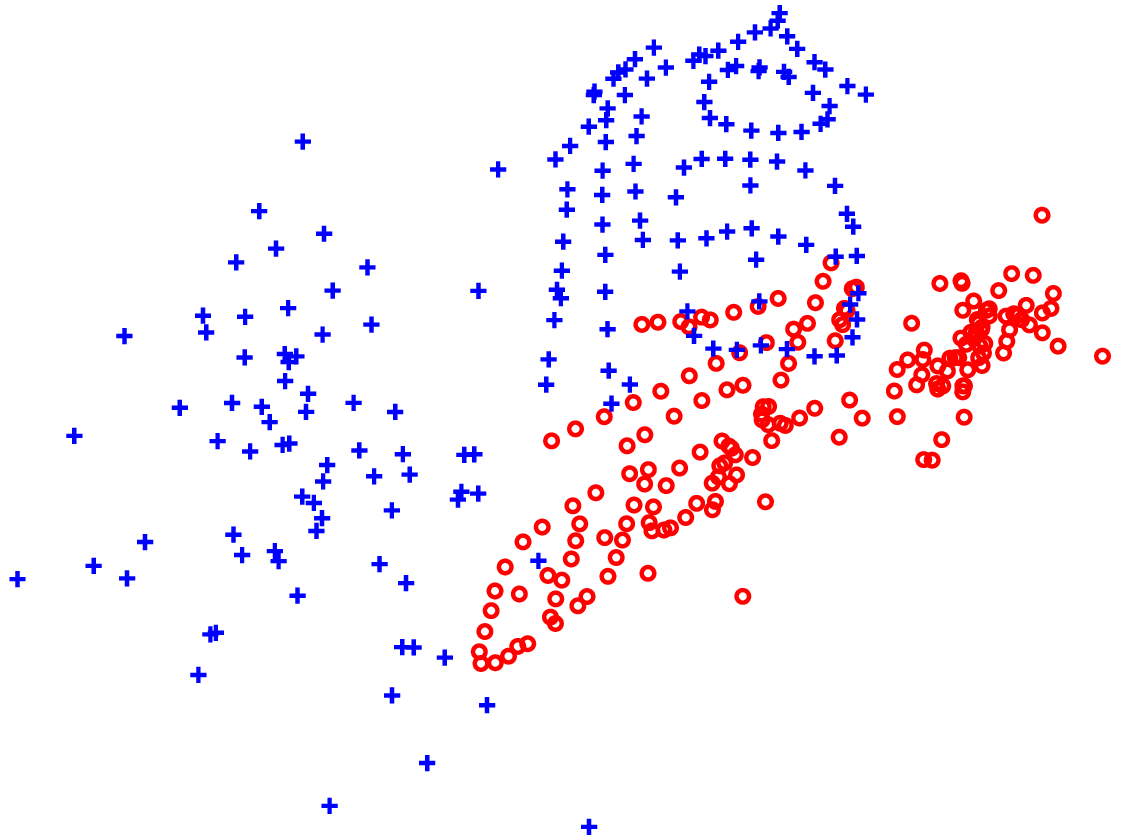} 
		}
		%		\includegraphics[width=\scale\linewidth]{figures/fu_dual_outlier_Go_icp_simi_exa}
		%		\\\hline
		%\includegraphics[width=\scale\linewidth]{figures/fish_dual_occlude_RPM_tri_simi_exa} &
		%\includegraphics[width=\scale\linewidth]{figures/fish_dual_occlude_RPM_tri_aff_exa} &
		%\includegraphics[width=\scale\linewidth]{figures/fish_dual_occlude_RPM_inner_simi_exa}&
		%\includegraphics[width=\scale\linewidth]{figures/fish_dual_occlude_RPM_cav_simi_exa} 
		%\includegraphics[width=\scale\linewidth]{figures/fish_dual_occlude_Go_icp_simi_exa}
		%\\\hline
		%		\subfigure[RPM-HTB (simi)]{\includegraphics[width=\scale\linewidth]{figures/fu_dual_occlude_RPM_tri_simi_exa}} &
		%		\subfigure[RPM-HTB (aff)]{\includegraphics[width=\scale\linewidth]{figures/fu_dual_occlude_RPM_tri_aff_exa}} &
		%\subfigure[RPM-PA]{\includegraphics[width=\scale\linewidth]{figures/fu_dual_occlude_RPM_inner_simi_exa}}&		
		%		\subfigure[RPM-CAV]{\includegraphics[width=\scale\linewidth]{figures/fu_dual_occlude_RPM_cav_simi_exa}}
		%		\subfigure[Go-ICP]{\includegraphics[width=\scale\linewidth]{figures/fu_dual_occlude_Go_icp_simi_exa}}
	\end{tabular}
	\caption{
		Example of registration results  by different methods		
		%		 RPM-HTB using similarity or affine transformation, RPM-PA \cite{lian2021polyhedral} and RPM-CAV \cite{RPM_model_occlude_PR} 
		in the  separate outliers and inliers test,
		where $n_p$ is chosen as ground truth for all the methods.
		%		and  occlusion+outlier (bottom 2 rows) tests.
		\label{rot_2D_syn_match_exa}}
\end{figure*}

\section{Experiments
	\label{sec:exp}
}

We implement the proposed   algorithm (denoted as  RPM-HTB) under   Matlab R2020b
and compare it with other methods  on a computer with quad-core $3.5$ GHz CPUs.
% and $64$G RAM.
For the competing methods only outputting point correspondence,
the generated point correspondence is used to compute the best affine transformation between  two point sets.
The matching  error is defined   as  the root-mean-square 
%(RMS)
 distance
%mean of the Euclidean distances 
between the  transformed model inliers and  their  corresponding scene inliers.
%
%Finally, we compare with a heuristic variant of RPM-HTB where   $\epsilon_0=0$ and the maximum  branching depth is set as $10$ instead of infinity.
%For RPM-HTB,
The maximum branching depth of RPM-HTB is chosen as $10$.

%\subsection{2D case}
\subsection{Case one: spatial transformation is linear w.r.t. its parameters
	\label{subsec:exp_case_one}
}

%Since 3D affine transformation has too many parameters,
%causing our method to run too slowly,
%thus, we only test our method for the 2D case.

%\subsubsection{2D synthesized datasets 
%	\label{subsec:2d_syn_test}}

\begin{figure*} [th]
	\centering
	\newcommand\scaleGd{0.203}
	\begin{tabular}{@{\hspace{0mm}}c@{\hspace{0mm}}c@{}c@{} c@{} c @{} }
		\includegraphics[width=\scaleGd\linewidth]{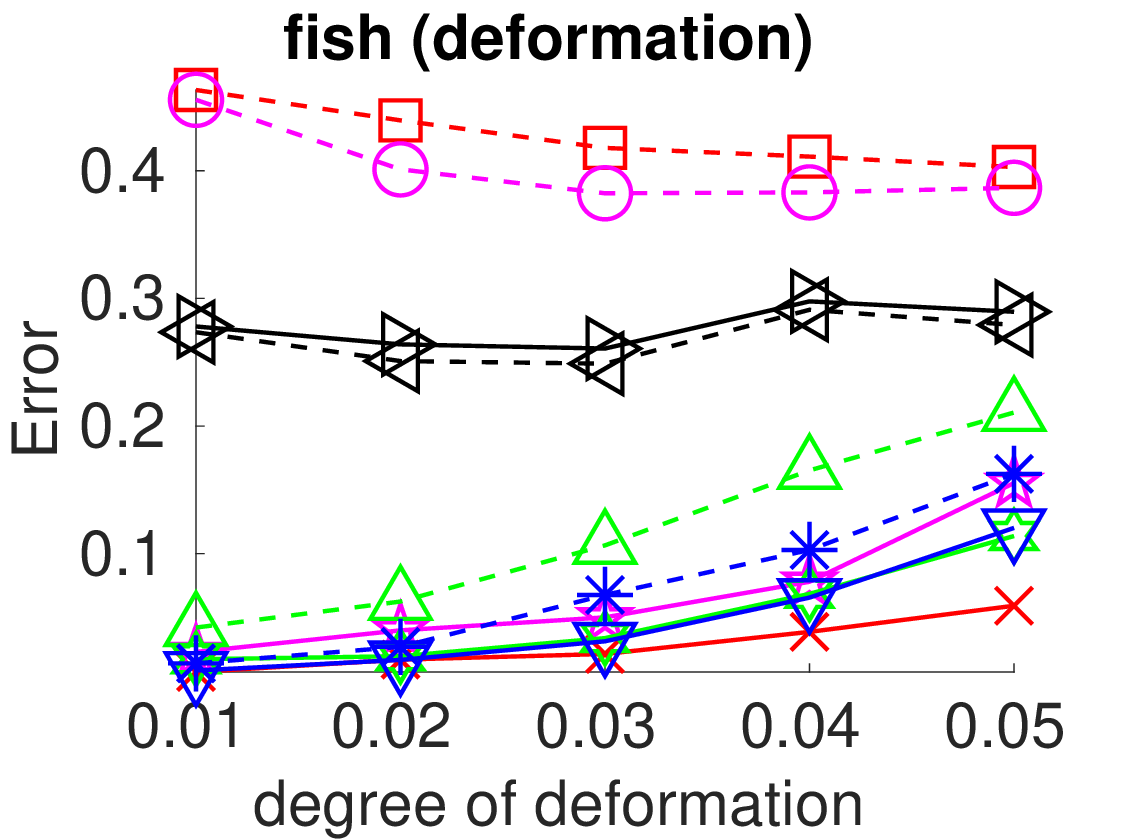}&
		\includegraphics[width=\scaleGd\linewidth]{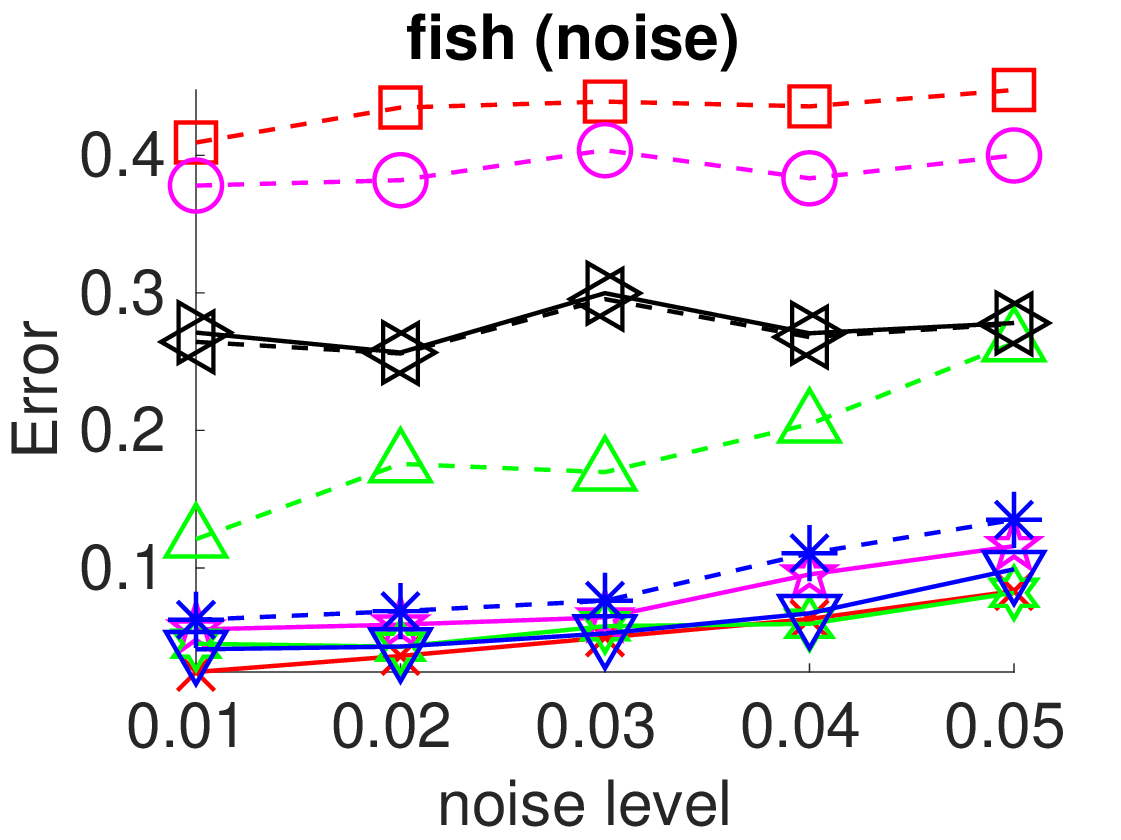}&
		\includegraphics[width=\scaleGd\linewidth]{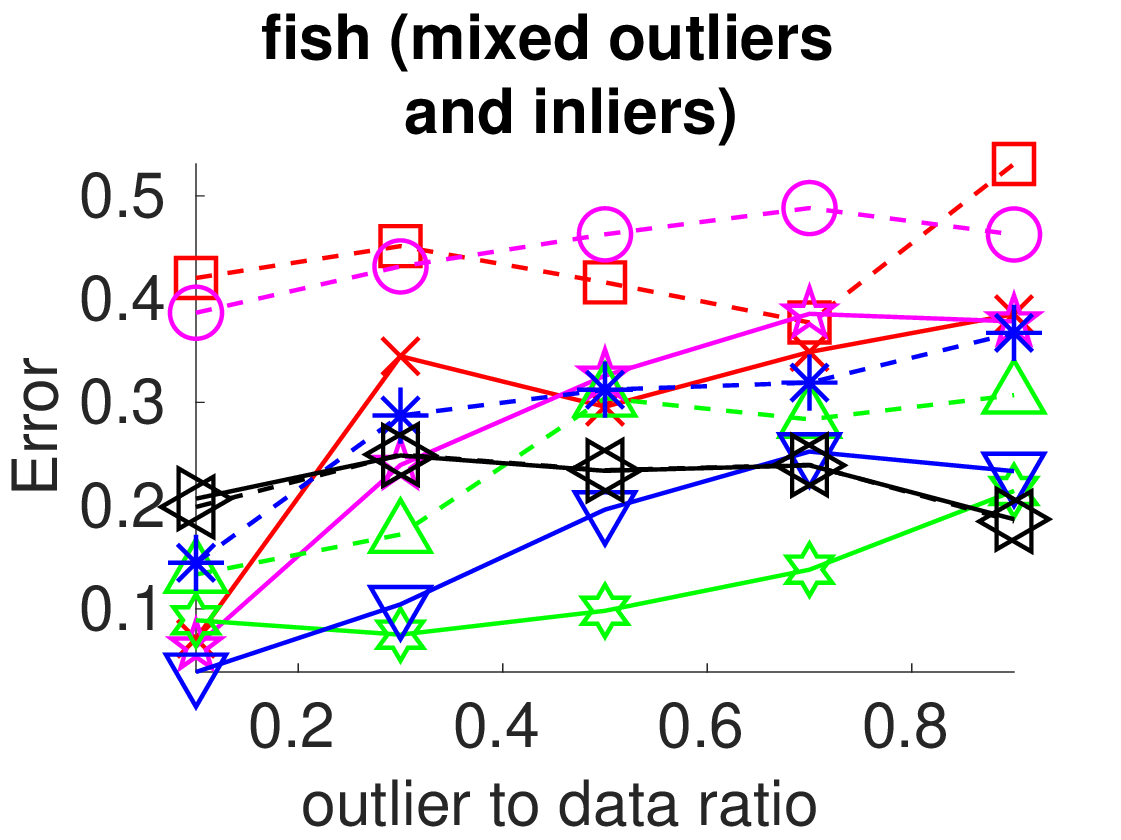}&			
		\includegraphics[width=\scaleGd\linewidth]{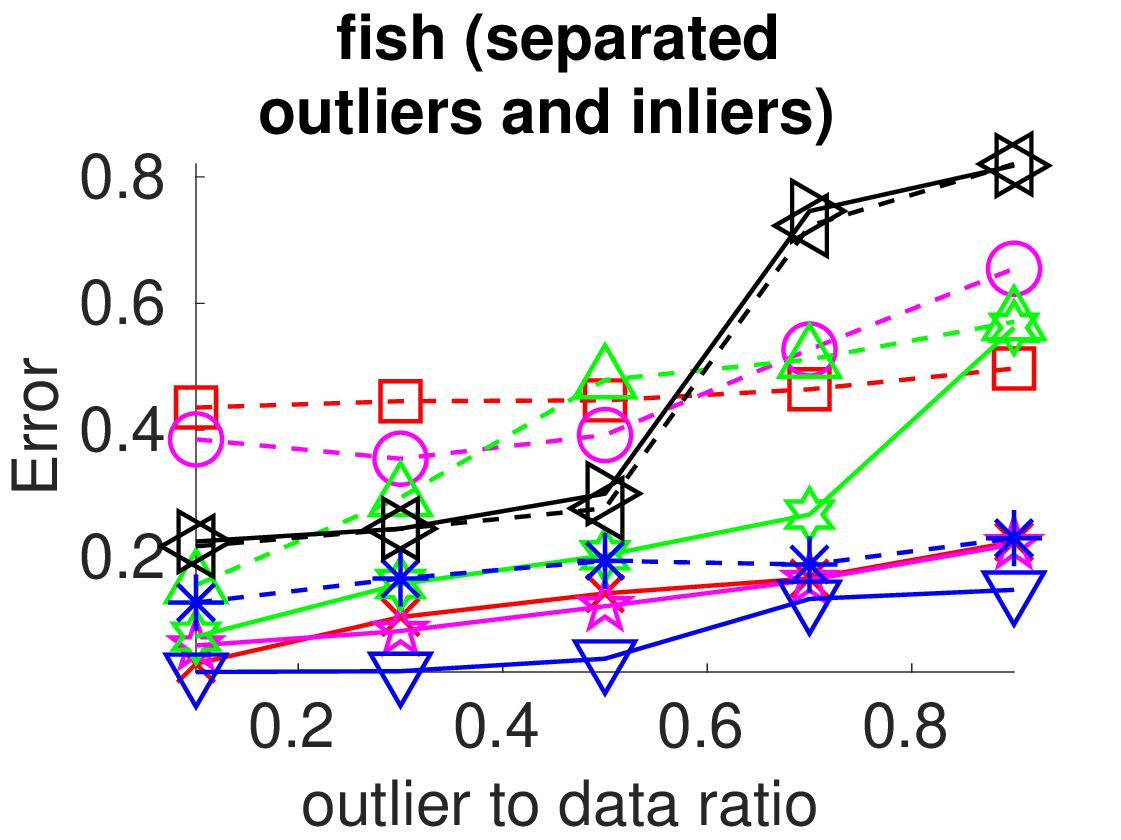}&
		\includegraphics[width=\scaleGd\linewidth]{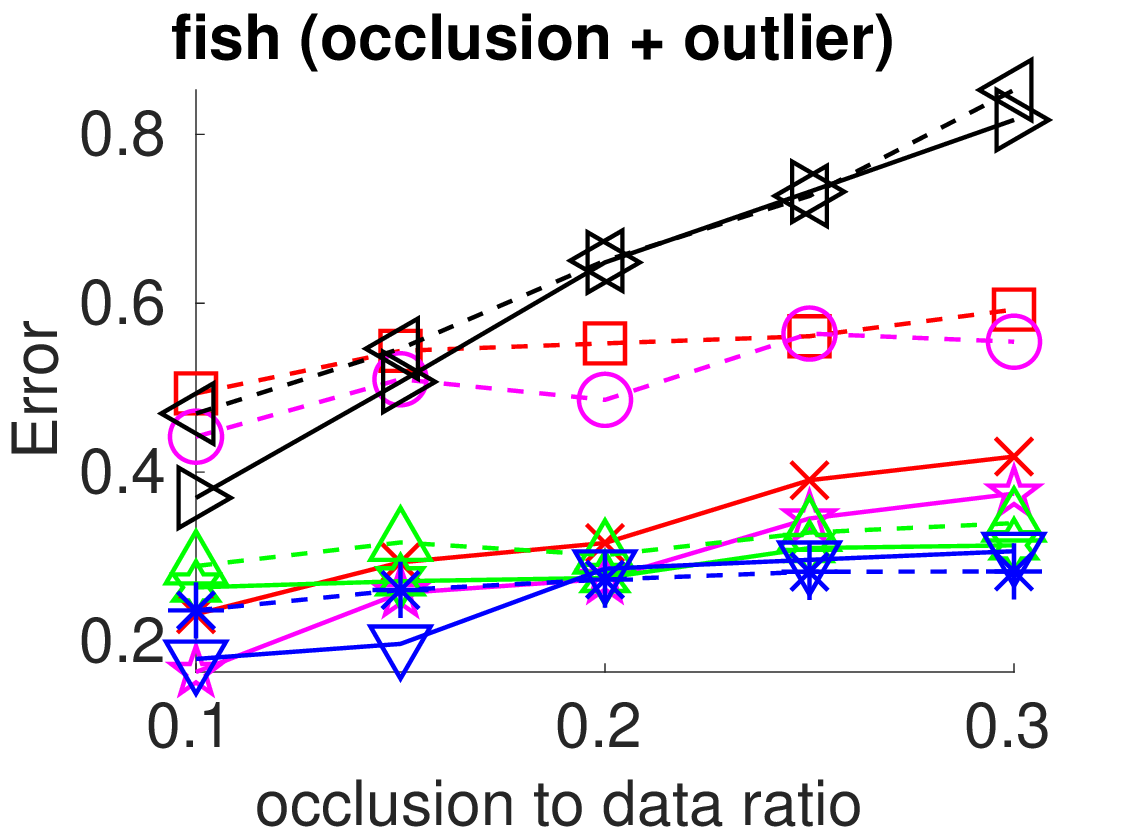}\\
		\includegraphics[width=\scaleGd\linewidth]{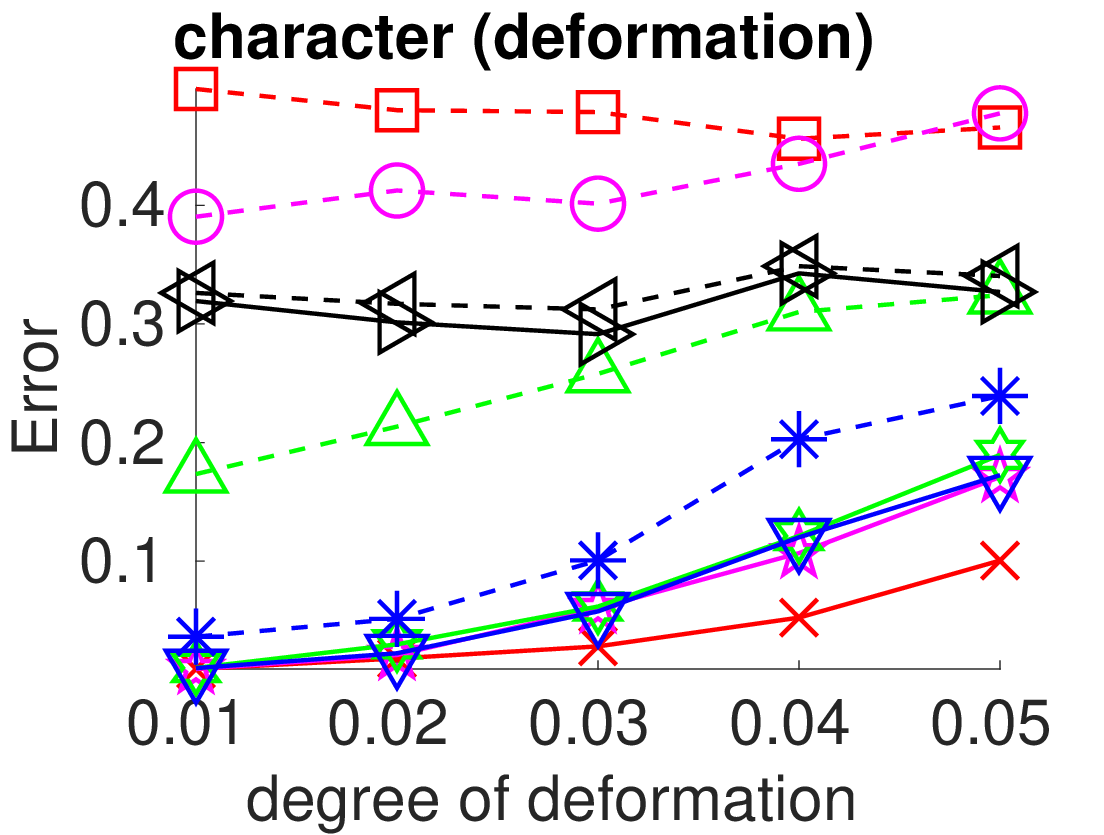}&	
		\includegraphics[width=\scaleGd\linewidth]{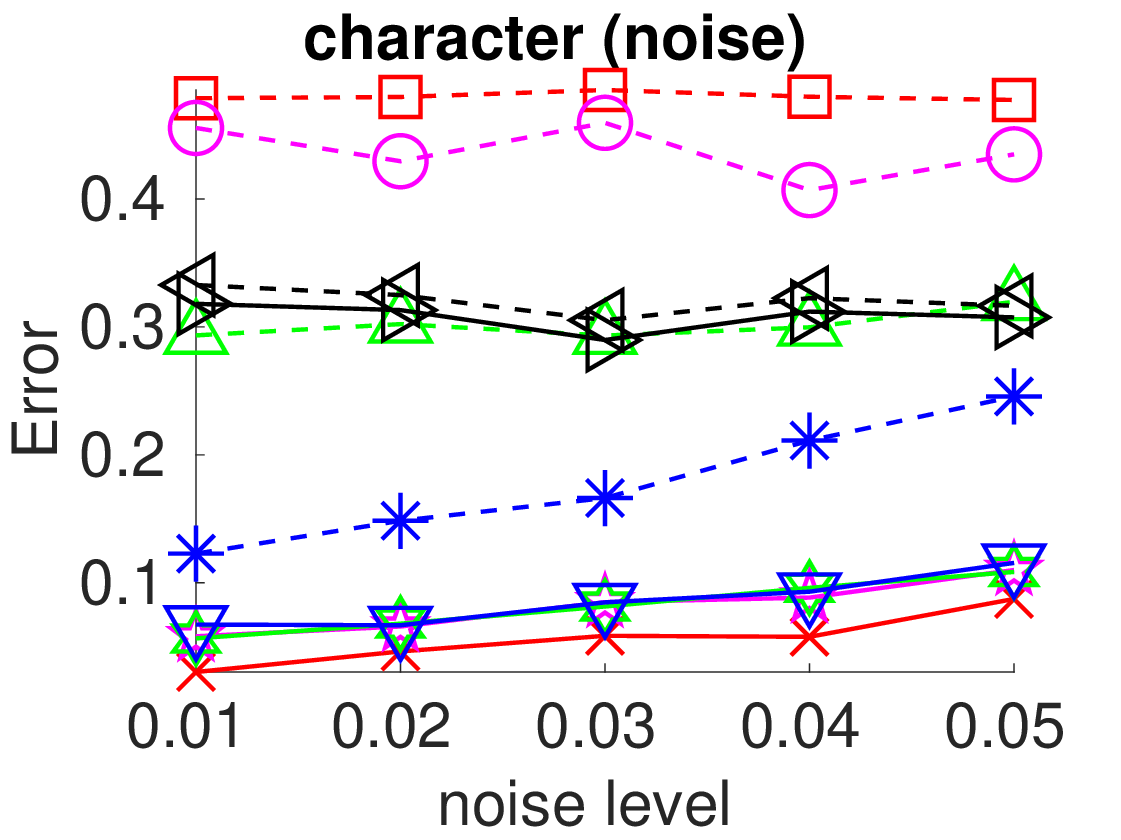}&		
		\includegraphics[width=\scaleGd\linewidth]{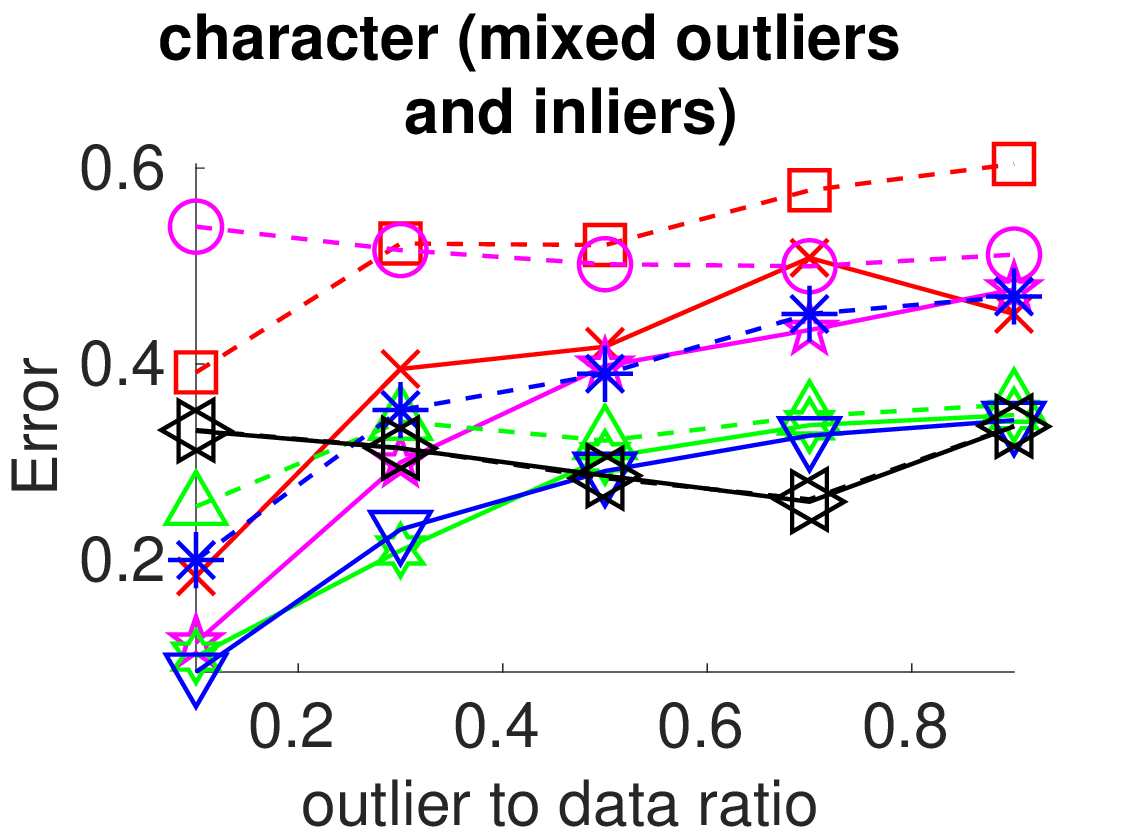}&
	\includegraphics[width=\scaleGd\linewidth]{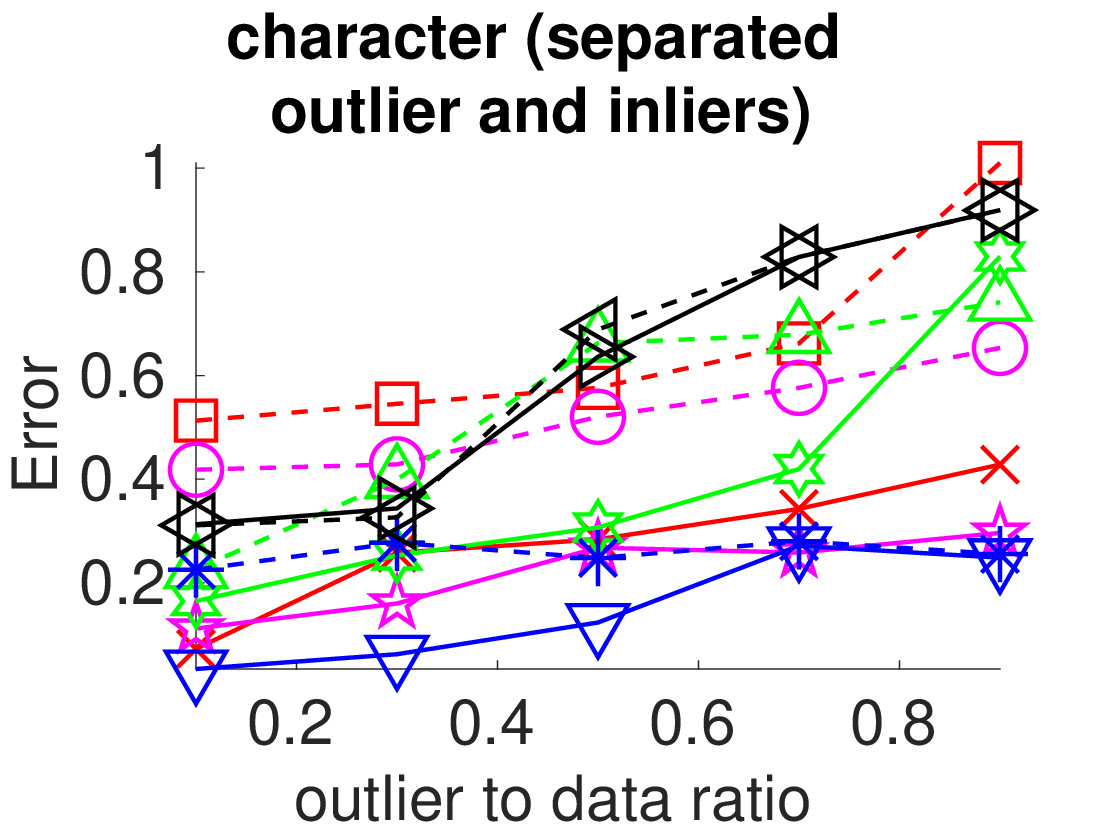}&		
		\includegraphics[width=\scaleGd\linewidth]{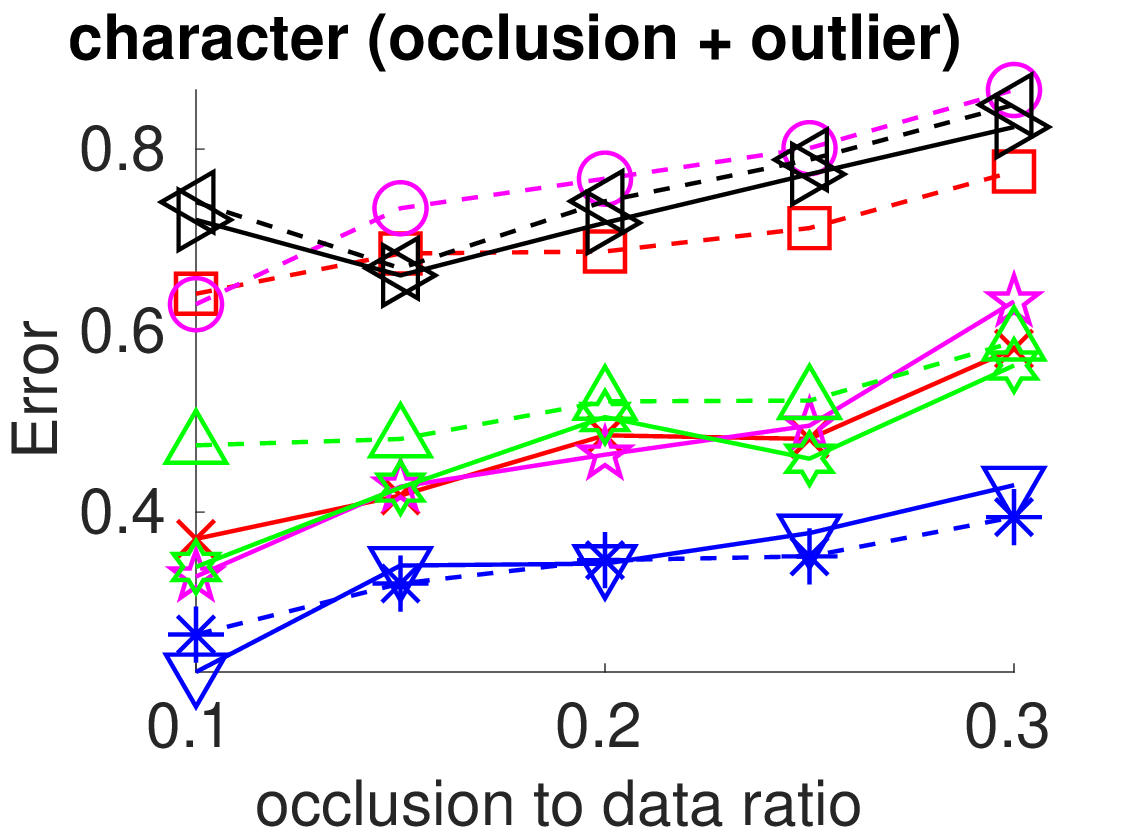}
	\end{tabular}
	\begin{tabular}{@{\hspace{-1.5mm}}c}	\includegraphics[width=.7\linewidth]{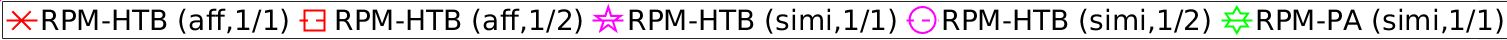}\\
	\includegraphics[width=.6\linewidth]{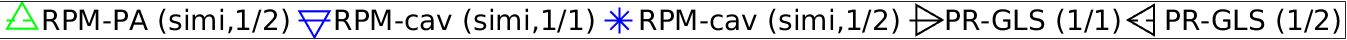}
	\end{tabular}	
\begin{tabular}{@{\hspace{-0mm}}c@{\hspace{0mm}}c@{}c@{}c@{} c }
	\includegraphics[width=\scaleGd\linewidth]{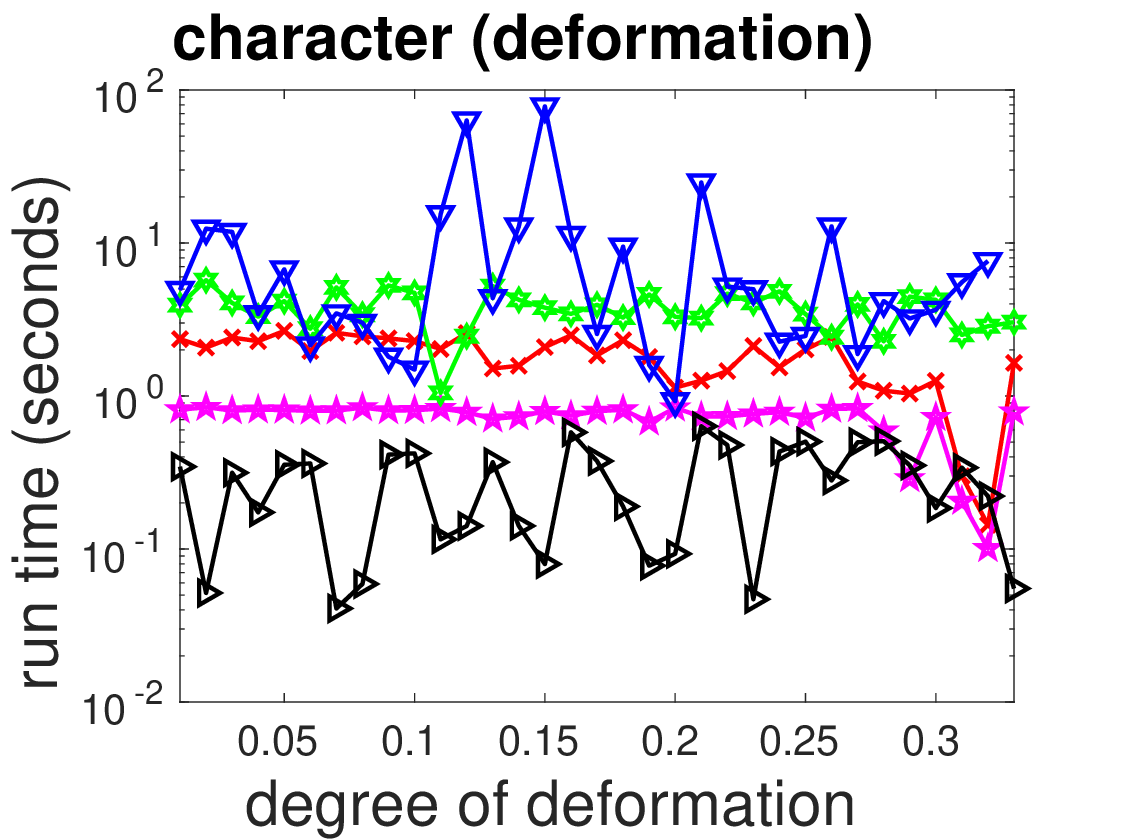}&
	\includegraphics[width=\scaleGd\linewidth]{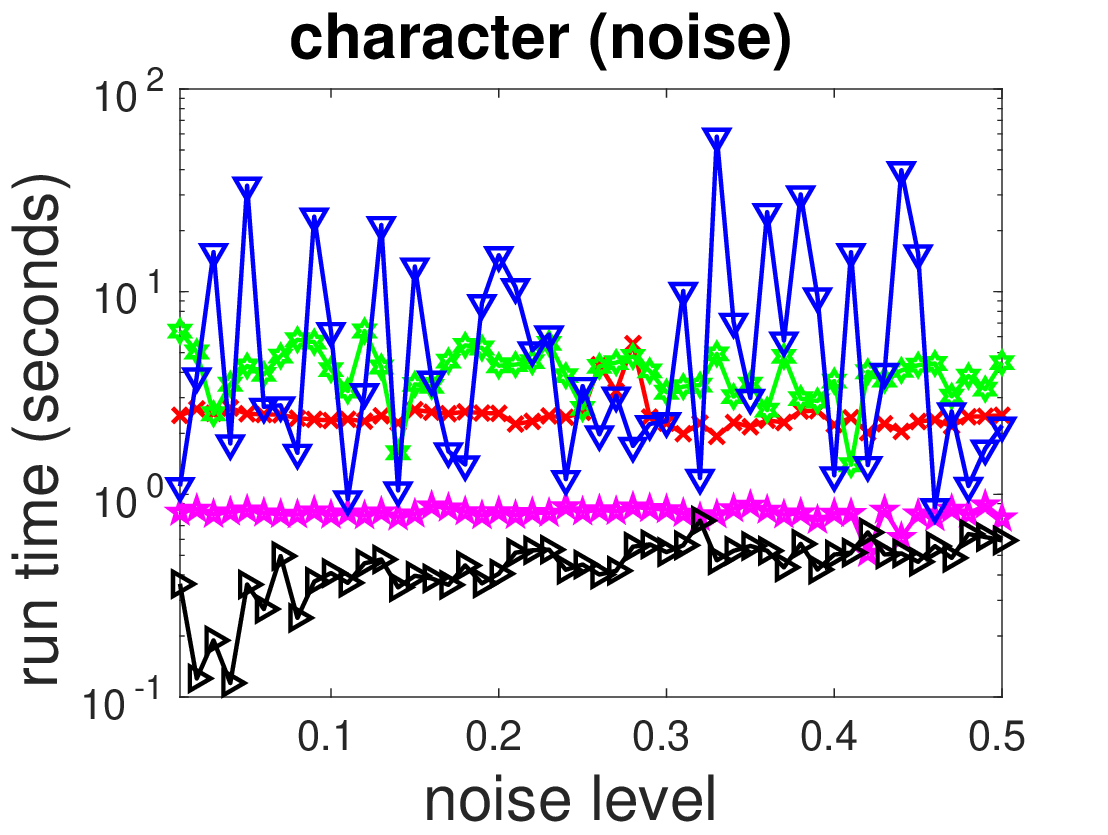}&
	\includegraphics[width=\scaleGd\linewidth]{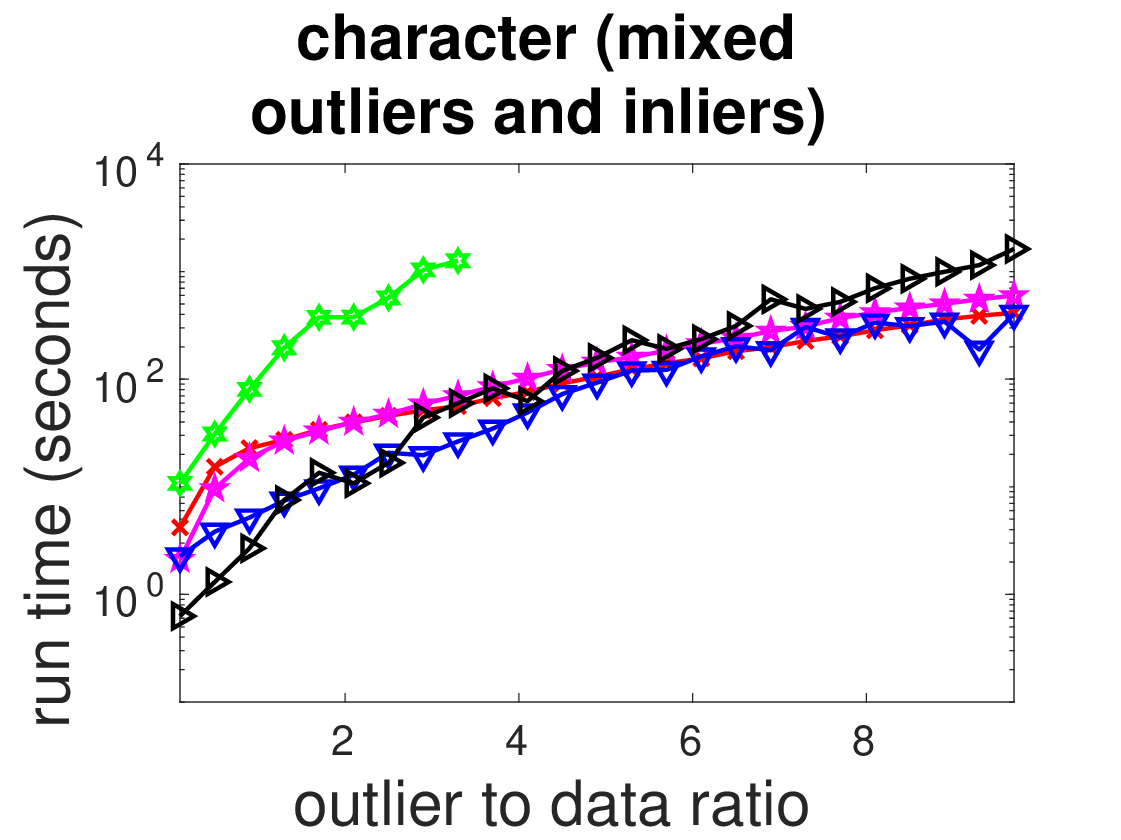}&
	\includegraphics[width=\scaleGd\linewidth]{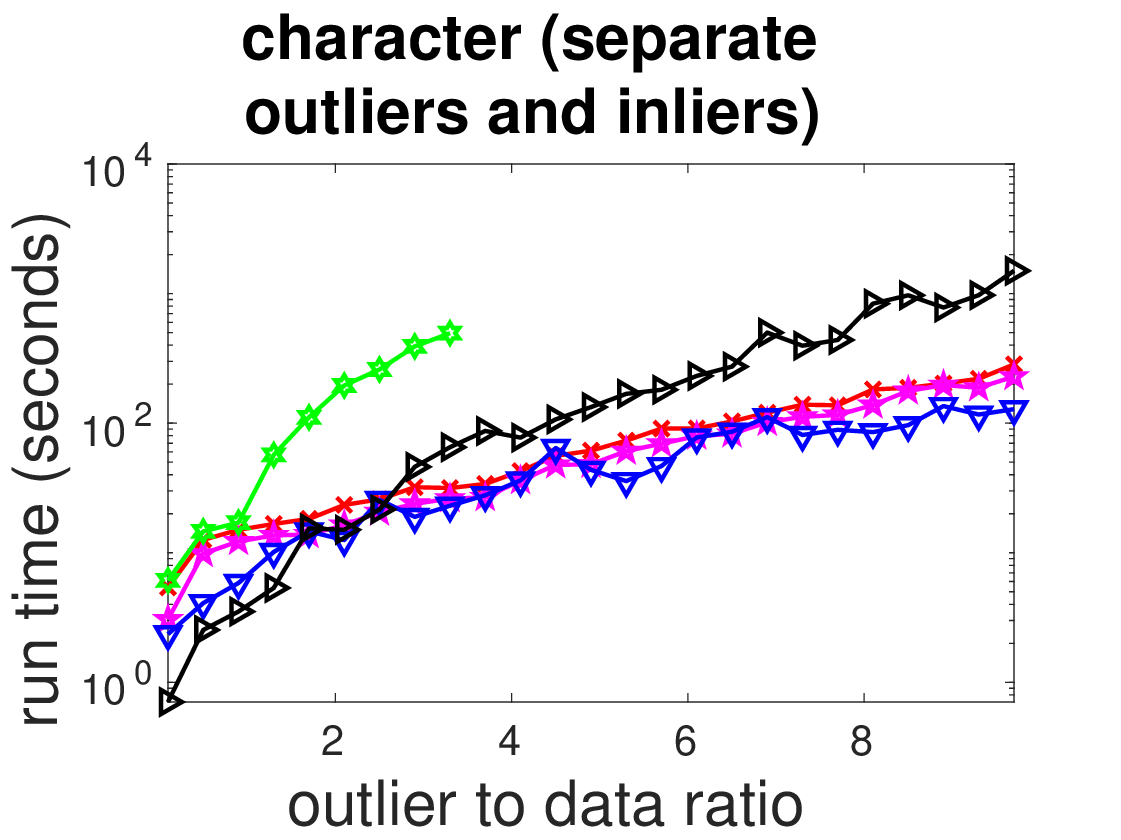}	&
	\includegraphics[width=\scaleGd\linewidth]{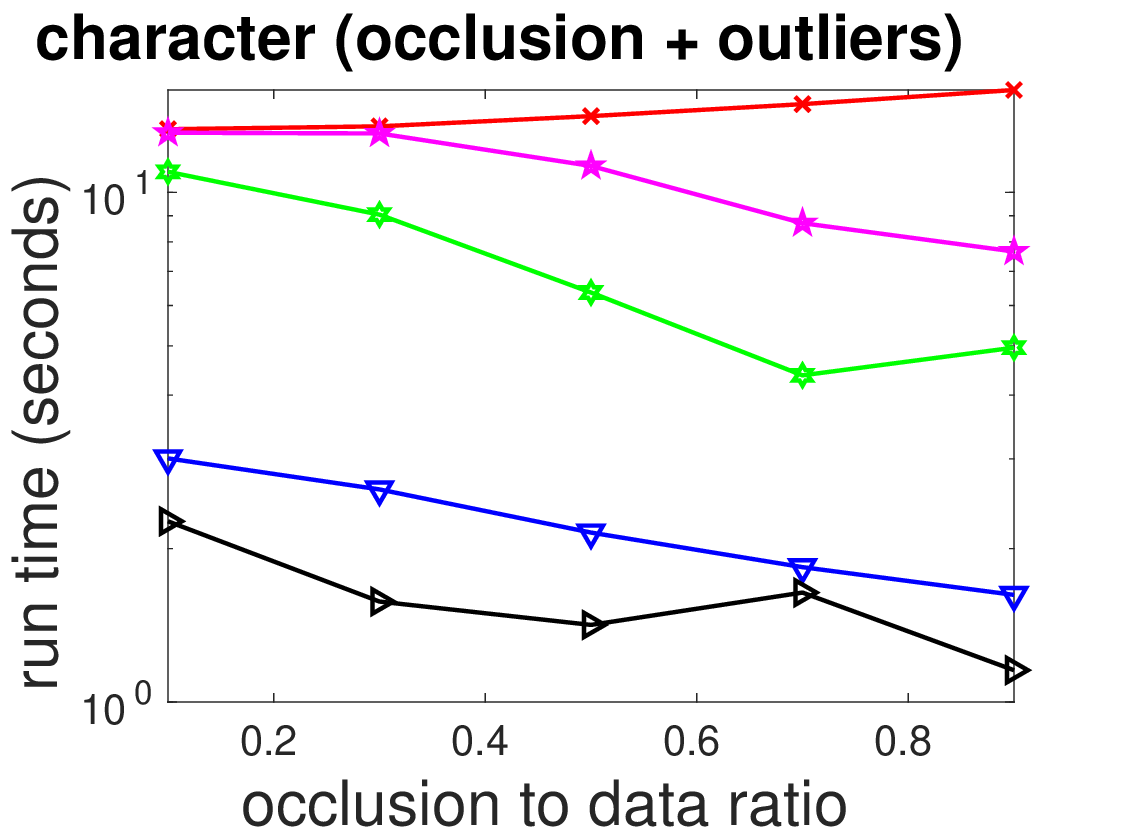}		
\end{tabular}
	\begin{tabular}{@{\hspace{-1.5mm}}c}	\includegraphics[width=.7\linewidth]{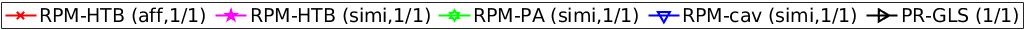}
\end{tabular}	
	%	\vspace{-2mm}
	\caption{Average  registration errors (top 2 rows) and run times (bottom row) by RPM-HTB,
%		with  maximum branching depth of $10$,
		 RPM-PA \cite{lian2021polyhedral},  RPM-CAV \cite{RPM_model_occlude_PR} and GLS
	 under  different $n_p$ values (chosen from $1/2$ to $1/1$ the ground truth value)
		over 100 random trials
		for the 2D deformation, noise, mixed outliers and inliers, separate outliers and inliers and occlusion+outlier tests.
For the two types of outlier tests,
only part of RPM-PA’s results is reported since RPM-PA requires too much memory when problem size is too big.
		\label{2D_simi_sta}
	}

\end{figure*}

%In this subsection,
We compare   with RPM-PA \cite{lian2021polyhedral} and RPM-CAV \cite{RPM_model_occlude_PR},
%and Go-ICP \cite{Go-ICP_pami}.
both of which are based on global optimization,
can handle  partial overlap %ping point sets and
%only utilizes point coordinates information and 
%allowing the corresponding transformations to take arbitrary values.
and allow arbitrary similarity transformation
between two point sets,
making them good candidates for comparison.
%2D similarity transformation is chosen for both RPM-PA and RPM-CAV.
We also compare with PR-GLS \cite{PR_GLS},
which is robust to  outliers, rotation and occlusion.

%The source code of Go-ICP is generously provided by the author. %freely available.
%The range of scale $s$ in our method is set as $0.5\le s\le 1.5$.

%Two choices of transformations:
% which satisfy the requirement as stated in the beginning of section \ref{subsec:scheme1}:
2D similarity  and affine transformations  are respectively tested for RPM-HTB.
% Both of them satisfy the requirements that the transformation is linear with respect to its parameters.
% are considered for our method.
For the former,
we have 
%$
%T( \mathbf x_i| \boldsymbol\theta)= \begin{bmatrix}
%\theta_1 x_i^1  - \theta_2 x_i^2 +\theta_3 \\
%\theta_2 x_i^1 +\theta_1 x_i^2  +\theta_4
%\end{bmatrix}
%$.
%where   ${\boldsymbol\theta}=\begin{bmatrix} \theta_1, \ldots , \theta_4 \end{bmatrix}^\top$.
%Therefore, %the Jacobian matrix
\[
\mathbf J(\mathbf x_i)=\begin{bmatrix}x_i^1&- x_i^2&1&0\\ x_i^2&x_i^1&0&1\end{bmatrix},\quad 
\mathbf B_2=\mathbf B([1,3,4],:)
\]
%constitute the unique  rows of $\mathbf \Xi$ 
%not equal to scaled versions of $\mathbf 1_{n_xn_y}^\top$.
%
For the latter,
we have 
%$
%T( \mathbf x_i| \boldsymbol\theta)= \begin{bmatrix}
%\theta_1 x_i^1  + \theta_2 x_i^2 +\theta_5\\
%\theta_3 x_i^1 +\theta_4 x_i^2  +\theta_6
%\end{bmatrix}.
%$
%where
%${\boldsymbol\theta}=\begin{bmatrix} \theta_1, \ldots, \theta_6 \end{bmatrix}^\top$.
%with $[\theta_1,\ldots, \theta_4 ]^\top$ 
%being the parameters of the linear part of the transformation and
%$[\theta_5, \theta_6]^\top$ representing  translation. %the Jacobian matrix
\begin{gather}
\mathbf J(\mathbf x_i)=\begin{bmatrix}
x_i^1& x_i^2&0&0&1&0\\
0&0&x_i^1& x_i^2&0&1
\end{bmatrix}, \quad 
\mathbf B_2=\mathbf B([1,2,5,8,11],:) \notag
\end{gather}
%constitute the unique rows of $\mathbf \Xi$ 
%not equal to scaled versions of $\mathbf 1_{n_xn_y}^\top$.
%For RPM-HTB,
%we set   $\epsilon_0=8$.

\subsubsection{2D synthetic data 
	\label{sec:2D_synth_test}
}
Synthetic data are easy to generate and can be used to test specific aspects of an algorithm. 
%Following \cite{lian2021polyhedral},
We perform 5 types of tests 
%to evaluate performances of different methods:
to evaluate  a method's robustness to various types of disturbances.
%\begin{inparaenum}[\upshape i\upshape)]
%(deformation, noise, mixed outliers and inliers, separate outliers and inliers, separate outliers and occluded inliers),
% against outliers. 
% 1) \textbf{Occlusion test}. 
\begin{inparaenum}[\upshape i\upshape)]
	\item \textit{Deformation test}:
The prototype shape is non-rigidly deformed to generate the scene  set.
%%	Here, the non-rigid deformation is generated by using the Gaussian radial basis function.	
	\item \textit{Noise test}:	
The prototype shape is perturbed by positional noise to generate the scene point set.
%%	Here, the noise follows the normal  distribution.
	\item  \textit{Mixed outliers and}  \textit{inliers test}:
Random outliers are respectively superimposed on  
the prototype shape to generate the two point sets.
\item \textit{Separate outliers and}  \textit{inliers test}:
Random outliers are respectively added to
different sides of the prototype shape to generate the two point sets.
%	%so as to simulate outlier disturbance. %, respectively, 
	\item
\textit{Occlusion+outlier test},
%	%are used to evaluate  performances of different methods.
	First, 
%%	equal degree of occlusions are applied to 
the prototype shape is respectively occluded 
	to generate the two point sets.
%%	We simulate occlusion by first finding the shortest Hamiltonian circle of the prototype point set 
%%	(by solving a traveling salesman problem) 
%%	and then retaining a segment of the circle starting at a random point.
	Then,
	%a fixed number of 
random outliers (outlier to data ratio is fixed to 0.5) 
are respectively added to different sides of the two point sets.
%%	so as to simulate outlier disturbance.
%	% as illustrated in column 2 of Fig. \ref{three_test_data_exa}.
%	% This test is a combination of the previous two tests 
%	% where there are 
%	% but varying degrees of occlusions on both point sets, 
\end{inparaenum}
Fig. \ref{rot_2D_test_data_exa}
illustrates these tests.
To evaluate a method's ability at handling arbitrary rotation and scaling,
%Different from \cite{RPM_model_occlude}, 
%for all the tests,
%disturbances of
random rotation and scaling within range $[0.5,1.5]$ are also performed  
%to the prototype shape 
when generating the two  point sets.
%a moderate amount of nonrigid deformation is applied to the prototype shape 
%when generating the scene point set.
%
%Finally, random rotation and scaling within range $[0.5,1.5]$
%is also applied when generating the model point sets
%so as to test a method's ability to cope with arbitrary similarity transformation.
%
Fig. \ref{rot_2D_syn_match_exa} shows
examples of registration  by different methods.

The  registration errors  by different methods are presented in the top 2 rows of 
Fig.\ \ref{2D_simi_sta}.
The results indicate that
compared with other methods,
RPM-HTB  %and RPM-CAV
with  $n_p$ value  chosen as the ground truth 
are robust to deformation, noise and
outliers in case when outliers are separate from inliers,
but are not robust when outliers are mixed with inliers.
This is because 
%our method's stopping criterion is based on branching depth instead of duality gap. Thus 
our method is not $\epsilon-$globally optimal as stated in Sec. \ref{sec:converge_2d},
which affects our method's robustness.
%
%performs the best for the deformation and noise tests,
%but slightly worse than other methods for the mixed outliers and inliers test.
%%
%Apparently, the separate outliers and inliers test and the occlusion+outlier test are more challenging  than the mixed outliers and inliers test, as is  evident by the  higher errors of the  methods in these tests.
%Despite the challenges,
%RPM-HTB  performs better than other methods except for  RPM-CAV.
%Compared with other methods,
PR-GLS performs poorly for almost all the tests.
%except for the mixed outlier and inlier test.
This is because our tests contain arbitrary rotations and PR-GLS is only a heuristic method.
%
%We speculate the reason for the poor performance of RPM-HTB for the mixed outliers and inliers test is that our method is not $\epsilon-$globally optimal, thus affecting its ability to distinguish inliers from outliers.
%
%RPM-HTB and its variant
%with  $n_p$ value  chosen as the ground truth
%performs the best.
In terms of different transformation choices,
affine transformation
enables RPM-HTB to perform  better for the deformation  test,
whereas similarity transformation enables RPM-HTB to 
perform better for the two types of outlier tests and the occlusion+outlier test.
 % than our method using similarity transformation.
In terms of different choices of $n_p$ value,
RPM-HTB  with $n_p$ value
close to  the ground truth  performs much better.
%
%is more sensitive to different values of $n_p$ 
%than RPM-inner.
%
%For the 2D outlier and occlusioin + outlier  tests presented in Sec. 7.1 of the main paper,

%than our method with $n_p$ chosen far away from the ground truth value.
%Examples of matching results by different methods are shown in Fig. \ref{two_test_exa}. 

%\begin{table}[h]
%	\renewcommand{\tabcolsep}{0.17cm}
%	\centering
%	\caption{
%		Average run time (in seconds).
%	}
%	\label{time_2d}
%	\small
%	\begin{tabular}{cccccc} 
%		\hline
%		& deform. &noise &
%\begin{tabular}{@{}c@{}}mixed \\ outl./inl.\end{tabular}				
%		&\begin{tabular}{@{}c@{}}separate\\
%outl./inl.\end{tabular}	
%		 &occl.
%		\\\hline		
%		RPM-HTB (aff.)  & 7.1 & 7.2 &16.8  & 16.0 &14.1 \\\hline
%		RPM-HTB (simi.)  & 3.9 & 4.0 &22.4 &  10.6 & 6.1 \\\hline
%		RPM-PA & 8.1   & 8.3  &33.7   &  9.5    &  5.8 \\\hline
%		RPM-CAV  & 2.7 &  2.8 &2.7  &    2.9& 1.6  \\\hline
%		PR-GLS & 0.4 &0.4 &1.4 &2.1 &1.5
%	\end{tabular} 
%\end{table} 

The average run times %(in seconds) 
by different methods
are presented in the bottom row of Fig. \ref{2D_simi_sta}.
%10.08 or
%757.81
%for RPM-HTB using similarity or affine  transformation,
%2.24
%for RPM-CAV %\textcolor{red}{(where GPU speedup is used)}
%and 11.30
%for Go-ICP.
%The result indicates that  
RPM-HTB is only less efficient that the best performing method PR-GLS 
for  the deformation and noise tests.
It is almost as efficient as the best performing method RPM-CAV  for the two types of outlier tests.
In terms of different transformation choices,
RPM-HTB employing similarity transformation is more efficient than it empolying affine transformation
owing to  similarity  transformation's  less number of paramters.

%
%Since our method using affine transformation has similar matching errors as our method using similarity transformation,
%%and also since it runs quite slowly,
%it will not be tested in the next section.

From the two types of outlier tests,
we can also  see how different methods scale with problem size.
%Fig. 
%reports the experimental result.
RPM-HTB and RPM-CAV both scale the best with problem size,
followed by  PR-GLS and then RPM-PA.
%which partly due to the method needing to cut an increasingly complex polygon.
%It is against intuition that PR-GLS, a heuristic method,  scales poorly than global optimization based methods.

%\begin{figure}[h]
%	\centering
%	
%	
%	\includegraphics[width=.7\linewidth]{figures/time2d_outlier.eps}
%	\caption{\textcolor{black}{
%		Average	run times of different methods in the separate outliers and inliers test when using the character shape.
%For RPM-PA, % and PR-GLS,
%only part of the results is shown since it requires increasingly larger  memory
%% and the latter becomes too slow  
% with the increase of  problem size.	}	
%		\label{runtime2d_outlier}	}
%\end{figure}

\begin{figure}[!t]
	\setlength\arrayrulewidth{1pt}
	
	\centering
	\newcommand\scale{0.9}
		\includegraphics[width=0.9\linewidth]{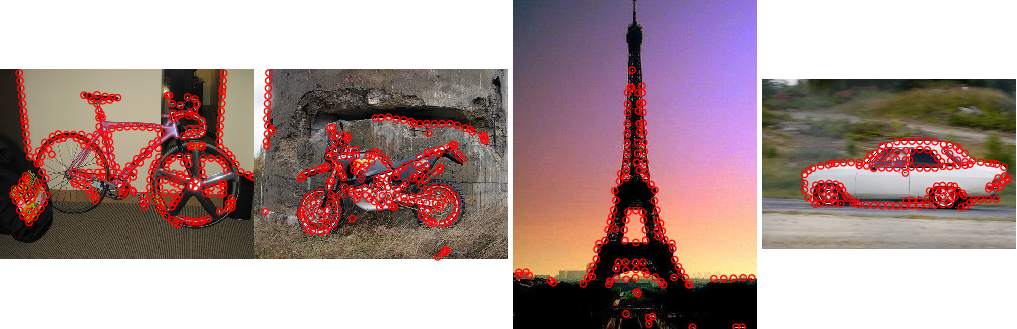}

	\caption{
		 Model  images with model point sets superimposed.		
		\label{rot_2D_canny_model}}
\end{figure}

\begin{figure*}[!t]
	\setlength\arrayrulewidth{1pt}
	
	\centering
	\newcommand\scale{0.5}
	
	\begin{tabular}{@{\hspace{-1mm}}c@{\hspace{-0mm}}|@{\hspace{.0mm}}c }
		\includegraphics[width=\scale\linewidth]{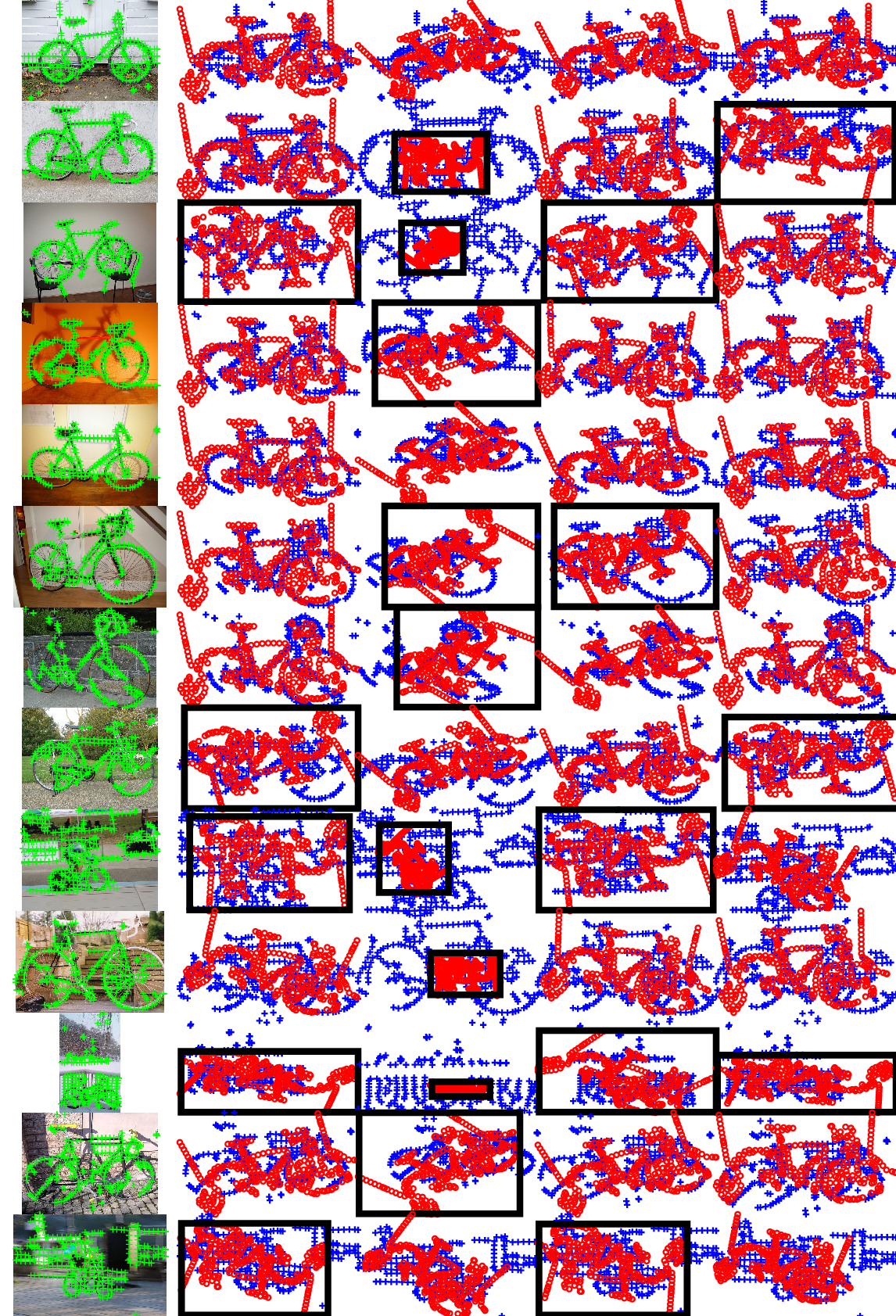} &
		\includegraphics[width=\scale\linewidth]{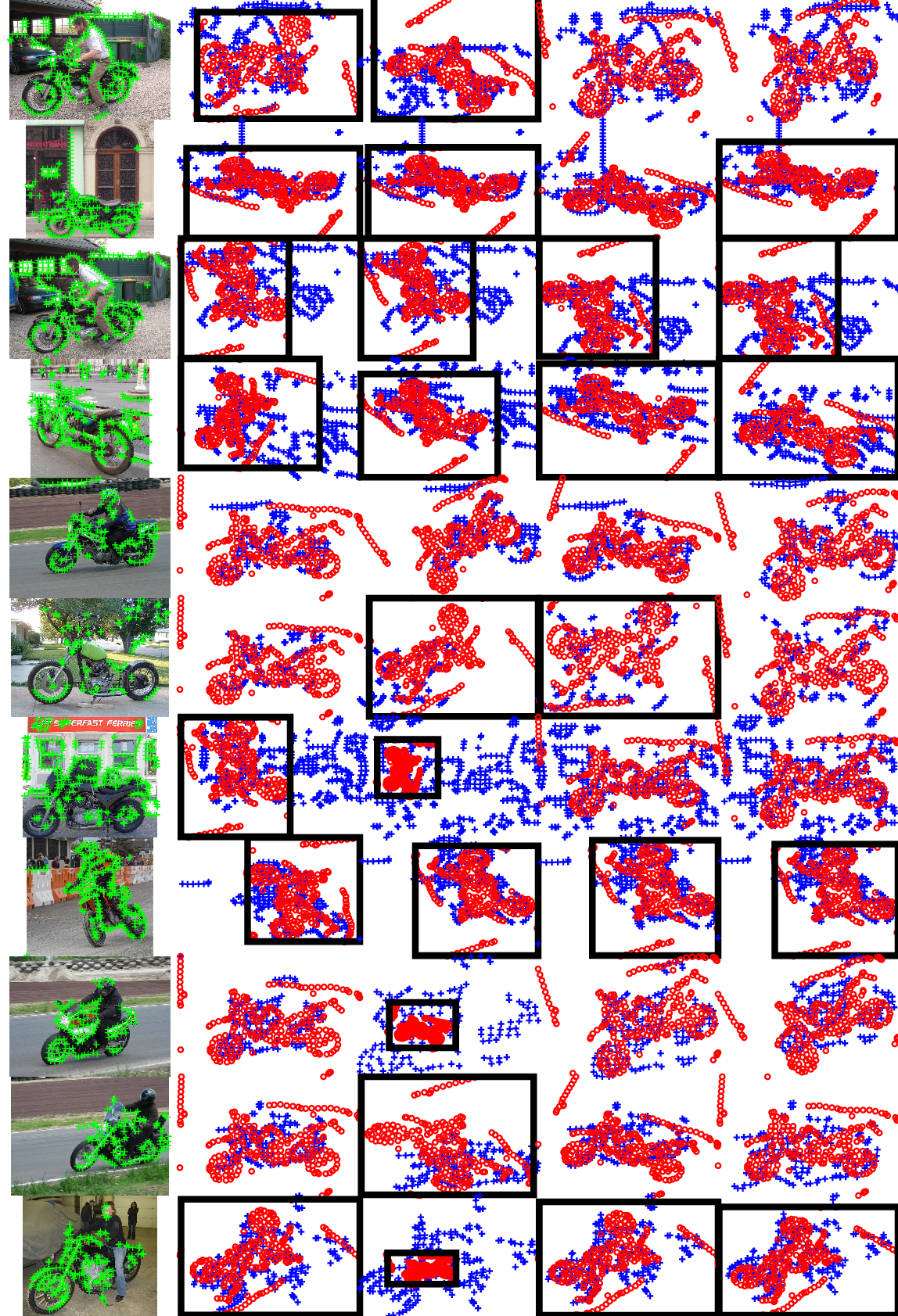} 
		%		\vspace{-1mm}
		%		\\
		%		%	\end{tabular}		\\
	%	%	\begin{tabular}{@{\hspace{-20mm}}c@{}|@{}c }	
		%		\hline
		%		&
		%		\includegraphics[width=\scale\linewidth]{figures/eiffel_5_methods} &
		%		\includegraphics[width=\scale\linewidth]{figures/moto_5_methods} 		
	\end{tabular}
%\small
%\begin{tabular}{@{\hspace{-5mm}}c @{\hspace{0 mm}} c@{\hspace{0 mm}}  c@{\hspace{0 mm}}  c@{\hspace{0 mm}}   c@{\hspace{30mm}}}	
%	(a) scene & (b) RPM-HTB (simi.) & (c) RPM-HTB & (d) Go-ICP & (e) FRS
%\end{tabular}
	\caption{
%		Left grid cell: model  images with model point sets superimposed.
%		The remaining  cells:
				For each cell: 
		scene images with scene point sets superimposed, registration results by RPM-HTB using similarity or affine transformations, RPM-PA \cite{lian2021polyhedral} and RPM-CAV \cite{RPM_model_occlude_PR}.
		The $n_p$ value for each method is chosen as $0.9$ the minimum of the cardinalities of two point sets.
		The failure cases are encircled by black boxes.
		\label{rot_2D_canny}}
\end{figure*}

\subsubsection{2D real data
	%	point sets extracted from images
}
Point sets extracted from images constitute  a more realistic type of data  for evaluating  registration methods.
We use the Canny edge detector to extract 2D point sets  from several images coming from   the Caltech-256 \cite{caltech_database} and VOC2007   \cite{pascal-voc-2007}
datasets.
The extracted  point sets are then used to test different registration  methods, as illustrated in Fig. \ref{rot_2D_canny_model} and \ref{rot_2D_canny}.
To evaluate a method's ability at handling arbitrary rotations,
we rotate a model point set by $180^\circ$ before empolying a registration method to align it with a scene point set.

Since PR-GLS has high registration errors  in the previous section,
it is not tested.
Fig. \ref{rot_2D_canny} reports
the registration  results  by the remaining methods.
RPM-HTB empolying similarity transformation, RPM-PA 
and RPM-CAV 
performs the best.
%followed by Go-ICP.
%In comparison,
%RPM-CAV  does not fit the wheels of  bicycles tightly 
%%fails to align majority of the cars
%and 
%aligns some eiffel towers tiltedly. 
RPM-HTB  empolying affine transformation  performs poorly.
This is because affine transformation contains too much deformation freedom,
causing the resulting registration method  not robust to background  clutters.
Also, most  of the methods fail in the car test.
This is because the car test contains heavy background clutters, 
posing a serious challenge for the registration methods.
% and majority of the methods do not perform well in this test.

\begin{figure*}[t]
	\setlength\arrayrulewidth{1pt}
	
	  \ContinuedFloat
	      \captionsetup{list=off,format=cont}
	  
	\centering
	\newcommand\scale{0.52}
	
	\begin{tabular}{@{\hspace{-1mm}}c@{\hspace{-0mm}}|@{\hspace{.3mm}}c@{} }
%		\includegraphics[width=0.1\linewidth]{figures/bike_car_eiffel_moto}&		
%		\includegraphics[width=\scale\linewidth]{figures/bike_5_methods} &
%		\includegraphics[width=\scale\linewidth]{figures/car_5_methods} \vspace{-1mm}
%		\\
		%	\end{tabular}		\\
	%	\begin{tabular}{@{\hspace{-20mm}}c@{}|@{}c }	
%		\hline
%		&
		\includegraphics[width=\scale\linewidth]{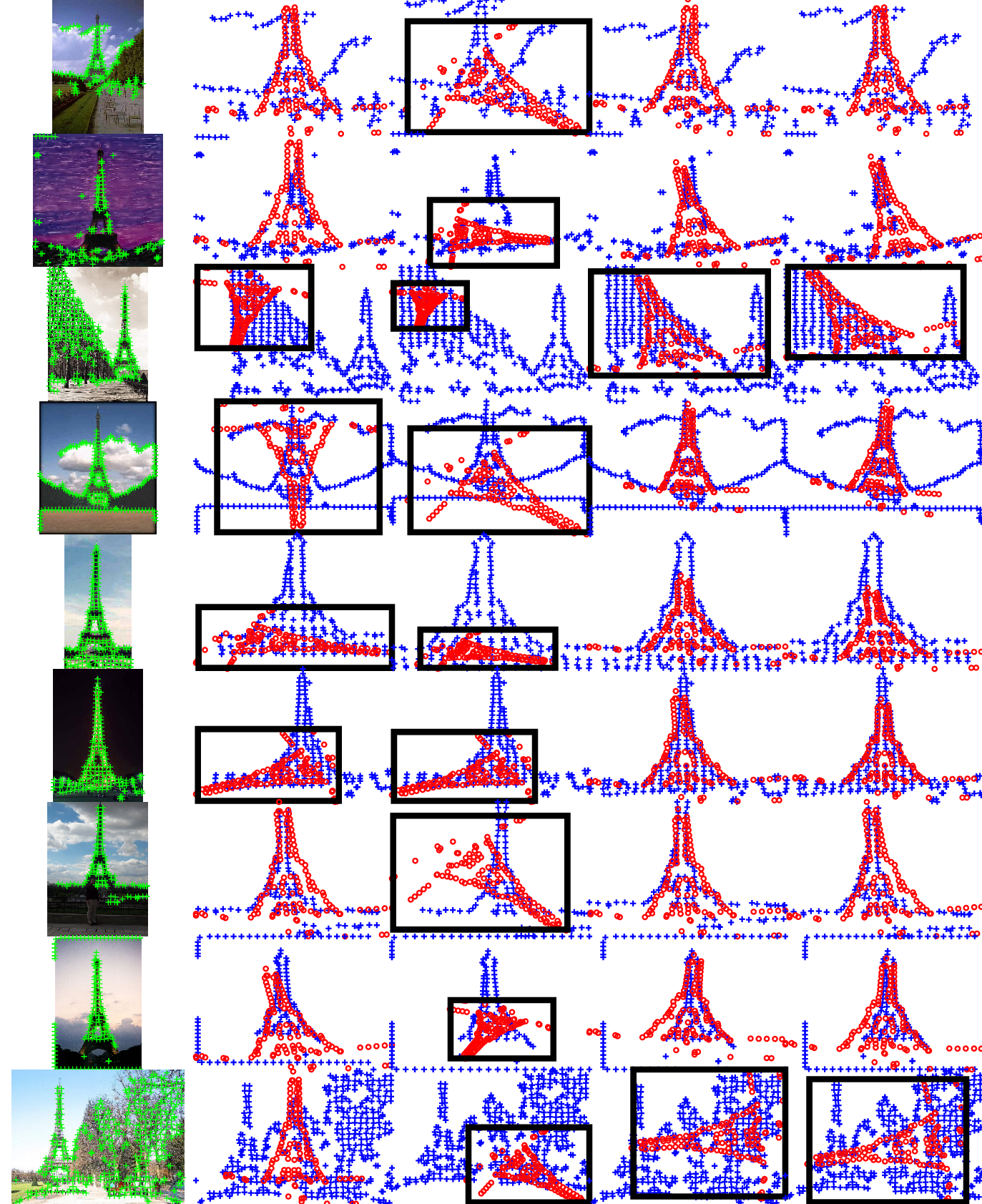} &
		\includegraphics[width=\scale\linewidth]{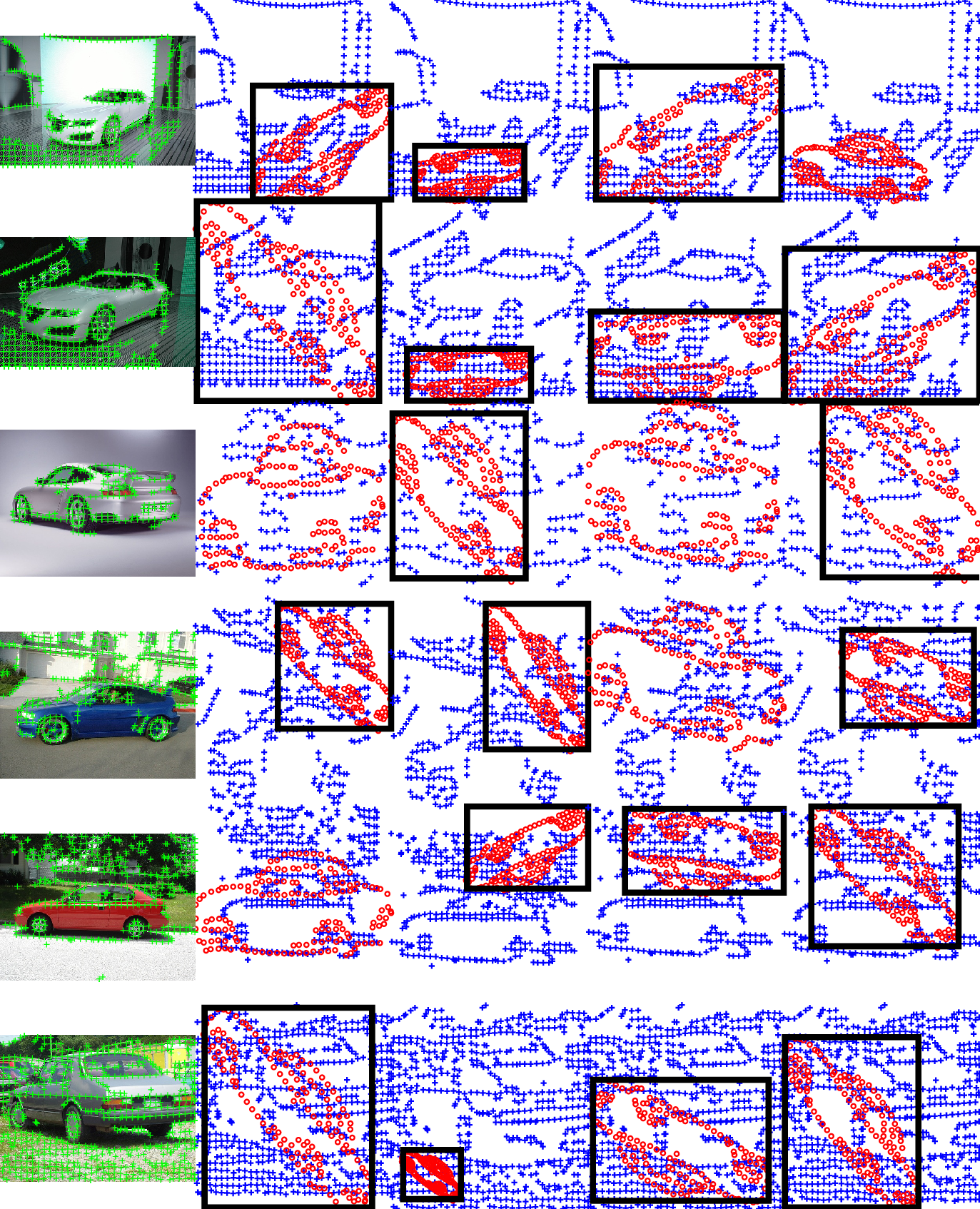} 		
	\end{tabular}
	\caption{ 
%For each    cell:
%		scene images with scene point sets superimposed, registration results by RPM-HTB using similarity or affine transformations, RPM-PA \cite{lian2021polyhedral} and RPM-CAV \cite{RPM_model_occlude_PR}.
%		The $n_p$ value for each method is chosen as $0.9$ the minimum of the cardinalities of two point sets.
%		\label{rot_2D_canny}
	}
\end{figure*}

\subsection{Case two: 3D rigid transformation}

%\subsubsection{3D synthesized datasets}

%In this section,
Since RPM-PA and  RPM-CAV are not efficient for the 3D case,
they are not  tested.
%In his section,
Instead,
we  compare with Go-ICP \cite{Go-ICP_pami}, FRS \cite{BnB_consensus_project}, GORE \cite{GORE_outlier_removal} and TEASER++ \cite{TEASER}, 
which are based on global optimization
%only utilizes point coordinate information 
and 
allows arbitrary rigid transformation between two point sets,
making them  good candidates for comparison.
%For our algorithm,
%we set   $\epsilon_0=6$.
%Finally, we compare with a heuristic variant of RPM-HTB where   $\epsilon_0=0$ and the maximum  branching depth is set as $10$ instead of infinity.
For GORE and TEASER++,
the FPFH feature descriptor is used to extract putative matches and
the number of putative matches is set to 500.

\subsubsection{3D synthetic data}
\label{sec:3D_synth_test}

\begin{figure*} [t]

	\centering
	\newcommand\scale{0.105}		
	\begin{tabular}{@{\hspace{-3mm}}c@{}|@{}c@{}|@{}c@{}|@{}c@{}|@{}c@{}|@{} c@{}|@{} c@{} |c@{}|c }						
		\includegraphics[width=\scale\linewidth]{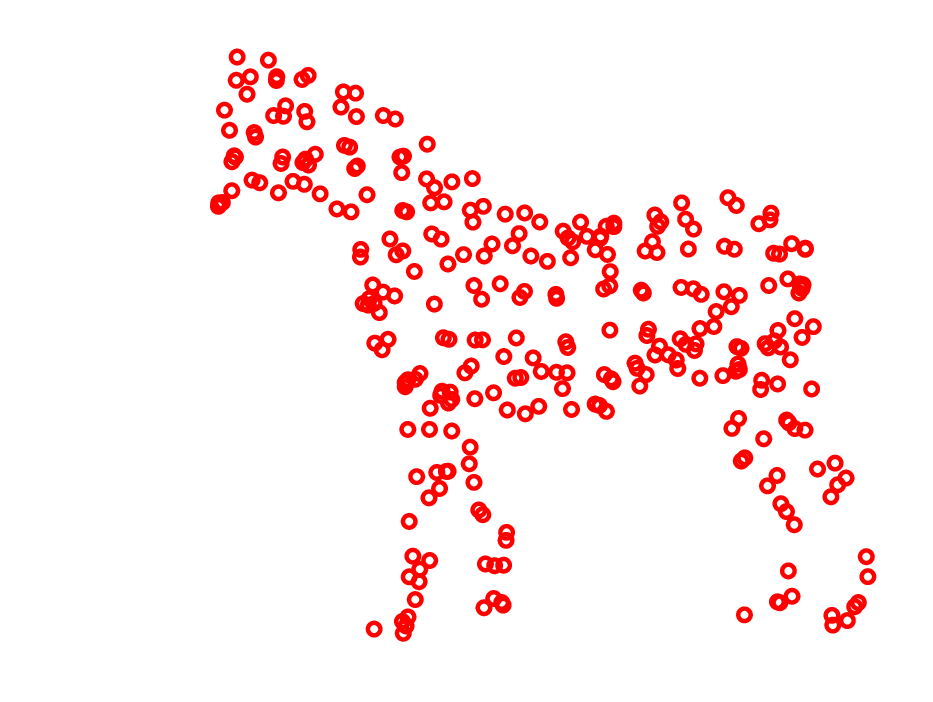}&
		\includegraphics[width=\scale\linewidth]{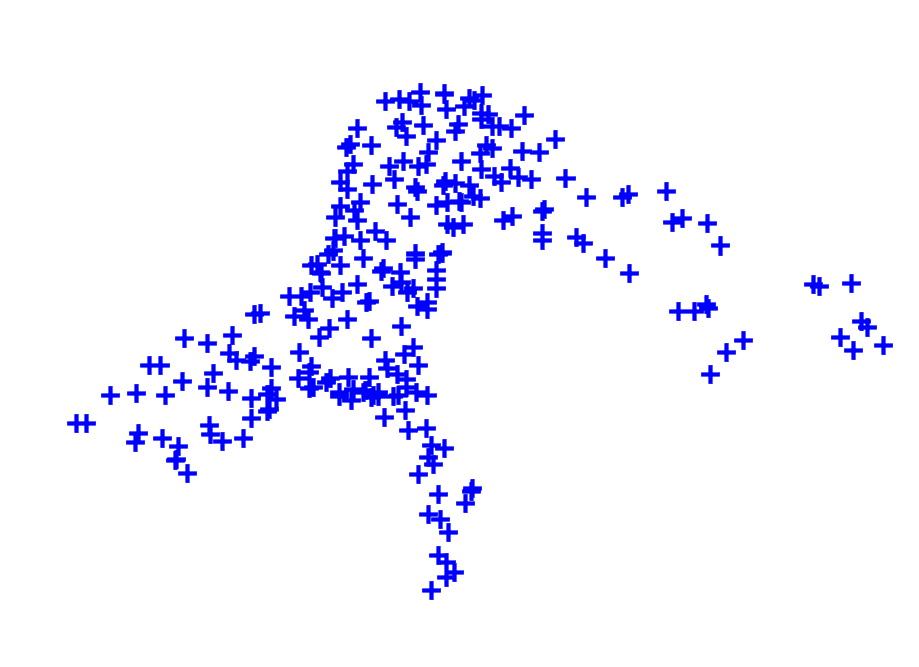}&
		\includegraphics[width=\scale\linewidth]{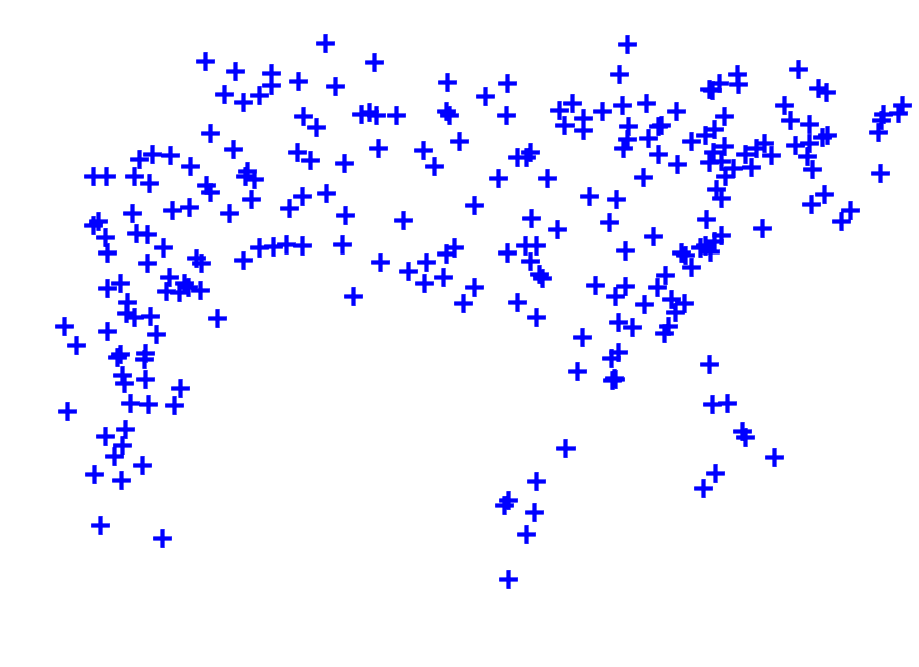}&
		\includegraphics[width=\scale\linewidth]{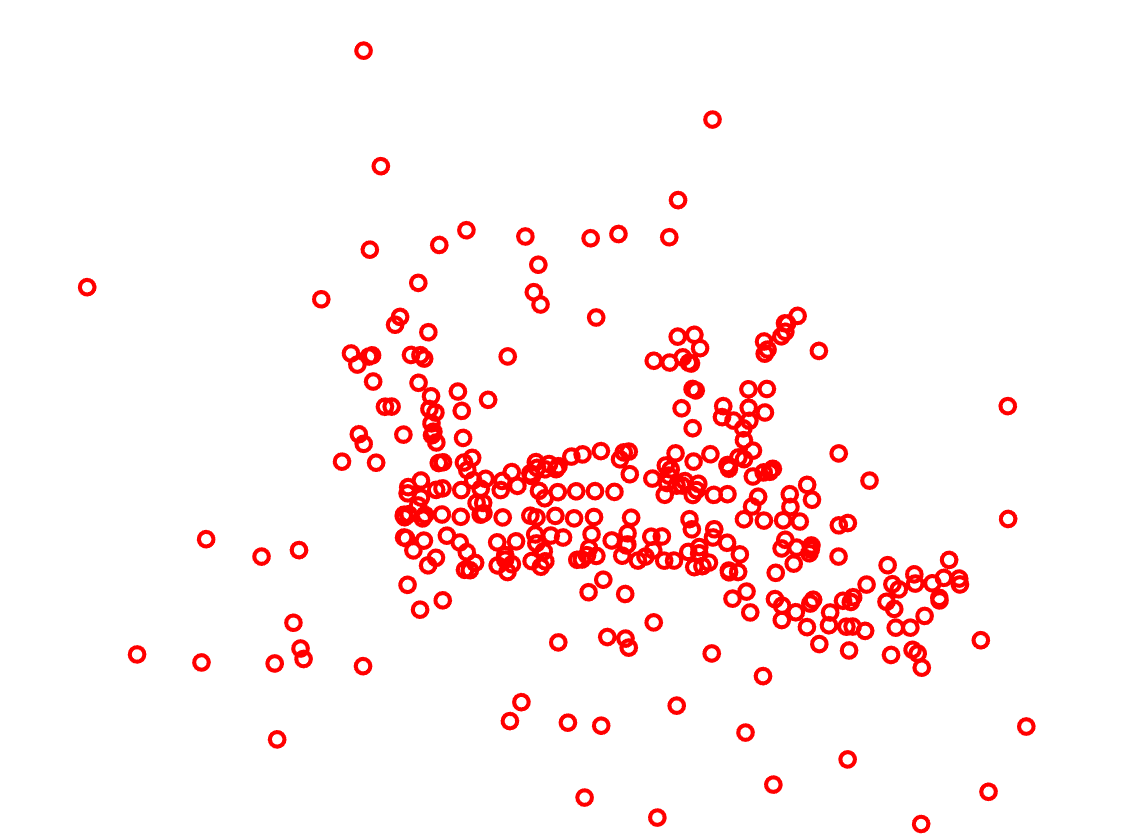}&	
		\includegraphics[width=\scale\linewidth]{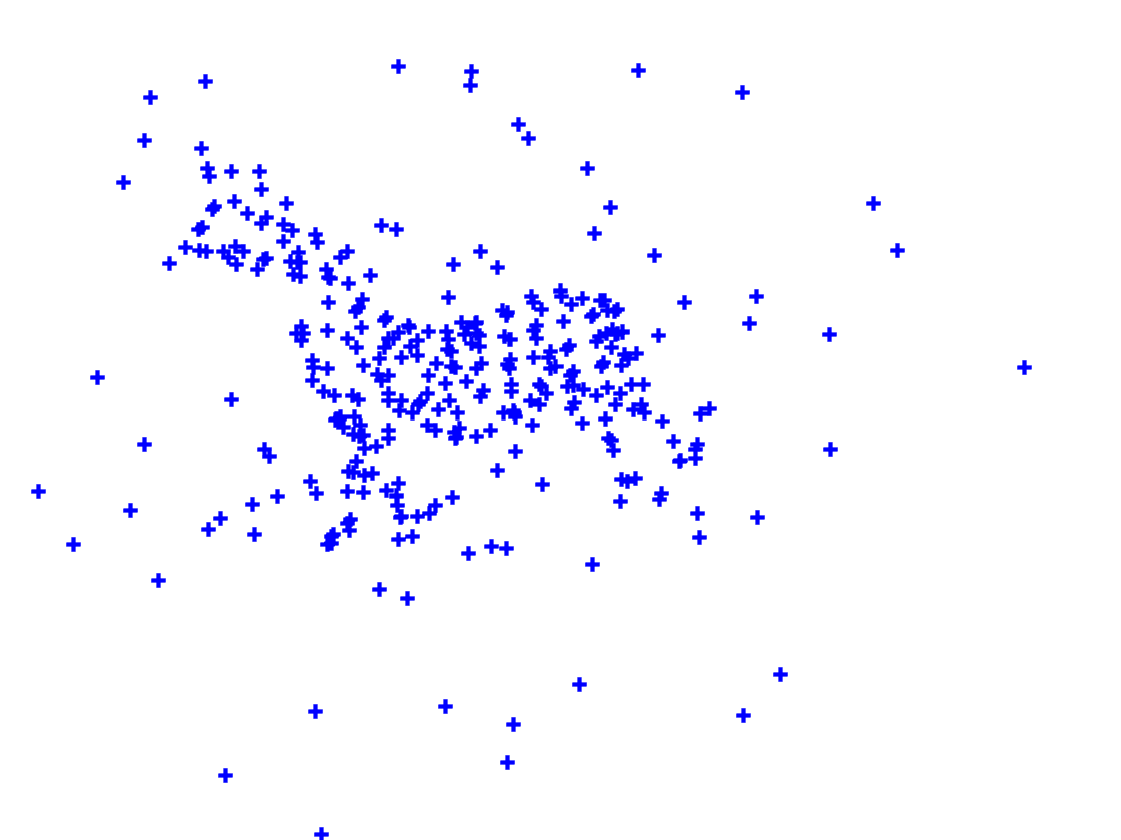}&					
		\includegraphics[width=\scale\linewidth]{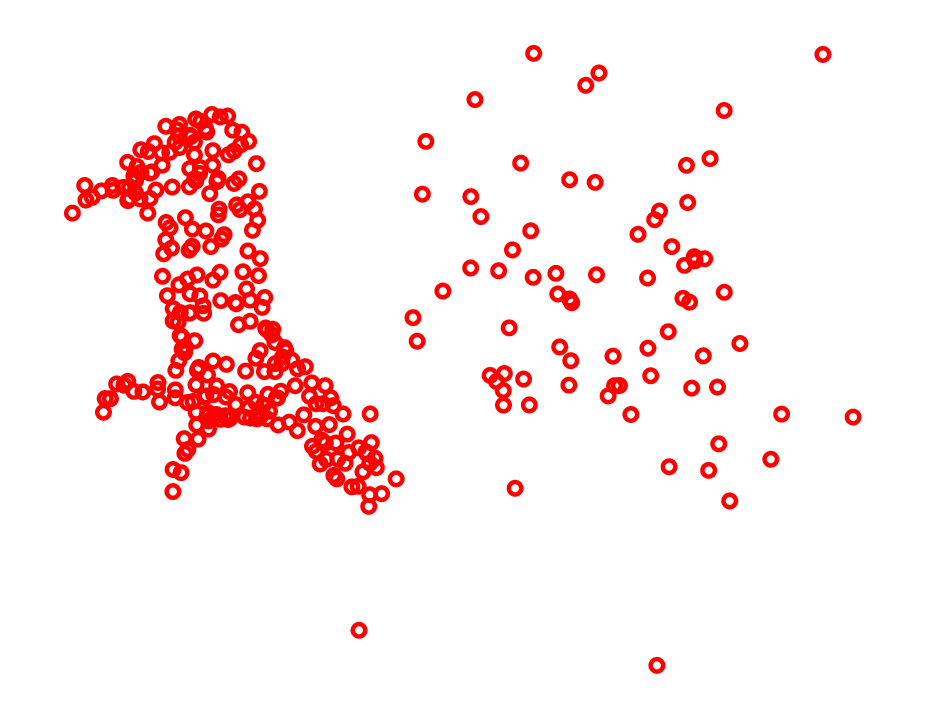}&
		\includegraphics[width=\scale\linewidth]{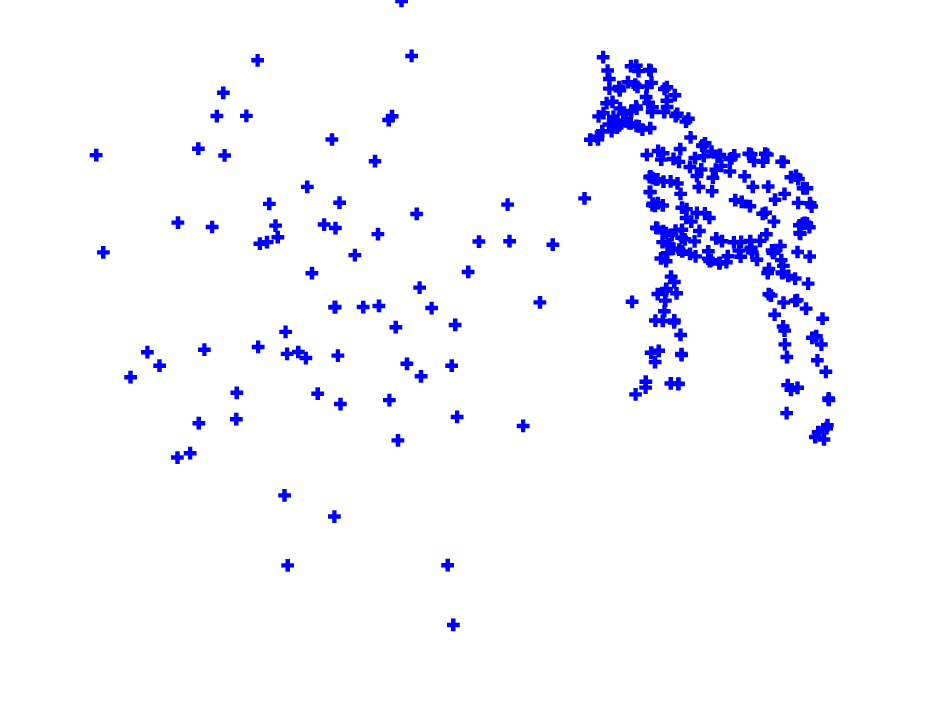}&
		\includegraphics[width=\scale\linewidth]{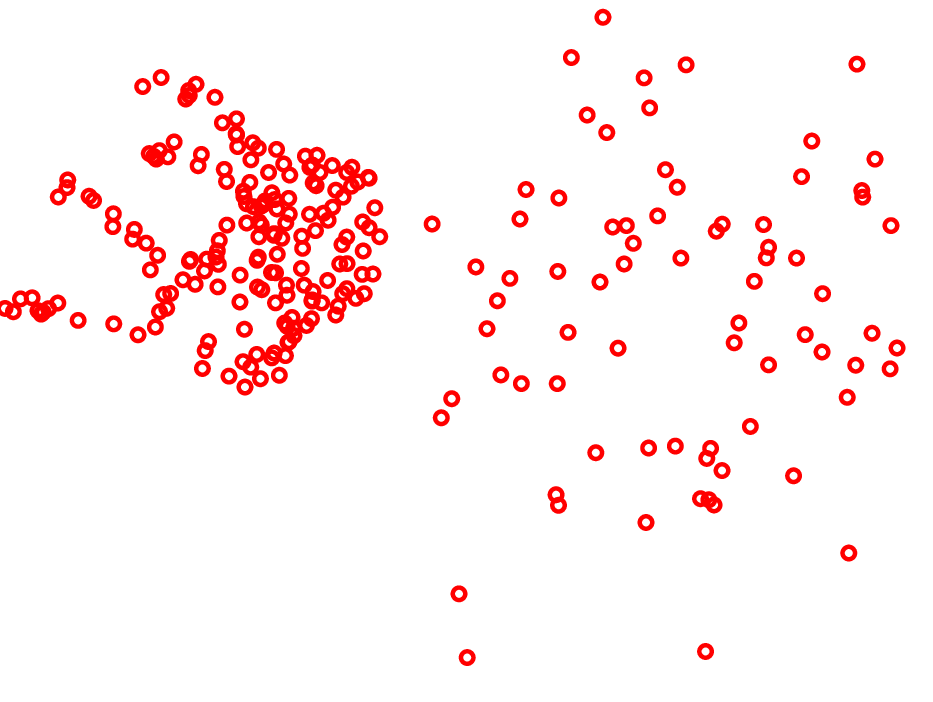}&
		\includegraphics[width=\scale\linewidth]{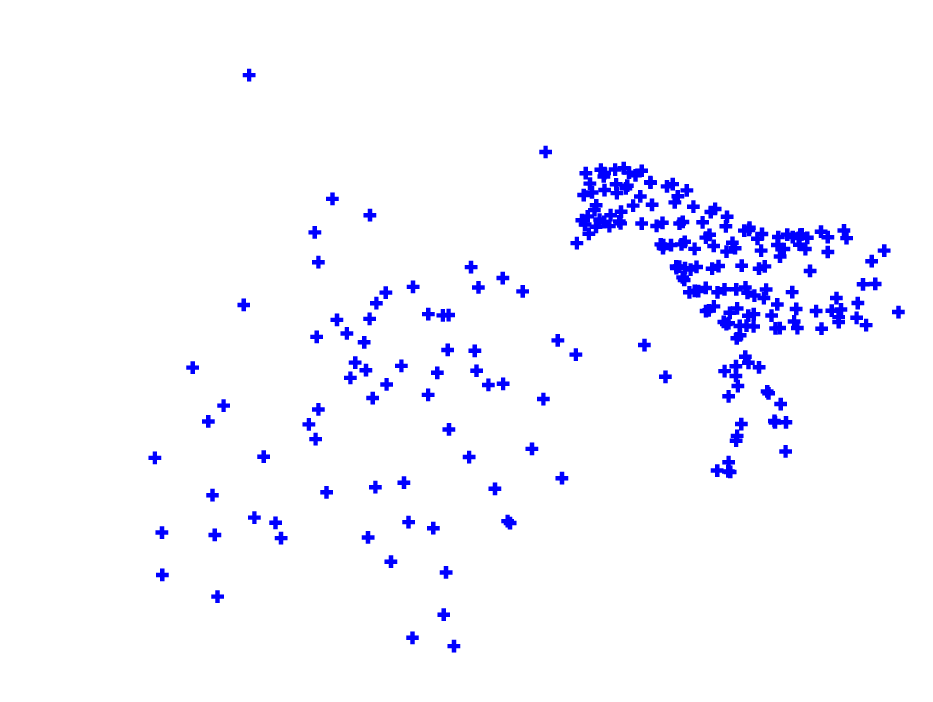} 
		\\\hline
		%	  \vspace{1mm}
		%
		\subfigure[]{		
			\includegraphics[width=\scale\linewidth]{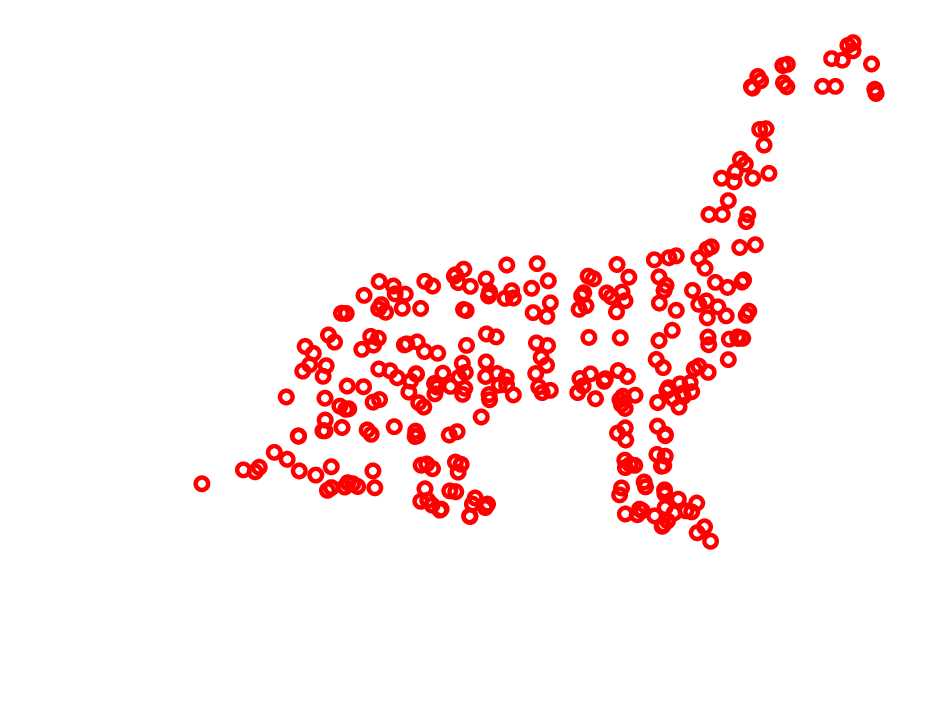}} &
		\subfigure[]{		
			\includegraphics[width=\scale\linewidth]{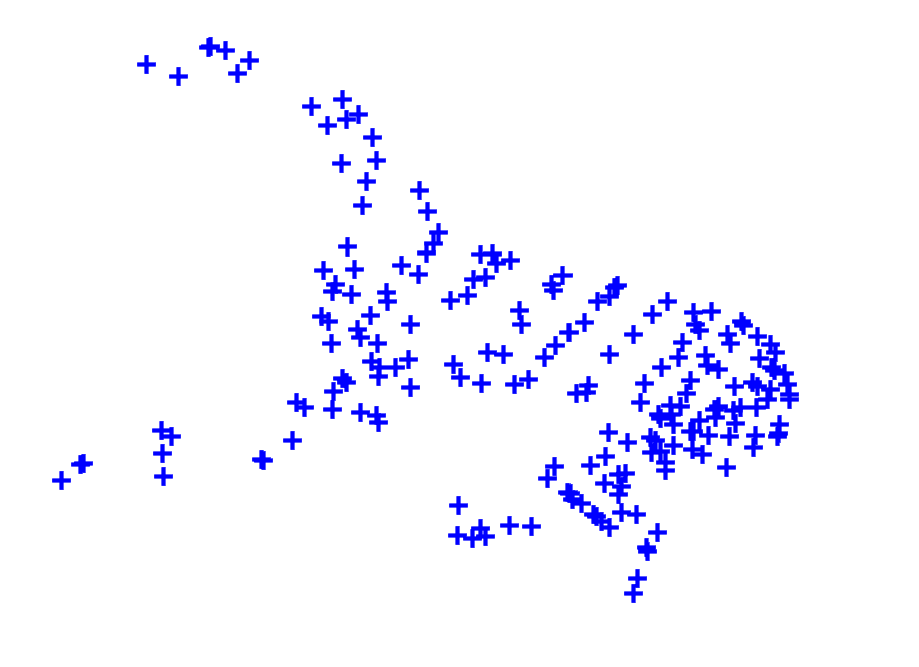}}&
		\subfigure[]{		
			\includegraphics[width=\scale\linewidth]{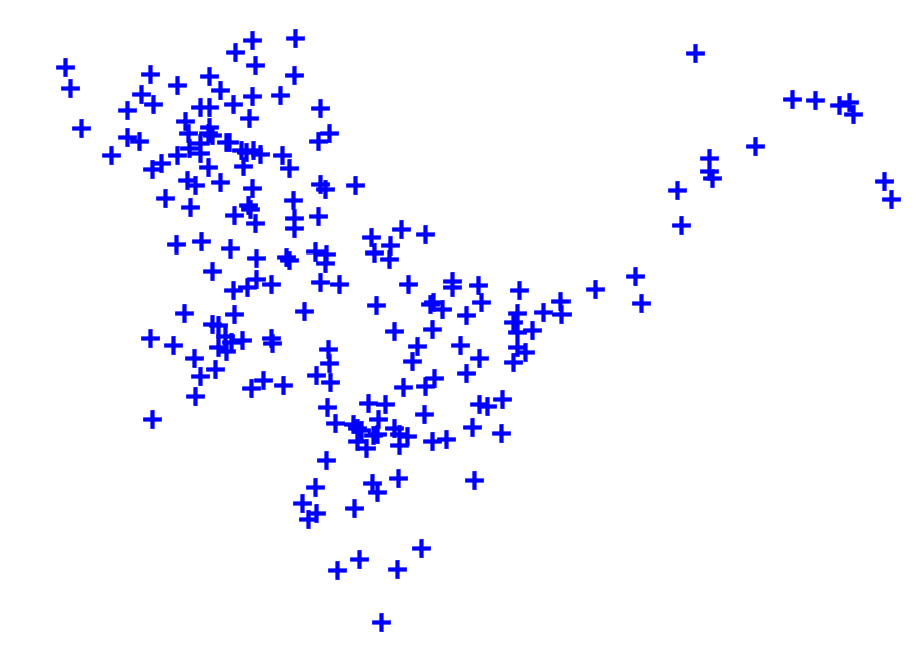}}&
		\subfigure[]{\includegraphics[width=\scale\linewidth]{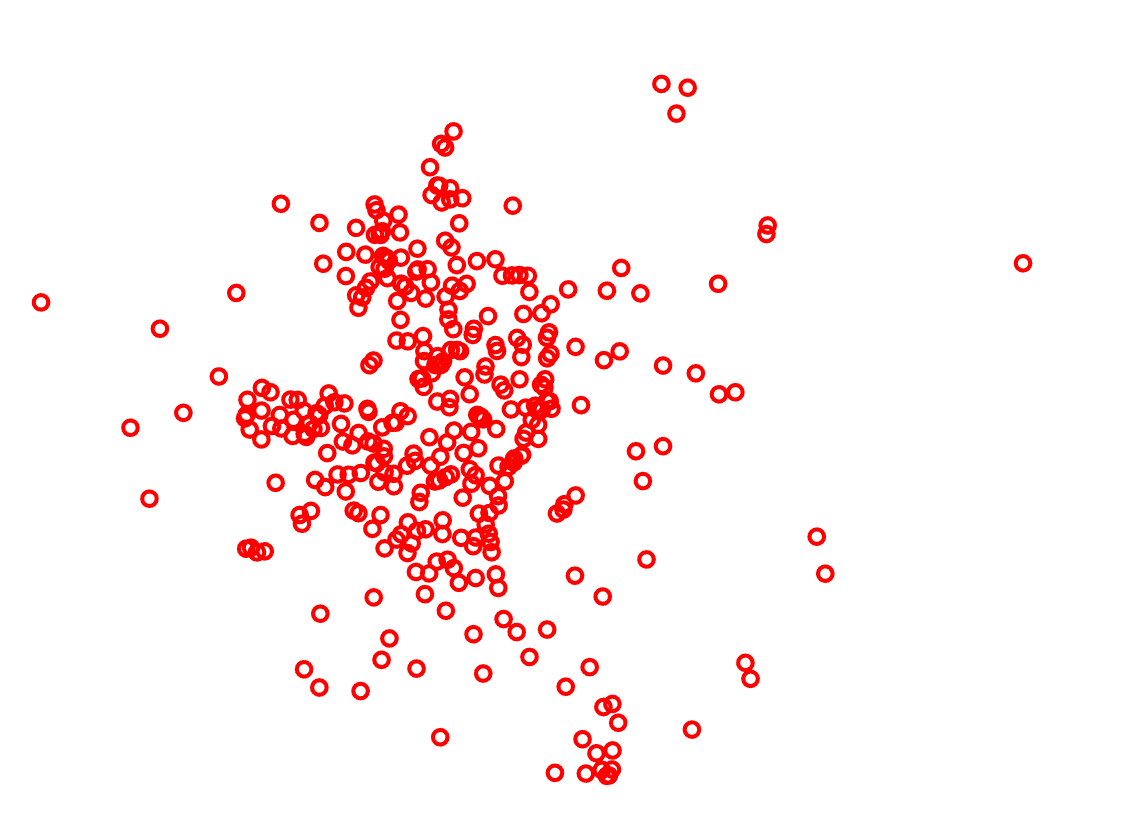}}&
		\subfigure[]{\includegraphics[width=\scale\linewidth]{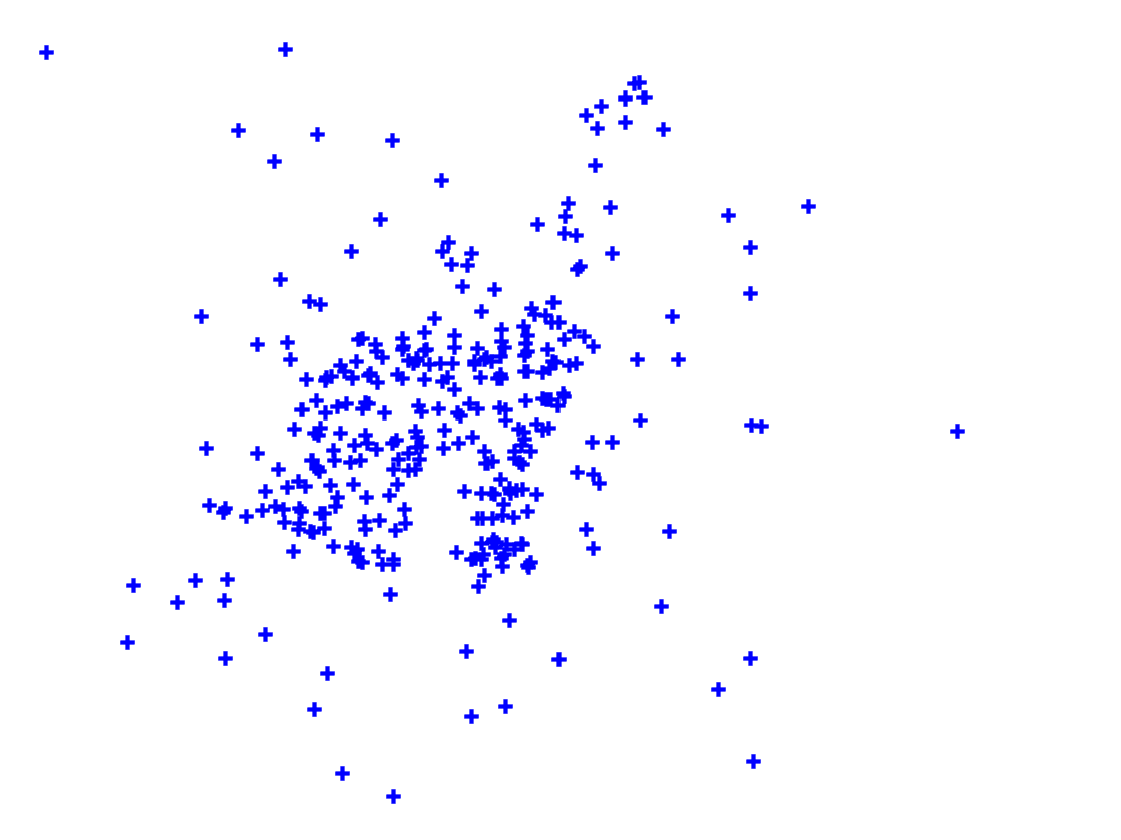}}&			
		\subfigure[]{		
			\includegraphics[width=\scale\linewidth]{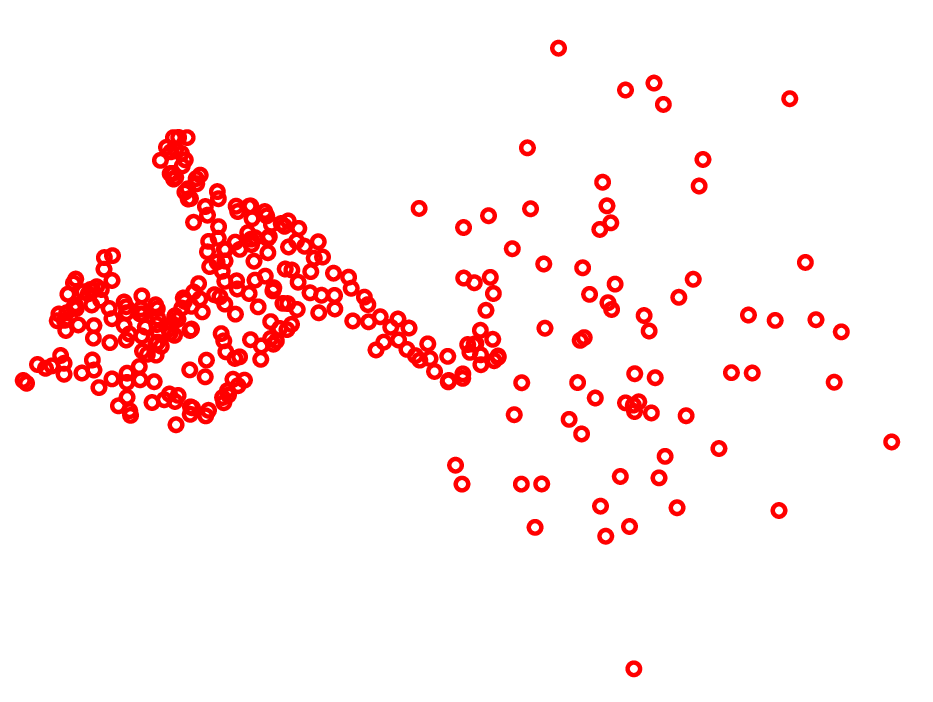}}&
		\subfigure[]{		
			\includegraphics[width=\scale\linewidth]{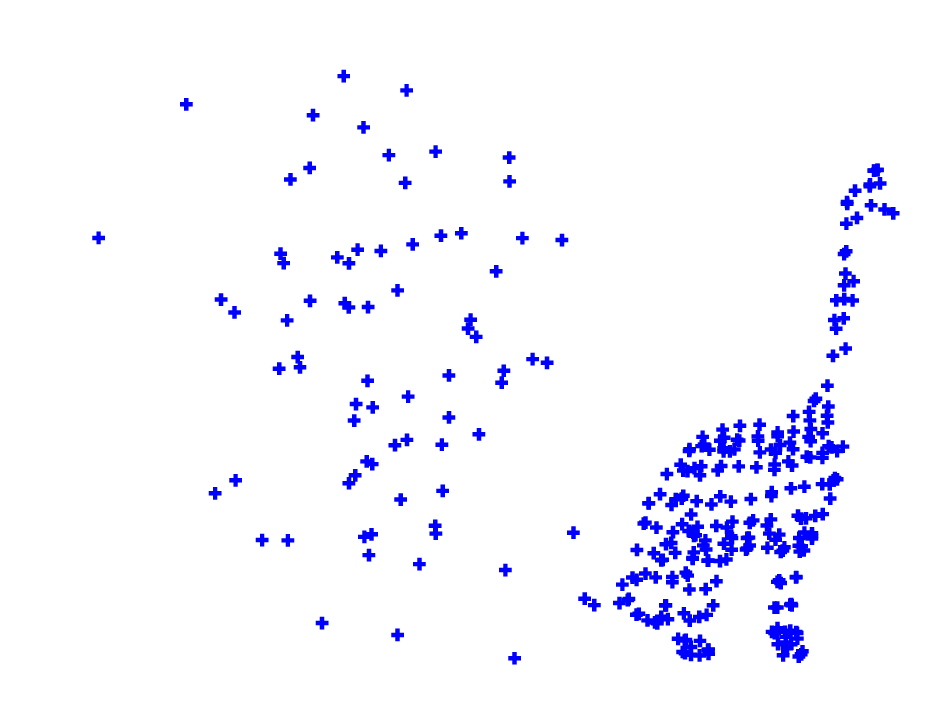}} &
		\subfigure[]{		
			\includegraphics[width=\scale\linewidth]{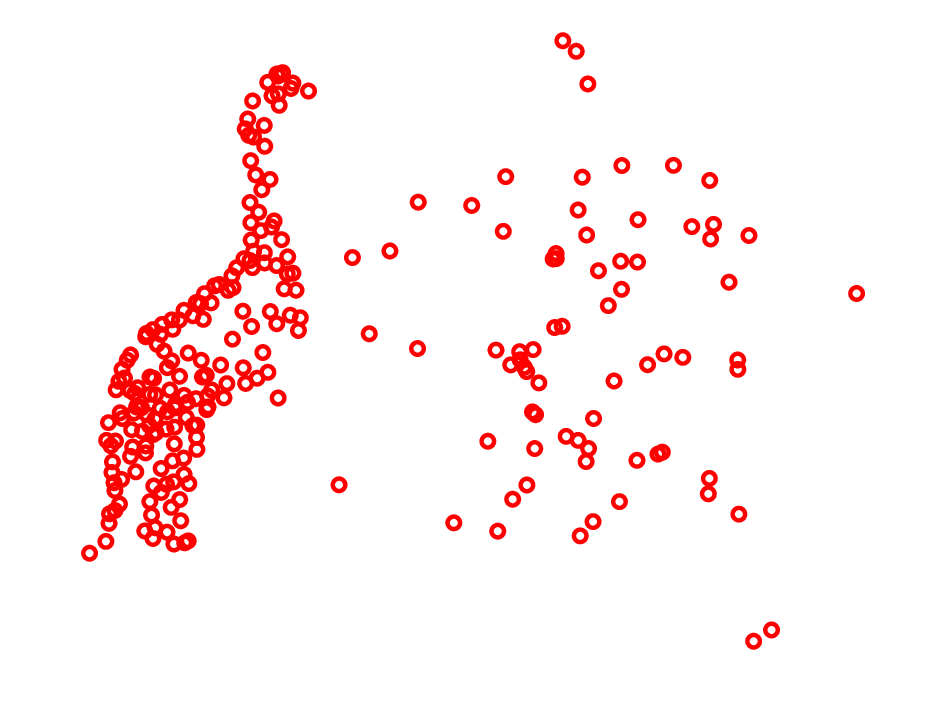}} &
		\subfigure[]{
			\includegraphics[width=\scale\linewidth]{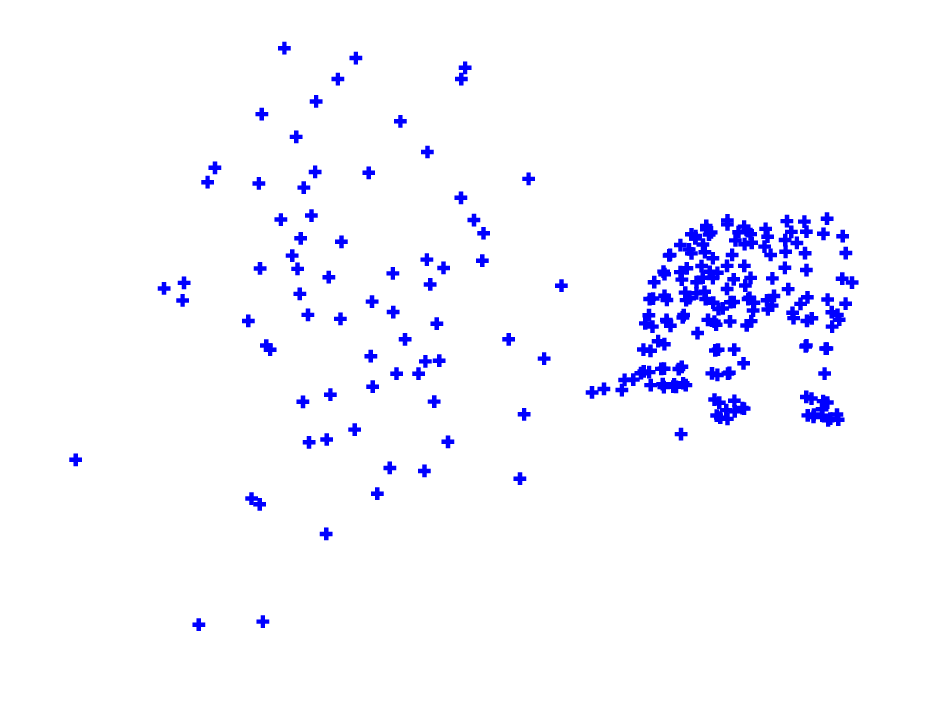}}
	\end{tabular}	
	\caption{
		(a) to (c): model point sets 
		and 
		examples of scene point sets  in the deformation and  noise  tests, % (columns 2 to 3), 
		respectively.		
		(d) to (i):
		examples of model and scene point sets  in the 
		mixed outliers and inliers test ((d), (e)),
		separate outliers and inliers test ((f), (g)) and occlusion+outlier test  ((h), (i)), respectively.
		Here, the model points are marked by red circles and
		the scene points  by blue crosses.
		\label{rot_3D_test_data_exa}}
%\end{figure*} 
%\begin{figure*}[t]
	\newcommand{\height}{2cm}
	\centering
	\newcommand\Scale{0.19}
	\begin{tabular}{@{\hspace{-0mm}}c@{}|@{\hspace{-0mm}} c@{}|@{} c@{}|@{}  c@{}|@{}c }	
		\includegraphics[height=\height,width=\Scale\linewidth]{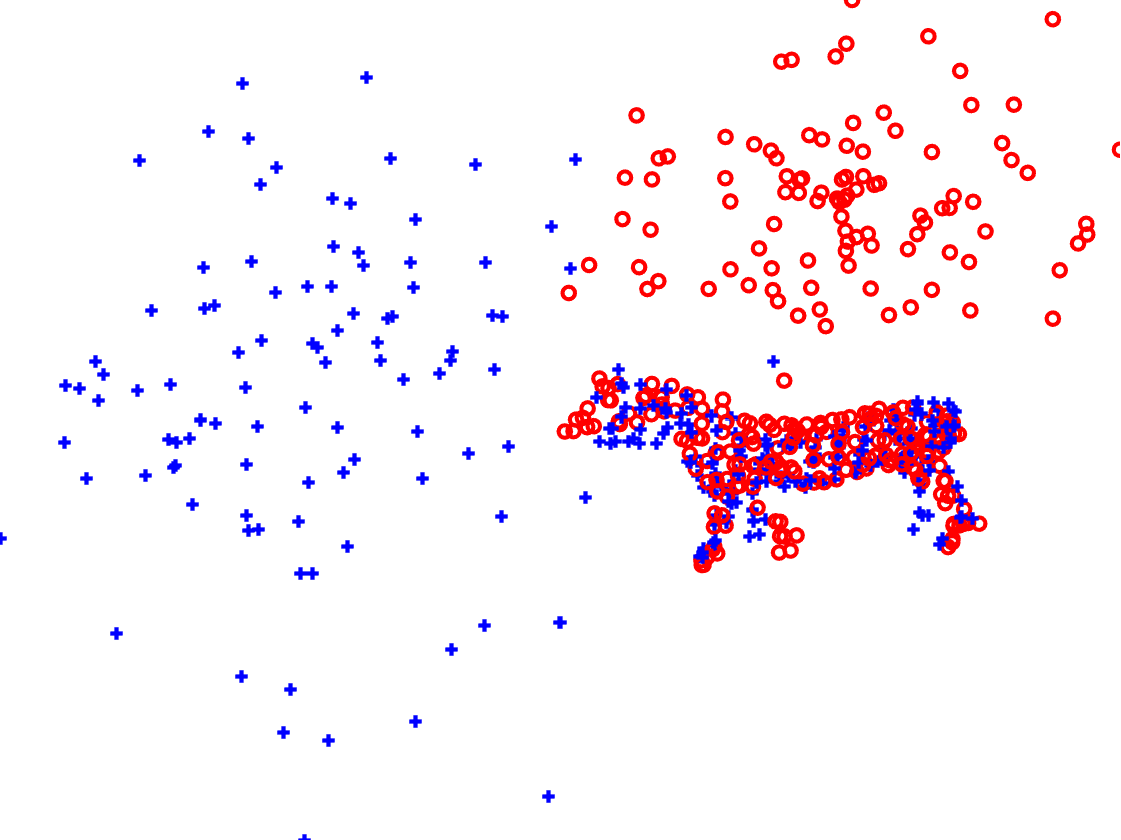}&
		\includegraphics[height=\height,width=\Scale\linewidth]{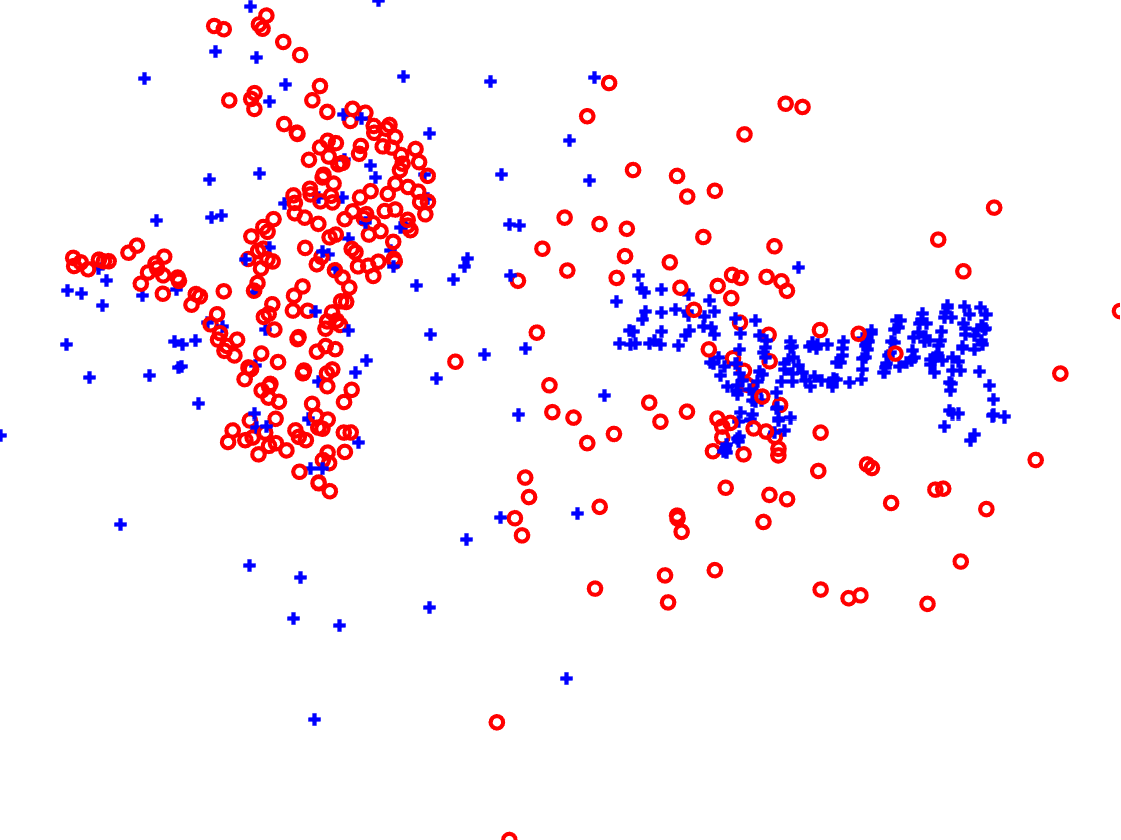}&
		\includegraphics[height=\height,width=\Scale\linewidth]{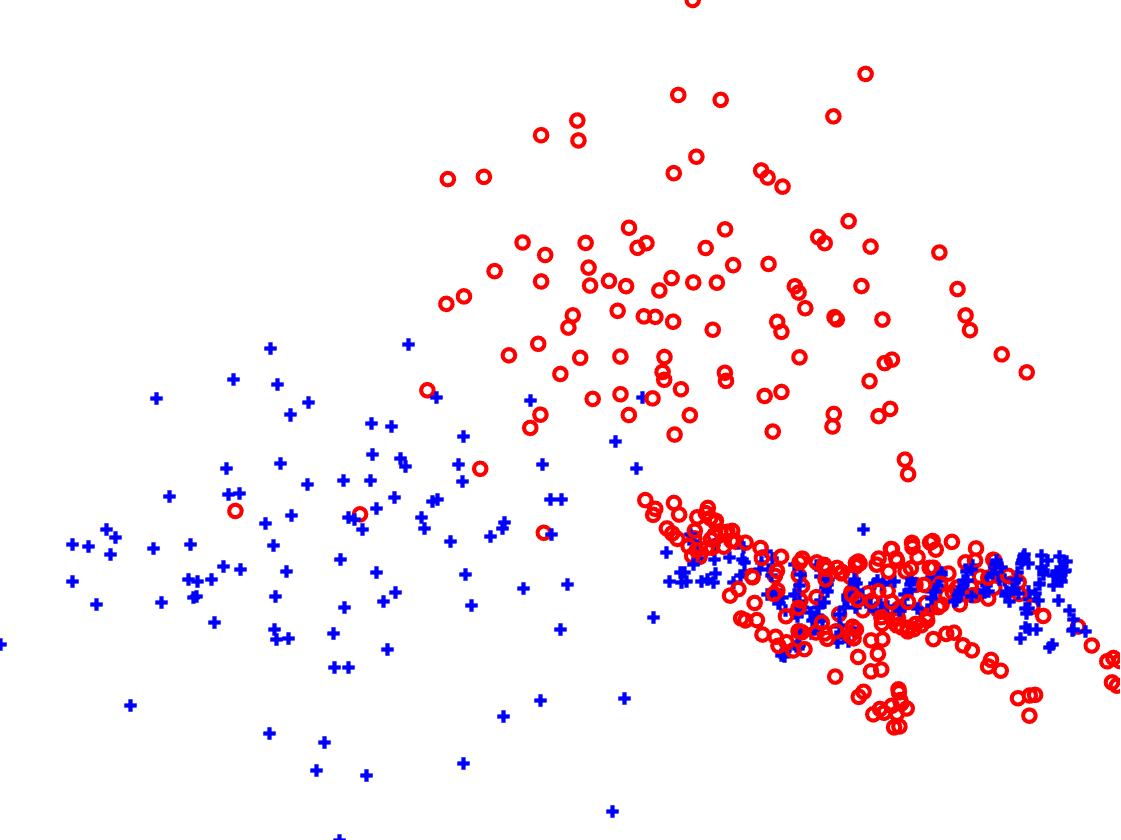}&
		\includegraphics[height=\height,width=\Scale\linewidth]{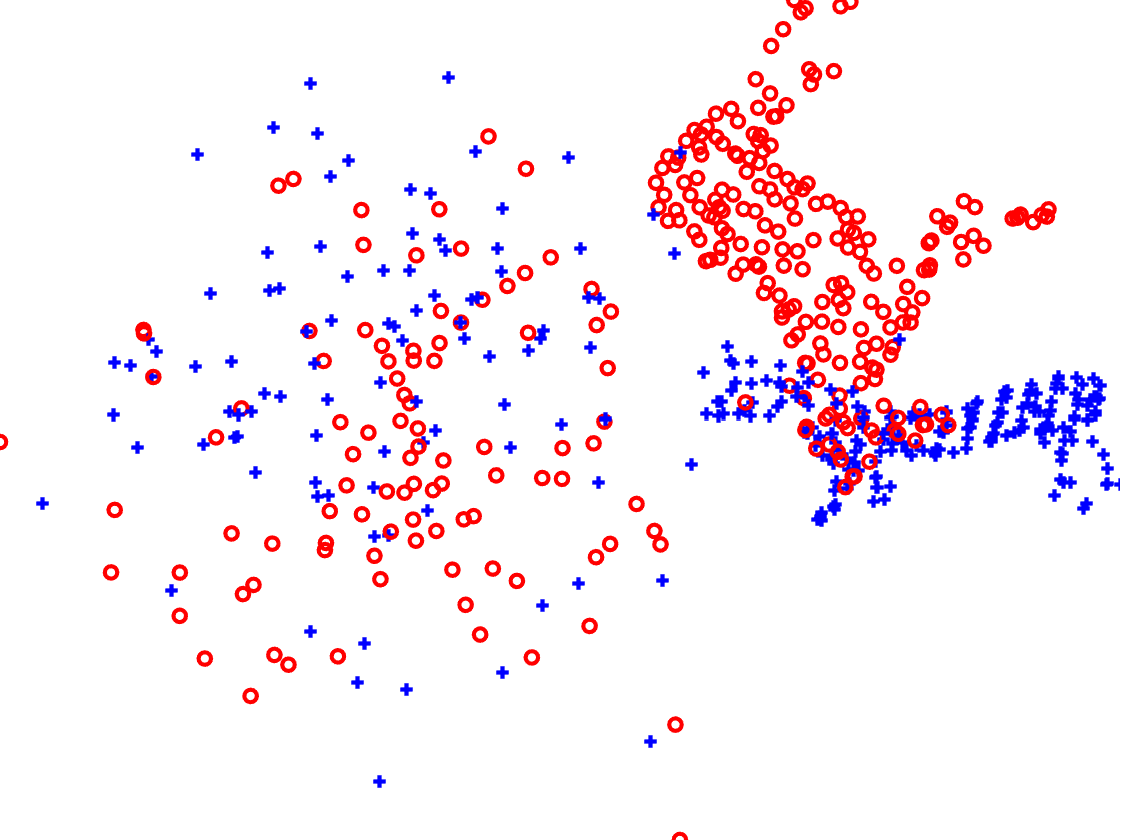}&	
		\includegraphics[height=\height,width=\Scale\linewidth]{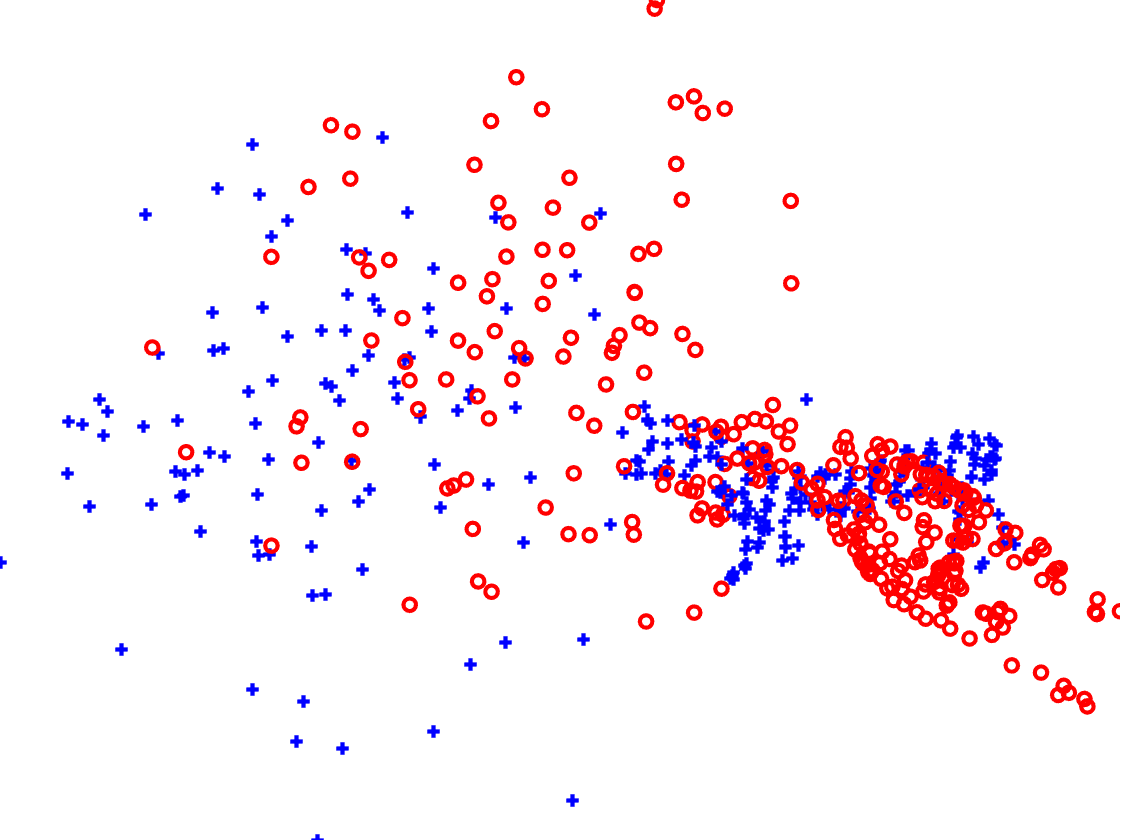}
		\\
		\hline
		\subfigure[RPM-HTB]{
			\includegraphics[height=\height,width=\Scale\linewidth]{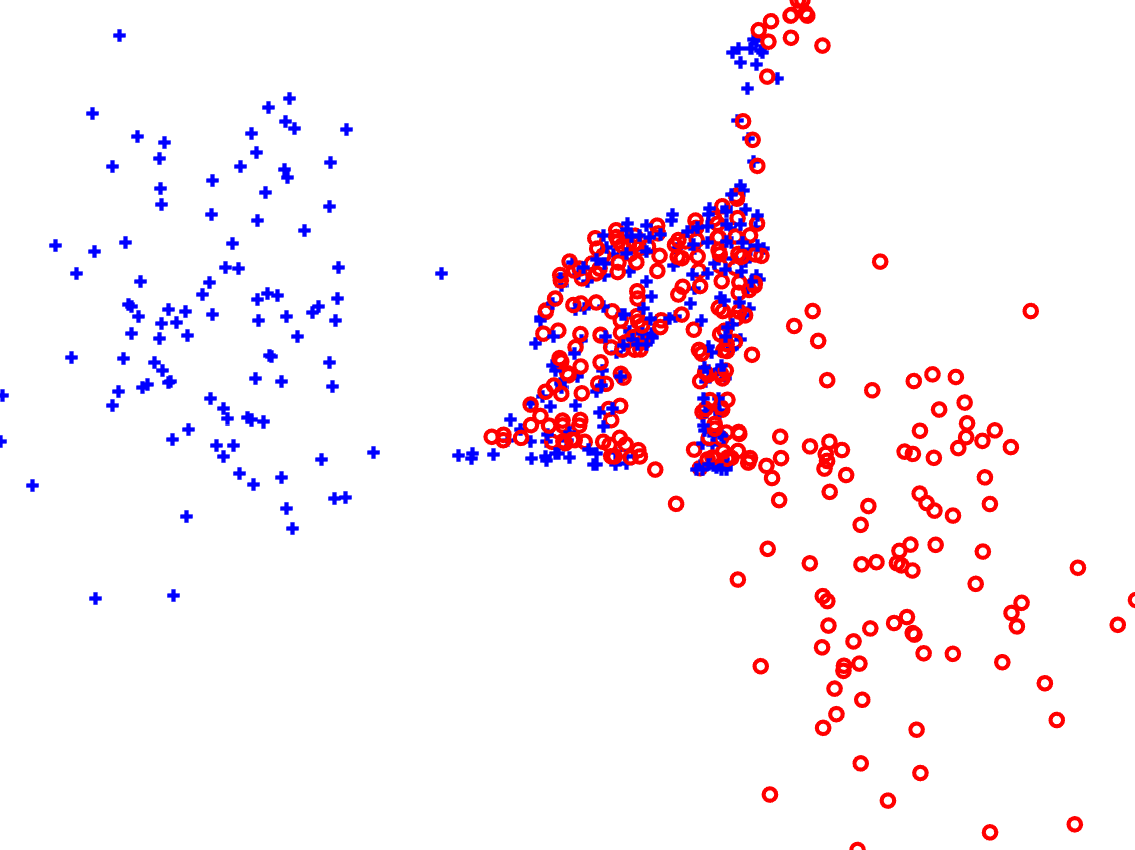}
		}&\subfigure[Go-ICP]	{
			\includegraphics[height=\height,width=\Scale\linewidth]{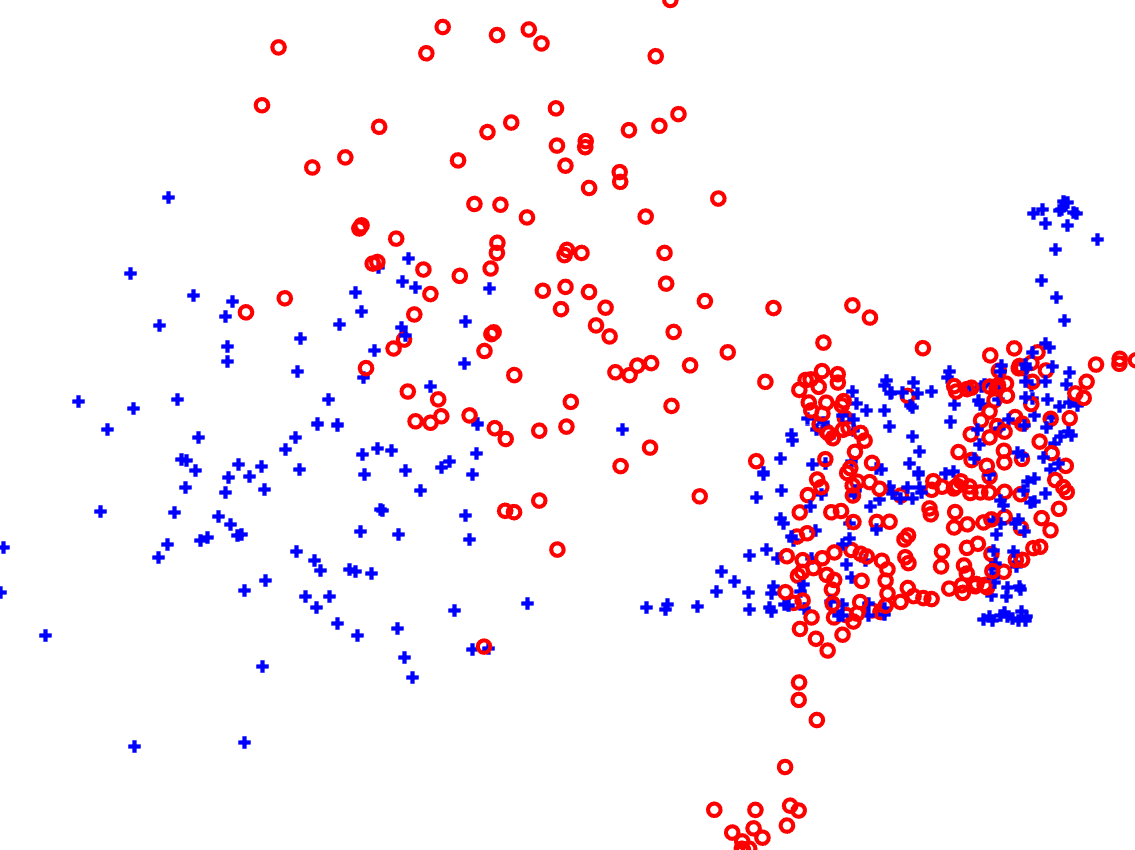}
		}&\subfigure[FRS]{
			\includegraphics[height=\height,width=\Scale\linewidth]{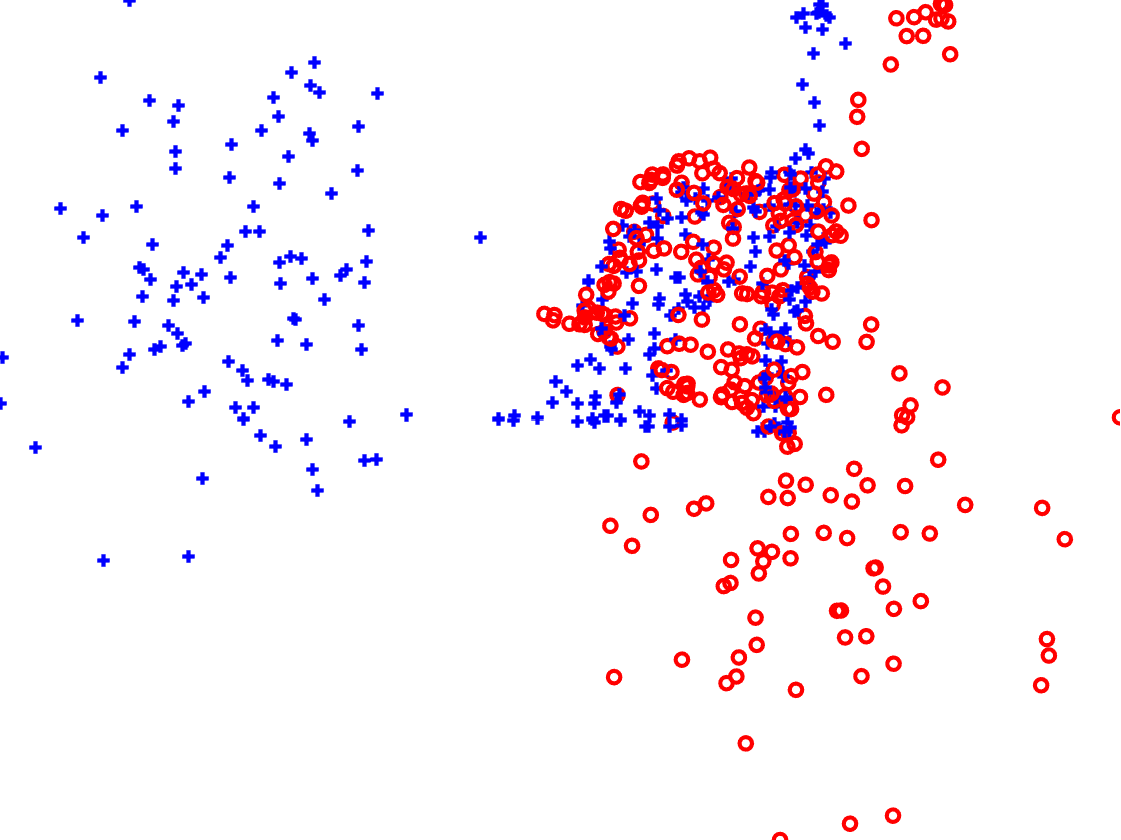}
		}&\subfigure[GORE]{
			\includegraphics[height=\height,width=\Scale\linewidth]{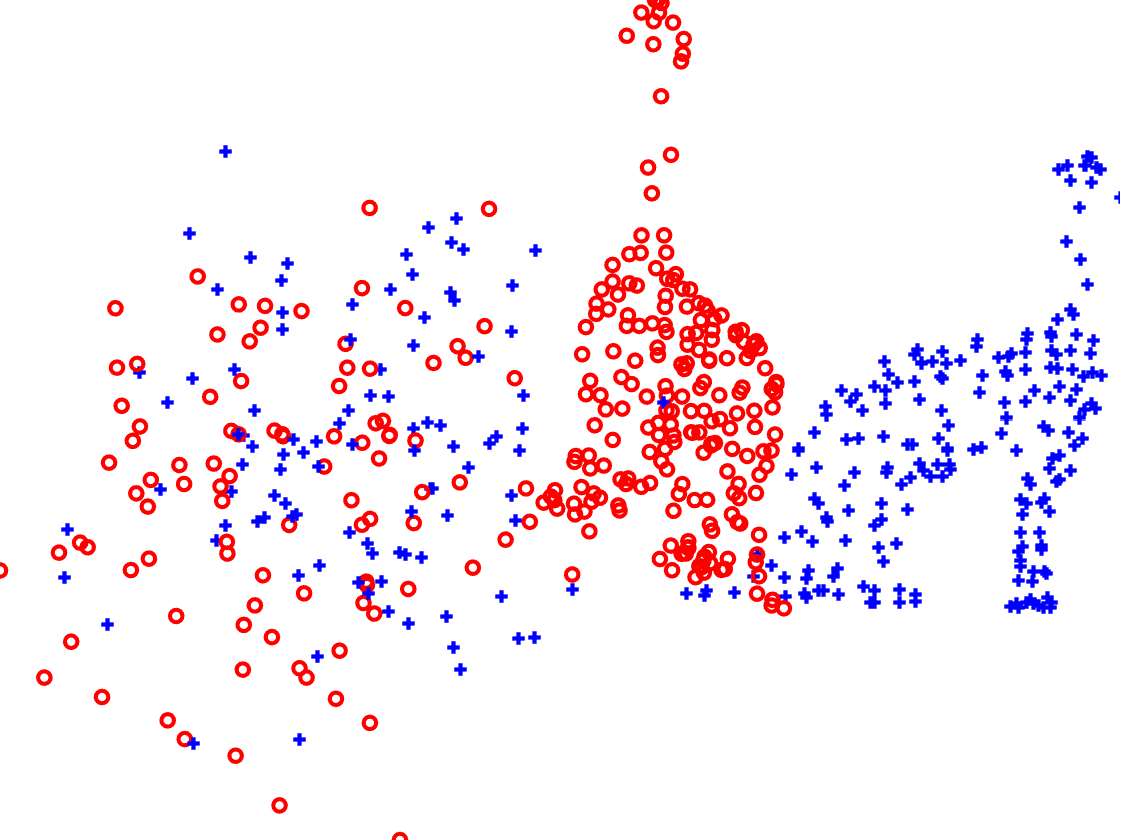}
		}&	\subfigure[TEASER++]{
			\includegraphics[height=\height,width=\Scale\linewidth]{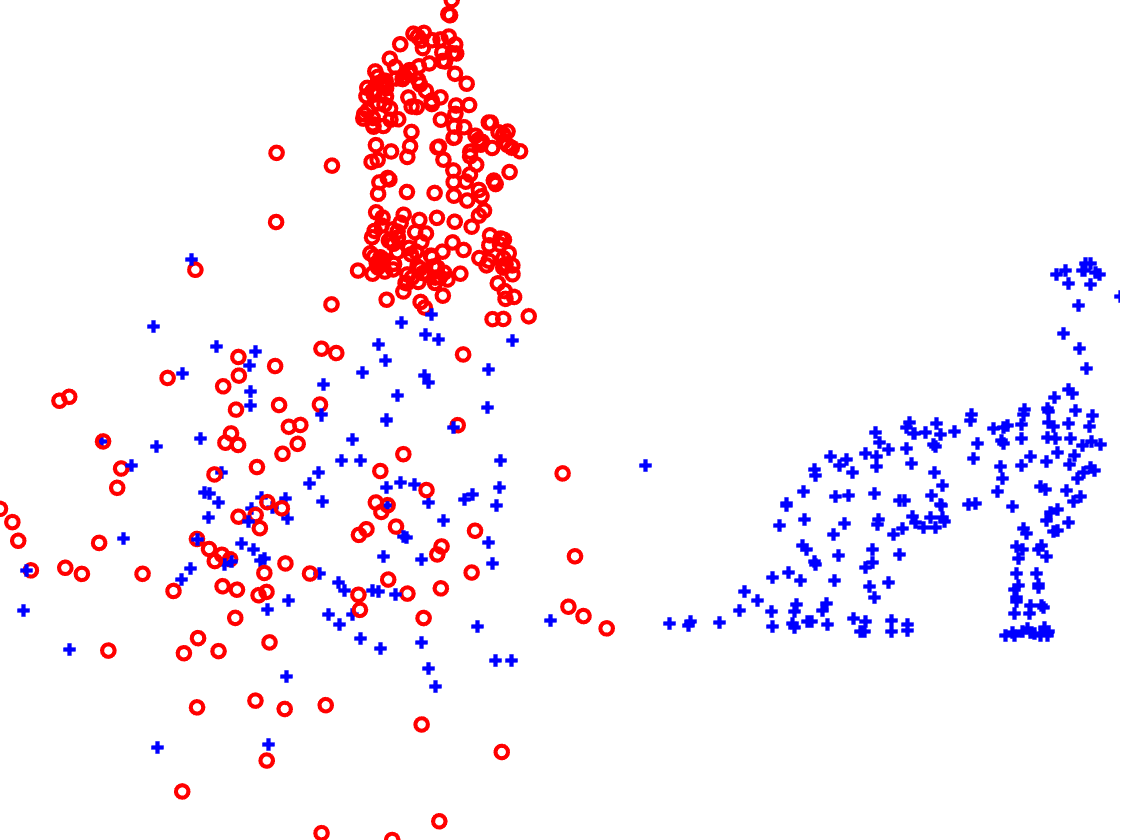}	
		}
		%		\\\hline
		%\includegraphics[width=\scale\linewidth]{figures/horse_dual_occlude_RPM_tri_exa}	&	
		%\includegraphics[width=\scale\linewidth]{figures/horse_dual_occlude_Go_icp_exa}&
		%\includegraphics[width=\scale\linewidth]{figures/horse_dual_occlude_FRS_exa}&
		%\includegraphics[width=\scale\linewidth]{figures/horse_dual_occlude_gore_exa}&	
		%\includegraphics[width=\scale\linewidth]{figures/horse_dual_occlude_teaser_exa}
		%		\\\hline
		%		\subfigure[RPM-HTB]{		\includegraphics[width=\scale\linewidth]{figures/dino_dual_occlude_RPM_tri_exa}}&
		%		\subfigure[Go-ICP]	{	\includegraphics[width=\scale\linewidth]{figures/dino_dual_occlude_Go_icp_exa}}&
		%		\subfigure[FRS]{		\includegraphics[width=\scale\linewidth]{figures/dino_dual_occlude_FRS_exa}}&
		%\subfigure[GORE]{\includegraphics[width=\scale\linewidth]{figures/dino_dual_occlude_gore_exa}}&	
		%\subfigure[TEASER++]{
			%\includegraphics[width=\scale\linewidth]{figures/dino_dual_occlude_teaser_exa}}
	\end{tabular}	
	\caption{
		Examples of  registration results  by different methods in  the separate  outliers and inliers test,
		where the 		$n_p$ values of 
		RPM-HTB  and Go-ICP 
		are both chosen as  the ground truth.
		\label{rot_3D_syn_match_exa}}
\end{figure*}

\begin{figure*}[t]
%%%%%%%%%%%%%%%%%%%%%%%%%%%%%%%%
	\centering
	\newcommand\scaleGd{0.204}
	\begin{tabular}{@{\hspace{-0mm}}c@{\hspace{0mm}}c@{}c@{}c@{} c }
		\includegraphics[width=\scaleGd\linewidth]{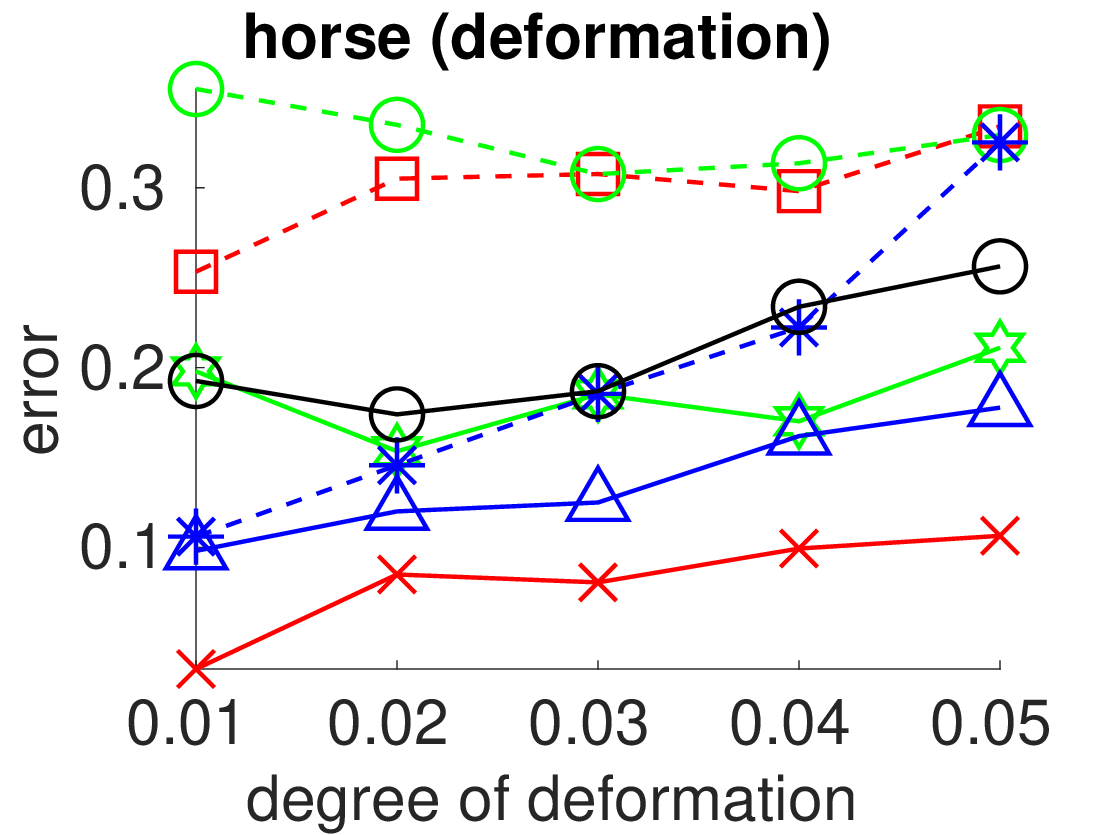}&	
		\includegraphics[width=\scaleGd\linewidth]{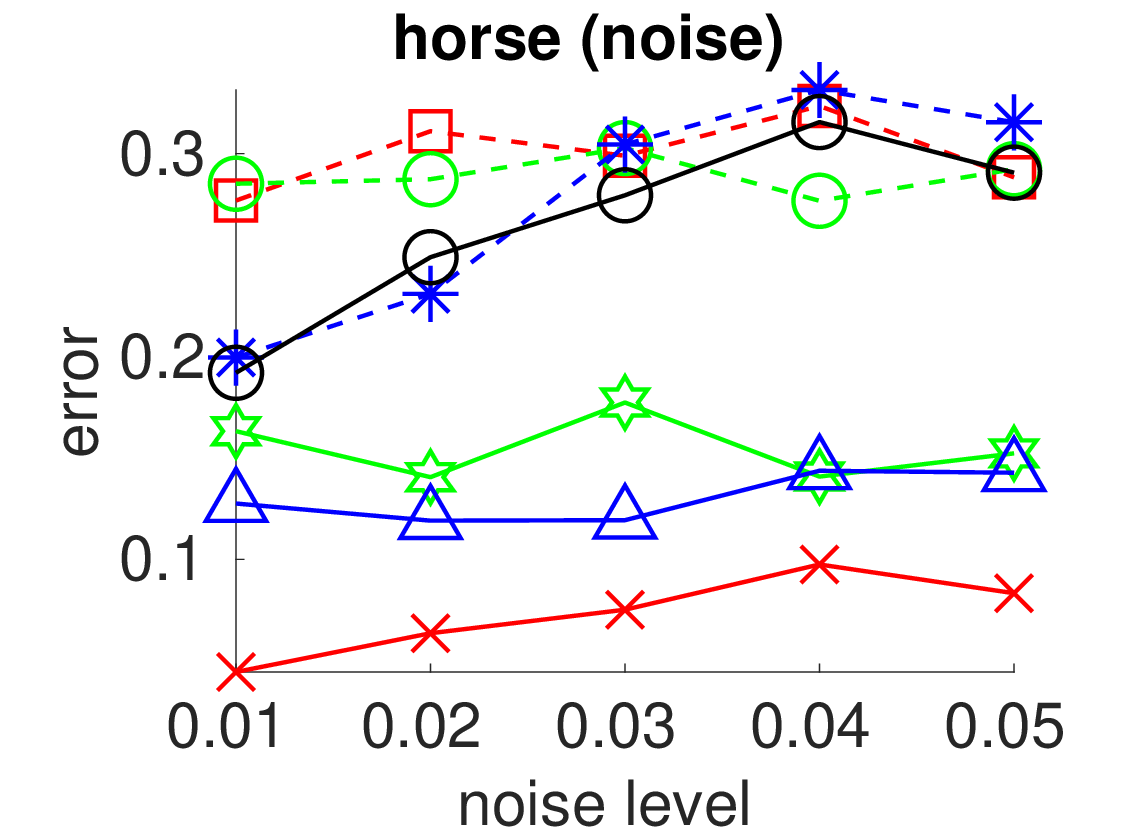}&		
		\includegraphics[width=\scaleGd\linewidth]{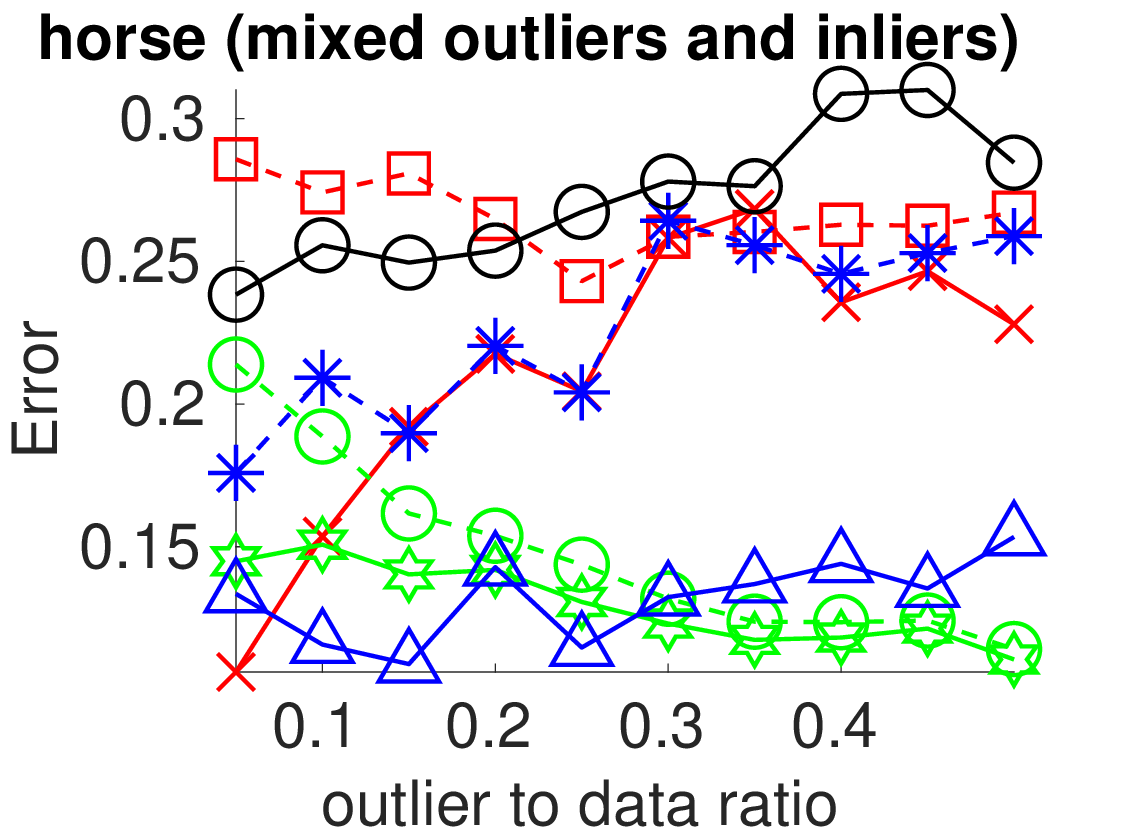}&
					
		\includegraphics[width=\scaleGd\linewidth]{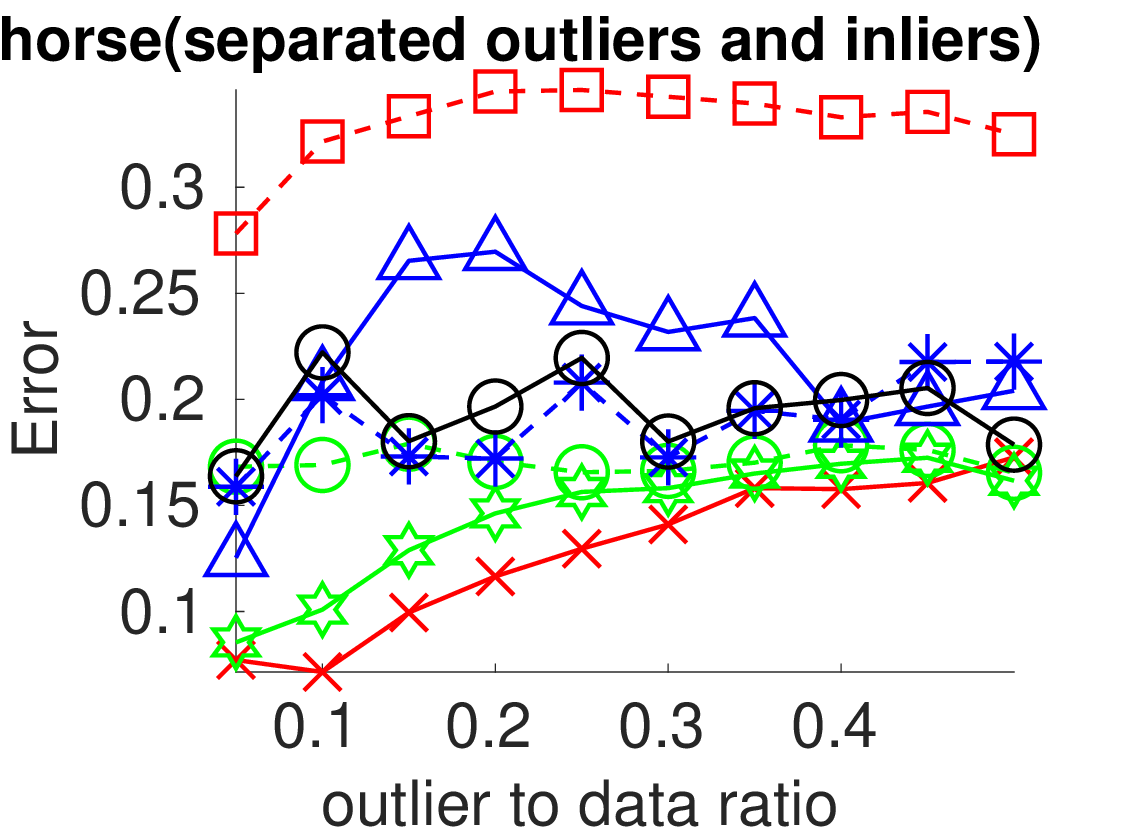}&
		\includegraphics[width=\scaleGd\linewidth]{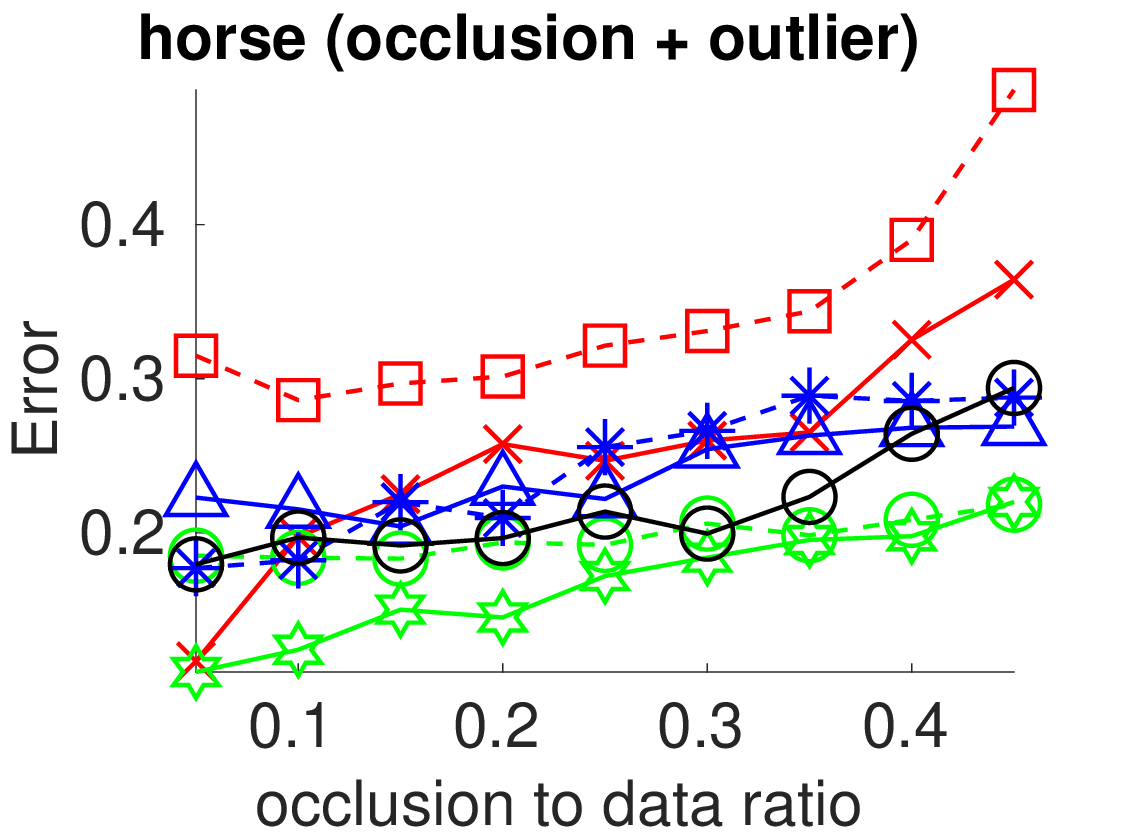}\\
		
		\includegraphics[width=\scaleGd\linewidth]{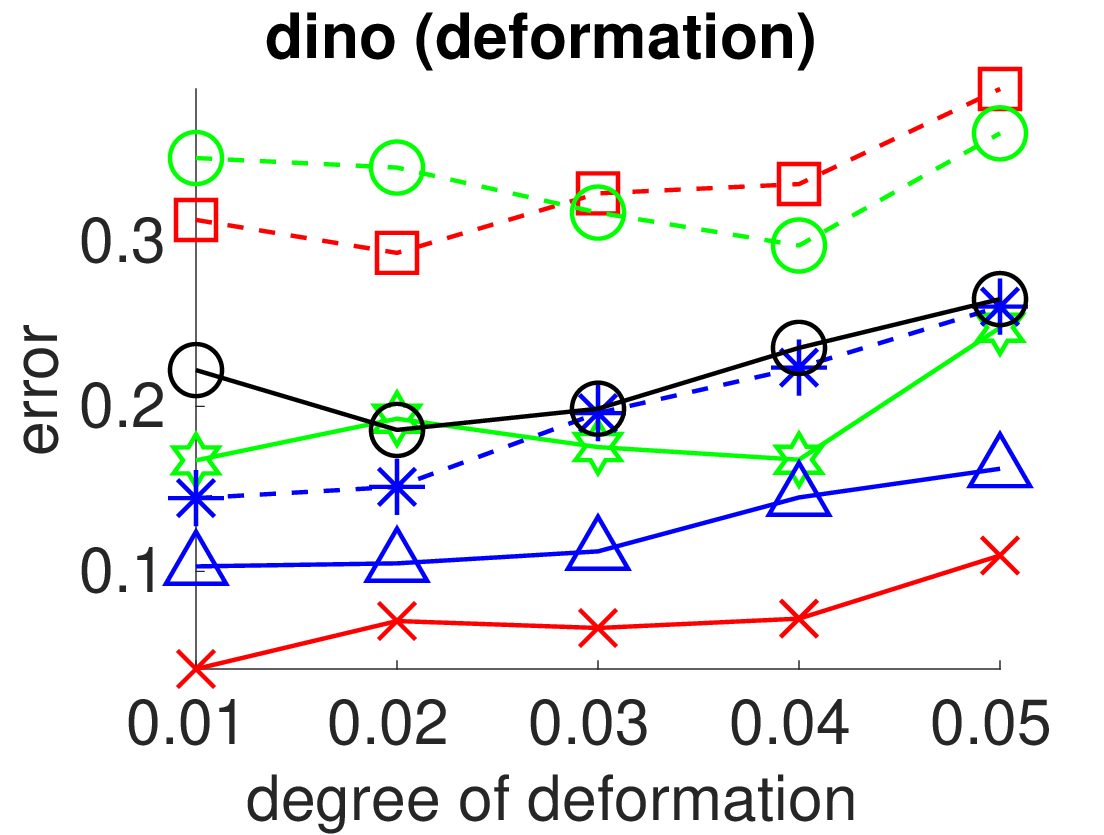}&
		\includegraphics[width=\scaleGd\linewidth]{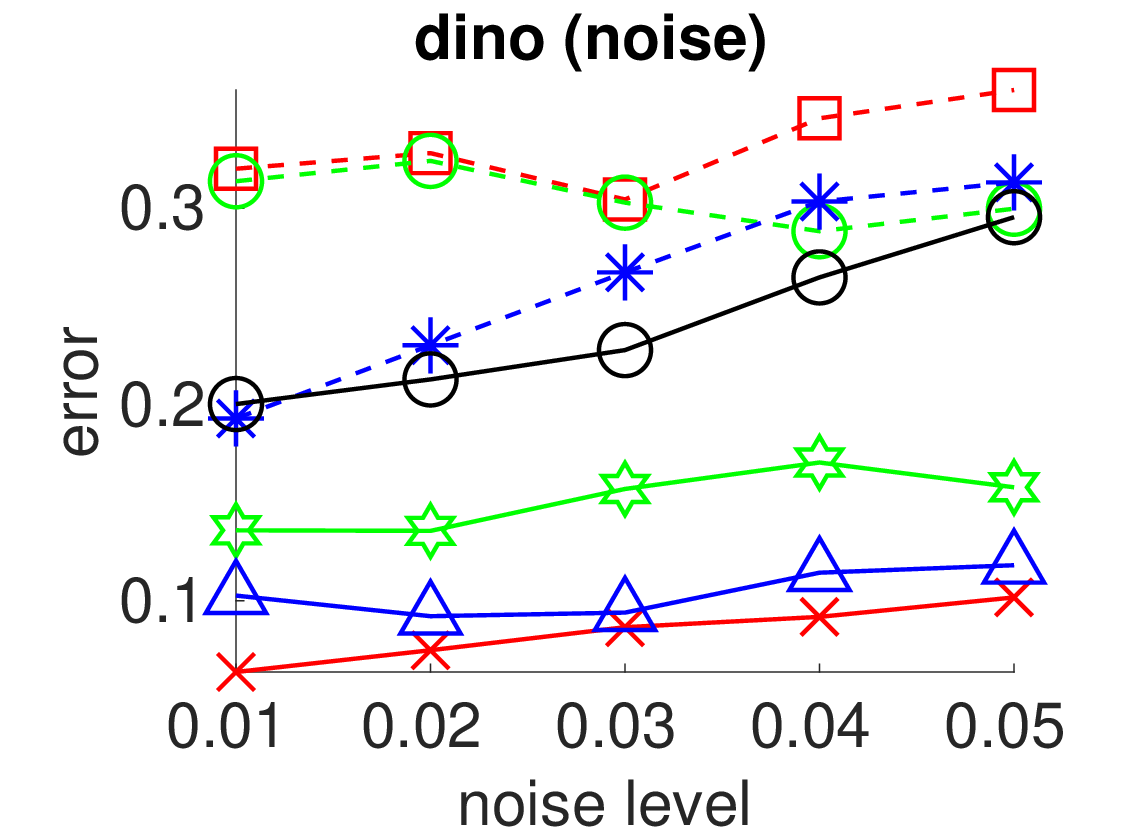}&
			\includegraphics[width=\scaleGd\linewidth]{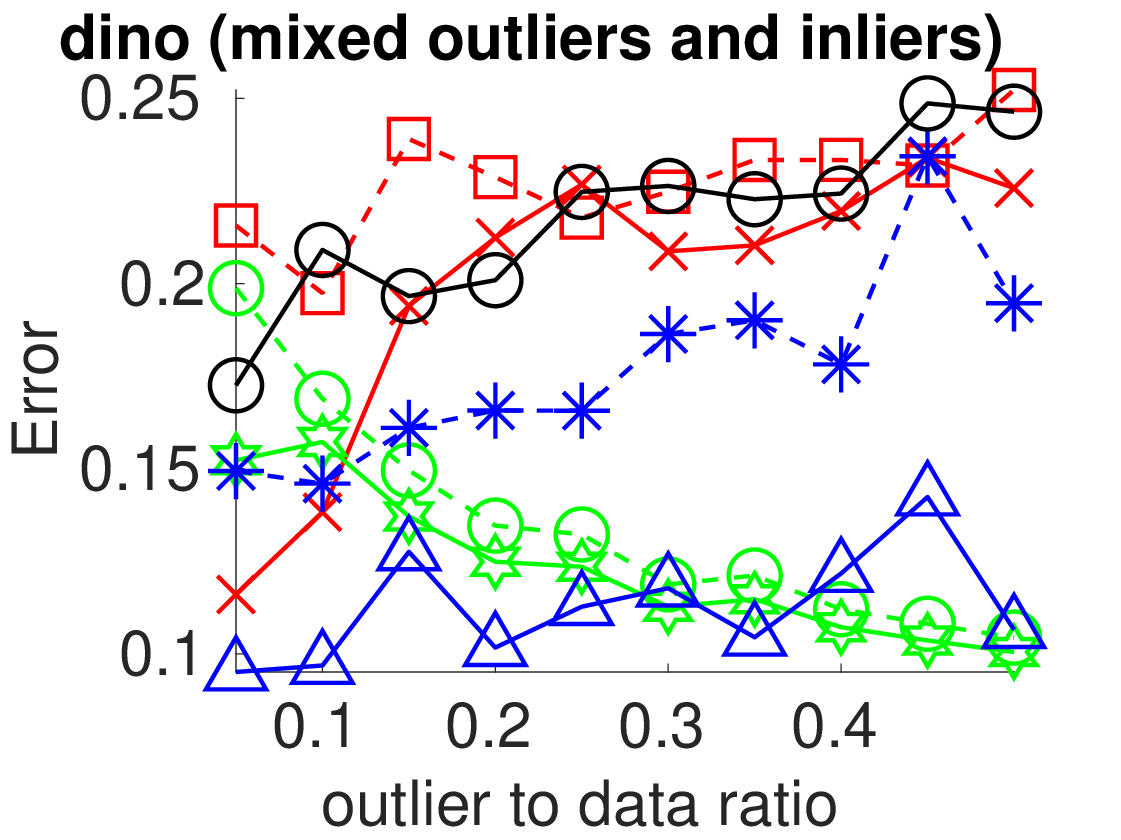}&		
		\includegraphics[width=\scaleGd\linewidth]{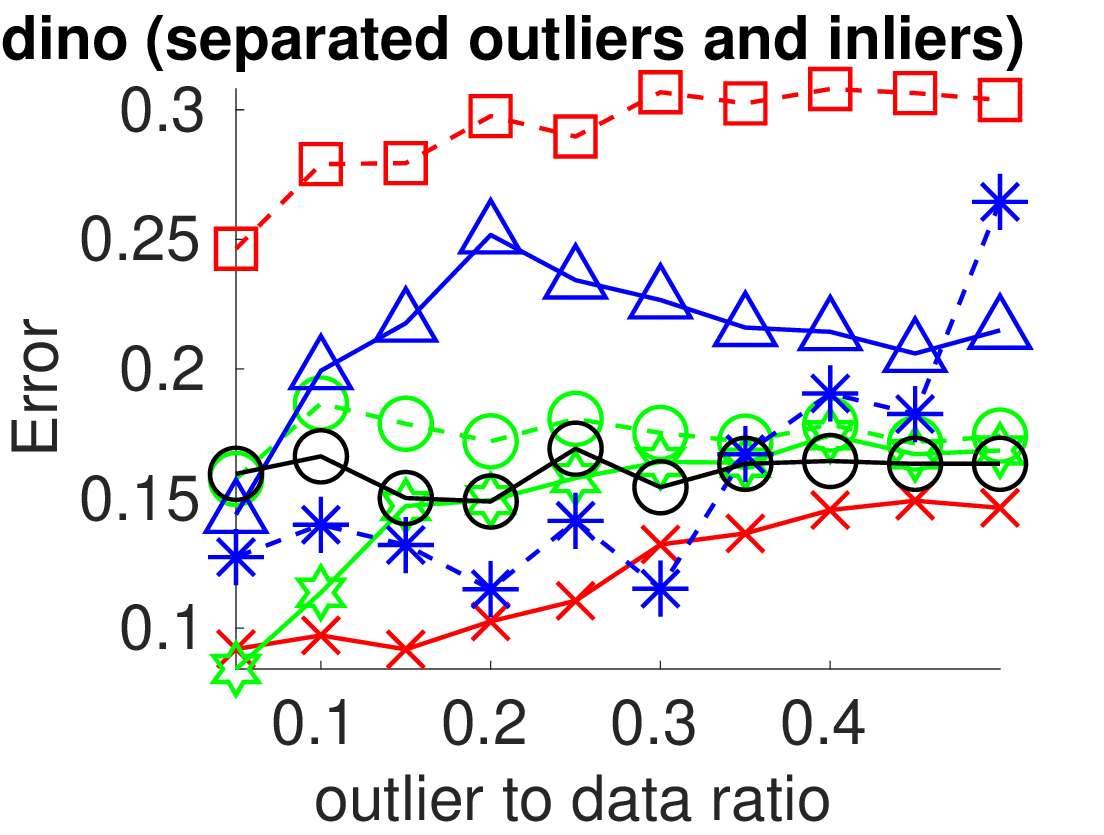}&
		\includegraphics[width=\scaleGd\linewidth]{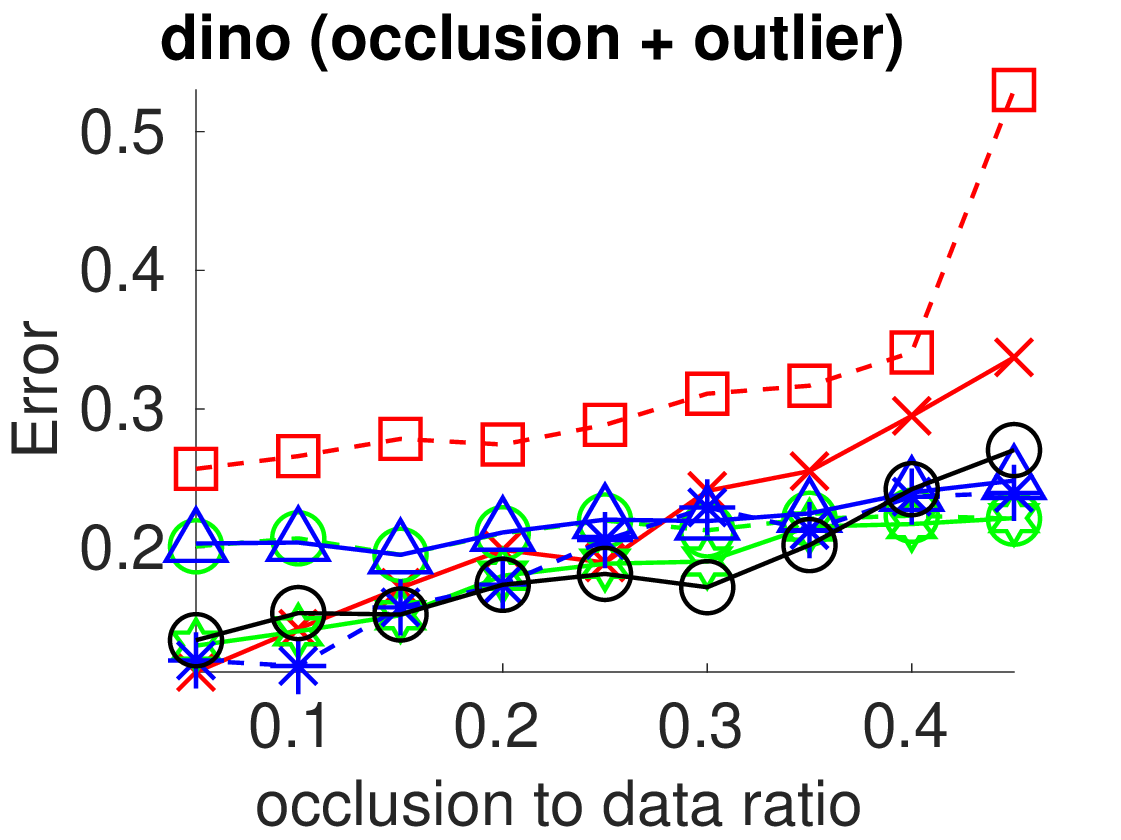} 
	\end{tabular}
	\begin{tabular}{@{\hspace{-0.1mm}}c}
	\includegraphics[width=.6\linewidth]{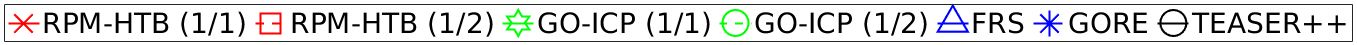}		
\end{tabular}	
\begin{tabular}{@{\hspace{-0mm}}c@{\hspace{0mm}}c@{}c@{}c@{} c }
\includegraphics[width=\scaleGd\linewidth]{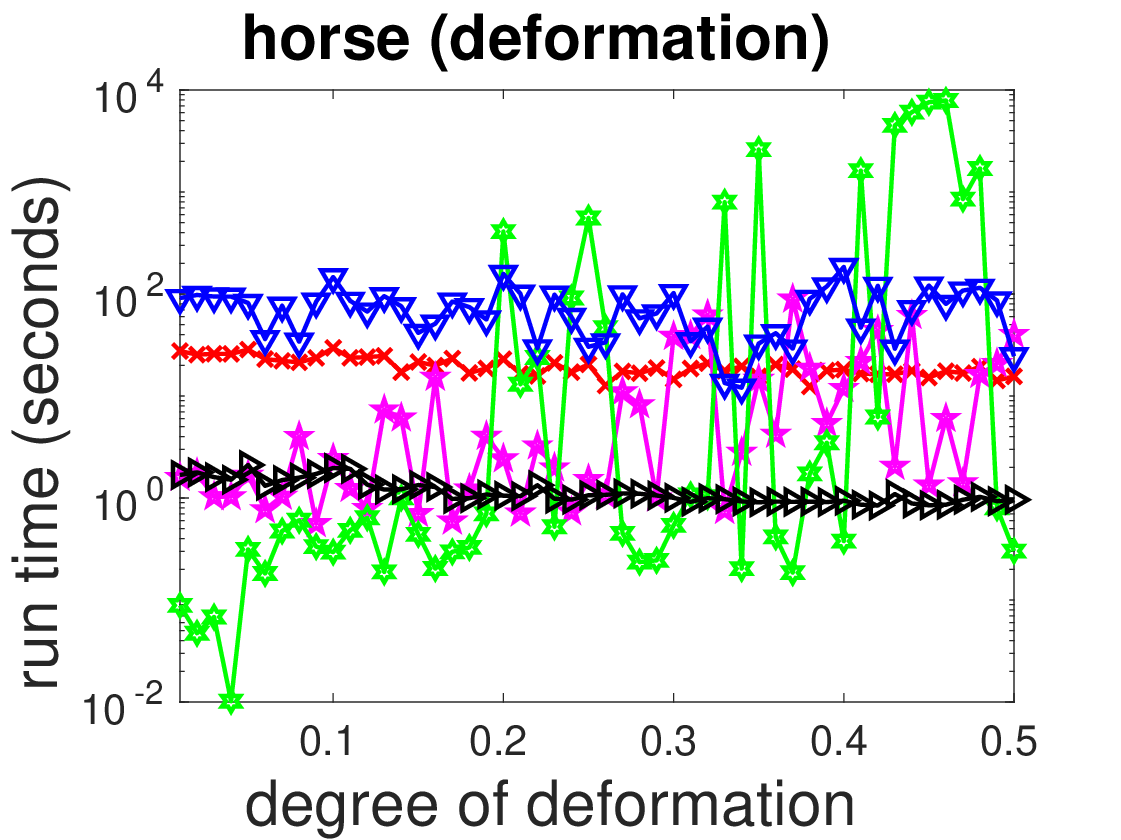}&
\includegraphics[width=\scaleGd\linewidth]{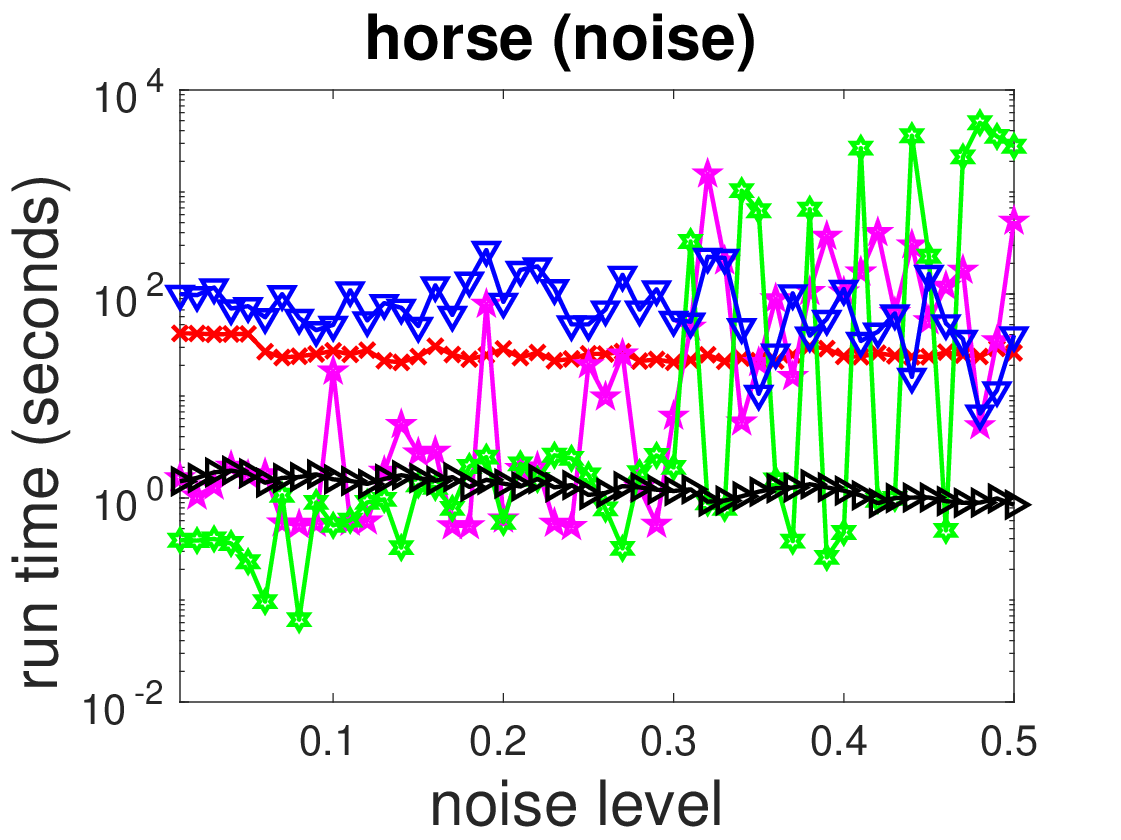}&
\includegraphics[width=\scaleGd\linewidth]{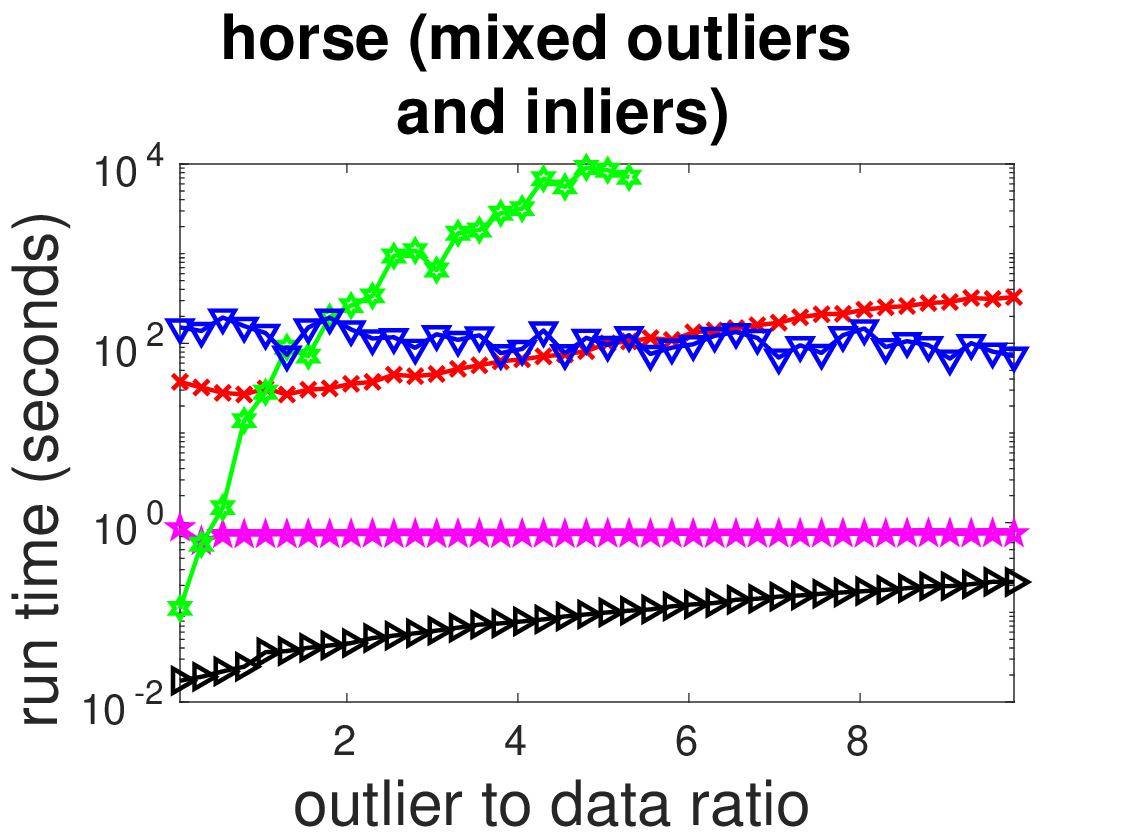}&
\includegraphics[width=\scaleGd\linewidth]{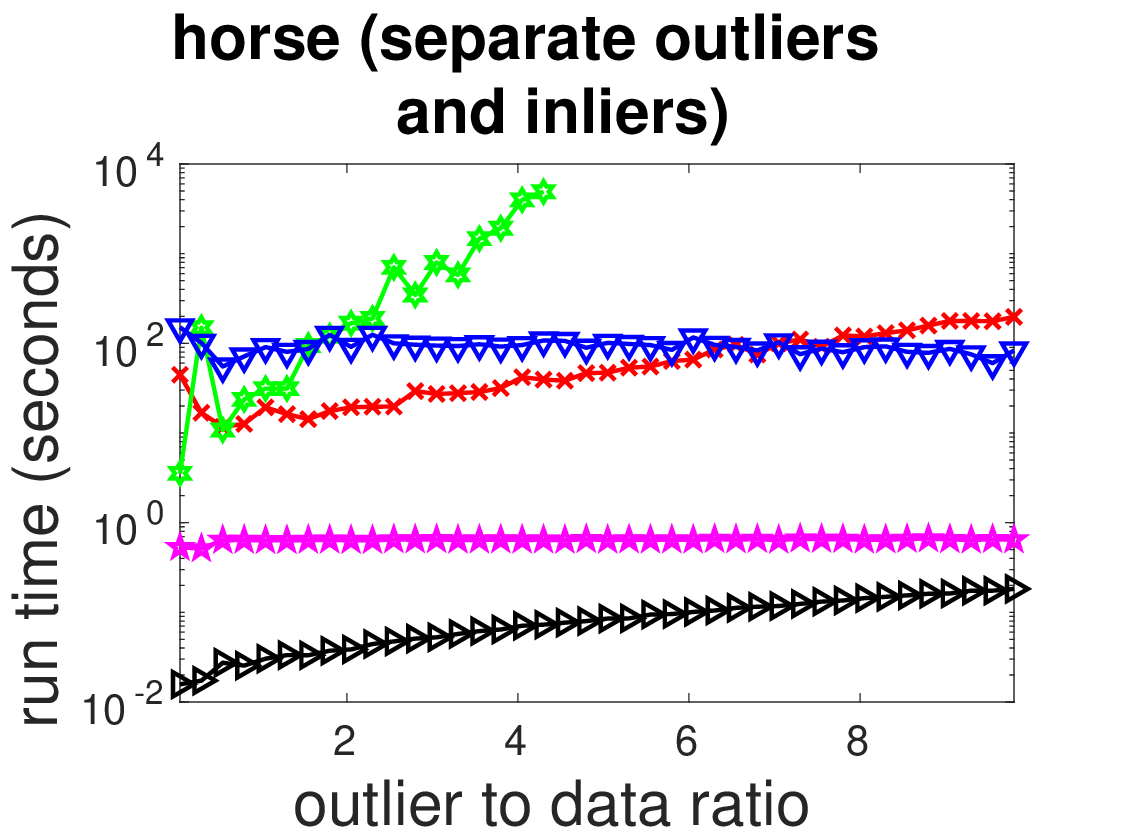}	&
\includegraphics[width=\scaleGd\linewidth]{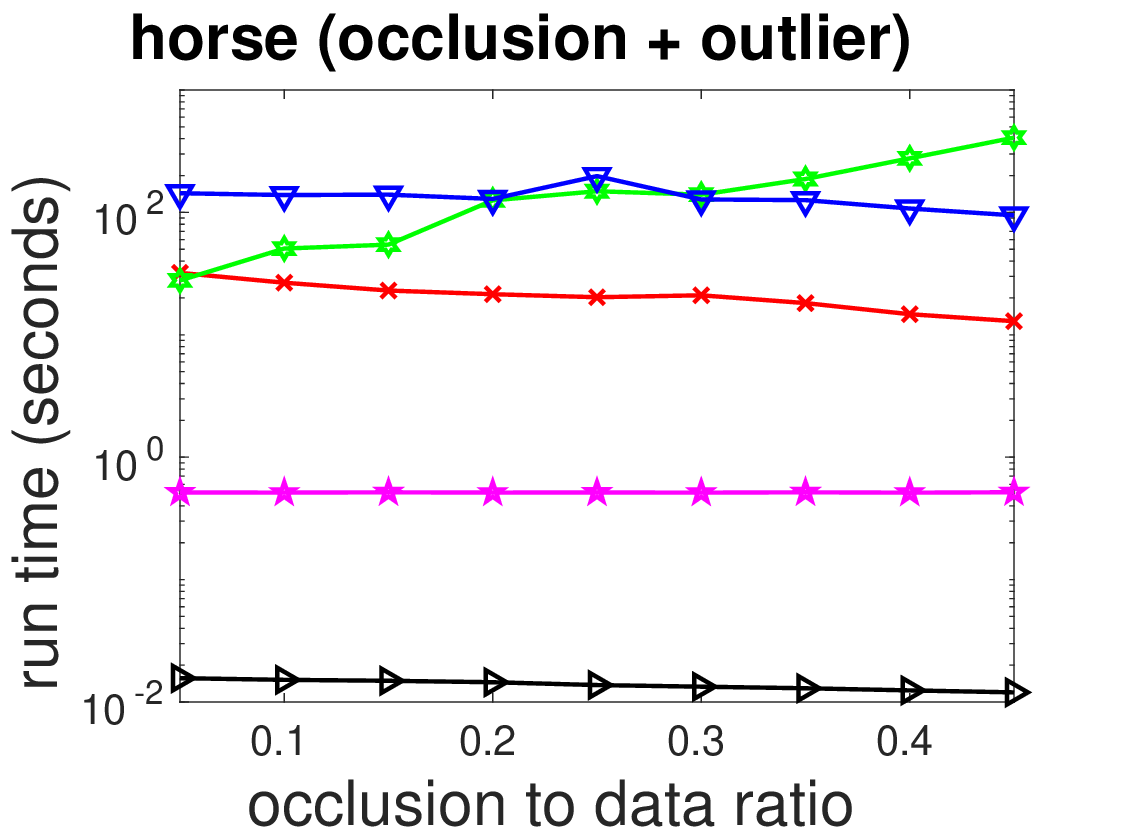}		
	\end{tabular}
	\begin{tabular}{@{\hspace{-0.1mm}}c}
	\includegraphics[width=.4\linewidth]{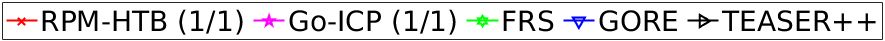}		
\end{tabular}
	\caption{Average  registration errors (top 2 rows) and run times (bottom row) of RPM-HTB %$with  maximum branching depth of  $10$
		and Go-ICP under different $n_p$ values (chosen from $1/2$ to $1/1$ the ground truth value), 
		FRS, GORE and TEASER++
		over 100 random trials
		for the 3D deformation, noise,
		mixed outliers and inliers,
		 separate  outliers and inliers
		 and occlusion+outlier tests.
For the two types of outlier tests,
only part of FRS’s results is reported since FRS becomes too slow when problem size is too big.		 
		\label{3D_rigid_sta}}
\end{figure*}

Analogous to the experimental setup in Sec.\ \ref{subsec:exp_case_one},
5 types of tests  are conducted 
to evaluate a  method's robustness to various types of disturbances:
% against outliers:
\begin{inparaenum}[\upshape i\upshape)]
	\item \textit{Deformation test},
	\item \textit{Noise test},	
	\item \textit{Mixed outliers and} \textit{inliers test},
	\item \textit{Separate outliers and} \textit{inliers test}
	%Normally distributed random outliers are added to 
	%different sides of the prototype shape to generate  two point sets
	%so as to simulate outlier disturbance. %, respectively, 
	and
	\item \textit{Occlusion+outlier test},
\end{inparaenum}
%as illustrated in  Fig. \ref{rot_3D_test_data_exa}.
%Different from Section \ref{subsec:3Dtest_nonrot}, 
%random rotation around the z-axis and uniform scaling within range $[0.5,1.5]$
%is  applied to the prototype shape when generating the model point sets.
%so as to test a method's ability to cope with arbitrary similarity transformation.
%The same two 3D prototype shapes as used in subsection \ref{subsec:regu3Dtest}
%% as shown in the left column of Fig.  \ref{three_test_data_exa},
%are used as the prototype shape, respectively.
%
as is illustrated in Fig. \ref{rot_3D_test_data_exa}.
 Examples of registration  by different methods  are shown  in Fig. \ref{rot_3D_syn_match_exa}.

The  registration errors  by different  methods are presented in the top 2 rows of
Fig. \ref{3D_rigid_sta}.
The results indicate that
compared with other methods,
RPM-HTB with  $n_p$ value  chosen as the ground truth
is robust to deformation, noise and  outliers in case when outliers are separate from  inliers,
but is not robust when outliers are mixed with inliers.
%performs overall the best.
%demonstrating its robustness to various types of disturbances.
In comparison,
due to use of features,
GORE and TEASER++ are not good at handling  deformation, noise
and outliers in case when outliers are mixed with inliers.
%
%For the occlusion+outlier test,
%RPM-HTB  performs similarly to Go-ICP until
%occlusion becomes severe.
%%
%This indicates that our method is more suitable  for the problems where  occlusion is not  severe.
%
In terms of different choices of $n_p$,
%our method is relatively sensitive to the value of $n_p$:
RPM-HTB  with $n_p$ value close to the ground truth performs much better.
%

%\begin{table}[h]
%	\renewcommand{\tabcolsep}{0.2cm}
%	\centering
%	\caption{
%		Average run  time (in seconds).
%	}
%	\label{time_3d}
%	\small
%	\begin{tabular}{cccccc} 
%		\hline
%& deform. &noise &
%\begin{tabular}{@{}c@{}}mixed \\ outl./inl.\end{tabular}				
%&\begin{tabular}{@{}c@{}}separate\\
%	outl./inl.\end{tabular}	
%&occl.
%\\\hline
%		%		RPM-HTB ($\epsilon_0=6$, depth=$\infty$) & 6.5 & 6.1   & 185.5 & 16.1\\\hline		
%		RPM-HTB   & 38.2 & 39.7 & 28.1  & 25.9 &24.1\\\hline
%		Go-ICP  & 0.9 & 3.5 &0.6 &   0.5 &0.5\\\hline
%		FRS  & 0.8  &  0.3 &1.2 &   239.2 & 163.6\\\hline		
%		GORE &    90.7 & 89.4 & 116.7 &92.3 & 115.2\\\hline
%		TEASER++&		1.4 &1.4 &0.02 &3.4& 0.7 \\\hline
%	\end{tabular} 
%\end{table} 

The average run  times by different methods
are presented in the bottom row of Fig. \ref{3D_rigid_sta}.
TEASER++ is the fastest, 
followed by Go-ICP and  then  RPM-HTB.
GORE is slow in most instances.
%for most of the tests.
FRS quickly becomes inefficient when registration problem becomes challenging (e.g., when  outliers increases in the two types of outlier tests).

From the two types of outlier tests,
we can also  see how different methods scale with problem size.
Go-ICP and GORE  scale the best with problem size,
followed by TEASER++ and  RPM-HTB.
FRS scales the worst with problem size.

%One note that GORE is generally less  efficient than RPM-HTB.
%One also notes that FRS's running time is rather irregular compared with other methods.
%although RPM-HTB is more efficient than GORE under small to medium problem size.

%which partl

%\begin{figure}[h]
%	\centering
%	
%	
%
%	\caption{\textcolor{black}{
%			Average	run times of different methods in the separate outliers and inliers test when using the horse shape.
%			For FRS,
%			only part of the results is shown since it becomes too slow  with the increase of  problem size.	}	
%		\label{runtime3d_outlier}	}
%\end{figure}

\subsubsection{3DMatch dataset}
\begin{figure*}[t]
	\centering
	\newcommand\scale{0.215} 
	\begin{tabular}{@{\hspace{-1mm}}c@{\hspace{-3mm}}c@{\hspace{-3mm}}c@{\hspace{-3mm}}c@{\hspace{-3mm}}c}	
\includegraphics[width=\scale\linewidth]{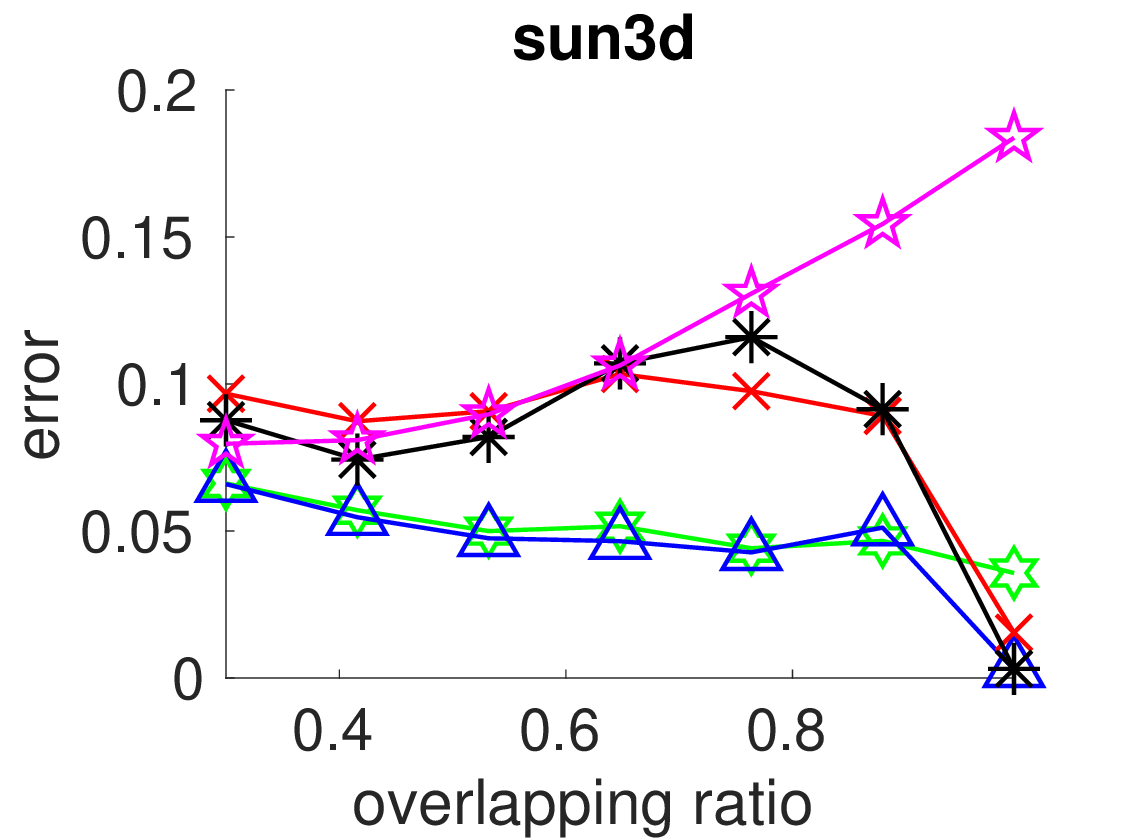}	&
\includegraphics[width=\scale\linewidth]{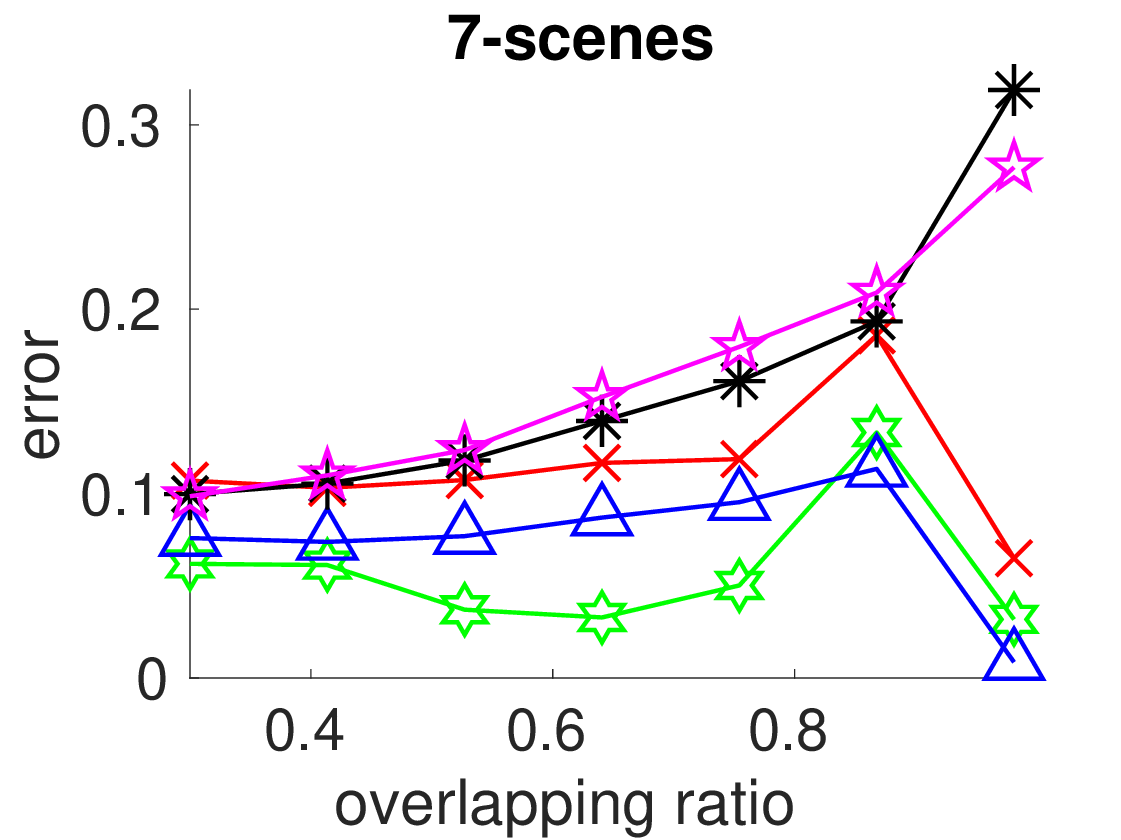} &
\includegraphics[width=\scale\linewidth]{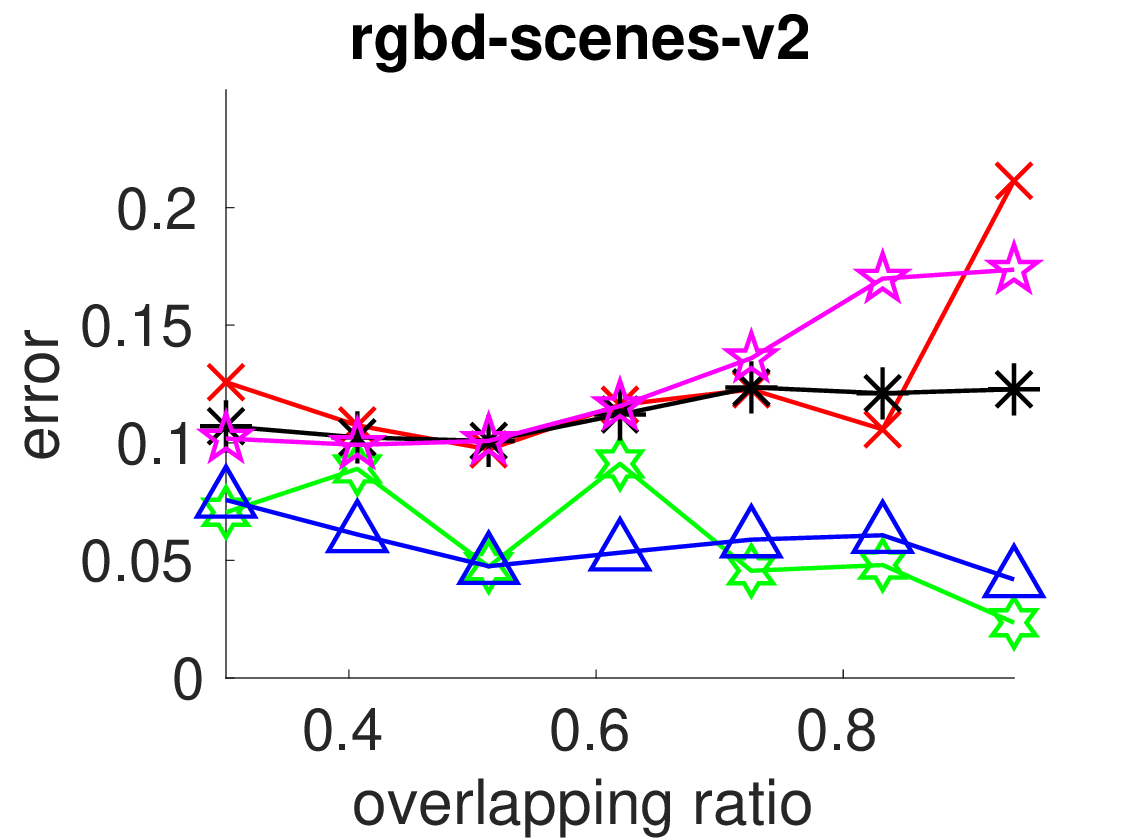} &
%\end{tabular}\\
%\begin{tabular}{c@{} c }
	\includegraphics[width=\scale\linewidth]{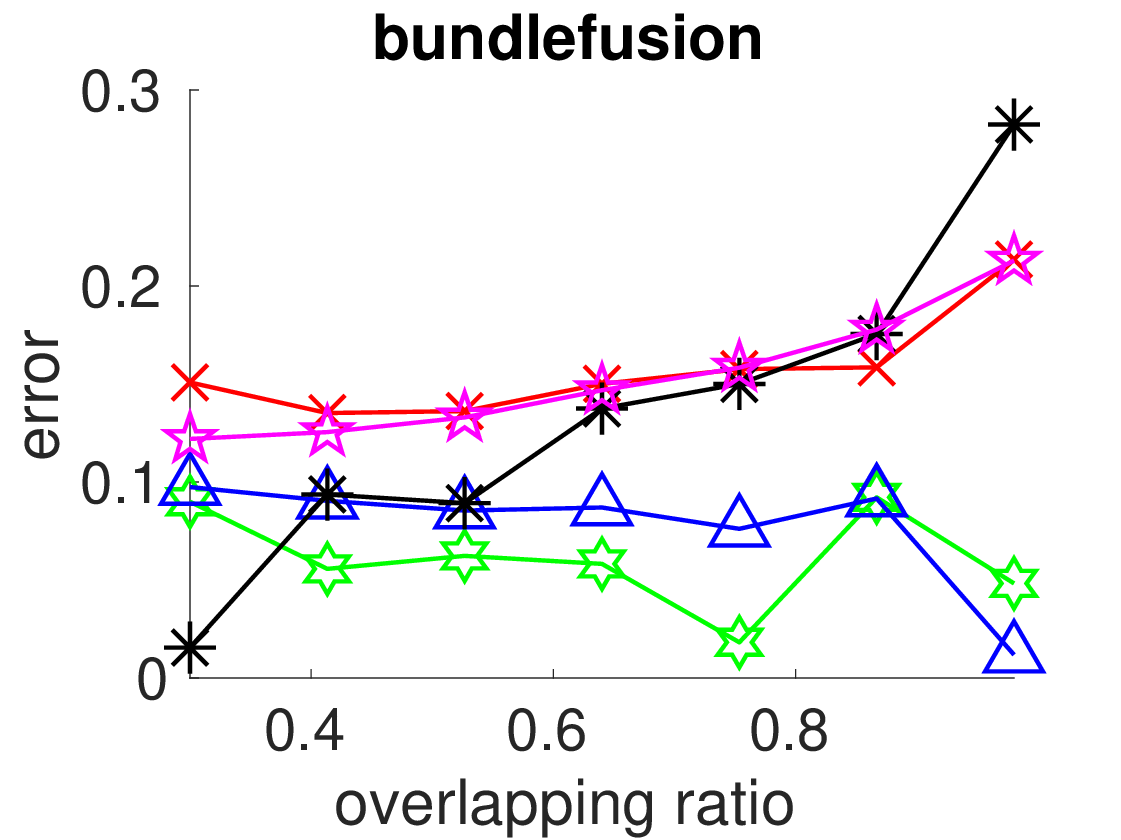}	&
		\includegraphics[width=\scale\linewidth]{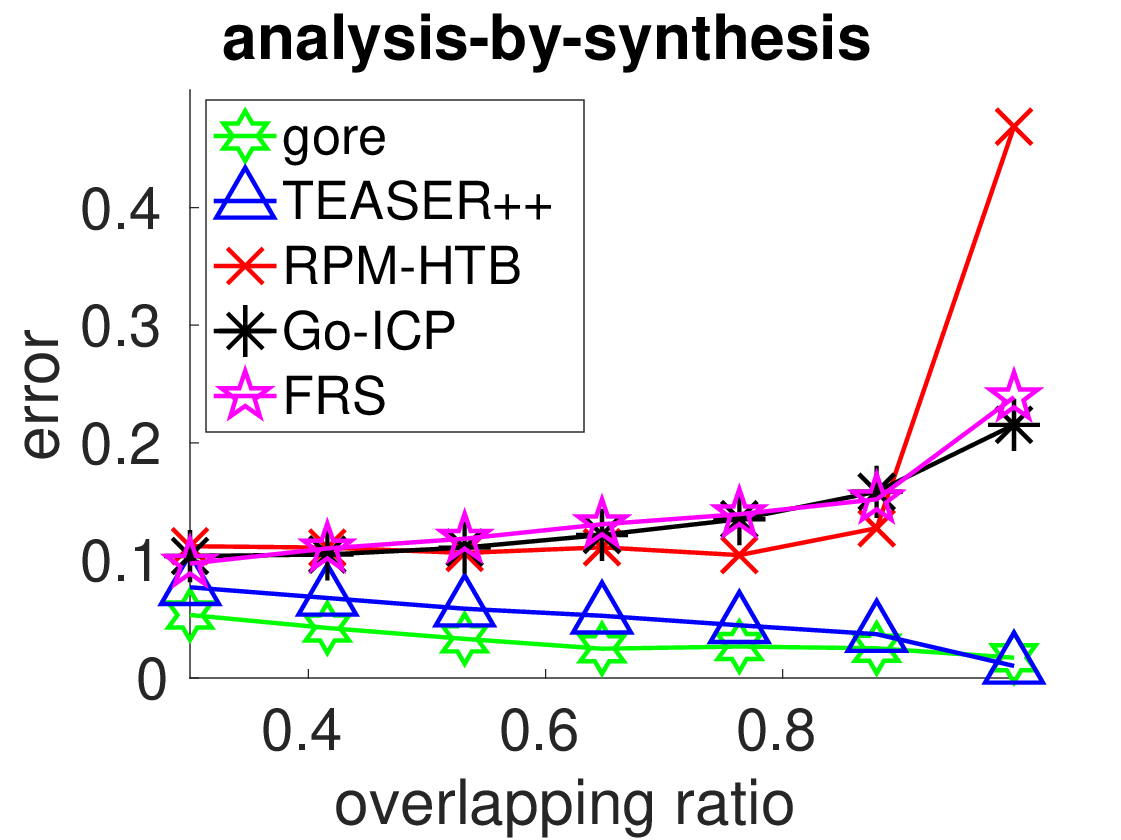}
		\end{tabular}

	\caption{
%		Top-left subfigure:
%average matching errors by different methods versus different RGB-D reconstruction dataset.
%The remaining subfigures: 
Average registration errors by different methods 
%versus different overlapping ratios 
on the five RGB-D reconstruction datasets. 
%in the 3DMatch benchmark.		 		
\label{3DMatch_tests}	}
\end{figure*}

\begin{figure*}[h]
	\centering
	\newcommand\scale{0.9}	
\includegraphics[width=\scale\linewidth]{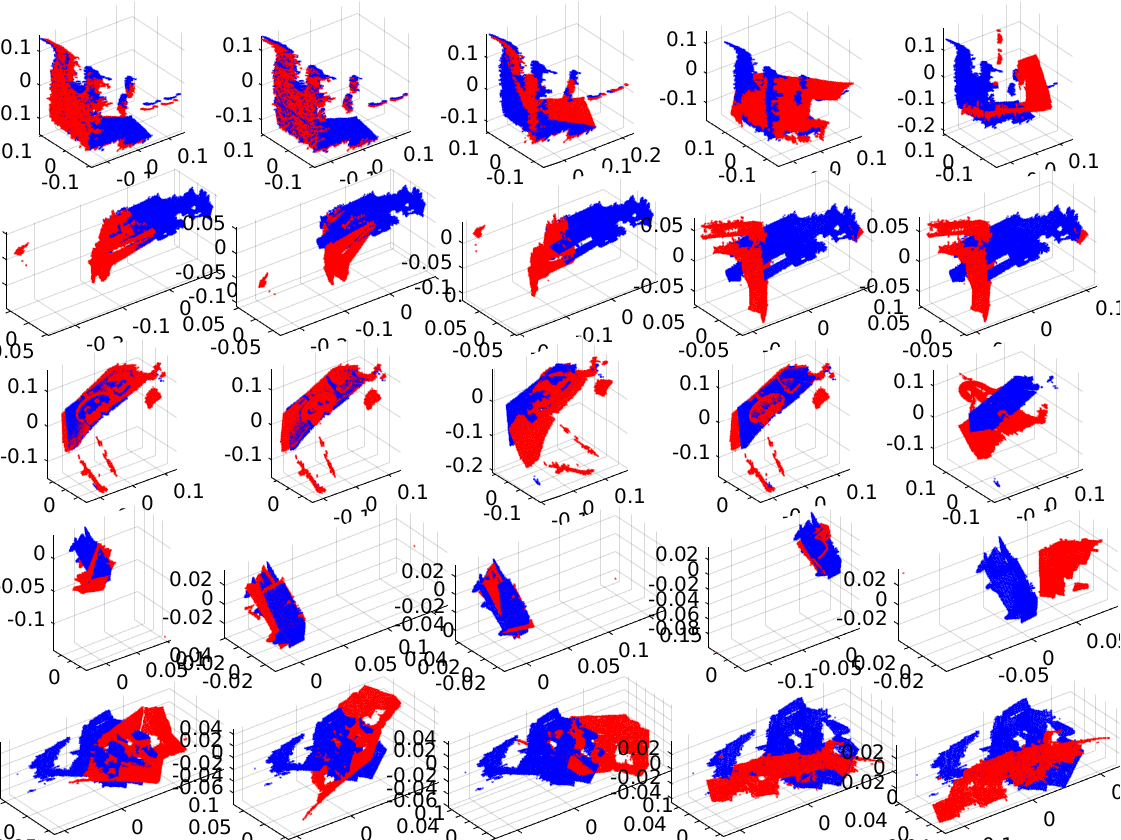}
\small
\begin{tabular}{@{\hspace{-10mm}}c @{\hspace{17 mm}} c@{\hspace{12 mm}}  c@{\hspace{16 mm}}  c@{\hspace{22 mm}}   c}	
(a) GORE & (b) TEASER++ & (c) RPM-HTB & (d) Go-ICP & (e) FRS
\end{tabular}

	\caption{
Examples of registration results generated by different methods
%(from left to right) GORE, TEASER++, RPM-HTB, Go-ICP and FRS 
on (from top to bottom) the five RGB-D reconstruction dataset: sun3d, 7-scenes, rgbd-scenes-v2, rgbd-scenes-v2 and analysis-by-synthesis.
%where the 		$n_p$ value of 
%RPM-HTB  and Go-ICP 
%is  chosen as  the ground truth.
\label{3DMatch_exa}	}
\end{figure*}

3DMatch benchmark \cite{3Dmatch_dataset} contains scans of 62  scenaries coming from five existing RGB-D reconstruction datasets (sun3d, 7-scenes, rgbd-scenes-v2, bundlefusion and analysis-by-synthesis).
%where 54  scenaries are used for training and 8 scenaries are used for evaluation.
%3DMatch use 5 existing RGB-D reconstruction dataset to train the model.
%contains 5 datasets  of 62 scenaries,
%where 54 scenaries are used for training.
%D3Feat provides files recording the  pairing of overlapping scenary fragments.
%
%D3Feat provides convenient ways of organizing 
%matching pairs for these scene fragments.
D3Feat \cite{D3feat_dataset} provides a unified file structure and format 
recording the  matching of these scenary scans. %fragments.
For sake of convenience,
we use the scan pairs provided by D3Feat to evaluate  performances of different  methods. 
The matching errors  by different methods are reported in Fig. \ref{3DMatch_tests}.
The results indicate that RPM-HTB performs overall 
better than Go-ICP and FRS,
but worse than GORE and TEASER++.
%We speculate that
The good performances of  GORE and TEASER++  partly comes from  their use of features, which can  significantly boost their abilities at matching complex structures.
%robustness against clutters.
In comparison, the remaining methods only use  point position information.
Besides, the 3DMatch benchmark contains purely rigid transformations, which conforms well with the premises of GORE and TEASER++.
Examples of registration by different methods are presented  in Fig. \ref{3DMatch_exa}.

\subsubsection{Living room dataset}

%We tested the performance of the proposed method on large-scale field data
%using ”livingRoom.mat” from the MATLAB example ”3-D Point Cloud Registration and Stitching”, which consists of a series of 3D point sets obtained by
%continuously scanning a living room.

We use the large-scale field data contained in  ”livingRoom.mat” from the MATLAB example  ”3-D Point Cloud Registration and Stitching” to test  performances of different methods.
The dataset consists of a series of 3D point sets obtained by  scanning a living room.
%
%Following \cite{},
%We use a sequence of point clouds  
%depicting various parts of a living room
%%captured via kinect and 
% provided by Matlab
%to evaluate the performances of different  methods,
%%captured using kinect and  
%% are used,
%as illustrated in Fig. \ref{stitch_result}.
%
All the points lie in a
cube with an edge length of 4 metres. They were down-sampled with a gridsize
of 0.2 metre.

%Since the original point sets
%% (with size ranging from 2 to 3 units)
%are dense and  also contaminated by  noise, following \cite{lian2019low},
%%which can significantly slow down the registration methods,
%point sets are downsampled by using a
%box grid filter with grid size chosen as $0.2$ units
%%which merges  points in the same box to a single point
%% provided in Matlab 
%before being fed  to a registration method.

Following \cite{lian2019low},
we use the following approach to get the ground-truth:
%To quantitatively evaluate the performances of different methods,
We first use an ICP variant  equipped with point-to-plane metric \cite{ICP_point_to_plane}
to compute the rigid transformations between adjacent frames of point clouds.
As the adjacent-frame point clouds differ only slightly,
the transformations computed in this way can  be regarded as the ground-truth.
We then compute transformations between frames of point clouds in large separation  by compositing  adjacent-frame transformations.
%Fig. \ref{stitch_result} shows the result of merging all the point clouds by using these transformations.
%
We next get  ground-truth point correspondence by finding  mutually nearest neighbors  between the transformed model points and scene points.
%Since both point sets may contain outliers,
%point pairs with distances  larger than a threshold ($0.2$ is chosen in this paper) are not considered to be  corresponding.

%\begin{figure}[t]
%	%		\renewcommand{\arraystretch}{0} %remover vertical space in tabular
%	\centering
%	\includegraphics[width=.4\linewidth]{figures/stitch_result.eps} %ICP_stitch_result.png}
%	\caption{Result of merging all the point clouds by using  adjacent-frame transformations and composited transformations.
%		\label{stitch_result}}
%\end{figure}  

\begin{figure}[h]
	\centering

	\includegraphics[width=.5\linewidth]{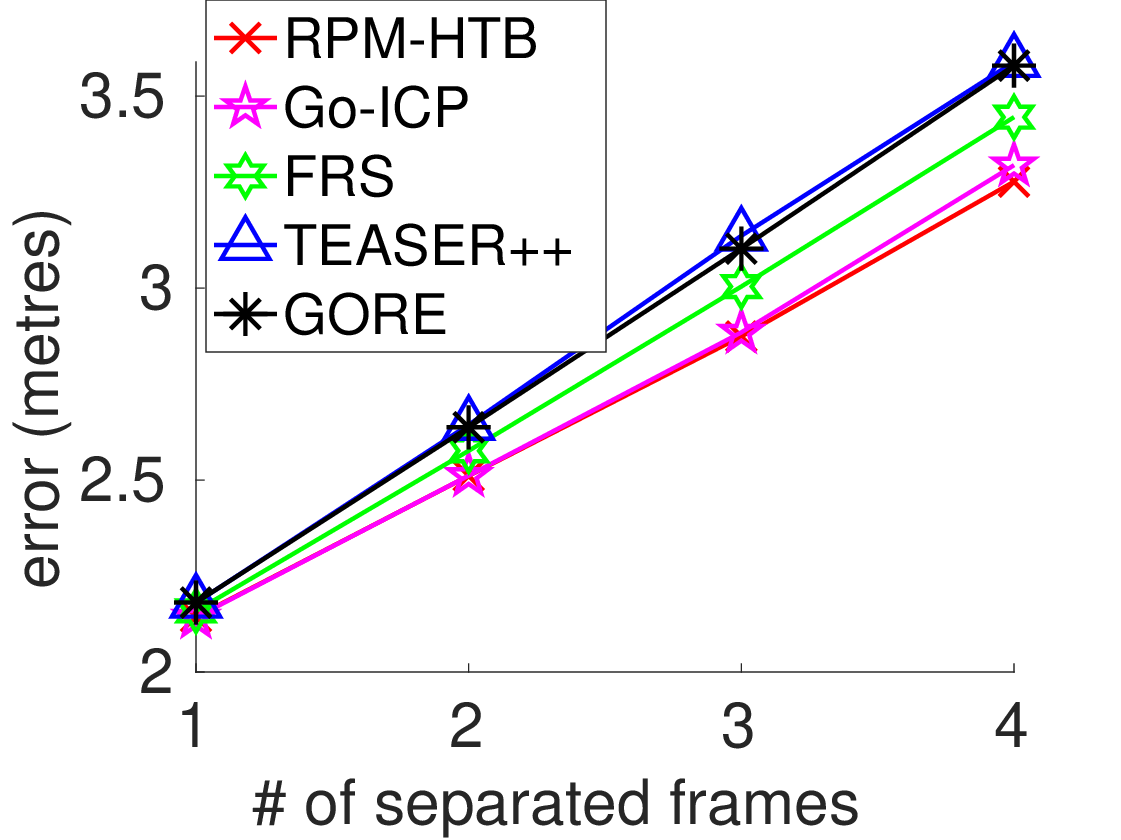}
	\caption{
		Average	registration errors by different methods on the living room dataset.
		\label{err_livingRoom}	}
\end{figure}

\begin{figure*}[h]
	\newcommand\scale{.95}	
	\centering
		\includegraphics[width=\scale\linewidth]{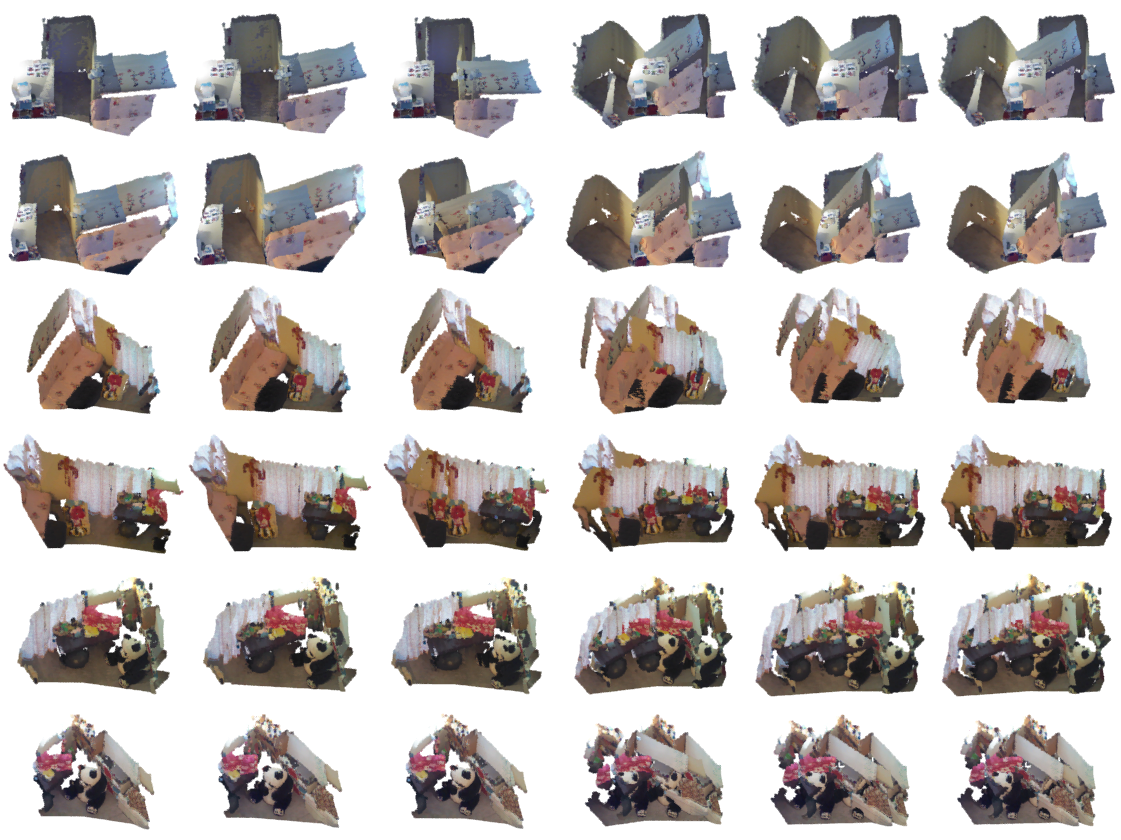}	
		\small
\begin{tabular}{@{\hspace{1mm}}c @{\hspace{9 mm}} c@{\hspace{15 mm}}  c@{\hspace{16 mm}}  c@{\hspace{13 mm}}   c@{\hspace{13 mm}}c}	
(a) ground-truth &(b) RPM-HTB & (c) Go-ICP & (d) FRS & (e) TEASER++ & (f) GORE
\end{tabular}
	%ICP_stitch_result.png}
	\caption{Registration   examples generated
by different methods
%		(from left to right)
		%		model point cloud,
		%		scene point cloud and registration results by
%		RPM-HTB, Go-ICP and FRS, TEASER++ and GORE
		on the living room dataset.
		%	The $n_p$ value for RPM-HTB and Go-ICP is chosen as $0.9$ the minimum of the cardinalities of two point sets.		
		\label{livingRoomTest}}
\end{figure*}

Since the pose difference between adjacent frames is small,
we set
tighter initial range %$\underline{\boldsymbol{\theta}}_0=-3 \mathbf 1_{n_\theta}$,
%%$\overline{\boldsymbol{\theta}}_0=3 \mathbf 1_{n_\theta}$,
$\left[  -\frac{\pi}{3} \mathbf 1_3,
\frac{\pi}{3} \mathbf 1_3\right ]$ for $\mathbf r$ in our method so as to speed up convergence.
%The maximum branching depth of our method is set as $10$.
%By using the transformation yielded by a method,  
%we define the alignment error as  mean of  squared Euclidean distances  between the transformed model inliers  and their corresponding scene inliers.
The registration errors  by different methods are reported in Fig. \ref{err_livingRoom},
where the $n_p$ value  is chosen as $0.9$ the minimum of the cardinalities of two point sets for RPM-HTB and Go-ICP.
Our method performs  better than other methods,
especially in the challenging 
cases where the frame separation is large.
% when matching problem becomes difficult (i.e., for point clouds with large frame separation).
Examples of registration results by different methods are presented in
Fig. \ref{livingRoomTest}.
One can see that RPM-HTB is  more robust to partial  overlap 
than other methods.
%Our method performs particularly well when there are distinct common geometry features between two point sets
%(e.g., the first row  in  Fig. \ref{livingRoomTest}).
%The running time by different methods is listed in Table \ref{time_livingRoom}.
%Our method is roughly 3  times more efficient than RPM-PF.
%Go-ICP is the slowest,
%especially in the case when matching problem becomes difficult (i.e., for point clouds with large frame separation).

%\begin{table}[h]
%	\renewcommand{\tabcolsep}{0.07cm}
%	\centering
%	\caption{
%		Average	registration errors.	
%	}
%	\label{err_livingRoom}
%	\small
%	\begin{tabular}{|c|c|c|c|c|} 
%		\hline
%		frame separation& 1 &2 &3 &4   \\\hline
%		RPM-HTB &  4.7188 &   \textbf{6.4257}  & \textbf{8.4673}  & \textbf{11.0388}\\\hline
%		Go-ICP & \textbf{4.7116} &  6.4299  &  8.5328 &  11.2994\\\hline
%		FRS & 4.7909  &  6.7814  &  9.2652  & 12.1526\\\hline
%	\end{tabular} 
%\end{table} 

\section{Conclusion}

\label{sec:conclude}
In this paper,
we presented a  BnB based point set registration algorithm
which can align partially overlapping point sets 
and can be rendered   invariant to the corresponding transformation.
%The lower bound function is obtained by relaxing  the
% trilinear and bilinear terms of the RPM objective function.
The method has the  merits that 
the lower bound can be  efficiently computed
via linear assignment
and
the dimensionality of 
the branching space  equals  the number of transformation parameters.
These merits enable the  method to be computationally   efficient
and  scale well with problem size.
%The method is also quite efficient and 
%
The use of the BnB algorithm enables the proposed method 
%Experimental results also showed that the method 
to be robust
%and efficiency of the proposed method in comparison  with the state-of-the-art approaches,
%particularly for tasks involving
to
 non-rigid deformations,
positional noise
and outliers in case when outliers are not mixed with  inliers
when compared with the state-of-the-art approaches.

Nevertheless,
%The weak points of
 the proposed method 
  can only tackle  transformations with few parameters due to 
exponential complexity of the proposed BnB algorithm
with respect to  the dimensionality of the branching space.
%for point set registration, due to the problem formulation can only work for few parameters, 
Besides, the method's robustness is substantially impacted by the parameter setting for $n_p$.
%for the proposed method,
This is because 
 branching depth instead of duality gap is adopted as the stopping criterion for the proposed method due to 
slow convergence of the  duality gap.
% of  the proposed method.
%
In the future, we will  explore better  methods capable of generating tighter lower bounds.
We will also develope adaptive $n_p$ setting scheme.

% which is not suitable for real-world tasks. An adaptive $n_p$ setting is necessary.

% Therefore, it would be beneficial to include a discussion of these limitations and suggestions for future research in the conclusion of this paper.

%Nevertheless,
%for the proposed method,
%the duality gap shrinks  slowly.
%% as the algorithm progresses.
%Thus, branching depth instead of duality gap is adopted as the stopping criterion for the proposed BnB algorithm.
%This choice  impedes its ablity to attain its full potential of  robustness.
%In the future,
%%we will explore better ways of setting the tolerance error $\epsilon$.
%we will  explore better  schemes capable of generating tighter lower bounds to address this issue.

\subsection*{ACKNOWLEDGMENTS}
%We  thank Dr. Lei Zhang of the department of computing at Hong Kong Polytechnic  University for his valuable suggestions during the initial writing of the paper.
This work was supported by
 National Natural Science Foundation of China under Grants  61773002 and  U19A2073, 
 the Fundamental Research Program of Shanxi Province, China under Grant 202103021223381
 and
Scientific and Technological Innovation Programs of Higher Education Institutions in Shanxi Province,  China under Grant 2022L517.

%and scientific and technologial innovation programs of higher education instituions in Shanxi (grant number 2019L0911).

%{\small
%\bibliographystyle{ieee_fullname}
%\bibliography{egbib}

%	\bibliographystyle{ieee}
\bibliographystyle{ieee_fullname}
%\bibliography{egbib}

%	\bibliographystyle{ieee}
\bibliography{DP_SC_CVPR}
%}

%\bibliographystyle{elsarticle-num}
%\bibliography{../DP_SC_rotate/DP_SC_CVPR}

%\begin{appendix}
%	content...
%	\section{ee}
%	\section{ff}
%\end{appendix}
\appendix

\section{Convergence properties of bilinear and trilinear monomials}\label{bil_tri_converge}

%\section*{Appendix}
%Let $(xy)_l\triangleq
%\frac{1}{2}(\underline{x}+\overline{x})y+ \frac{1}{2}(\underline{y}+\overline{y})x-\frac{1}{2}(\underline{x}\underline{y}+\overline{x}\overline{y})$.
\begin{proposition}
	Assume $\underline{x}\le x\le \overline{x}$ and
	$\underline{y}\le y\le \overline{y}$,
	%	Assume $\overline{y}-\underline{y}\ge 0$ is  finite.
	if 
	$|\overline{x}-\underline{x}|$ or 
	$|\overline{y}-\underline{y}|$
	is sufficiently small,
then	we will have  $|xy-(xy)_l|\le \epsilon$
	for any  real number $\epsilon>0$.
\end{proposition}
{\proof
	Since
	\begin{align}
	(xy)_{l}& \triangleq \frac{1}{2} \sum_i (xy)_{li} %\notag\\
	%	(xy)_{l} \triangleq
	%\frac{1}{2}((xy)_{l1} + (xy)_{l2}) \\ 
=	\frac{1}{2}(\underline{x}+\overline{x})y+ \frac{1}{2}(\underline{y}+\overline{y})x-\frac{1}{2}(\underline{x}\underline{y}+\overline{x}\overline{y}) 
	%	\le xy
	%	\label{bil_single_lb}
	\end{align}
It follows that
	\begin{align}
	|xy-(xy)_l| &=%\notag\\
	\frac{1}{2}	\left| (x-\underline{x})(y-\underline{y})
	+(\overline{x}-x)(\overline{y}-y) \right|
	\notag	\\
	&\le \frac{1}{2}| (x-\underline{x})(y-\underline{y})|
	+\frac{1}{2}|(\overline{x}-x)(\overline{y}-y) |
	\notag \\
	&\le 
	|\overline{x}-\underline{x}|\cdot|\overline{y}-\underline{y}|% =  M(\overline{x}-\underline{x})
	\end{align}
	Therefore,	
	if
	$|\overline{x}-\underline{x}|$ or 
	$|\overline{y}-\underline{y}|$
	is sufficiently small,
	we will have  $|xy-(xy)_l|\le \epsilon$
	for any $\epsilon>0$.

	%	 $\overline{y}-\underline{y}=0$,
	%	for any $\overline{x}-\underline{x}$,
	%	we  have  $xy-(xy)_l=0\le \epsilon$. 
	%	Otherwise,  $\overline{y}-\underline{y}>0$.
	%	Let $\overline{x}-\underline{x}\le \epsilon/(\overline{y}-\underline{y})$, 
	%	then we will have 
	%	$xy-(xy)_l\le \epsilon$.
}

%Based on the symmetry of the form of $(xy)_l$,
%we now have the following result:
%If the range of $x$ or $y$ is small enough,
%we will have  $xy-(xy)_l\le \epsilon$
%for any $\epsilon>0$.

\begin{proposition}
	Assume $\underline{x}\le x\le \overline{x}$,
$\underline{y}\le y\le \overline{y}$ and 
	$\underline{z}\le z\le \overline{z}$,
if the ranges of any two variables
	that makes up $xyz$ 
	are sufficiently small,
	%	Assume $\overline{x}-\underline{x}$ is finite.
	%	If $\overline{y}-\underline{y}$
	%	and $\overline{z}-\underline{z}$
	%	are sufficiently small,
	we will have $|xyz-(xyz)_l|\le \epsilon$ for any real number $\epsilon >0$.
\end{proposition}
{\proof
	In the case  $\underline{x}\le 0, \underline{y}\ge 0, \underline{z}\le 0, \overline{z}\ge 0$,	
	we have the following inequalities (which has been  presented in Sec. 3.2 of the main paper):
	%	\eqref{tril_lb}:
	\begin{align}
	(xyz)_{l1}\triangleq&\overline{y}\overline{z}x+ \overline{x}\overline{z}y+ \overline{x}\overline{y}z -2\overline{x}\overline{y}\overline{z}\le xyz \label{tril_lb_1}\\%checked
	(xyz)_{l2}\triangleq&\overline{y}\underline{z}x+ \underline{x}\overline{z}y+ \underline{x}\overline{y}z -\underline{x}\overline{y}\underline{z}-\underline{x}\overline{y}\overline{z}\le xyz \label{tril_lb_2}\\%checked
	(xyz)_{l3}\triangleq&\overline{y}\underline{z}x+ \underline{x}\underline{z}y+ \underline{x}\underline{y}z -\underline{x}\overline{y}\underline{z}-\underline{x}\underline{y}\underline{z}\le xyz  
%	\\ %checked
		\end{align}
\begin{align}	
	(xyz)_{l4}\triangleq&\underline{y}\overline{z}x+ \overline{x}\underline{z}y+ \overline{x}\underline{y}z -\overline{x}\underline{y}\overline{z}-\overline{x}\underline{y}\underline{z}\le xyz  %checked
	\\
	(xyz)_{l5}\triangleq&\underline{y}\underline{z}x+ \overline{x}\underline{z}y+ \underline{x}\underline{y}z -\overline{x}\underline{y}\underline{z}-\underline{x}\underline{y}\underline{z}\le xyz  
	\\%checked	
	(xyz)_{l6}\triangleq&\underline{y}\overline{z}x+ \underline{x}\overline{z}y+ \phi/(\overline{z}-\underline{z})z -\phi\underline{z}/(\overline{z}-\underline{z}) \notag\\
	&-\underline{x}\overline{y}\overline{z}-\overline{x}\underline{y}\overline{z}+\overline{x}\overline{y}\underline{z} \le xyz
	\label{tril_lb_6} 
	\end{align}
	where
	$
	\phi =\underline{x}\overline{y}\overline{z}-\overline{x}\overline{y}\underline{z}-\underline{x}\underline{y}\overline{z}+\overline{x}\underline{y}\overline{z} %checked
	$.

	For the first inequality \eqref{tril_lb_1},
	we have
	\begin{align}
	|xyz- (xyz)_{l1}|  %\notag\\	
&	=\left|
	xz(y-\overline{y})
	+ x\overline{y}(z - \overline{z})
	+ \overline{x}\overline{z}(\overline{y}-y)
	+\overline{x}\overline{y}(\overline{z}-z)
	\right| \notag \\
&	\le 
	|xz(y-\overline{y})|
	+	|x\overline{y}(z-\overline{z})|
	+ |\overline{x}\overline{z}(\overline{y}-y)|
	+ |\overline{x}\overline{y}(\overline{z}-z)
	| 
	\end{align}
	Therefore, if $|\overline{y}-\underline{y}|$
	and $|\overline{z}-\underline{z}|$
	are sufficiently small,
	we will have $|xyz-(xyz)_{l1}|\le \epsilon$.
	
	Similarly, we can prove
	\begin{align}
&	|xyz- (xyz)_{l1}|  %\notag\\
	\le |xz(y-\overline{y})| +|\overline{y}z(x-\overline{x})|
	+|\overline{y}\overline{z}(\overline{x}-x)|
	+|\overline{x}\overline{z}(\overline{y}-y)|
	\end{align}
	Therefore, if $|\overline{x}-\underline{x}|$
	and $|\overline{y}-\underline{y}|$
	are sufficiently small,
	we will have $|xyz-(xyz)_{l1}|\le \epsilon$.	
	
	Similarly,
	one can   also  prove
	\begin{align}
&	|xyz- (xyz)_{l1}|  %\notag\\
	\le |xy(z-\overline{z})| +
	|y\overline{z}(x-\overline{x})|+
	%|(xz-\overline{x}\overline{z})y| +
	|\overline{y}\overline{z}(\overline{x}-x)|+
	|\overline{x}\overline{y}(\overline{z}-z)| 
	\end{align}
	Therefore, if $|\overline{x}-\underline{x}|$
	and $|\overline{z}-\underline{z}|$
	are sufficiently small,
	we will have $|xyz-(xyz)_{l1}|\le \epsilon$.

In conclusion,
	we will have $|xyz-(xyz)_{l1}|\le \epsilon$
	if the ranges of any two variables
	that makes up $xyz$
	are sufficiently small.
	For other inequalities \eqref{tril_lb_2} to \eqref{tril_lb_6},
	we can prove similar results.
	
	Based on  the above results, therefore, for the average of the inequalities  \eqref{tril_lb_1} to \eqref{tril_lb_6},
	we will have $|xyz-(xyz)_{l}|\le \epsilon$ if the ranges of any two variables
	are sufficiently small.
	
	For other cases of the bounds on $x,y,z$,
	we can prove similar results.
	
}

\section{Derivation of Eq. \eqref{E_p_for_upper_bound} in the main text}\label{append_E_wrt_p}
%\begin{gather}
%	E(\mathbf p,\boldsymbol \theta)=\boldsymbol\theta^\top \left[\text{mat} ( \mathbf K\mathbf B_2\mathbf p   
%	){+} \mathbf C\right] \boldsymbol{\theta}  
%	{-}
%	2\boldsymbol\theta^\top \mathbf A\mathbf p  {+} 
%	\boldsymbol \rho^\top \mathbf p  
%\end{gather} 

Based on  Eq. \eqref{energy_case_one_generic},
given a feasible $\mathbf p$, $E$ is apparently a convex quadratic function of $\boldsymbol\theta$.
Thus, the optimal $\boldsymbol\theta$ minimizing $E$ can be obtained by solving the equation  $\frac{\partial E}{\partial \boldsymbol\theta}=0$,
where for convenient, we choose Eq. \ref{energy_case_one_final} as the form of $E$.
Consequently,  the  optimal $\boldsymbol\theta$ can be obtained as:
\begin{equation}
	\boldsymbol{\hat\theta}=
	\left[\text{mat} ( \mathbf K\mathbf B_2\mathbf p   
	){+} \mathbf C\right]^{-1} \mathbf A \mathbf p
\end{equation}
By substituting  $\boldsymbol{\hat\theta}$ into Eq. \ref{energy_case_one_final}, $\boldsymbol\theta$ is eliminated and we get an energy function in only one variable $\mathbf p$ as shown in Eq. \ref{E_p_for_upper_bound}.
%
%\begin{gather}
%	E(\mathbf p)= -\mathbf p^\top \mathbf A^\top [\text{mat} (\mathbf K \mathbf B_2 \mathbf p)+\mathbf C]^{-1} \mathbf A\mathbf p + \boldsymbol{\rho}^\top \mathbf p
%	\label{E_p}
%\end{gather}

\end{document}